\PassOptionsToPackage{table}{xcolor}
\documentclass[a4paper,fleqn]{cas-dc}

\usepackage[numbers]{natbib}

\usepackage{booktabs}
\usepackage{adjustbox}
\usepackage{array}
\usepackage{ragged2e}
\usepackage{tikz}
\usepackage{ulem}
\usepackage{graphicx}
\usepackage{subcaption}
\usepackage{hyperref}
\hypersetup{colorlinks,citecolor={blue},urlcolor={blue},breaklinks=true}
\usepackage{xcolor}
\usepackage[table]{xcolor}
\usepackage{multirow}
\usetikzlibrary{positioning,arrows.meta,calc,shadows,shadows.blur,fit,backgrounds}

\usepackage{tabularx}
\newlength{\tabcellpad}
\newcolumntype{Q}[1]{>{\kern\tabcellpad}m{#1}<{\endgraf\kern\tabcellpad}}
\usepackage{enumitem}
\usepackage{hhline}
\usepackage{balance}

\newcommand{\headerbreak}[1]{%
  \begin{minipage}[c][0.7cm][c]{\linewidth}
    \centering\textbf{#1}
  \end{minipage}%
}

\newlist{tabitem}{itemize}{1}
\setlist[tabitem]{
  leftmargin=*,
  label=\textbullet,
  labelsep=0.4em,
  itemsep=0pt,
  parsep=0pt,
  topsep=0pt,
  partopsep=0pt,
  nosep,
  before=\vspace{\tabcellpad}\justifying
}

\begin{document}
\let\WriteBookmarks\relax
\def\floatpagepagefraction{1}
\def\textpagefraction{.001}

\shorttitle{Facial Expression Recognition in the Deep Learning Era}    

\shortauthors{S. Georgiou et al.}  

\title [mode = title]{Facial Expression Recognition in the Deep Learning Era: A Systematic Multi-Criteria Review of Methods, Models, Datasets, Performance, Challenges, and Future Research Directions}  

\nonumnote{This work has received funding from the European Union's Horizon Europe research and innovation programme under Grant Agreement No. 101189557 project TORNADO (foundaTion mOdels for Robots that haNdle smAll, soft and Deformable Objects) and No. 101168042 project TRIFFID (auTonomous Robotic aId For increasing First responders Efficiency). The views and opinions expressed in this paper are those of the authors only and do not necessarily reflect those of the European Union or the European Commission.} 

\author[1]{Spyridon Georgiou}[orcid=0009-0006-4733-996X]
\ead{sgeorgiou@hua.gr}
\credit{Investigation, Methodology, Writing - original draft}

\author[1]{Aggelos Psiris}[orcid=0009-0005-1059-9509]
\ead{aggelospsiris@hua.gr}
\credit{Investigation, Methodology, Supervision, Writing - review \& editing}

\author[2]{Spyridon Evangelatos}[orcid=0000-0001-9119-1751]
\ead{sevangelatos@netcompany.com}
\credit{Writing - review \& editing}

\author[3]{Thomas Lagkas}[orcid=0000-0002-0749-9794]
\ead{tlagkas@cs.duth.gr}
\credit{Writing - review \& editing}

\author[4]{Vasileios Argyriou}[orcid=0000-0003-4679-8049]
\ead{vasileios.argyriou@kingston.ac.uk}
\credit{Writing - review \& editing}

\author[5]{Panagiotis Sarigiannidis}[orcid=0000-0001-6042-0355]
\ead{psarigiannidis@uowm.gr}
\credit{Writing - review \& editing}

\author[1]{Iraklis Varlamis}[orcid=0000-0002-0876-8167]
\ead{varlamis@hua.gr}
\credit{Investigation, Methodology, Supervision, Writing - review \& editing}

\author[1,6]{Georgios Th. Papadopoulos}[orcid=0000-0003-1686-421X]
\ead{g.th.papadopoulos@hua.gr}
\credit{Conceptualization, Investigation, Methodology, Supervision, Writing - review \& editing, Funding acquisition}

\affiliation[1]{
  organization={Department of Informatics and Telematics, Harokopio University of Athens},
  city={Athens},
  country={Greece}
}

\affiliation[2]{
  organization={Research \& Innovation Development Department, Netcompany-Intrasoft S.A.},
  city={Luxembourg},
  country={Luxembourg}
}

\affiliation[3]{
  organization={Department of Informatics, Faculty of Science, Democritus University of Thrace},
  city={Kavala},
  country={Greece}
}

\affiliation[4]{
  organization={Department of Networks and Digital Media, Kingston University},
  city={Kingston upon Thames},
  country={United Kingdom}
}

\affiliation[5]{
  organization={Department of Electrical and Computer Engineering, University of Western Macedonia},
  city={Kozani},
  country={Greece}
}

\affiliation[6]{
  organization={Archimedes Research Unit, Athena Research Center},
  city={Athens},
  country={Greece}
}

\cortext[8]{Corresponding author: Georgios Th. Papadopoulos}

\begin{abstract}
Facial Expression Recognition (FER) has progressed rapidly over the last decade, driven by the transition from handcrafted descriptors and shallow classifiers to deep convolutional, attention-based, vision--language, and foundation-model architectures, and by the parallel growth of large-scale in-the-wild benchmarks covering categorical, dimensional, compound, micro-expression, action unit (AU), and intensity-estimation tasks. Despite this progress, and although recent surveys have broadened the coverage of the field, existing reviews largely center on the categorical recognition of basic expressions, organize the literature along a single axis, or focus on specific tasks, modalities, or applications, while the relationships between these axes remain largely unexamined. This survey addresses this gap through a recent, cross-task, and cross-criteria synthesis of deep learning-based FER, with explicit links to the wider Facial Affect Recognition (FAR) domain, building on the analysis of  primary deep learning-based FER studies that were systematically selected and uniformly characterized. Its main contributions are: a) A description of the evolution of FER research into five distinct phases, from handcrafted features and classical machine learning to attention-based, vision--language, and foundation-model approaches, together with the key milestone works of each phase; b) A multi-criteria taxonomy that systematically analyzes the literature along seven complementary criteria, namely the targeted recognition task, the input modality, the face-analysis pre-processing pipeline, the underlying neural network architecture, the employed learning strategy, the acquisition setting, and the targeted application domain, together with a cross-criteria analysis of how the four method-level criteria (task, modality, architecture, and learning strategy) are jointly instantiated in the reviewed studies; c) A per-criterion comparative analysis of the literature, accompanied by critical insights into the relative strengths and limitations of the various categories of approaches under in-the-wild conditions; d) A task-organized review of the public datasets available for FER, together with a structured discussion of the corresponding annotation schemes, modalities, and evaluation protocols; e) A systematic compilation of the performance metrics adopted in the FER literature and a per-task quantitative comparison of representative recent deep learning methods on the most widely adopted benchmarks; and f) A structured discussion of the current challenges and the most promising future research directions of the field, spanning data and annotation; generalization and robustness; representation and temporal modeling; computation and deployment; evaluation; and trustworthiness, fairness, and ethical aspects.
\end{abstract}

\begin{keywords}
Facial expression recognition \sep Facial affect recognition \sep Deep learning \sep Vision--language and foundation models \sep Systematic survey
\end{keywords}

\maketitle

\section{Introduction}
\label{sec:Introduction}

Facial Expression Recognition (FER) is the automatic analysis of the face in images or video to infer what a person's face expresses. Its foundations trace back to Ekman and Friesen, who established a small set of basic emotions (happiness, sadness, anger, fear, surprise, disgust) universally recognized from the face~\citep{ekman1971constants}, and later formalized the Facial Action Coding System (FACS) in terms of measurable action units (AUs)~\citep{ekman1997what}. In this survey, FER is understood as the recognition of affect-related information from observable facial behavior, irrespective of the form of the output, i.e., categorical labels (basic or compound expressions~\citep{du2014compound,li2019blended}), AU activations, expression intensity, or continuous valence--arousal values as perceived from the face~\citep{mollahosseini2017affectnet,kollias2018aff}, targeting static images, dynamic sequences, and short-duration micro-expressions (micro-expression recognition, MER); categorical recognition from static images and from dynamic sequences is referred to as static FER (SFER) and dynamic FER (DFER), respectively. This task-level definition is consistent with the major FER benchmarks and challenges, in which dimensional labels (e.g., AffectNet, Aff-Wild2) and AU-intensity labels (e.g., DISFA) are assigned by annotators on the basis of the displayed facial behavior~\citep{mollahosseini2017affectnet,kollias2023abaw,kollias2018aff,mavadati2013disfa}. FER is thus distinguished from the broader Facial Affect Recognition (FAR), which aims at inferring the underlying, not directly observable, affective, cognitive, or clinical state of a person (e.g., felt emotion, pain, depression, engagement), typically by combining facial cues with additional modalities and contextual information. Hence, the distinction between the two lies in the target of inference (the displayed expression versus the latent state), rather than in the output representation (categorical, dimensional, or ordinal). FER is the focus of this survey, while FAR is addressed through the works that build on facial expression analysis for inferring such latent states (Sections~\ref{sec:scope} and~\ref{sec:app_domains}), which constitute the links of this review to the wider FAR domain. The adopted terminology and the relationships between FER, FAR, and the related terms used in the literature are summarized in Table~\ref{tab:terminology} (Section~\ref{sec:scope}). Throughout the manuscript, the acronym DFER denotes exclusively dynamic FER, and all abbreviations are listed in Table~\ref{tab:abbreviations}.

FER is important because the face is one of the most informative non-verbal channels of communication, conveying affective and intentional cues that complement, and at times override, the verbal channel~\citep{sariyanidi2014automatic}. Reliable machine perception of such cues underpins a wide range of real-world applications~\citep{picard2001toward,kumar2023artificial}, including human--computer and human--robot interaction; healthcare and clinical decision support (e.g., pain, depression, autism); intelligent transportation and driver monitoring, where detecting stress, drowsiness, or distraction can prevent accidents; education and e-learning; entertainment, gaming, and advertising; and security, surveillance, and lie detection. The proliferation of cameras across mobile devices, vehicles, public spaces, and immersive environments has further amplified its societal relevance.

Although recognizing facial expressions appears effortless for humans, automated FER from in-the-wild imagery remains challenging~\citep{li2022deep}. First, facial expressions exhibit high intra-class variability across individuals, cultures, ages, and genders, alongside low inter-class variability, with several emotions (e.g., fear vs.\ surprise) sharing strongly overlapping facial configurations. Second, real-world acquisition introduces severe nuisance factors, such as extreme head poses, partial occlusions (e.g., hair, glasses, masks, hands), non-uniform illumination, motion blur, and low resolution, which degrade recognition reliability. Third, expression labeling is intrinsically subjective and noisy, with annotators disagreeing on ambiguous samples, while most public datasets are heavily imbalanced, with neutral and happy faces predominating. Fourth, micro-expressions last fractions of a second and involve very subtle muscular activations, making their localization and recognition exceptionally demanding. Finally, FER systems are expected to generalize across datasets, demographics, and cultures, while respecting privacy, fairness, and ethical constraints, and remaining lightweight enough for edge deployment.

Over the past decade, Deep Learning (DL) has reshaped FER. Deep Convolutional Neural Networks (CNNs) replaced handcrafted descriptors (e.g., local binary patterns (LBP), histograms of oriented gradients (HOG), Gabor filters) and shallow classifiers with end-to-end learned representations, yielding consistent gains under both lab-controlled and in-the-wild conditions~\citep{li2020deep,he2016deep}. Recurrent and 3D convolutional architectures then enabled modeling of the temporal dynamics underlying dynamic expressions and AU activations. Vision Transformers and self-attention~\citep{dosovitskiy2020image,vaswani2017attention} further improved the reported performance by capturing long-range spatio-temporal dependencies. In parallel, generative paradigms (generative adversarial networks, GANs, and, more recently, diffusion models) alleviated data scarcity via augmentation, pose/identity normalization, and occlusion synthesis. Finally, Vision--Language Models (VLMs), Multi-modal Large Language Models (MLLMs), and self- or weakly supervised foundation models~\citep{radford2021clip} now enable open-vocabulary, instruction-following, explanation-aware FER beyond closed-set classification. Within this broader evolution, this survey focuses primarily on recent (2023--2026) deep learning-based FER works.

Against this background, this work provides a systematic, cross-task, and cross-criteria review of \textbf{recent deep learning-based Facial Expression Recognition}, with explicit links to the wider Facial Affect Recognition landscape. Its main contributions are as follows:
\begin{itemize}[leftmargin=*, itemsep=2pt, topsep=2pt, parsep=0pt]
    \item A description of the \textbf{evolution of FER research}, organized into a sequence of distinct phases, from handcrafted features and classical machine learning to attention-based, vision--language, and foundation-model approaches, together with the key milestone works associated with each phase.
    \item A \textbf{multi-criteria taxonomy} of the FER literature, which systematically analyzes the field along seven complementary criteria, each addressing a distinct question about a FER method, namely the targeted recognition task, the input modality, the adopted face-analysis pre-processing pipeline, the underlying neural network architecture, the employed learning strategy, the acquisition setting, and the targeted application domain.
    \item A \textbf{cross-criteria analysis} of 167 primary studies, which quantifies how the four method-level criteria (task, modality, architecture, and learning strategy) are jointly instantiated in the literature and identifies densely populated and under-explored combinations of them.
    \item A \textbf{per-criterion comparative analysis}, accompanied by critical insights, which contrasts the various categories of approaches and clarifies their relative strengths and limitations under in-the-wild conditions.
    \item A \textbf{task-organized review of the public datasets} available for FER, together with a structured discussion of the corresponding annotation schemes, modalities, and evaluation protocols.
    \item A systematic compilation of the \textbf{performance metrics} adopted in the FER literature and a \textbf{per-task quantitative comparison of representative recent deep learning methods} on the most widely adopted benchmarks, accompanied by critical observations on methodological trends and on recurring patterns among the highest reported results.
    \item A structured discussion of the \textbf{current challenges} and the most promising \textbf{future research directions} of the field.
\end{itemize}

\begin{table*}[!t]
  \caption{Comparative analysis of recent surveys in the field of Facial Expression Recognition (FER) and the wider Facial Affect Recognition (FAR) domain (FAA: facial affect analysis; check Table~\ref{tab:terminology} for the adopted terminology and Table~\ref{tab:abbreviations} for the abbreviations).}
  \label{tab:surveys}
  \centering
  \scriptsize

  \setlength{\aboverulesep}{0pt}
  \setlength{\belowrulesep}{0pt}
  
  \setlength{\tabcolsep}{4pt} 
  
  \renewcommand{\arraystretch}{1.3} 

  \setlength{\tabcellpad}{1pt}
  \rowcolors{2}{gray!20}{gray!2}

  \begin{adjustbox}{max width=\textwidth, max totalheight=0.93\textheight}%
  \begin{tabular}{@{}|
    >{\raggedright\arraybackslash}Q{2.4cm}|
    Q{2.7cm}| %
    >{\centering\arraybackslash}Q{0.8cm}|
    >{\raggedright\arraybackslash}Q{4.2cm}|
    >{\raggedright\arraybackslash}Q{5.2cm}|
    >{\raggedright\arraybackslash}Q{5.0cm}|
  @{}}
    \toprule
    \rowcolor{gray!50}
    \headerbreak{Article} &
    \headerbreak{Scope} &
    \headerbreak{DL focus} &
    \headerbreak{Methodology} &
    \headerbreak{Primary contributions} &
    \headerbreak{Limitations} \\
    \midrule

    \citet{sariyanidi2014automatic} (IEEE TPAMI, 2014) &
    Registration, representation, and recognition for facial affect analysis &
    Partial &
    \begin{tabitem}
      \item Component-wise structural analysis of FAA pipelines
      \item Coverage of registration, representation, and recognition stages
      \item Pre-deep-learning literature analysis
    \end{tabitem} &
    \begin{tabitem}
      \item Systematic decomposition of FAA into fundamental components
      \item Analysis of how each component addresses key challenges
      \item Joint treatment of basic and non-basic emotions and AUs
    \end{tabitem} &
    \begin{tabitem}
      \item Predates the dominance of deep learning
      \item Focused on handcrafted features
      \item Outdated datasets and benchmarks
    \end{tabitem} \\
    \midrule

    \citet{li2020deep} (IEEE TAC, 2022) &
    Deep facial expression recognition from images and videos &
    Yes &
    \begin{tabitem}
      \item Categorization of the literature by data pre-processing, network design, and training strategy
      \item Coverage of both lab-controlled and in-the-wild settings
      \item Analysis of the main public categorical FER benchmarks
    \end{tabitem} &
    \begin{tabitem}
      \item Comprehensive review of deep FER architectures
      \item Discussion of how to handle identity bias and expression-unrelated variations
      \item Discussion of open challenges and future research directions
    \end{tabitem} &
    \begin{tabitem}
      \item Limited to the categorical model
      \item Excludes AUs and dimensional affect
      \item Coverage up to 2022
    \end{tabitem} \\
    \midrule

    \citet{canal2022survey} (Inf. Sci., 2022) &
    State-of-the-art literature review on facial emotion recognition &
    Partial &
    \begin{tabitem}
      \item Systematic PRISMA-style review across well-established databases
      \item Analysis of 94 methods extracted from 51 papers
      \item Statistical comparison of classical and DL-based pipelines
    \end{tabitem} &
    \begin{tabitem}
      \item Workflow-level analysis (pre-processing, feature extraction, classification)
      \item Quantitative comparison of classical vs.\ neural pipelines
      \item Discussion of open challenges and future research directions
    \end{tabitem} &
    \begin{tabitem}
      \item Gives equal weight to classical and DL methods
      \item No taxonomy of modern DL architectures
      \item Limited treatment of DFER, AUs, and micro-expressions
    \end{tabitem} \\
    \midrule

    \citet{li2022deep} (IEEE TAC, 2022) &
    Deep learning for micro-expression recognition &
    Yes &
    \begin{tabitem}
      \item Pipeline-based taxonomy (datasets, pre-processing, networks, evaluation)
      \item Analysis of small-scale benchmarks and protocols
      \item Coverage of the 2016--2022 literature
    \end{tabitem} &
    \begin{tabitem}
      \item First focused survey on deep MER
      \item Systematic framework and pipeline-step taxonomy
      \item Discussion of practical importance, key skills, and ethics
    \end{tabitem} &
    \begin{tabitem}
      \item Restricted to micro-expressions
      \item Excludes macro-expression recognition, AUs, and dimensional affect
      \item Coverage up to 2022
    \end{tabitem} \\
    \midrule

    \citet{jampour2022multiview} (IEEE TAC, 2022) &
    Multi-view facial expression recognition &
    Partial &
    \begin{tabitem}
      \item Categorization of approaches by how they handle head-pose variation
      \item Joint analysis of classical and deep methodologies
      \item Coverage of single- and multi-view evaluation protocols
    \end{tabitem} &
    \begin{tabitem}
      \item First multi-view FER-specific survey
      \item Systematic categorization across pose-variation strategies
      \item Bridges traditional and DL methodologies
    \end{tabitem} &
    \begin{tabitem}
      \item Single-task scope (multi-view FER)
      \item Limited direct quantitative comparison
      \item Limited coverage of recent Transformer-based methods
    \end{tabitem} \\
    \midrule

    \citet{liu2022graph} (IEEE TAC, 2023) &
    Graph-based facial affect analysis &
    Yes &
    \begin{tabitem}
      \item Categorization of graph-based FAA into representations and GNN variants
      \item Analysis of algorithm evolution and applications
      \item Inclusion of representative studies across FAA tasks
    \end{tabitem} &
    \begin{tabitem}
      \item First systematic review dedicated to graph-based FAA
      \item In-depth, novel categorization with emphasis on state-of-the-art GNNs
      \item Discussion of graph-based affective representations
    \end{tabitem} &
    \begin{tabitem}
      \item Single-architecture focus (graphs)
      \item No holistic comparison with CNN\slash Transformer paradigms
      \item No dataset\slash task-level meta-analysis
    \end{tabitem} \\
    \midrule

    \citet{adyapady2023comprehensive} (Multimed. Syst., 2023) &
    Comprehensive review of FER techniques &
    Yes &
    \begin{tabitem}
      \item Narrative review of algorithm development over the past decades
      \item Coverage of advantages and constraints per method family
      \item Inclusion of ethical and privacy considerations
    \end{tabitem} &
    \begin{tabitem}
      \item Broad overview of FER algorithms from visual cues
      \item Explicit discussion of ethical concerns regarding privacy
      \item Discussion of open challenges and future research directions
    \end{tabitem} &
    \begin{tabitem}
      \item Narrative-style analysis
      \item No multi-criteria taxonomy
      \item Limited treatment of Transformers and VLMs
    \end{tabitem} \\
    \midrule

    \citet{saadi2024driver} (Expert Syst. Appl., 2024) &
    FER for driver state monitoring &
    Yes &
    \begin{tabitem}
      \item Application-specific categorization of driver facial expression recognition methods
      \item Analysis of driver-oriented datasets and evaluation settings
      \item Coverage of 2018--2024 literature
    \end{tabitem} &
    \begin{tabitem}
      \item First survey dedicated to driver facial expression recognition
      \item Overview of emotion classes and acquisition protocols in driving
      \item Discussion of open challenges and future research directions
    \end{tabitem} &
    \begin{tabitem}
      \item Application-specific (driving)
      \item Excludes broader FAA tasks
      \item Coverage up to 2024
    \end{tabitem} \\
    \midrule

    \citet{khan2024advances} (Information, 2024) &
    Methods, benchmarks, models, and datasets for deep FER &
    Yes &
    \begin{tabitem}
      \item Timeline-based analysis of deep FER methods and datasets
      \item Categorization by backbone family and benchmark
      \item Coverage of major in-the-wild categorical benchmarks
    \end{tabitem} &
    \begin{tabitem}
      \item Timeline of deep FER architectural evolution
      \item Joint discussion of methods, benchmarks, and datasets
      \item Discussion of overfitting and expression-unrelated variations
    \end{tabitem} &
    \begin{tabitem}
      \item Centered on categorical FER
      \item Limited coverage of AUs, micro-expressions, and dimensional affect
      \item No per-criterion comparative analysis
    \end{tabitem} \\
    \midrule

    \citet{wang2024survey_fer} (arXiv, 2024) &
    FER of static and dynamic emotions &
    Yes &
    \begin{tabitem}
      \item Joint model-oriented and challenge-focused categorization
      \item Separate analysis of SFER and DFER pipelines
      \item Coverage of recent Transformer- and attention-based methods
    \end{tabitem} &
    \begin{tabitem}
      \item Joint review of SFER and DFER
      \item Discussion of key challenges (e.g., uncertainty, key-frame sampling)
      \item Discussion of open challenges and future research directions
    \end{tabitem} &
    \begin{tabitem}
      \item Focused on macro-expression recognition (SFER\slash DFER)
      \item Limited treatment of AUs, micro-expressions, and VLM\slash foundation models
      \item Not peer-reviewed
    \end{tabitem} \\
    \midrule

    \citet{rehman2025comprehensive} (CMC, 2025) &
    FER from static facial images &
    Partial &
    \begin{tabitem}
      \item Categorization into machine learning, deep, and hybrid approaches
      \item Coverage of pre-processing, augmentation, and feature extraction
      \item Performance comparison on public benchmarks
    \end{tabitem} &
    \begin{tabitem}
      \item Joint comparison of classical, deep, and hybrid pipelines
      \item Overview of FER datasets
      \item Discussion of limitations and future research directions
    \end{tabitem} &
    \begin{tabitem}
      \item Restricted to static images and basic emotions
      \item Limited treatment of Transformers; no VLMs
      \item Narrative-style analysis
    \end{tabitem} \\
    \midrule

    \citet{shehu2025emotion} (Neurocomputing, 2025) &
    Emotion categorization (basic expressions and neutral) from facial expressions &
    Partial &
    \begin{tabitem}
      \item Narrative review of posed and non-posed datasets
      \item Review of emotion categorization tools and common deep backbones
      \item Analysis of robustness to adversarial and emotion-irrelevant perturbations
    \end{tabitem} &
    \begin{tabitem}
      \item Critical discussion of pixel-centric end-to-end deep networks
      \item Review of AU-, landmark-, attention-, and lateralization-based methods
      \item Discussion of open challenges and future research directions
    \end{tabitem} &
    \begin{tabitem}
      \item Restricted to categorical recognition; micro-expressions excluded
      \item No treatment of AU detection, valence--arousal, compound, or intensity tasks
      \item No coverage of VLM/MLLM-based FER
    \end{tabitem} \\
    \midrule

    \citet{sarvakar2025revolutionizing} (Discov. Artif. Intell., 2025) &
    FER models, methodologies, and datasets &
    Partial &
    \begin{tabitem}
      \item Classical vs.\ deep learning taxonomy of FER approaches
      \item Datasets grouped by image/sequence and lab/in-the-wild setting
      \item Coverage of 2024--2025 methods
    \end{tabitem} &
    \begin{tabitem}
      \item Timeline of FER methods and datasets
      \item Comparison of 60 models and 21 datasets
      \item Discussion of open challenges and future research directions
    \end{tabitem} &
    \begin{tabitem}
      \item Centered on categorical FER
      \item Single-axis (architecture) categorization; no cross-criteria analysis
      \item Limited coverage of VLM/MLLM-based FER
    \end{tabitem} \\
    \midrule

    \citet{kus2026systematic} (Intell. Syst. Appl., 2026) &
    Vision Transformers and XAI in multi-modal emotion recognition (including facial input) &
    Yes &
    \begin{tabitem}
      \item PRISMA-based systematic review of 67 studies (2021--2025)
      \item Categorization by fusion level (early, late, hybrid)
      \item Review of multi-modal (visual, audio, EEG, text) datasets
    \end{tabitem} &
    \begin{tabitem}
      \item Integrated review of ViT-based models and XAI methods
      \item Comparative assessment of model families
      \item Discussion of privacy, cost, and explainability challenges
    \end{tabitem} &
    \begin{tabitem}
      \item Restricted to multi-modal ViT/XAI-based emotion recognition
      \item Mainly classification-based; micro-expression, AU, compound, and intensity tasks largely not covered
      \item VLM/MLLM-based FER discussed only as a future direction
    \end{tabitem} \\
    \midrule

    \textbf{Current survey} &
    Recent (2023--2026), cross-task, and cross-criteria review of deep learning-based FER, with links to the wider FAR landscape &
    Yes &
    \begin{tabitem}
      \item Criteria-based selection and uniform characterization of 167 primary deep learning-based FER studies, out of the 325 cited works
      \item Seven-criterion taxonomy (task, modality, pre-processing, architecture, learning strategy, acquisition setting, application domain)
      \item Per-criterion comparative analysis of the literature
      \item Cross-criteria analysis of the joint instantiation of the four method-level taxonomy axes
      \item Task-organized review of public datasets and quantitative performance comparison
    \end{tabitem} &
    \begin{tabitem}
      \item Research-evolution phases 
      \item Systematic compilation of FER performance metrics and quantitative comparisons per task
      \item Cross-task and cross-criteria synthesis, including VLM/MLLM-based approaches, with identification of under-explored combinations
      \item Discussion of open challenges and future research directions
    \end{tabitem} &
    \begin{tabitem}
      \item Reliance on reported results; no controlled re-evaluation under a common protocol
      \item Heterogeneous protocols preclude strict quantitative comparison and meta-analysis
      \item Partially overlapping (non-orthogonal) taxonomy criteria
      \item Criteria-driven, prioritized corpus; partial coverage of 2026 works
      \item Per-stage counts not logged; no formal double screening
    \end{tabitem} \\
    \bottomrule
  \end{tabular}%
  \end{adjustbox}
\end{table*}

A considerable number of survey papers have addressed FER and the wider FAR domain, and recent surveys~\citep{rehman2025comprehensive,shehu2025emotion,sarvakar2025revolutionizing,kus2026systematic} have considerably narrowed the gap in the coverage of recent deep learning-based FER. Table~\ref{tab:surveys} compares the present work with representative recent surveys, with respect to their scope, focus on deep learning, adopted methodology, primary contributions, and main limitations. As can be observed, existing surveys generally exhibit the following limitations: a) They are typically narrow in scope, adopting a single-task perspective, focusing only on macro-expression recognition~\citep{li2020deep,khan2024advances,wang2024survey_fer}, micro-expressions~\citep{li2022deep}, multi-view FER~\citep{jampour2022multiview}, or driver facial expression recognition~\citep{saadi2024driver}, while the most recent broad surveys still largely center on the categorical recognition of basic expressions~\citep{rehman2025comprehensive,shehu2025emotion,sarvakar2025revolutionizing}, in one case restricted to static images~\citep{rehman2025comprehensive}; b) They are frequently restricted to a single architectural family, such as graph-based approaches~\citep{liu2022graph}, or to a specific combination of model family and modality setting, such as Vision Transformer (ViT)- and explainable AI (XAI)-based multi-modal emotion recognition~\citep{kus2026systematic}; c) They predate the widespread adoption of modern paradigms (CNNs, Transformers, VLMs, foundation models)~\citep{sariyanidi2014automatic} or extend only up to the early Transformer era~\citep{li2020deep,canal2022survey,adyapady2023comprehensive}; and d) They commonly adopt a narrative-style discussion~\citep{adyapady2023comprehensive,rehman2025comprehensive,shehu2025emotion,sarvakar2025revolutionizing}, without a systematic multi-criteria taxonomy or a per-criterion comparative analysis of methods, modalities, learning strategies, acquisition settings, and application domains. Moreover, none of them examines how the different axes of the field (e.g., recognition task, architecture family, learning strategy, and input modality) relate to each other. In contrast, the distinctive contribution of the present survey does not lie in breadth alone, but in a recent, cross-task, and cross-criteria synthesis, since it focuses primarily on the most recent (2023--2026) deep learning advances, including VLM/MLLM-based approaches; it jointly covers six FER tasks, with per-task quantitative comparisons; and it characterizes each of the 167 primary studies along a seven-criterion taxonomy, analyzing not only each criterion separately, through a per-criterion comparative analysis with critical insights, but also the relationships between the criteria (Section~\ref{sec:cross_criteria}) and their evolution over time (Fig.~\ref{fig:bibliometrics}(c)--(d)). This synthesis is complemented by a task-organized treatment of datasets, evaluation protocols, challenges, and future research directions.

The remainder of the manuscript is organized as follows. Section~\ref{sec:bibliometrics} describes the adopted systematic literature review methodology. Section~\ref{sec:Evolution} presents the evolution of FER research through a set of distinct phases and the corresponding milestone works. Section~\ref{sec:Taxonomy} introduces the key criteria used to systematically analyze the FER literature and the resulting main categories of methods, along with a quantitative analysis of how the four method-level criteria are jointly instantiated in the reviewed studies. Sections~\ref{sec:recognition_tasks}--\ref{sec:app_domains} analyze the FER literature in detail according to the targeted recognition task, the input modality, the face-analysis pre-processing pipeline, the neural network architecture, the learning strategy, the acquisition setting, and the application domain, respectively. Section~\ref{sec:Datasets} reviews the main public datasets available for FER, and Section~\ref{sec:performance_metrics} discusses the corresponding performance metrics and provides a quantitative comparative analysis of representative recent methods. Moreover, Section~\ref{sec:challenges} systematically discusses the current challenges and Section~\ref{sec:future_directions} outlines the most promising future research directions. Section~\ref{sec:conclusion} concludes the manuscript.

\section{Literature review methodology}
\label{sec:bibliometrics}

To identify and map the deep learning-based FER literature thoroughly, while also detecting key concepts and trends, a systematic approach was followed to ensure the coverage and relevance of the selected works. In particular, a structured literature review methodology was adopted, consisting of the main (iterative) steps described below. In the context of this work, the term ``systematic'' refers to the consistent application of explicit and uniform criteria at every stage of the selection process, as well as to the uniform characterization of every retained study along the same set of criteria, rather than to the reporting of record counts at each stage of the database search; this choice is explained in Sections~\ref{sec:data_extraction} and~\ref{sec:review_limitations}.

\definecolor{darkblue}{RGB}{44,92,136}
\definecolor{lightblue}{RGB}{124,177,212}
\definecolor{orange}{RGB}{227,146,75}
\definecolor{lightgrey}{RGB}{204,204,204}
\definecolor{bgbox}{RGB}{250,250,250}

\begin{figure*}[!t]
    \centering
    \resizebox{\textwidth}{!}{%
%
    }
    \vspace{0.2cm}
  \caption{Key bibliometric statistics of the analyzed deep learning-based FER literature: (a) Article types, (b) Most frequent publication venues (venues with at least three primary studies), (c) Number of primary studies per publication year, broken down by (first-listed) recognition task, (d) Share of the architecture families of the proposed models per period ($n$: number of primary studies per period), and (e) Number of primary studies per (first-listed) recognition task; ``other affective states'' denotes studies that analyze facial expressions for inferring latent affective or clinical states, e.g., pain, depression, or engagement (Section~\ref{sec:scope}). All panels are computed over the 167 primary deep learning-based FER studies that constitute the analyzed corpus, i.e., the contextual references are excluded (Sections~\ref{sec:selection} and~\ref{sec:data_extraction}); *2026: partial coverage of the publication year (Section~\ref{sec:search_strategy}).}
  \label{fig:bibliometrics}
\end{figure*}

\subsection{Scope and objectives formulation}
\label{sec:scope}
The goal of this survey was to review the deep learning-based FER literature, i.e., approaches that address facial expression recognition tasks (namely, categorical macro-expression recognition, dimensional valence--arousal estimation, compound expression recognition, micro-expression recognition, action unit detection, and expression intensity estimation) using deep learning methodologies and relying on facial visual input (and, where applicable, complementary multi-modal signals such as audio, text, or physiological streams). In particular, the focus was on identifying the most recent advancements, with emphasis on works published in 2023--2026 (which account for 93 of the 167 primary studies analyzed, see Section~\ref{sec:data_extraction}), as well as milestone works that made substantial contributions to the field, with the specific objectives of tracking: a) The main categories of methods based on multiple criteria, b) The publicly available datasets, c) The key trends and the evolution of the field, and d) The current challenges and future research directions. In line with the definition adopted in Section~\ref{sec:Introduction}, all six tasks are considered at the level of the displayed facial behavior, i.e., their targets describe what the face expresses, rather than the underlying affective state of the person. Works that analyze facial expressions for inferring such a latent state (e.g., pain, depression, autism-related traits, engagement, or driver state) were also retained, provided that their facial analysis component relies on the above tasks or on the corresponding representations (check `IC2' in Section~\ref{sec:eligibility}); they are grouped as ``other affective states'' in the corpus statistics (Figs.~\ref{fig:bibliometrics} and~\ref{fig:crosscriteria}) and constitute the links of this review to FAR. The relationships between FER, FAR, and the related terms that are used in the literature, along with the terminology adopted in this survey, are summarized in Table~\ref{tab:terminology}, and the abbreviations used throughout the manuscript are listed in Table~\ref{tab:abbreviations}.

\begin{table*}[!t]
  \caption{Terminology adopted in this survey and relationships between the related terms used in the literature, ordered from the broadest to the most specific concept (affective computing $\supset$ FAR $\supset$ FER). Unless otherwise stated, FER and the six task names listed below are used throughout the manuscript; the remaining terms appear only when reporting the scope of cited works (e.g., Table~\ref{tab:surveys}) or when referring to multi-modal settings.}
  \label{tab:terminology}
  \centering
  \scriptsize

  \setlength{\aboverulesep}{0pt}
  \setlength{\belowrulesep}{0pt}
  \setlength{\tabcolsep}{4pt}
  \renewcommand{\arraystretch}{1.3}
  \renewcommand{\tabularxcolumn}[1]{m{#1}}

  \rowcolors{2}{gray!20}{gray!2}

  \begin{tabularx}{\textwidth}{@{}|>{\raggedright\arraybackslash}m{3.4cm}|X|@{}}
  \toprule
  \rowcolor{gray!50}
  \headerbreak{Term} & \headerbreak{Definition and usage in this survey} \\
  \midrule
  Affective computing & Research field concerned with the recognition, interpretation, and synthesis of human affect by computing systems, from any modality (e.g., face, voice, text, physiological signals, behavior)~\citep{picard2001toward}. \\ \midrule
  Affect recognition / emotion recognition & Inference of affective states from any modality or combination of modalities, not necessarily including the face; used in this survey only for multi-modal or non-face-specific settings. \\ \midrule
  Facial Affect Analysis (FAA) & Umbrella term adopted by part of the surveyed literature~\citep{sariyanidi2014automatic,liu2022graph}, covering the analysis of facial affect in a broad sense (including both FER and FAR tasks); used in this survey only when referring to these works (Table~\ref{tab:surveys}). \\ \midrule
  Facial Affect Recognition (FAR) & Inference of the underlying, not directly observable, affective, cognitive, or clinical state of a person (e.g., felt emotion, pain, depression, engagement) from the face, typically complemented by additional modalities and contextual information. It encompasses FER, whose outputs are commonly used as intermediate cues; in this survey, it is addressed through the works that analyze facial expressions for inferring such states (``other affective states'', Section~\ref{sec:scope}). \\ \midrule
  Facial Expression Recognition (FER), also expression recognition or facial expression analysis & Recognition of affect-related information from observable facial behavior, irrespective of the output representation; it comprises six tasks (Section~\ref{sec:taxonomy_criteria}), namely categorical macro-expression recognition, dimensional valence--arousal (VA) estimation (as perceived from the face), compound expression recognition (CFER), micro-expression recognition (MER), action unit (AU) detection, and expression intensity estimation. It constitutes the focus of this survey. \\ \midrule
  Facial emotion recognition / emotion categorization & Terms used in part of the literature for the assignment of the displayed facial expression to discrete expression categories (typically named after the basic emotions); treated in this survey as synonyms of categorical (macro-)expression recognition. \\ \midrule
  Static FER (SFER) / dynamic FER (DFER) & Categorical macro-expression recognition from single images (SFER) or from video sequences (DFER), i.e., the categorical recognition task instantiated on static or dynamic input (Section~\ref{sec:taxonomy_criteria}). In this survey, DFER denotes exclusively dynamic FER; driver facial expression recognition, which part of the driver-monitoring literature also abbreviates as DFER~\citep{saadi2024driver}, is always written in full (Section~\ref{ssec:app_driver}). \\
  \bottomrule
  \end{tabularx}
\end{table*}

\begin{table*}[!htbp]
  \caption{Abbreviations used in this survey. Names of methods and datasets are given as in the original publications and are not listed. The acronym DFER always denotes dynamic facial expression recognition; driver facial expression recognition is always written in full.}
  \label{tab:abbreviations}
  \centering
  \scriptsize

  \setlength{\aboverulesep}{0pt}
  \setlength{\belowrulesep}{0pt}
  \setlength{\tabcolsep}{4pt}
  \renewcommand{\arraystretch}{1.3}
  \renewcommand{\tabularxcolumn}[1]{m{#1}}

  \rowcolors{2}{gray!20}{gray!2}

  \begin{tabularx}{\textwidth}{@{}|>{\raggedright\arraybackslash}m{1.7cm}|X|>{\raggedright\arraybackslash}m{1.7cm}|X|@{}}
  \toprule
  \rowcolor{gray!50}
  \headerbreak{Abbreviation} & \headerbreak{Definition} & \headerbreak{Abbreviation} & \headerbreak{Definition} \\
  \midrule
   ABAW & Affective Behavior Analysis in-the-wild (challenge series) & LBP & Local binary patterns \\
  \midrule
   ADAS & Advanced driver-assistance system & LLM & Large language model \\
  \midrule
   AR/VR & Augmented/virtual reality & LoRA & Low-rank adaptation \\
  \midrule
   ASD & Autism spectrum disorder & LOSO & Leave-one-subject-out \\
  \midrule
   AU & Action unit & LSTM & Long short-term memory \\
  \midrule
   AUD & Action unit detection (task code in Tables~\ref{tab:datasets_fer_lab} and~\ref{tab:datasets_fer_itw}) & MAE & Masked autoencoder \\
  \midrule
   CCC & Concordance correlation coefficient & MER & Micro-expression recognition \\
  \midrule
   CFER & Compound (facial) expression recognition & MFCC & Mel-frequency cepstral coefficients \\
  \midrule
   CLAHE & Contrast-limited adaptive histogram equalization & MLLM & Multi-modal large language model \\
  \midrule
   CNN & Convolutional neural network & MMD & Maximum mean discrepancy \\
  \midrule
   CORAL & Correlation alignment & MSE & Mean squared error \\
  \midrule
   DA & Domain adaptation & MSX & Multi-spectral dynamic imaging \\
  \midrule
   DETR & Detection Transformer & NICU & Neonatal intensive care unit \\
  \midrule
   DFER & Dynamic facial expression recognition & NIR & Near-infrared \\
  \midrule
   DL & Deep learning & PCC & Pearson correlation coefficient \\
  \midrule
   DOI & Digital Object Identifier & PEFT & Parameter-efficient fine-tuning \\
  \midrule
   ECG & Electrocardiography & PPG & Photoplethysmography \\
  \midrule
   EDA & Electrodermal activity & PRISMA & Preferred Reporting Items for Systematic Reviews and Meta-Analyses \\
  \midrule
   EEG & Electroencephalography & RGB & Red--green--blue (visible-spectrum) imagery \\
  \midrule
   EMA & Exponential moving average & RMSE & Root mean square error \\
  \midrule
   EOG & Electrooculography & RNN & Recurrent neural network \\
  \midrule
   ERI & Emotional reaction intensity & rPPG & Remote photoplethysmography \\
  \midrule
   FAA & Facial affect analysis & SAGR & Sign agreement metric \\
  \midrule
   FACS & Facial Action Coding System & SFER & Static facial expression recognition \\
  \midrule
   FAR & Facial affect recognition & SIFT & Scale-invariant feature transform \\
  \midrule
   FEIE & Facial expression intensity estimation & SSD & Single-shot detector \\
  \midrule
   FER & Facial expression recognition & SSL & Self-supervised learning \\
  \midrule
   FLOPs & Floating-point operations & SVM & Support vector machine \\
  \midrule
   FPS & Frames per second & TCN & Temporal convolutional network \\
  \midrule
   GAN & Generative adversarial network & TIR & Thermal infrared \\
  \midrule
   GCN & Graph convolutional network & TSN & Temporal segment network \\
  \midrule
   GNN & Graph neural network & TV-L1 & Total variation--L1 (optical flow) \\
  \midrule
   GRU & Gated recurrent unit & UAR & Unweighted average recall \\
  \midrule
   GSR & Galvanic skin response & UDA & Unsupervised domain adaptation \\
  \midrule
   HCI & Human--computer interaction & UF1 & Unweighted F1-score \\
  \midrule
   HOG & Histogram of oriented gradients & VA & Valence--arousal \\
  \midrule
   HRI & Human--robot interaction & VIS & Visible-spectrum imagery \\
  \midrule
   ICC & Intra-class correlation coefficient & ViT & Vision Transformer \\
  \midrule
   IoT & Internet of Things & VLM & Vision--language model \\
  \midrule
   IR & Infrared & WAR & Weighted average recall \\
  \midrule
   KAN & Kolmogorov--Arnold network & XAI & Explainable artificial intelligence \\
  \midrule
   KD & Knowledge distillation &  &  \\
  \bottomrule
  \end{tabularx}
\end{table*}

\subsection{Information sources and search strategy}
\label{sec:search_strategy}
To ensure broad and thorough coverage of the relevant literature, the search strategy involved querying several major scientific databases, including IEEE Xplore, Scopus, DBLP, arXiv, and Web of Science, complemented by Google Scholar for citation searching. The actual search was performed by combining targeted keywords (e.g., ``facial expression recognition'', ``facial affect'', ``facial emotion'', ``action unit'', ``micro-expression'', ``valence'', ``arousal'', ``FACS'', etc.) and Boolean operators (i.e., ``AND'', ``OR'', ``NOT''), while explicitly excluding loosely related modalities and tasks (e.g., ``face identification'', ``face verification'', ``face anti-spoofing'', ``speech emotion'', etc.) that fall outside the scope of facial-input-based FER. To ensure contemporary relevance, the search focused primarily on works published in recent years (2023--2026); however, certain seminal and milestone studies introduced earlier were also included. The query used in the Scopus database was as follows:

\smallskip
{\ttfamily\spaceskip=0.525em plus 0.5em minus 0.05em\relax\noindent{\textcolor{cyan}{TITLE-ABS-KEY} ( \textcolor{brown}{"facial expression recognition"} \textcolor{blue}{OR} \textcolor{brown}{"facial affect"} \textcolor{blue}{OR} \textcolor{brown}{"facial emotion"} \textcolor{blue}{OR} ( (\textcolor{brown}{"action unit"} \textcolor{blue}{OR} \textcolor{brown}{"micro-expression"} \textcolor{blue}{OR} \textcolor{brown}{"valence"} \textcolor{blue}{OR} \textcolor{brown}{"arousal"} \textcolor{blue}{OR} \textcolor{brown}{"FACS"}) \textcolor{blue}{AND} (\textcolor{brown}{"face"} \textcolor{blue}{OR} \textcolor{brown}{"facial"}) ) ) \textcolor{blue}{AND NOT} \textcolor{cyan}{TITLE-ABS-KEY} ( \textcolor{brown}{"face identification"} \textcolor{blue}{OR} \textcolor{brown}{"face verification"} \textcolor{blue}{OR} \textcolor{brown}{"face recognition"} \textcolor{blue}{OR} \textcolor{brown}{"anti-spoofing"} \textcolor{blue}{OR} \textcolor{brown}{"deepfake detection"} \textcolor{blue}{OR} \textcolor{brown}{"speech emotion"} \textcolor{blue}{OR} \textcolor{brown}{"eeg"} \textcolor{blue}{OR} \textcolor{brown}{"ecg"} \textcolor{blue}{OR} \textcolor{brown}{"text emotion"} ) \textcolor{blue}{AND} \textcolor{cyan}{PUBYEAR} > \textcolor{magenta}{2022} \textcolor{blue}{AND} \textcolor{cyan}{PUBYEAR} < \textcolor{magenta}{2027} \textcolor{blue}{AND} ( \textcolor{cyan}{SRCTYPE} (\textcolor{magenta}{j}) \textcolor{blue}{OR} \textcolor{cyan}{SRCTYPE} (\textcolor{magenta}{p}) )}\par}
\smallskip

\noindent{The same keywords and Boolean structure were used for the remaining databases, adapted to the query syntax and to the searchable fields of each platform; restrictions that cannot be expressed in a query string (e.g., publication years and content types in IEEE Xplore) were applied through the filters of the corresponding search interface. The resulting queries were the following:}

\smallskip
{\noindent\textit{IEEE Xplore} (Command Search; searched fields: document title, abstract, and author keywords; filters: publication years 2023--2026, content types ``Journals'' and ``Conferences''):\par}
{\footnotesize\ttfamily\spaceskip=0.525em plus 0.5em minus 0.05em\relax\noindent{( \textcolor{cyan}{"Document Title"}:\allowbreak\textcolor{brown}{"facial expression recognition"} \textcolor{blue}{OR} \textcolor{cyan}{"Abstract"}:\allowbreak\textcolor{brown}{"facial expression recognition"} \textcolor{blue}{OR} \textcolor{cyan}{"Author Keywords"}:\allowbreak\textcolor{brown}{"facial expression recognition"} \textcolor{blue}{OR} \textcolor{cyan}{"Document Title"}:\allowbreak\textcolor{brown}{"facial affect"} \textcolor{blue}{OR} \textcolor{cyan}{"Abstract"}:\allowbreak\textcolor{brown}{"facial affect"} \textcolor{blue}{OR} \textcolor{cyan}{"Author Keywords"}:\allowbreak\textcolor{brown}{"facial affect"} \textcolor{blue}{OR} \textcolor{cyan}{"Document Title"}:\allowbreak\textcolor{brown}{"facial emotion"} \textcolor{blue}{OR} \textcolor{cyan}{"Abstract"}:\allowbreak\textcolor{brown}{"facial emotion"} \textcolor{blue}{OR} \textcolor{cyan}{"Author Keywords"}:\allowbreak\textcolor{brown}{"facial emotion"} \textcolor{blue}{OR} ( ( \textcolor{cyan}{"Document Title"}:\allowbreak\textcolor{brown}{"action unit"} \textcolor{blue}{OR} \textcolor{cyan}{"Abstract"}:\allowbreak\textcolor{brown}{"action unit"} \textcolor{blue}{OR} \textcolor{cyan}{"Author Keywords"}:\allowbreak\textcolor{brown}{"action unit"} \textcolor{blue}{OR} \textcolor{cyan}{"Document Title"}:\allowbreak\textcolor{brown}{"micro-expression"} \textcolor{blue}{OR} \textcolor{cyan}{"Abstract"}:\allowbreak\textcolor{brown}{"micro-expression"} \textcolor{blue}{OR} \textcolor{cyan}{"Author Keywords"}:\allowbreak\textcolor{brown}{"micro-expression"} \textcolor{blue}{OR} \textcolor{cyan}{"Document Title"}:\allowbreak\textcolor{brown}{"valence"} \textcolor{blue}{OR} \textcolor{cyan}{"Abstract"}:\allowbreak\textcolor{brown}{"valence"} \textcolor{blue}{OR} \textcolor{cyan}{"Author Keywords"}:\allowbreak\textcolor{brown}{"valence"} \textcolor{blue}{OR} \textcolor{cyan}{"Document Title"}:\allowbreak\textcolor{brown}{"arousal"} \textcolor{blue}{OR} \textcolor{cyan}{"Abstract"}:\allowbreak\textcolor{brown}{"arousal"} \textcolor{blue}{OR} \textcolor{cyan}{"Author Keywords"}:\allowbreak\textcolor{brown}{"arousal"} \textcolor{blue}{OR} \textcolor{cyan}{"Document Title"}:\allowbreak\textcolor{brown}{"FACS"} \textcolor{blue}{OR} \textcolor{cyan}{"Abstract"}:\allowbreak\textcolor{brown}{"FACS"} \textcolor{blue}{OR} \textcolor{cyan}{"Author Keywords"}:\allowbreak\textcolor{brown}{"FACS"} ) \textcolor{blue}{AND} ( \textcolor{cyan}{"Document Title"}:\allowbreak\textcolor{brown}{"face"} \textcolor{blue}{OR} \textcolor{cyan}{"Abstract"}:\allowbreak\textcolor{brown}{"face"} \textcolor{blue}{OR} \textcolor{cyan}{"Author Keywords"}:\allowbreak\textcolor{brown}{"face"} \textcolor{blue}{OR} \textcolor{cyan}{"Document Title"}:\allowbreak\textcolor{brown}{"facial"} \textcolor{blue}{OR} \textcolor{cyan}{"Abstract"}:\allowbreak\textcolor{brown}{"facial"} \textcolor{blue}{OR} \textcolor{cyan}{"Author Keywords"}:\allowbreak\textcolor{brown}{"facial"} ) ) ) \textcolor{blue}{NOT} ( \textcolor{cyan}{"Document Title"}:\allowbreak\textcolor{brown}{"face identification"} \textcolor{blue}{OR} \textcolor{cyan}{"Abstract"}:\allowbreak\textcolor{brown}{"face identification"} \textcolor{blue}{OR} \textcolor{cyan}{"Author Keywords"}:\allowbreak\textcolor{brown}{"face identification"} \textcolor{blue}{OR} \textcolor{cyan}{"Document Title"}:\allowbreak\textcolor{brown}{"face verification"} \textcolor{blue}{OR} \textcolor{cyan}{"Abstract"}:\allowbreak\textcolor{brown}{"face verification"} \textcolor{blue}{OR} \textcolor{cyan}{"Author Keywords"}:\allowbreak\textcolor{brown}{"face verification"} \textcolor{blue}{OR} \textcolor{cyan}{"Document Title"}:\allowbreak\textcolor{brown}{"face recognition"} \textcolor{blue}{OR} \textcolor{cyan}{"Abstract"}:\allowbreak\textcolor{brown}{"face recognition"} \textcolor{blue}{OR} \textcolor{cyan}{"Author Keywords"}:\allowbreak\textcolor{brown}{"face recognition"} \textcolor{blue}{OR} \textcolor{cyan}{"Document Title"}:\allowbreak\textcolor{brown}{"anti-spoofing"} \textcolor{blue}{OR} \textcolor{cyan}{"Abstract"}:\allowbreak\textcolor{brown}{"anti-spoofing"} \textcolor{blue}{OR} \textcolor{cyan}{"Author Keywords"}:\allowbreak\textcolor{brown}{"anti-spoofing"} \textcolor{blue}{OR} \textcolor{cyan}{"Document Title"}:\allowbreak\textcolor{brown}{"deepfake detection"} \textcolor{blue}{OR} \textcolor{cyan}{"Abstract"}:\allowbreak\textcolor{brown}{"deepfake detection"} \textcolor{blue}{OR} \textcolor{cyan}{"Author Keywords"}:\allowbreak\textcolor{brown}{"deepfake detection"} \textcolor{blue}{OR} \textcolor{cyan}{"Document Title"}:\allowbreak\textcolor{brown}{"speech emotion"} \textcolor{blue}{OR} \textcolor{cyan}{"Abstract"}:\allowbreak\textcolor{brown}{"speech emotion"} \textcolor{blue}{OR} \textcolor{cyan}{"Author Keywords"}:\allowbreak\textcolor{brown}{"speech emotion"} \textcolor{blue}{OR} \textcolor{cyan}{"Document Title"}:\allowbreak\textcolor{brown}{"eeg"} \textcolor{blue}{OR} \textcolor{cyan}{"Abstract"}:\allowbreak\textcolor{brown}{"eeg"} \textcolor{blue}{OR} \textcolor{cyan}{"Author Keywords"}:\allowbreak\textcolor{brown}{"eeg"} \textcolor{blue}{OR} \textcolor{cyan}{"Document Title"}:\allowbreak\textcolor{brown}{"ecg"} \textcolor{blue}{OR} \textcolor{cyan}{"Abstract"}:\allowbreak\textcolor{brown}{"ecg"} \textcolor{blue}{OR} \textcolor{cyan}{"Author Keywords"}:\allowbreak\textcolor{brown}{"ecg"} \textcolor{blue}{OR} \textcolor{cyan}{"Document Title"}:\allowbreak\textcolor{brown}{"text emotion"} \textcolor{blue}{OR} \textcolor{cyan}{"Abstract"}:\allowbreak\textcolor{brown}{"text emotion"} \textcolor{blue}{OR} \textcolor{cyan}{"Author Keywords"}:\allowbreak\textcolor{brown}{"text emotion"} )}\par}

\smallskip
{\noindent\textit{Web of Science} (Core Collection, Advanced Search; the Topic field \texttt{TS} covers title, abstract, author keywords, and Keywords Plus):\par}
{\footnotesize\ttfamily\spaceskip=0.525em plus 0.5em minus 0.05em\relax\noindent{( \textcolor{cyan}{TS}=( \textcolor{brown}{"facial expression recognition"} \textcolor{blue}{OR} \textcolor{brown}{"facial affect"} \textcolor{blue}{OR} \textcolor{brown}{"facial emotion"} \textcolor{blue}{OR} ( (\textcolor{brown}{"action unit"} \textcolor{blue}{OR} \textcolor{brown}{"micro-expression"} \textcolor{blue}{OR} \textcolor{brown}{"valence"} \textcolor{blue}{OR} \textcolor{brown}{"arousal"} \textcolor{blue}{OR} \textcolor{brown}{"FACS"}) \textcolor{blue}{AND} (\textcolor{brown}{"face"} \textcolor{blue}{OR} \textcolor{brown}{"facial"}) ) ) \textcolor{blue}{NOT} \textcolor{cyan}{TS}=( \textcolor{brown}{"face identification"} \textcolor{blue}{OR} \textcolor{brown}{"face verification"} \textcolor{blue}{OR} \textcolor{brown}{"face recognition"} \textcolor{blue}{OR} \textcolor{brown}{"anti-spoofing"} \textcolor{blue}{OR} \textcolor{brown}{"deepfake detection"} \textcolor{blue}{OR} \textcolor{brown}{"speech emotion"} \textcolor{blue}{OR} \textcolor{brown}{"eeg"} \textcolor{blue}{OR} \textcolor{brown}{"ecg"} \textcolor{blue}{OR} \textcolor{brown}{"text emotion"} ) ) \textcolor{blue}{AND} \textcolor{cyan}{PY}=(\textcolor{magenta}{2023-2026}) \textcolor{blue}{AND} \textcolor{cyan}{DT}=(\textcolor{magenta}{Article} \textcolor{blue}{OR} \textcolor{magenta}{Proceedings Paper})}\par}

\smallskip
{\noindent\textit{arXiv} (API query syntax; searched fields: title and abstract; arXiv records carry no document-type field):\par}
{\footnotesize\ttfamily\spaceskip=0.525em plus 0.5em minus 0.05em\relax\noindent{( ( \textcolor{cyan}{ti}:\allowbreak\textcolor{brown}{"facial expression recognition"} \textcolor{blue}{OR} \textcolor{cyan}{abs}:\allowbreak\textcolor{brown}{"facial expression recognition"} \textcolor{blue}{OR} \textcolor{cyan}{ti}:\allowbreak\textcolor{brown}{"facial affect"} \textcolor{blue}{OR} \textcolor{cyan}{abs}:\allowbreak\textcolor{brown}{"facial affect"} \textcolor{blue}{OR} \textcolor{cyan}{ti}:\allowbreak\textcolor{brown}{"facial emotion"} \textcolor{blue}{OR} \textcolor{cyan}{abs}:\allowbreak\textcolor{brown}{"facial emotion"} \textcolor{blue}{OR} ( ( \textcolor{cyan}{ti}:\allowbreak\textcolor{brown}{"action unit"} \textcolor{blue}{OR} \textcolor{cyan}{abs}:\allowbreak\textcolor{brown}{"action unit"} \textcolor{blue}{OR} \textcolor{cyan}{ti}:\allowbreak\textcolor{brown}{"micro-expression"} \textcolor{blue}{OR} \textcolor{cyan}{abs}:\allowbreak\textcolor{brown}{"micro-expression"} \textcolor{blue}{OR} \textcolor{cyan}{ti}:\allowbreak\textcolor{brown}{"valence"} \textcolor{blue}{OR} \textcolor{cyan}{abs}:\allowbreak\textcolor{brown}{"valence"} \textcolor{blue}{OR} \textcolor{cyan}{ti}:\allowbreak\textcolor{brown}{"arousal"} \textcolor{blue}{OR} \textcolor{cyan}{abs}:\allowbreak\textcolor{brown}{"arousal"} \textcolor{blue}{OR} \textcolor{cyan}{ti}:\allowbreak\textcolor{brown}{"FACS"} \textcolor{blue}{OR} \textcolor{cyan}{abs}:\allowbreak\textcolor{brown}{"FACS"} ) \textcolor{blue}{AND} ( \textcolor{cyan}{ti}:\allowbreak\textcolor{brown}{"face"} \textcolor{blue}{OR} \textcolor{cyan}{abs}:\allowbreak\textcolor{brown}{"face"} \textcolor{blue}{OR} \textcolor{cyan}{ti}:\allowbreak\textcolor{brown}{"facial"} \textcolor{blue}{OR} \textcolor{cyan}{abs}:\allowbreak\textcolor{brown}{"facial"} ) ) ) \textcolor{blue}{ANDNOT} ( \textcolor{cyan}{ti}:\allowbreak\textcolor{brown}{"face identification"} \textcolor{blue}{OR} \textcolor{cyan}{abs}:\allowbreak\textcolor{brown}{"face identification"} \textcolor{blue}{OR} \textcolor{cyan}{ti}:\allowbreak\textcolor{brown}{"face verification"} \textcolor{blue}{OR} \textcolor{cyan}{abs}:\allowbreak\textcolor{brown}{"face verification"} \textcolor{blue}{OR} \textcolor{cyan}{ti}:\allowbreak\textcolor{brown}{"face recognition"} \textcolor{blue}{OR} \textcolor{cyan}{abs}:\allowbreak\textcolor{brown}{"face recognition"} \textcolor{blue}{OR} \textcolor{cyan}{ti}:\allowbreak\textcolor{brown}{"anti-spoofing"} \textcolor{blue}{OR} \textcolor{cyan}{abs}:\allowbreak\textcolor{brown}{"anti-spoofing"} \textcolor{blue}{OR} \textcolor{cyan}{ti}:\allowbreak\textcolor{brown}{"deepfake detection"} \textcolor{blue}{OR} \textcolor{cyan}{abs}:\allowbreak\textcolor{brown}{"deepfake detection"} \textcolor{blue}{OR} \textcolor{cyan}{ti}:\allowbreak\textcolor{brown}{"speech emotion"} \textcolor{blue}{OR} \textcolor{cyan}{abs}:\allowbreak\textcolor{brown}{"speech emotion"} \textcolor{blue}{OR} \textcolor{cyan}{ti}:\allowbreak\textcolor{brown}{"eeg"} \textcolor{blue}{OR} \textcolor{cyan}{abs}:\allowbreak\textcolor{brown}{"eeg"} \textcolor{blue}{OR} \textcolor{cyan}{ti}:\allowbreak\textcolor{brown}{"ecg"} \textcolor{blue}{OR} \textcolor{cyan}{abs}:\allowbreak\textcolor{brown}{"ecg"} \textcolor{blue}{OR} \textcolor{cyan}{ti}:\allowbreak\textcolor{brown}{"text emotion"} \textcolor{blue}{OR} \textcolor{cyan}{abs}:\allowbreak\textcolor{brown}{"text emotion"} ) ) \textcolor{blue}{AND} \textcolor{cyan}{submittedDate}:[\textcolor{magenta}{202301010000} \textcolor{blue}{TO} \textcolor{magenta}{202612312359}]}\par}

\smallskip
{\noindent\textit{DBLP} (title search; DBLP combines the words of a query with AND, matches word prefixes, and uses \texttt{|} as the OR operator; one query per inclusion term, with the results restricted to 2023--2026 through the year facet):\par}
{\footnotesize\ttfamily\spaceskip=0.525em plus 0.5em minus 0.05em\relax\noindent{\textcolor{brown}{facial expression recognition}; \ \textcolor{brown}{facial affect}; \ \textcolor{brown}{facial emotion}; \ \textcolor{brown}{action unit face}\textcolor{blue}{|}\textcolor{brown}{facial}; \ \textcolor{brown}{micro-expression face}\textcolor{blue}{|}\textcolor{brown}{facial}; \ \textcolor{brown}{valence face}\textcolor{blue}{|}\textcolor{brown}{facial}; \ \textcolor{brown}{arousal face}\textcolor{blue}{|}\textcolor{brown}{facial}; \ \textcolor{brown}{FACS face}\textcolor{blue}{|}\textcolor{brown}{facial}}\par}
{\noindent{Since DBLP supports neither abstract-level search nor an exclusion operator, the exclusion terms of the Scopus query were applied during the title/abstract screening (EC1).}\par}
\smallskip

\noindent{Google Scholar was used in a complementary manner, i.e., for backward and forward citation searching rather than as a primary source of records, since its result sets are not fully reproducible. The search was refined iteratively by adjusting the employed keywords to retrieve further relevant works. In addition, the reference list of each selected article was screened to identify additional relevant studies. Since the exclusion terms of the queries also discard records that merely mention one of these terms in their title, abstract, or keywords (e.g., multi-modal FER methods that combine facial and physiological signals, or FER methods that mention face recognition as a pre-training source), relevant works of this kind were identified through the backward and forward citation searching. The searches were conducted and iteratively refined throughout the course of the study, rather than being executed on a single date, and the dates of the individual searches were not recorded; as a consequence, the coverage of the most recently published literature, and in particular of works appearing within 2026, is inevitably partial.}

\subsection{Eligibility criteria}
\label{sec:eligibility}
The criteria that were applied for deciding whether a candidate work would be retained for in-depth analysis are reported in Table~\ref{tab:eligibility}. They comprise seven inclusion criteria (IC1--IC7), which a work had to satisfy cumulatively, and six exclusion criteria (EC1--EC6), each of which was sufficient for discarding a work. A further category (CR1) covers the works that are cited in this manuscript in order to support the analysis, without belonging to the analyzed corpus; these correspond to dataset and challenge papers, prior surveys, non-deep FER works, and background references on generic deep learning methods, pre-processing techniques, statistical and evaluation metrics, and software tools.

\begin{table*}[!t]
  \caption{Inclusion (IC) and exclusion (EC) criteria applied for the selection of the analyzed studies, along with the definition of the contextual references (CR1).}
  \label{tab:eligibility}
  \centering
  \scriptsize

  \setlength{\aboverulesep}{0pt}
  \setlength{\belowrulesep}{0pt}
  \setlength{\tabcolsep}{4pt}
  \renewcommand{\arraystretch}{1.3}
  \renewcommand{\tabularxcolumn}[1]{m{#1}}

  \rowcolors{2}{gray!20}{gray!2}

  \begin{tabularx}{\textwidth}{@{}|>{\centering\arraybackslash}m{1.1cm}|X|@{}}
  \toprule
  \rowcolor{gray!50}
  \headerbreak{Code} & \headerbreak{Criterion} \\
  \midrule
  IC1 & The work proposes or systematically evaluates a deep learning-based method. \\ \midrule
  IC2 & The work addresses at least one of the six recognition tasks that fall within the scope of this review (Section~\ref{sec:scope}), namely categorical macro-expression recognition, dimensional valence--arousal estimation, compound expression recognition, micro-expression recognition, action unit detection, or expression intensity estimation, or analyzes facial expression information (e.g., by means of expression, AU, or intensity recognition models or representations) for inferring a latent affective, cognitive, or clinical state, such as pain, depression, autism-related traits, engagement, or driver state (``other affective states'', check Section~\ref{sec:scope}). \\ \midrule
  IC3 & The method operates on facial visual input, optionally complemented by additional modalities (i.e., audio, text, or physiological signals). \\ \midrule
  IC4 & The work introduces a substantial methodological or experimental contribution, i.e., a new model or architecture, a new learning strategy, a new pre-processing pipeline, a new dataset or benchmark, or a new evaluation protocol or comparative analysis. \\ \midrule
  IC5 & The work is published in a peer-reviewed venue or is available as an archival preprint. \\ \midrule
  IC6 & The work is written in English and its full text is accessible. \\ \midrule
  IC7 & The work was published within the targeted period 2023--2026; works published earlier are retained only if they were identified, through backward citation searching, as seminal or milestone contributions of the field. \\ \midrule
  EC1 & The work targets a task that lies outside the scope of this review (e.g., face identification or verification, face anti-spoofing, deepfake detection, speech-only or physiological-signal-only emotion recognition). \\ \midrule
  EC2 & The method does not operate on facial visual input. \\ \midrule
  EC3 & The method is not deep learning-based (non-deep FER works, including pre-deep-learning seminal works, are cited as contextual references, see CR1). \\ \midrule
  EC4 & The work constitutes an incremental re-parameterization of an already covered approach, without a substantial additional contribution. \\ \midrule
  EC5 & The work is a preliminary version that is superseded by a more complete article of the same authors. \\ \midrule
  EC6 & The record is a duplicate, is not written in English, or its full text is not accessible. \\ \midrule
  CR1 & \textit{Contextual references}: Works that do not satisfy IC1--IC4, or whose main contribution is a dataset, a challenge, or a survey, but are cited in order to support the analysis (i.e., dataset and challenge papers, prior surveys, non-deep FER works, and background references on generic deep learning methods, pre-processing techniques, statistical and evaluation metrics, and software tools); they are not part of the analyzed corpus and are not counted in any of the statistics reported for it (Figs.~\ref{fig:bibliometrics} and~\ref{fig:crosscriteria}, Section~\ref{sec:cross_criteria}). \\
  \bottomrule
  \end{tabularx}
\end{table*}

\subsection{Study selection and analysis procedure}
\label{sec:selection}
The retained works were not selected by means of an automated filtering procedure; instead, each candidate article was manually studied and assessed, and this in-depth study constitutes the basis of the analysis that is reported in the remainder of this manuscript. In particular, the records retrieved from the different databases were first merged and duplicate entries were removed manually, by matching the Digital Object Identifier (DOI) of the records and, whenever a DOI was not available, their title, author list, and publication year; in the case that an archival preprint and the corresponding peer-reviewed version were both retrieved, only the latter was retained. Articles not written in English and works without full-text access were also removed at this stage (EC6). Each remaining candidate was subsequently assessed on the basis of its title and abstract against the scope of the review (IC1--IC3), and works addressing adjacent, but out-of-scope, problems were discarded at this stage (EC1--EC3). Each surviving article was then read in full and was retained only if it introduced a substantial methodological or experimental contribution (IC4), as opposed to an incremental re-parameterization of an already covered approach (EC4), and was published in a peer-reviewed venue or as an archival preprint (IC5); whenever multiple articles described successive versions of the same method, only the most complete one was kept (EC5). For every retained study, the proposed approach was characterized along the seven criteria of Section~\ref{sec:taxonomy_criteria}, together with the employed datasets, the followed evaluation protocol, and the reported results; it is this per-study characterization that supports the research phases of Section~\ref{sec:Evolution}, the cross-criteria analysis of Section~\ref{sec:cross_criteria}, the per-criterion comparative analyses of Sections~\ref{sec:recognition_tasks}--\ref{sec:app_domains}, the dataset review of Section~\ref{sec:Datasets}, and the quantitative analysis of Section~\ref{sec:performance_metrics}. Since the number of eligible recent works substantially exceeds the number of studies that can be analyzed at this level of detail, a prioritization step was finally applied among the eligible works, favoring studies whose citation count was high relative to that of the remaining eligible works of the same publication year, so that recent works would not be penalized, as well as studies published in established computer vision, pattern recognition, multimedia, and affective computing venues, while explicitly preserving coverage across all recognition tasks, input modalities, pre-processing pipelines, neural network architectures, learning strategies, acquisition settings, and application domains, so that the analyzed corpus reflects the breadth of the field rather than a single research direction. The screening, the full-text assessment, and the coding of the studies were performed by the authors contributing to the Investigation and Methodology roles of the CRediT statement, while the resulting assessments were iteratively cross-checked and discussed within the author team until consensus was reached. The overall procedure, along with the criteria that were applied at each stage, is summarized in Fig.~\ref{fig:corpus}. This process resulted in 167 primary deep learning-based FER studies, which constitute the analyzed corpus; in addition, 159 works are cited as contextual references (i.e., 39 dataset and challenge papers, 20 prior surveys, 16 non-deep FER works, and 84 background references on generic deep learning methods, pre-processing techniques, statistical and evaluation metrics, and software tools), leading to a total of 326 references. The contextual references are cited only in order to support the analysis; they were therefore not subject to the selection procedure described above, and none of them is counted in the statistics that are reported for the analyzed corpus (Figs.~\ref{fig:bibliometrics} and~\ref{fig:crosscriteria}, Section~\ref{sec:cross_criteria}).

\begin{figure*}[!t]
    \centering
    \adjustbox{max width=\textwidth, max totalheight=0.75\textheight, center}{%
    \begin{tikzpicture}[font=\rmfamily]
\tikzset{
  corpusstage/.style={draw=darkblue, line width=0.9pt, fill=lightblue!12, rounded corners=3pt, align=left, inner sep=6pt, font=\small},
  corpusexcl/.style={draw=orange, line width=0.9pt, fill=orange!10, rounded corners=3pt, dash pattern=on 3pt off 2pt, text width=7.40cm, align=left, inner sep=6pt, font=\small},
  corpusout/.style={draw=darkblue, line width=1.4pt, fill=lightblue!30, rounded corners=3pt, text width=8.60cm, align=left, inner sep=7pt, font=\small},
  corpusctx/.style={draw=darkblue!70, line width=1.0pt, fill=lightgrey!25, rounded corners=3pt, dash pattern=on 3pt off 2pt, text width=7.40cm, align=left, inner sep=7pt, font=\small},
  corpustot/.style={draw=darkblue, line width=1.2pt, fill=bgbox, rounded corners=3pt, minimum width=21.0cm, align=center, inner sep=7pt, font=\small},
  corpusnote/.style={text width=21.0cm, align=left, font=\footnotesize, text=black!75},
  corpusarrow/.style={-{Latex[length=2.6mm,width=2.0mm]}, draw=darkblue, line width=0.9pt},
  corpusarrowx/.style={-{Latex[length=2.6mm,width=2.0mm]}, draw=orange, line width=0.9pt},
}

\node[corpusstage, text width=8.60cm] (s1) at (0.00,0.00) {\textbf{Identification (database search)}\\[1pt] IEEE Xplore, Scopus, Web of Science, DBLP, and arXiv; targeted publication period 2023--2026; iteratively refined keyword and Boolean queries (Section~\ref{sec:search_strategy})};
\node[corpusstage, text width=7.40cm] (s2) at (11.60,0.00) {\textbf{Identification (complementary sources)}\\[1pt] Backward and forward citation searching (Google Scholar); screening of the reference list of every retained article; seminal and milestone works published before 2023};
\node[corpusstage, text width=8.60cm] (s3) at (0.00,-4.30) {\textbf{De-duplication and formal eligibility checks}\\[1pt] Duplicate removal by DOI and by title/author/year matching; preprints superseded by their peer-reviewed versions (EC6)};
\node[corpusstage, text width=8.60cm] (s4) at (0.00,-6.50) {\textbf{Title/abstract screening}\\[1pt] Assessment against the scope of the review (IC1--IC3, EC1--EC3)};
\node[corpusstage, text width=8.60cm] (s5) at (0.00,-8.90) {\textbf{Full-text assessment}\\[1pt] Assessment of the methodological or experimental contribution and of the publication type (IC4--IC5, EC4--EC5)};
\node[corpusstage, text width=8.60cm] (s6) at (0.00,-11.40) {\textbf{Prioritization among eligible studies}\\[1pt] Citation impact relative to the eligible works of the same publication year; publication in established venues; preserved coverage across all seven criteria (Section~\ref{sec:taxonomy_criteria})};
\node[corpusstage, text width=8.60cm] (s7) at (0.00,-14.10) {\textbf{Per-study characterization}\\[1pt] Recognition task, input modality, pre-processing pipeline, architecture family, learning strategy, acquisition setting, and application domain; employed datasets, evaluation protocols, and reported results};

\node[corpusexcl] (e1) at (11.60,-4.30) {\textbf{Removed}\\[1pt] Duplicate records; articles not written in English; articles without full-text access};
\node[corpusexcl] (e2) at (11.60,-6.50) {\textbf{Excluded}\\[1pt] Out-of-scope task (face identification/verification, face anti-spoofing, deepfake detection, speech- or physiological-signal-only emotion recognition); no facial visual input; not deep learning-based};
\node[corpusexcl] (e3) at (11.60,-8.90) {\textbf{Excluded}\\[1pt] No substantial methodological or experimental contribution; preliminary version superseded by a more complete article of the same authors};
\node[corpusexcl] (e4) at (11.60,-11.40) {\textbf{Eligible but not retained}\\[1pt] Number of eligible recent works exceeding what can be analyzed at the level of detail adopted in this survey};

\node[corpusout] (out) at (0.00,-17.20) {\textbf{Analyzed corpus (167 primary deep learning-based FER studies)}\\[1pt] 93 published in 2023--2026; 74 milestone studies published up to 2022; each characterized along the seven criteria (Sections~\ref{sec:Evolution}--\ref{sec:performance_metrics})};
\node[corpusctx] (ctx) at (11.60,-17.20) {\textbf{Contextual references (159 cited works)}\\[1pt] 39 dataset and challenge papers; 20 prior surveys; 16 non-deep FER works; 84 background references on generic deep learning methods, pre-processing techniques, statistical and evaluation metrics, and software tools; cited for context only, not part of the analyzed corpus};
\node[corpustot] (tot) at (5.80,-20.50) {\textbf{All cited references (326 = 167 + 159)}};

\draw[corpusarrow] (s1.south) -- (s3.north);
\draw[corpusarrow] (s2.south) -- ++(0,-1.20) -| (s3.north);
\draw[corpusarrow] (s3.south) -- (s4.north);
\draw[corpusarrow] (s4.south) -- (s5.north);
\draw[corpusarrow] (s5.south) -- (s6.north);
\draw[corpusarrow] (s6.south) -- (s7.north);
\draw[corpusarrow] (s7.south) -- (out.north);

\draw[corpusarrowx] (s3.east) -- (e1.west);
\draw[corpusarrowx] (s4.east) -- (e2.west);
\draw[corpusarrowx] (s5.east) -- (e3.west);
\draw[corpusarrowx] (s6.east) -- (e4.west);

\draw[corpusarrow] (out.south) -- ++(0,-0.70) -| (tot.north);
\draw[corpusarrow] (ctx.south) -- ++(0,-0.70) -| (tot.north);

\node[corpusstage, text width=0.75cm, minimum height=0.42cm, inner sep=0pt] at (-4.55,-21.95) {};
\node[anchor=west, font=\small, text width=5.8cm, align=left] at (-3.90,-21.95) {Selection or analysis stage};
\node[corpusexcl, text width=0.75cm, minimum height=0.42cm, inner sep=0pt] at (2.65,-21.95) {};
\node[anchor=west, font=\small, text width=5.8cm, align=left] at (3.30,-21.95) {Records removed, excluded, or not retained};
\node[corpusout, text width=0.75cm, minimum height=0.42cm, inner sep=0pt] at (9.85,-21.95) {};
\node[anchor=west, font=\small, text width=5.8cm, align=left] at (10.50,-21.95) {Analyzed corpus (primary studies)};
\node[corpusctx, text width=0.75cm, minimum height=0.42cm, inner sep=0pt] at (-4.55,-22.80) {};
\node[anchor=west, font=\small, text width=5.8cm, align=left] at (-3.90,-22.80) {Contextual references (outside the corpus)};
\draw[corpusarrow] (2.20,-22.80) -- ++(0.9,0);
\node[anchor=west, font=\small, text width=5.8cm, align=left] at (3.30,-22.80) {Flow of retained records};
\draw[corpusarrowx] (9.40,-22.80) -- ++(0.9,0);
\node[anchor=west, font=\small, text width=5.8cm, align=left] at (10.50,-22.80) {Flow of removed or excluded records};
    \end{tikzpicture}%
    }
    \vspace{0.2cm}
  \caption{Overview of the corpus construction and study selection procedure that was followed. Each stage box reports the decision rules and the eligibility criteria (Table~\ref{tab:eligibility}) that were applied at the corresponding stage, while the bottom row quantifies the resulting analyzed corpus and the contextual references. The numbers of records retrieved and excluded at the individual stages were not logged at search time and are therefore not reported (Sections~\ref{sec:data_extraction} and~\ref{sec:review_limitations}); all reported quantities refer to works cited in this manuscript, rather than to intermediate records of the database searches.}
  \label{fig:corpus}
\end{figure*}

Key bibliometric statistics of the review are shown in Fig.~\ref{fig:bibliometrics}, while the detailed analysis of the identified FER works is provided in Sections~\ref{sec:Evolution}--\ref{sec:performance_metrics}.

\subsection{Bibliometric analysis}
\label{sec:data_extraction}
Fig.~\ref{fig:bibliometrics} reports statistics computed over the 167 primary deep learning-based FER studies that constitute the analyzed corpus, whereas the 159 contextual references (i.e., dataset and challenge papers, prior surveys, non-deep FER works, and background references on generic deep learning methods, pre-processing techniques, statistical and evaluation metrics, and software tools) are excluded from all of its panels. Regarding the article types and venues (Fig.~\ref{fig:bibliometrics}(a)--(b)), 108 primary studies (64.7\%) are journal articles, 53 (31.7\%) are conference papers, and 6 (3.6\%) are archival preprints, while IEEE Transactions on Affective Computing is the most frequent venue (32 studies), followed by CVPR (12), CVPR Workshops (10), Pattern Recognition (9), and Neurocomputing (7). Regarding publication trends (Fig.~\ref{fig:bibliometrics}(c)), 93 primary studies (55.7\%) were published in 2023--2026, while the largest annual counts are observed in 2022 (35 studies) and 2024 (32 studies). Regarding the distribution across recognition tasks (Fig.~\ref{fig:bibliometrics}(e)), categorical expression recognition is the primary task of the majority of the studies (104, 62.3\%), followed by micro-expression recognition (22), other affective states (12), AU detection (10), and valence--arousal estimation (10), whereas expression intensity estimation (5) and compound expression recognition (4) are considerably less represented; notably, research on compound expressions is mainly reflected in dataset and challenge papers (Section~\ref{sec:Datasets}). Regarding the architectural evolution (Fig.~\ref{fig:bibliometrics}(d)), CNN-based designs account for 73.0\% of the primary studies published up to 2022, but only for 31.2--36.0\% of those published from 2023 onwards. Transformer-based and hybrid (e.g., CNN--Transformer) designs together reach their highest share in 2023 (60.0\%), while VLM/MLLM-based approaches appear from 2023 onwards and account for 22.2\% of the primary studies published in 2025--2026. It should be noted that these statistics describe the analyzed corpus, which was selected with priority given to recent and highly cited works; hence, they reflect research activity trends and do not indicate the relative performance of the corresponding approaches, which is discussed in Section~\ref{sec:performance_metrics}.

It should be noted that the searches were conducted and refined iteratively over the course of the study, without recording the dates of the individual searches, and that the number of records that were retrieved and excluded at each individual screening stage was not logged at the time. These intermediate counts are not reported here, since they could only be reconstructed retrospectively and would, in any case, strongly depend on the exact query formulation and on the indexing state of each database at the moment of retrieval; for a field evolving as rapidly as deep learning-based FER, they would additionally be outdated already at the time of publication. Instead, the criteria and the decision rules that were applied at each stage are reported in full in Sections~\ref{sec:eligibility} and~\ref{sec:selection}, while all quantities that are reported for the analyzed corpus (Fig.~\ref{fig:corpus}) refer to works that are cited in this manuscript, rather than to intermediate records of the database searches. In this sense, the rigor of the adopted methodology lies in the explicit and uniform criteria that governed the manual selection of the studies and in the uniform per-study characterization that followed, which constitute the basis of the entire analysis presented in the subsequent sections.

\subsection{Limitations of this review}
\label{sec:review_limitations}
This subsection provides an explicit self-assessment of the present review. Its limitations concern both the construction of the analyzed corpus and the analysis performed on it; for each of them, the adopted mitigation and the resulting implication for the interpretation of the findings are stated.
\begin{itemize}[leftmargin=*, itemsep=2pt, topsep=2pt, parsep=0pt]
    \item \uline{Reliance on reported results}: The quantitative analysis of Section~\ref{sec:performance_metrics} relies on the results and on the ablations reported in the original publications, rather than on a controlled re-evaluation of the methods under a common protocol (i.e., identical data splits, pre-processing, pre-training data, and training budget). The reported values were checked against the original publications, and the metadata of Tables~\ref{tab:perf_sfer}--\ref{tab:perf_intensity} (e.g., pre-training source, input resolution) is given only to the extent that it is reported there. Consequently, the observations of Section~\ref{ssec:perf_discussion} describe associations in the reported results, rather than causal effects.
    \item \uline{Heterogeneous evaluation protocols}: The reviewed studies differ in class sets (e.g., 7- vs.\ 8-class AffectNet), data splits and validation schemes (e.g., leave-one-subject-out (LOSO) vs.\ composite MER protocols), metrics, and pre-training data, while almost all tabulated values are obtained within-dataset (Section~\ref{ssec:perf_quantitative}). This is mitigated by reporting the protocol of every entry, by computing the bold entries only among results that share the standard protocol of each benchmark, and by collecting the available cross-dataset results separately (Table~\ref{tab:perf_cross}). Nevertheless, a strict quantitative comparison, a ranking of methods, or a pooled statistical synthesis (meta-analysis) of the reported values is not possible, and none is attempted.
    \item \uline{Partially overlapping taxonomy criteria}: The seven criteria of the taxonomy are complementary but not orthogonal, since some design choices are strongly associated in practice (e.g., text-grounded inputs and VLM/MLLM backbones); this is mitigated by assigning each study to a single primary category per criterion and by restricting the description of a method in each section to the aspect that corresponds to that criterion (Section~\ref{sec:taxonomy_criteria}); in addition, the cross-criteria analysis of Section~\ref{sec:cross_criteria} covers the four method-level criteria only.
    \item \uline{Curated and time-bounded corpus}: The analyzed corpus is criteria-driven and prioritized, rather than exhaustive; among the works satisfying the inclusion criteria, those exhibiting higher citation impact and those published in established venues were favored, subject to preserving coverage across all seven criteria. Consequently, the statistics reported in Fig.~\ref{fig:bibliometrics} and Fig.~\ref{fig:crosscriteria} describe the analyzed corpus and should be interpreted as indicative of research activity trends, rather than as an exhaustive census of the field. Moreover, only works written in English and with accessible full text were considered (IC6), which may introduce a language and access bias, and Google Scholar was used only in a complementary manner, since its result sets are not fully reproducible. Finally, since the searches were conducted and refined iteratively over the course of the study, rather than being executed on a single date, and their dates were not recorded (Section~\ref{sec:search_strategy}), the coverage of the most recently published literature, and in particular of works appearing within 2026, is inevitably partial.
    \item \uline{Documentation of the selection process}: As discussed in Section~\ref{sec:data_extraction}, the number of records that were retrieved, de-duplicated, and excluded at each individual screening stage was not logged and is therefore not reported. The screening and the coding of the studies were performed within the author team, without a formal independent double-screening protocol and the computation of an inter-rater agreement statistic. In addition, no formal quality or risk-of-bias appraisal of the individual primary studies (e.g., with respect to reproducibility or code availability) was performed. This is mitigated by reporting the eligibility criteria and the decision rules of every stage explicitly (Table~\ref{tab:eligibility}, Section~\ref{sec:selection}, and Fig.~\ref{fig:corpus}) and by cross-checking all assessments within the author team until consensus was reached; consequently, the analyzed corpus is reproducible at the level of its selection criteria, but not at the level of the individual screening decisions, and the findings of the primary studies are taken as reported.
\end{itemize}

\section{FER research evolution}
\label{sec:Evolution}

Although facial expression analysis has a longer research history than many contemporary deep learning topics, the way in which facial expressions are represented and recognized from RGB imagery has undergone a profound transformation over the last two decades, gradually shifting from explicit anatomical coding schemes and shallow handcrafted descriptors towards integrated, data-driven, and increasingly general-purpose models that operate under unconstrained, multi-modal, and language-grounded settings. Before the detailed analysis of the literature, this section provides a concise historical perspective on this evolution, identifying the key paradigm shifts that have shaped the current landscape and the milestone works associated with each of them.

\begin{figure*}[!t]
    \centering
    \begin{tikzpicture}[
        font=\small,
        >=Stealth,
        header/.style={
            rectangle, rounded corners=5pt, line width=0.8pt,
            text width=2.4cm, align=center, minimum height=2.4cm, inner sep=4pt,
            blur shadow={shadow blur steps=5, shadow xshift=1.5pt, shadow yshift=-1.5pt, shadow opacity=20}
        },
        mbox/.style={
            rectangle, rounded corners=5pt, line width=0.8pt,
            text width=2.4cm, align=left, font=\scriptsize, inner sep=4pt, anchor=north,
            blur shadow={shadow blur steps=5, shadow xshift=1.5pt, shadow yshift=-1.5pt, shadow opacity=15}
        },
        arrowline/.style={
            ->, line width=1.8pt, draw=gray!50
        },
        connector/.style={
            dotted, thick, shorten >=2pt, shorten <=2pt
        }
    ]

        \node[header, top color=blue!15,   bottom color=blue!5,   draw=blue!40]   (h1) at (0, 0)    {\textbf{Phase 1}\\[2pt]Handcrafted features \&\\classical machine learning\\[3pt]\textit{pre-2013}};
        \node[header, top color=cyan!15,   bottom color=cyan!5,   draw=cyan!40]   (h2) at (3.4, 0)  {\textbf{Phase 2}\\[2pt]Deep CNNs \& in-the-wild static FER\\[3pt]\textit{2013--2017}};
        \node[header, top color=teal!15,   bottom color=teal!5,   draw=teal!40]   (h3) at (6.8, 0)  {\textbf{Phase 3}\\[2pt]Spatio-temporal\\modeling \& AU/micro-expression deep nets\\[3pt]\textit{2017--2020}};
        \node[header, top color=orange!15, bottom color=orange!5, draw=orange!40] (h4) at (10.2, 0) {\textbf{Phase 4}\\[2pt]Attention, Transformers \& uncertainty-aware FER\\[3pt]\textit{2020--2023}};
        \node[header, top color=red!15,    bottom color=red!5,    draw=red!40]    (h5) at (13.6, 0) {\textbf{Phase 5}\\[2pt]Vision--language, MLLMs \& foundation models\\[3pt]\textit{2023--present}};

        \draw[arrowline] ([xshift=5pt]h1.east) -- ([xshift=-5pt]h2.west);
        \draw[arrowline] ([xshift=5pt]h2.east) -- ([xshift=-5pt]h3.west);
        \draw[arrowline] ([xshift=5pt]h3.east) -- ([xshift=-5pt]h4.west);
        \draw[arrowline] ([xshift=5pt]h4.east) -- ([xshift=-5pt]h5.west);

        \node[mbox, top color=white, bottom color=blue!4, draw=blue!30] (m1) at (0, -1.8) {
            \textbf{1978:} Facial Action Coding System (FACS)~\cite{friesen1978facial,ekman1997what}\\[4pt]
            \textbf{1997--2010:} JAFFE, CK+, BU-3DFE controlled datasets~\cite{lyons1998japanese,lyons1997jaffe,lucey2010extended,yin20063d}\\[4pt]
            \textbf{2009:} LBP descriptors for FER with SVM/AdaBoost classifiers~\cite{shan2009facial}\\[4pt]
            \textbf{2013--2014:} Spontaneous micro-expression datasets (SMIC, CASME~II)~\cite{li2013spontaneous,yan2014casme}
        };

        \node[mbox, top color=white, bottom color=cyan!4, draw=cyan!30] (m2) at (3.4, -1.8) {
            \textbf{2013:} FER-2013 challenge ignites end-to-end deep FER~\cite{goodfellow2013challenges}\\[4pt]
            \textbf{2014:} Compound facial expressions of emotion formalized~\cite{du2014compound}\\[4pt]
            \textbf{2016:} FER+ crowdsourced label distribution~\cite{barsoum2016training}\\[4pt]
            \textbf{2017:} RAF-DB, AffectNet large-scale in-the-wild benchmarks~\cite{li2017reliable,mollahosseini2017affectnet}
        };

        \node[mbox, top color=white, bottom color=teal!4, draw=teal!30] (m3) at (6.8, -1.8) {
            \textbf{2016--2018:} Region-based deep AU detectors (DRML, EAC-Net)~\cite{zhao2016deep,li2017eacnet}\\[4pt]
            \textbf{2018:} Aff-Wild2 in-the-wild VA benchmark, later extended with expression and AU labels~\cite{kollias2018aff,kollias2022abaw}\\[4pt]
            \textbf{2019:} Blended/compound in-the-wild FER~\cite{li2019blended}\\[4pt]
            \textbf{2017--2020:} 3D-CNN/CNN--LSTM spatio-temporal FER and AU dynamics~\cite{li2020deep}
        };

        \node[mbox, top color=white, bottom color=orange!4, draw=orange!30] (m4) at (10.2, -1.8) {
            \textbf{2020:} Self-Cure Network for label noise (SCN)~\cite{wang2020suppressing}\\[4pt]
            \textbf{2020:} DFEW large-scale in-the-wild dynamic FER~\cite{jiang2020dfew}\\[4pt]
            \textbf{2021--2023:} TransFER, VTFF and POSTER Transformer-based FER~\cite{xue2021transfer,ma2021facial,zheng2023poster}\\[4pt]
            \textbf{2022:} FERV39k multi-scene and 4DME 4D micro-expression benchmarks~\cite{li20224dme,wang2022ferv39k}
        };

        \node[mbox, top color=white, bottom color=red!4, draw=red!30] (m5) at (13.6, -1.8) {
            \textbf{2023--2025:} CLIP-based FER and prompt learning (DFER-CLIP, MPA-FER)~\cite{zhao2023prompting,ma2025multimodal}\\[4pt]
            \textbf{2024:} LibreFace open toolkit \& HTNet micro-expression network~\cite{chang2024libreface,wang2024htnet}\\[4pt]
            \textbf{2024--2026:} Emotion-LLaMA, EMO-LLaMA, FaceLLM, EmoVerse~\cite{cheng2024emotionllama,xing2024emollama,shahreza2025facellm,li2025emoverse}\\[4pt]
            \textbf{2025:} POSTER++, Mamba-VA, FaceXFormer, MOL~\cite{mao2025posterpp,liang2025mamba,narayan2025facexformer,shao2025mol}
        };

        \draw[connector, draw=blue!50]   (h1.south) -- (m1.north);
        \draw[connector, draw=cyan!50]   (h2.south) -- (m2.north);
        \draw[connector, draw=teal!50]   (h3.south) -- (m3.north);
        \draw[connector, draw=orange!50] (h4.south) -- (m4.north);
        \draw[connector, draw=red!50]    (h5.south) -- (m5.north);

    \end{tikzpicture}
    \vspace{0.2cm}
    \caption{Main phases in deep learning-based facial expression recognition research and key/milestone works. Research has progressively shifted from anatomical coding and handcrafted descriptors towards large-scale, pre-trained, multi-modal, and language-grounded models.}
    \label{fig:FERtimeline}
\end{figure*}

\subsection{Research phases}
\label{ssec:FERphases}

The evolution of FER research can be roughly divided into five successive phases, each corresponding to a critical paradigm shift regarding how facial expressions are represented, how models are supervised, and how much data and computation are required. Overall, research has shifted from coding-based, lab-constrained, and task-specific designs towards large-scale, pre-trained, multi-modal, and language-grounded systems that transfer across tasks, datasets, and demographic groups. Two cross-cutting drivers underpin this trajectory, namely the steady growth in the volume and diversity of openly available facial imagery (from a few hundred posed lab samples to multi-million in-the-wild video corpora), and the progressive adoption of architectural and learning advances from the broader computer vision and natural language processing communities. The considered research phases are graphically illustrated in Fig.~\ref{fig:FERtimeline}, along with the milestone works associated with each of them, and are briefly summarized below, highlighting the main advancements and the research shift introduced at each stage.

\subsubsection{Phase 1: Handcrafted features and classical machine learning (pre-2013)}
\label{sssec:FERphase1}
Early FER relied on explicit anatomical coding, most notably the Facial Action Coding System (FACS)~\citep{friesen1978facial,ekman1997what}, and on handcrafted appearance and geometric descriptors (LBP, Gabor, scale-invariant feature transform (SIFT), landmark distances) fed to shallow classifiers such as support vector machines (SVMs) and AdaBoost~\citep{shan2009facial,sariyanidi2014automatic}. Posed, lab-controlled datasets, including JAFFE~\citep{lyons1998japanese}, which was originally created for research on facial-expression perception, CK+~\citep{lucey2010extended}, and BU-3DFE~\citep{yin20063d}, were adopted as benchmarks and enabled the first systematic comparisons across methods, while the introduction of spontaneous micro-expression corpora (SMIC, CASME~II) at the end of this period~\citep{li2013spontaneous,yan2014casme} opened the study of subtle, involuntary muscle activations. Overall, this phase was dominated by expert-designed features, which transferred poorly across subjects, illumination, and head pose, and performed modestly on unconstrained data.

\subsubsection{Phase 2: Deep CNNs and in-the-wild static FER (2013--2017)}
\label{sssec:FERphase2}
Deep CNN backbones (AlexNet, VGG, ResNet), pre-trained on ImageNet or face-recognition corpora and fine-tuned on expression data~\citep{lecun2002gradient,li2020deep,savchenko2022classifying}, rapidly displaced handcrafted pipelines. The FER-2013 challenge~\citep{goodfellow2013challenges} catalyzed end-to-end deep FER on weakly labeled web imagery, while large in-the-wild datasets such as FER+~\citep{barsoum2016training}, RAF-DB~\citep{li2017reliable}, and AffectNet~\citep{mollahosseini2017affectnet}, together with the compound-emotion formulation~\citep{du2014compound}, introduced naturalistic variation in pose, illumination, occlusion, demographics, and label noise. The central research question shifted from ``how do we describe a face?'' to ``how do we cope with label noise, demographic variance, and annotation ambiguity at scale?'', and in-the-wild evaluation became a defining characteristic of modern FER.

\subsubsection{Phase 3: Spatio-temporal modeling and AU/micro-expression deep nets (2017--2020)}
\label{sssec:FERphase3}
Attention turned from frame-level recognition to the temporal and structural characteristics of facial behavior. 3D CNNs, CNN--LSTM hybrids, and two-stream optical-flow architectures captured the neutral--onset--apex--offset evolution of expressions and AU intensity dynamics~\citep{li2020deep,jiang2020dfew,hochreiter1997long}, while region-based deep AU detectors such as DRML~\citep{zhao2016deep} and EAC-Net~\citep{li2017eacnet} exploited the local nature of muscular activations on BP4D~\citep{zhang2014bp4d} and DISFA~\citep{mavadati2013disfa}. Aff-Wild2~\citep{kollias2018aff} extended in-the-wild valence--arousal annotation to a large audio-visual corpus, later complemented with basic-expression and AU labels for the Affective Behavior Analysis in-the-wild (ABAW) challenges~\citep{kollias2022abaw}, blended in-the-wild recognition~\citep{li2019blended} pushed beyond the six-emotion paradigm, and micro-expression recognition matured around motion-magnified and apex-frame inputs~\citep{davison2016samm}. The period is best understood as one of specialization, consolidating task-specific, structure-aware designs that encoded facial-behavior priors into the model topology.

\begin{table*}[!t]
\centering
\caption{Key models, benchmarks, and methodological milestones in the evolution of facial expression recognition research, including its pre-deep-learning foundations. Year: Publication year of the cited reference. Phase: The phase of Fig.~\ref{fig:FERtimeline} whose paradigm a work introduced or exemplifies; since the phase boundaries are approximate and overlapping, a work may fall outside the nominal period of its phase (e.g., CASME~II, whose baseline relies on handcrafted spatio-temporal descriptors, and DRML, a 2016 deep AU detector assigned to Phase~3), while works published from 2023 onwards are assigned to Phase~5, except POSTER, which concludes the CNN--Transformer developments of Phase~4. Task focus follows the recognition tasks of Section~\ref{sec:taxonomy_criteria}; Architecture refers to the proposed model and is omitted (--) for datasets and coding systems.}
\label{tab:FERmilestones}
\renewcommand{\arraystretch}{1.3}
\setlength{\aboverulesep}{0pt}
\setlength{\belowrulesep}{0pt}
\footnotesize
\begin{adjustbox}{width=\textwidth}
\begin{tabular}{| m{2.7cm} | c | c | >{\centering\arraybackslash}m{2.5cm} | >{\centering\arraybackslash}m{2.0cm} | m{8.0cm} |}
\hline
\rowcolor{gray!50} \centering \textbf{Model / Work} & \textbf{Year} & \textbf{Phase} & \textbf{Task focus} & \textbf{Architecture} & \centering\arraybackslash \textbf{Key contribution} \\
\hline
\rowcolor{gray!2} FACS~\citep{friesen1978facial,ekman1997what} & 1978 & 1 & AU coding & -- & Anatomical coding of facial muscle activity into action units; foundation of subsequent FER and AU detection. \\
\hline
\rowcolor{gray!2} JAFFE~\citep{lyons1998japanese,lyons1997jaffe} & 1997 & 1 & Categorical (image) & -- & Posed expression dataset created for research on facial-expression perception and representation, later adopted as one of the earliest widely used lab-controlled FER benchmarks. \\
\hline
\rowcolor{gray!2} LBP-FER~\citep{shan2009facial} & 2009 & 1 & Categorical (image) & Handcrafted & Comprehensive study of local binary pattern descriptors and SVM/AdaBoost classifiers for FER. \\
\hline
\rowcolor{gray!2} CK+~\citep{lucey2010extended} & 2010 & 1 & Categorical \& AU detection & -- & Extended Cohn--Kanade dataset providing AU and expression labels under controlled conditions. \\
\hline
\rowcolor{gray!2} CASME~II~\citep{yan2014casme} & 2014 & 1 & MER & -- & Spontaneous micro-expression benchmark recorded at 200 fps that, together with SMIC, became a standard MER evaluation corpus. \\
\hline
\rowcolor{gray!20} FER-2013~\citep{goodfellow2013challenges} & 2013 & 2 & Categorical (image) & -- & Web-scale challenge that ignited end-to-end deep FER and standardized in-the-wild evaluation. \\
\hline
\rowcolor{gray!20} FER+~\citep{barsoum2016training} & 2016 & 2 & Categorical (image) & -- & Crowdsourced label-distribution learning that mitigated annotator subjectivity. \\
\hline
\rowcolor{gray!20} RAF-DB~\citep{li2017reliable} & 2017 & 2 & Categorical (image) & -- & Reliable crowd-sourcing and deep locality-preserving learning for in-the-wild FER. \\
\hline
\rowcolor{gray!20} AffectNet~\citep{mollahosseini2017affectnet} & 2017 & 2 & Categorical \& VA estimation & -- & In-the-wild benchmark of about 1M web images, of which about 0.4M were manually annotated with discrete expressions and continuous valence--arousal. \\
\hline
\rowcolor{gray!2} DRML~\citep{zhao2016deep} & 2016 & 3 & AU detection & CNN & Deep region and multi-label learning that introduced AU-specific local filters into a unified CNN. \\
\hline
\rowcolor{gray!2} EAC-Net~\citep{li2017eacnet} & 2018 & 3 & AU detection & CNN & Enhancing-and-cropping branches around facial landmarks focused on AU-bearing regions. \\
\hline
\rowcolor{gray!2} Aff-Wild2~\citep{kollias2018aff} & 2018 & 3 & Categorical, VA \& AU detection & -- & Large-scale in-the-wild audio-visual benchmark introduced with VA annotation and later extended with AU and basic-expression labels for the ABAW challenges~\citep{kollias2022abaw}. \\
\hline
\rowcolor{gray!20} DFEW~\citep{jiang2020dfew} & 2020 & 4 & Categorical (video) & -- & Large-scale in-the-wild dynamic FER benchmark covering thousands of unconstrained video clips. \\
\hline
\rowcolor{gray!20} SCN~\citep{wang2020suppressing} & 2020 & 4 & Categorical (image) & CNN & Self-Cure Network suppressing label noise via self-attention ranking and careful relabeling. \\
\hline
\rowcolor{gray!20} TransFER~\citep{xue2021transfer} & 2021 & 4 & Categorical (image) & Hybrid & Relation-aware local-patch learning with multi-attention dropping and multi-head self-attention on CNN feature maps. \\
\hline
\rowcolor{gray!20} VTFF~\citep{ma2021facial} & 2021 & 4 & Categorical (image) & Hybrid & Visual Transformers with attentional selective fusion, contributing to the adoption of Transformers as SFER backbones. \\
\hline
\rowcolor{gray!20} FERV39k~\citep{wang2022ferv39k} & 2022 & 4 & Categorical (video) & -- & Multi-scene in-the-wild dynamic FER benchmark, the largest at the time of its release; supports cross-scene generalization studies. \\
\hline
\rowcolor{gray!20} POSTER~\citep{zheng2023poster} & 2023 & 4 & Categorical (image) & Hybrid & Pyramid cross-fusion Transformer combining CNN local cues with global ViT reasoning. \\
\hline
\rowcolor{gray!2} MMA-DFER~\citep{chumachenko2024mma} & 2024 & 5 & Categorical (audio-visual) & Transformer & Multi-modal adaptation of pre-trained uni-modal (audio and visual) models for in-the-wild DFER. \\
\hline
\rowcolor{gray!2} Emotion-LLaMA~\citep{cheng2024emotionllama} & 2024 & 5 & Categorical (audio-visual) & MLLM & Audio-visual instruction-tuned MLLM for joint emotion recognition and reasoning. \\
\hline
\rowcolor{gray!2} MPA-FER~\citep{ma2025multimodal} & 2025 & 5 & Categorical (image) & VLM & Multi-modal prompt alignment providing fine-grained textual--visual semantics for in-the-wild SFER. \\
\hline
\rowcolor{gray!2} FaceLLM~\citep{shahreza2025facellm} & 2025 & 5 & Categorical (image) + other face tasks & MLLM & Multi-modal large language model for unified face understanding, including expression recognition. \\
\hline
\rowcolor{gray!2} POSTER++~\citep{mao2025posterpp} & 2025 & 5 & Categorical (image) & Hybrid & Simplified successor of POSTER (window-based cross-attention, image-to-landmark branch removed) with 8.4G floating-point operations (FLOPs) and 43.7M parameters; 92.21\% on RAF-DB and 67.49\% on AffectNet (7 cl.), Table~\ref{tab:perf_sfer}. \\
\hline
\rowcolor{gray!2} FaceXFormer~\citep{narayan2025facexformer} & 2025 & 5 & Categorical (image) + other face tasks & Transformer & Unified Transformer addressing FER, head pose estimation, landmark detection, face parsing, and attribute estimation within a single model. \\
\hline
\rowcolor{gray!2} Mamba-VA~\citep{liang2025mamba} & 2025 & 5 & VA estimation & Hybrid & State-space (Mamba) temporal modeling on MAE-pre-trained ViT features for continuous valence--arousal estimation in the wild. \\
\hline
\rowcolor{gray!2} EMO-LLaMA~\citep{xing2024emollama} & 2026 & 5 & Categorical (image, video) & MLLM & MLLM with facial priors and instruction tuning for fine-grained emotion understanding. \\
\hline
\end{tabular}
\end{adjustbox}
\end{table*}

\subsubsection{Phase 4: Attention, Transformers, and uncertainty-aware FER (2020--2023)}
\label{sssec:FERphase4}
The Vision Transformer~\citep{dosovitskiy2020image,vaswani2017attention} prompted a second architectural shift, as ViT, Swin, and hierarchical attention backbones captured long-range dependencies that convolutional kernels model only weakly. TransFER~\citep{xue2021transfer}, VTFF~\citep{ma2021facial}, and POSTER~\citep{zheng2023poster} established Transformer-based FER through relation-aware patch learning and pyramid cross-fusion of local and global cues. In parallel, the subjectivity of expression labels motivated uncertainty-aware methods, with the Self-Cure Network~\citep{wang2020suppressing} becoming a de facto baseline for noisy-label FER~\citep{zhang2022learn,ma2023transformer}. Large-scale dynamic in-the-wild benchmarks (DFEW~\citep{jiang2020dfew}, FERV39k~\citep{wang2022ferv39k}), the lab-controlled 4D micro-expression corpus 4DME~\citep{li20224dme}, and the ABAW challenges built on Aff-Wild2~\citep{kollias2018aff,kollias2022abaw} repositioned the field around dynamic, multi-task evaluation.

\subsubsection{Phase 5: Vision--language, MLLMs, and foundation models (2023--present)}
\label{sssec:FERphase5}
The current phase is defined by a transition from task-specific FER models towards generalist, language-grounded systems. CLIP-based methods, such as MPA-FER~\citep{ma2025multimodal} and the AU-guided micro-expression alignment of MER-CLIP~\citep{liu2025mer}, ground expression semantics in textual prompts, while MLLMs tailored to facial expression and emotion understanding (Emotion-LLaMA~\citep{cheng2024emotionllama}, EMO-LLaMA~\citep{xing2024emollama}, EmoVerse~\citep{li2025emoverse}, FaceLLM~\citep{shahreza2025facellm}) couple recognition with affective reasoning through natural-language interfaces. Unified face-analysis foundation models (FaceXFormer~\citep{narayan2025facexformer}), pyramid hybrids (POSTER++~\citep{mao2025posterpp}), efficient state-space architectures of the Mamba family (Mamba-VA~\citep{liang2025mamba}), Transformer-graph micro-expression models (MOL~\citep{shao2025mol}), and reproducible open-source releases, such as the LibreFace analysis toolkit~\citep{chang2024libreface} and the HTNet micro-expression network~\citep{wang2024htnet}, reposition FER towards systems pre-trained once at scale that are intended to transfer broadly with minimal supervision across modalities and demographics. These overarching trends extend those documented in earlier dedicated FER surveys~\citep{li2020deep,li2022deep,liu2022graph}.

\subsection{Key models and milestones}
\label{ssec:FERmodels}
Throughout the research phases described in Section~\ref{ssec:FERphases}, a number of key models, benchmarks, and methodological milestones have been introduced, which have both driven significant advances and served as the basis for numerous subsequent methods. The most representative of these milestone works, along with their main characteristics, are summarized in Table~\ref{tab:FERmilestones}.

\section{Key criteria and main categories of FER methods}
\label{sec:Taxonomy}

This section provides a systematic overview of the landscape of deep learning-based FER methods. To facilitate the analysis, a set of complementary criteria is defined, each focusing on a specific key aspect of a FER method and resulting in the classification of the literature into a corresponding set of main categories. The criteria and the resulting categories are illustrated in Fig.~\ref{f:FERtaxonomy} and detailed below, while the corresponding in-depth analysis of the literature works is provided in the subsequent sections. The selected criteria collectively span the conceptual definition of the recognition task, the nature of the input signal, the face-analysis pre-processing pipeline, the underlying neural network architecture, the adopted learning strategy, the acquisition setting, and the targeted application domain; together they reflect the dominant design choices that shape contemporary FER systems~\citep{li2020deep,li2022deep,liu2022graph,sariyanidi2014automatic}.

\begin{figure*}[!t]
    \centering
    \begin{tikzpicture}[
        root/.style={
            rectangle, rounded corners=5pt, draw=black!70, thick, fill=gray!15,
            text width=6.5cm, align=center, minimum height=0.8cm,
            font=\normalsize\bfseries,
            blur shadow={shadow blur steps=5, shadow xshift=1.5pt, shadow yshift=-1.5pt, shadow opacity=20}
        },
        crit/.style={
            rectangle, rounded corners=3pt, thick,
            text width=2.4cm, align=center, minimum height=1.0cm,
            inner sep=2pt, font=\small\bfseries,
            blur shadow={shadow blur steps=5, shadow xshift=1pt, shadow yshift=-1pt, shadow opacity=15}
        },
        item/.style={
            rectangle, rounded corners=2pt, thick,
            text width=2.4cm, align=center, font=\small,
            inner sep=2.5pt,
            blur shadow={shadow blur steps=4, shadow xshift=0.5pt, shadow yshift=-0.5pt, shadow opacity=10}
        },
        link/.style={
            draw=black!45, line width=1pt,
            -{Stealth[length=2.5mm, width=2mm, inset=0.5mm, round]},
            rounded corners=6pt
        }
    ]

        \node[crit, fill=blue!10, draw=blue!50]                            (c1) {Recognition\\task};
        \node[crit, fill=cyan!10, draw=cyan!50, right=0.6cm of c1]         (c2) {Input\\modality};
        \node[crit, fill=teal!10, draw=teal!50, right=1.6cm of c2]         (c3) {Pre-processing}; %
        \node[crit, fill=violet!10, draw=violet!50, right=0.6cm of c3]     (c4) {Neural network\\architecture};

        \path (c2.east) -- (c3.west) coordinate[midway] (gap_center);
        \node[root, above=1.3cm of gap_center] (root) {Facial expression recognition methods};

        \draw[thick, draw=black!45] (root.south) -- ++(0,-0.4cm) coordinate (branch1);
        \foreach \c in {c1,c2,c3,c4} {
            \draw[link] (branch1) -| (\c.north);
        }

        \node[item, fill=blue!4, draw=blue!30, below=0.35cm of c1] (g1_1) {Categorical\\macro-expression};
        \node[item, fill=blue!4, draw=blue!30, below=0.20cm of g1_1] (g1_2) {Dimensional\\VA estimation};
        \node[item, fill=blue!4, draw=blue!30, below=0.20cm of g1_2] (g1_3) {Compound\\expression};
        \node[item, fill=blue!4, draw=blue!30, below=0.20cm of g1_3] (g1_4) {Micro-\\expression};
        \node[item, fill=blue!4, draw=blue!30, below=0.20cm of g1_4] (g1_5) {Action unit\\detection};
        \node[item, fill=blue!4, draw=blue!30, below=0.20cm of g1_5] (g1_6) {Expression\\intensity};
        \draw[link] (c1.south) -- (g1_1.north);

        \node[item, fill=cyan!4, draw=cyan!30, below=0.35cm of c2] (g2_1) {Static 2D\\image};
        \node[item, fill=cyan!4, draw=cyan!30, below=0.20cm of g2_1] (g2_2) {Dynamic 2D\\video};
        \node[item, fill=cyan!4, draw=cyan!30, below=0.20cm of g2_2] (g2_3) {3D/4D\\mesh--depth};
        \node[item, fill=cyan!4, draw=cyan!30, below=0.20cm of g2_3] (g2_4) {Thermal/NIR\\imagery};
        \node[item, fill=cyan!4, draw=cyan!30, below=0.20cm of g2_4] (g2_5) {Audio-visual};
        \node[item, fill=cyan!4, draw=cyan!30, below=0.20cm of g2_5] (g2_6) {Physiological\\fusion};
        \node[item, fill=cyan!4, draw=cyan!30, below=0.20cm of g2_6] (g2_7) {Vision--\\language};
        \draw[link] (c2.south) -- (g2_1.north);

        \node[item, fill=teal!4, draw=teal!30, below=0.35cm of c3] (g3_1) {Face detection\\\& localization};
        \node[item, fill=teal!4, draw=teal!30, below=0.20cm of g3_1] (g3_2) {Landmark-based\\alignment};
        \node[item, fill=teal!4, draw=teal!30, below=0.20cm of g3_2] (g3_3) {Photometric\\normalization};
        \node[item, fill=teal!4, draw=teal!30, below=0.20cm of g3_3] (g3_4) {Spatial normalization\\\& augmentation};
        \node[item, fill=teal!4, draw=teal!30, below=0.20cm of g3_4] (g3_5) {Temporal sampling \& motion magnification};
        \node[item, fill=teal!4, draw=teal!30, below=0.20cm of g3_5] (g3_6) {Modality-\\specific};
        \draw[link] (c3.south) -- (g3_1.north);

        \node[item, fill=violet!4, draw=violet!30, below=0.35cm of c4] (g4_1) {2D/3D CNNs};
        \node[item, fill=violet!4, draw=violet!30, below=0.20cm of g4_1] (g4_2) {RNNs/LSTMs};
        \node[item, fill=violet!4, draw=violet!30, below=0.20cm of g4_2] (g4_3) {Transformers};
        \node[item, fill=violet!4, draw=violet!30, below=0.20cm of g4_3] (g4_4) {Graph\\networks};
        \node[item, fill=violet!4, draw=violet!30, below=0.20cm of g4_4] (g4_5) {GANs/\\diffusion};
        \node[item, fill=violet!4, draw=violet!30, below=0.20cm of g4_5] (g4_6) {VLMs/\\MLLMs};
        \node[item, fill=violet!4, draw=violet!30, below=0.20cm of g4_6] (g4_7) {Hybrid};
        \draw[link] (c4.south) -- (g4_1.north);

        \path (gap_center |- g2_7.south) ++(0, -0.9cm) coordinate (branch2);
        
        \draw[thick, draw=black!45] (branch1) -- (branch2);

        \node[crit, fill=red!10, draw=red!50, below=0.4cm of branch2]       (c6) {Acquisition\\setting};
        \node[crit, fill=green!10, draw=green!50, left=0.8cm of c6]         (c5) {Learning\\strategy};
        \node[crit, fill=orange!10, draw=orange!50, right=0.8cm of c6]      (c7) {Application\\domain};

        \foreach \c in {c5,c6,c7} {
            \draw[link] (branch2) -| (\c.north);
        }

        \node[item, fill=green!4, draw=green!30, below=0.35cm of c5] (g5_1) {Supervised};
        \node[item, fill=green!4, draw=green!30, below=0.20cm of g5_1] (g5_2) {Semi-\\supervised};
        \node[item, fill=green!4, draw=green!30, below=0.20cm of g5_2] (g5_3) {Self-\\supervised};
        \node[item, fill=green!4, draw=green!30, below=0.20cm of g5_3] (g5_4) {Weakly/few-/\\zero-shot};
        \node[item, fill=green!4, draw=green!30, below=0.20cm of g5_4] (g5_5) {Domain\\adaptation};
        \node[item, fill=green!4, draw=green!30, below=0.20cm of g5_5] (g5_6) {Uncertainty/\\noisy-label};
        \node[item, fill=green!4, draw=green!30, below=0.20cm of g5_6] (g5_7) {Knowledge\\distillation};
        \node[item, fill=green!4, draw=green!30, below=0.20cm of g5_7] (g5_8) {Metric/\\contrastive};
        \draw[link] (c5.south) -- (g5_1.north);

        \node[item, fill=red!4, draw=red!30, below=0.35cm of c6] (g6_1) {Controlled\\(laboratory)};
        \node[item, fill=red!4, draw=red!30, below=0.20cm of g6_1] (g6_2) {In-the-wild};
        \draw[link] (c6.south) -- (g6_1.north);

        \node[item, fill=orange!4, draw=orange!30, below=0.35cm of c7] (g7_1) {Healthcare\\\& clinical};
        \node[item, fill=orange!4, draw=orange!30, below=0.20cm of g7_1] (g7_2) {Education\\\& learning};
        \node[item, fill=orange!4, draw=orange!30, below=0.20cm of g7_2] (g7_3) {Driver \&\\transport};
        \node[item, fill=orange!4, draw=orange!30, below=0.20cm of g7_3] (g7_4) {Security \&\\moderation};
        \node[item, fill=orange!4, draw=orange!30, below=0.20cm of g7_4] (g7_5) {HRI \& HCI};
        \node[item, fill=orange!4, draw=orange!30, below=0.20cm of g7_5] (g7_6) {Entertainment\\\& media};
        \node[item, fill=orange!4, draw=orange!30, below=0.20cm of g7_6] (g7_7) {Marketing\\\& consumer};
        \draw[link] (c7.south) -- (g7_1.north);

    \end{tikzpicture}
    \caption{Key criteria and main resulting categories of deep learning-based facial expression recognition methods.}
    \label{f:FERtaxonomy}
\end{figure*}

\subsection{Taxonomy of key criteria}
\label{sec:taxonomy_criteria}
\begin{itemize}[leftmargin=*]
    \item \textbf{Recognition task}: Groups methods by the specific expression-level target (i.e., the representation of the displayed facial behavior, check Section~\ref{sec:scope}) they predict, which determines the output, supervision and evaluation protocol. The main FER tasks are:
    \begin{itemize}
        \item \textbf{Categorical macro-expression recognition}: Maps prolonged, prototypical expressions to a small set of universal categories (happiness, anger, sadness, surprise, fear, disgust, optionally contempt and neutral)~\citep{ekman1971constants,li2020deep}.
        \item \textbf{Dimensional valence--arousal (VA) estimation}: Regresses continuous valence and arousal, as perceived from the displayed facial behavior, on a circumplex plane~\citep{kollias2022abaw,liang2025mamba}.
        \item \textbf{Compound expression recognition}: Addresses composite, blended facial expressions (e.g., happily surprised, sadly fearful) beyond the basic categories via multi-label supervision~\citep{du2014compound,li2019blended}.
        \item \textbf{Micro-expression recognition}: Targets spontaneous, very brief ($<0.5$~s), low-intensity facial movements that surface when subjects attempt to conceal emotion~\citep{yan2014casme,davison2016samm}.
        \item \textbf{Action unit (AU) detection}: Identifies the activation of individual anatomical muscle groups defined by the Facial Action Coding System~\citep{friesen1978facial,mavadati2013disfa}.
        \item \textbf{Expression intensity estimation}: Estimates the ordinal or continuous strength of a displayed expression or AU~\citep{walecki2017deep,mavadati2013disfa}.
    \end{itemize}

    \item \textbf{Input modality}: Groups methods by the type and dimensionality of the input signal, which dictates the available facial cues (spatial appearance, dynamics, geometry, multi-spectral, audio, bio-signals, text) and largely shapes the architectural design space:
    \begin{itemize}
        \item \textbf{Static 2D RGB image}: Operates on a single isolated frame using spatial appearance and geometric cues~\citep{li2020deep,mollahosseini2017affectnet}.
        \item \textbf{Dynamic 2D RGB video}: Processes ordered frame sequences to exploit facial motion across the neutral--onset--apex--offset cycle~\citep{jiang2020dfew,wang2022ferv39k}.
        \item \textbf{3D/4D geometric and depth data}: Uses dense meshes, depth maps, or 4D mesh sequences to provide view-invariant geometric representations~\citep{yin20063d,li20224dme}.
        \item \textbf{Thermal and near-infrared imagery}: Captures heat-emission and reflectance patterns that are largely invariant to visible-light conditions~\citep{zhao2011facial,wang2025ctiferk}.
        \item \textbf{Audio-visual}: Jointly exploits facial video and synchronously recorded speech~\citep{kollias2023abaw,chumachenko2024mma}.
        \item \textbf{Multi-modal physiological-signal fusion}: Fuses facial imagery with bio-signals, such as remote photoplethysmography (rPPG), electroencephalography (EEG), or electrocardiography (ECG)~\citep{ouzar2022video,jin2024residual}.
        \item \textbf{Text-grounded vision--language inputs}: Couples the facial input with textual information on expression semantics (class prompts, AU descriptions, or instructions), used either at inference or as training-time supervision~\citep{ma2025multimodal,shahreza2025facellm}.
    \end{itemize}

    \item \textbf{Pre-processing}: Encompasses the upstream pipeline that converts raw imagery (or video, depth, thermal, audio, bio-signals) into a normalized, model-ready representation, directly affecting robustness to pose, lighting, occlusion, and demographic shift. The main components are:
    \begin{itemize}
        \item \textbf{Face detection and localization}: Identifies the spatial extent of the face within the scene using cascade or deep detectors~\citep{zhang2016mtcnn,deng2020retinaface}.
        \item \textbf{Facial landmark detection and geometric alignment}: Registers the detected face to a canonical frame via transformations driven by anatomical landmarks~\citep{kazemi2014one,bulat2017far}.
        \item \textbf{Illumination and photometric normalization}: Suppresses lighting and color nuisance variation through histogram, retinex, or learned operators~\citep{sariyanidi2014automatic,zuiderveld1994clahe}.
        \item \textbf{Spatial normalization and data augmentation}: Converts the aligned face into a fixed-resolution tensor and applies deterministic and stochastic spatial, photometric, occlusion-based, sample-mixing, and generative augmentation~\citep{cubuk2019autoaugment,yun2019cutmix}.
        \item \textbf{Temporal sampling, apex selection, and motion magnification}: Applies clip sampling, apex/onset--offset selection, optical-flow extraction, and motion magnification on video and 4D inputs~\citep{liong2018less,wu2012eulerian}.
        \item \textbf{Modality-specific pre-processing}: Provides dedicated normalization for non-RGB modalities in the geometric, spectral, spectro-temporal acoustic, and autonomic-temporal domains~\citep{feng2018joint,chaddad2023eeg}.
    \end{itemize}

    \item \textbf{Neural network architecture}: The neural family used as the feature backbone or prediction head largely determines representational capacity, efficiency, and inductive biases. The main architectural families are:
    \begin{itemize}
        \item \textbf{2D/3D Convolutional Neural Networks (CNNs)}: Hierarchical convolutional backbones (2D for SFER, 3D and two-stream for DFER/MER)~\citep{lecun2002gradient,li2020deep}.
        \item \textbf{Recurrent neural networks (RNNs, including long short-term memory (LSTM) networks)}: Stacked on top of CNN backbones to model the long-range temporal dependencies of expressions and AU intensities~\citep{hochreiter1997long,jiang2020dfew}.
        \item \textbf{Transformers}: Vision Transformers and hierarchical variants that tokenize the face and apply multi-head self-attention to capture long-range, holistic dependencies~\citep{dosovitskiy2020image,chen2024static}.
        \item \textbf{Graph neural networks}: Treat the face as a sparse topological graph over landmarks or AU regions~\citep{kipf2016semi,liu2022graph}.
        \item \textbf{Generative and diffusion models}: GANs and diffusion models for identity decoupling, pose-invariant expression synthesis, and AU-conditioned image editing~\citep{pumarola2018ganimation,preechakul2022diffusion}.
        \item \textbf{Vision--language and multi-modal large language models (VLMs/MLLMs)}: CLIP-based dual encoders and multi-modal LLMs that introduce textual semantics into FER pipelines via prompt learning or instruction tuning~\citep{ma2025multimodal,shahreza2025facellm}.
        \item \textbf{Hybrid architectures}: Combine CNN feature extractors with Transformer global reasoning and attention modules~\citep{zheng2023poster,mao2025posterpp}.
    \end{itemize}

    \item \textbf{Learning strategy}: The various strategies differ in the type and amount of supervision required and in how label noise, scarcity, and distribution shift are handled. The most commonly adopted strategies are:
    \begin{itemize}
        \item \textbf{Supervised learning}: Standard cross-entropy training on expression-labeled data~\citep{li2020deep,mollahosseini2017affectnet}.
        \item \textbf{Semi-supervised learning}: Combines a small labeled subset with abundant unlabeled facial data via pseudo-labeling and consistency regularization~\citep{zeng2022face2exp,li2025enhanced}.
        \item \textbf{Self-supervised learning}: Pre-trains on large unlabeled corpora using masked autoencoders or contrastive objectives~\citep{gao2024self,fan2023selfme}.
        \item \textbf{Weakly supervised and few/zero-shot learning}: Leverages coarse, partial, or language-based supervision~\citep{jung2025text} and learning from few labeled samples, e.g., through meta-learning on scarce micro-expression data~\citep{wang2024meta}, and, in a few cases, enables zero-shot recognition.
        \item \textbf{Domain adaptation}: Bridges source--target distribution gaps through adversarial alignment, prototype transfer, source-free, or distributionally robust optimization~\citep{chen2021cross,cui2025learning}.
        \item \textbf{Uncertainty-aware and noisy-label learning}: Explicitly models annotation noise and uncertainty via self-cure mechanisms, label-distribution learning, and robust losses~\citep{wang2020suppressing,zhang2022learn}.
        \item \textbf{Knowledge distillation and parameter-efficient adaptation}: Compact students distill knowledge from heavy teachers~\citep{sun2020dynamic,chang2024libreface}, or large pre-trained vision/vision--language models are adapted with few trainable parameters~\citep{chumachenko2024mma}.
        \item \textbf{Metric and contrastive learning}: Triplet, center, and contrastive losses shape an embedding space in which same-expression samples cluster and different-expression samples separate~\citep{farzaneh2021facial}.
    \end{itemize}

    \item \textbf{Acquisition setting}: Refers to the conditions under which FER data are captured, which strongly influence both task difficulty and the conclusions drawn from reported performance. The two main settings are:
    \begin{itemize}
        \item \textbf{Controlled (laboratory)}: Captures expressions under uniform lighting, frontal pose, and constrained subjects~\citep{lucey2010extended,yin20063d,mavadati2013disfa}.
        \item \textbf{In-the-wild}: Captures unconstrained imagery and video from web sources or naturalistic environments~\citep{mollahosseini2017affectnet,jiang2020dfew,kollias2023abaw}.
    \end{itemize}

    \item \textbf{Application domain}: FER methods are deployed across diverse domains, each imposing distinctive constraints on latency, robustness, fairness, modality availability, and affect granularity. The most common application domains are:
    \begin{itemize}
        \item \textbf{Healthcare and clinical assessment}: Decision support for pain, depression, neurological/neuropsychiatric, and autism-spectrum assessment~\citep{kumar2023artificial,islam2024facepsy}.
        \item \textbf{Education and learning environments}: Continuous estimation of learner engagement, attention, confusion, and frustration in classrooms and online learning~\citep{savchenko2022classifying,guo2022smartedu}.
        \item \textbf{Driver monitoring and intelligent transportation}: Advanced driver-assistance system (ADAS)-oriented estimation of driver affect, fatigue, distraction, takeover readiness, and road-rage onset~\citep{duongthang2025driver,gu2024emotake}.
        \item \textbf{Security, surveillance, and content moderation}: Auxiliary FER for surveillance, mask-aware recognition, content moderation, and AU-guided deepfake forensics~\citep{kaushik2022isecurehome,deepfakeau2025}.
        \item \textbf{Human--robot and human--computer interaction}: Affect-aware behavior for social robots, conversational agents, and assistive HCI tools~\citep{embodied2025msafnet}.
        \item \textbf{Entertainment, gaming, and immersive media}: Drives real-time expressive avatars, emotion-responsive game logic, and facial animation for augmented/virtual reality (AR/VR)/metaverse applications~\citep{bellenger2024avatar,narayan2025facexformer}.
        \item \textbf{Marketing and consumer experience analytics}: Aggregate, population-level inference of engagement, sentiment, and reaction strength towards advertising stimuli and consumer products~\citep{poria2017review,mao2025posterpp}.
    \end{itemize}
\end{itemize}

The seven criteria are complementary rather than mutually exclusive, since each of them answers a distinct question about an FER method, namely what is predicted (recognition task), what information is presented to the model (input modality), how this information is prepared (pre-processing), which network computes the mapping (architecture), how its parameters are obtained (learning strategy), under which conditions the data are captured (acquisition setting), and what the system is used for (application domain). Each primary study is assigned a single (primary) category per method-level criterion (Section~\ref{sec:cross_criteria}); a method may therefore be discussed under several criteria, but from a different perspective in each of them. This is most evident for language-grounded FER, where the textual input (Section~\ref{ssec:mod_vlm}), the VLM/MLLM backbone that processes it (Section~\ref{ssec:arch_vlm}), and the scheme that trains it (Sections~\ref{ssec:learn_supervised}, \ref{ssec:learn_weak_fewshot}, and~\ref{ssec:learn_kd_peft}) are distinct design choices; accordingly, each of these sections describes only the aspect that corresponds to its criterion. The same rule applies to the other pairs of criteria that are most easily confused, namely the acronyms SFER and DFER denote categorical recognition from static and dynamic input, respectively, i.e., a recognition task instantiated on a given input modality; a generative model is regarded as the architecture of a method when it performs the recognition itself (Section~\ref{ssec:arch_generative}), as a pre-processing component when it only synthesizes training samples (Section~\ref{ssec:prep_augment}), and as a learning strategy when it provides a pre-training objective (Section~\ref{ssec:learn_selfsupervised}); the ``other affective states'' of Section~\ref{sec:scope} are recognition targets, whereas the application domains of Section~\ref{sec:app_domains} describe the deployment context; and the acquisition setting is a property of the data on which a method is trained and evaluated, rather than of the method itself. Throughout Sections~\ref{sec:recognition_tasks}--\ref{sec:app_domains}, each method is therefore described in full only in the subsection of the criterion that its main contribution concerns, whereas the other subsections mention only the aspect that is relevant to their criterion, with a cross-reference. The input-modality and architecture criteria are nevertheless strongly associated in the reviewed literature, where 9 of the 10 studies with text-grounded input employ a VLM/MLLM backbone, the remaining one being a hybrid CNN--Transformer model (FER-Former~\citep{li2024fer}), while a VLM/MLLM backbone is also used with audio-visual input (Emotion-LLaMA~\citep{cheng2024emotionllama}) (Fig.~\ref{fig:crosscriteria}(c)). In practice, a given FER method is thus simultaneously characterized along all of the criteria (e.g., a vision--language Transformer trained with prompt learning for DFER on in-the-wild videos and deployed for driver monitoring). This multi-criteria perspective is therefore adopted throughout the remainder of the manuscript.

\begin{figure*}[!t]
    \centering
    \resizebox{\textwidth}{!}{%
%
    }
    \caption{Cross-criteria distribution of the 167 primary deep learning-based FER studies (Section~\ref{sec:data_extraction}): (a) Recognition task versus architecture family, (b) Recognition task versus learning strategy, and (c) Input modality versus architecture family (the ``other affective'' task category comprises the studies that analyze facial expressions for inferring latent affective or clinical states, such as pain, depression, or engagement, check Section~\ref{sec:scope}). Each cell reports the number of studies exhibiting the corresponding combination of categories; every study is counted once per panel and is assigned to its primary category along each criterion, so that the row and column totals ($\Sigma$) sum to 167 and are consistent with the task and architecture distributions of Fig.~\ref{fig:bibliometrics}(c)--(e).}
    \label{fig:crosscriteria}
\end{figure*}

\subsection{Cross-criteria analysis}
\label{sec:cross_criteria}
Although the criteria of Section~\ref{sec:taxonomy_criteria} are presented separately, the reviewed literature populates their combinations very unevenly. Fig.~\ref{fig:crosscriteria} reports the number of the 167 primary studies (Section~\ref{sec:data_extraction}) for each combination of a) Recognition task and architecture family, b) Recognition task and learning strategy, and c) Input modality and architecture family; every study is assigned to a single (primary) category per criterion, so that the task and architecture marginals coincide with those of Fig.~\ref{fig:bibliometrics}(c)--(e). The analysis is restricted to the four criteria that characterize the method itself and are reported consistently in the primary studies; pre-processing, acquisition setting, and application domain are analyzed qualitatively in Sections~\ref{sec:preprocessing}, \ref{sec:acquisition_settings}, and~\ref{sec:app_domains}, since the corresponding choices are either shared by most methods (e.g., face detection and alignment), a property of the employed datasets, or stated only in a minority of the studies. Four observations can be made. First, the corpus is strongly concentrated, where categorical recognition combined with CNN-, Transformer-, or hybrid backbones accounts for 81 of the 167 studies (48.5\%), while 22 of the 49 task--architecture combinations are not represented at all. Second, the more recent architecture families are adopted unevenly across tasks; VLM/MLLM-based designs are encountered almost exclusively in categorical recognition (9 of 10 studies), with a single MER study and none on valence--arousal estimation, compound expression recognition, AU detection, or expression intensity estimation, whereas thermal/near-infrared (NIR) inputs are addressed by CNN-based models in 3 of 4 studies (the fourth being a CNN--Transformer hybrid) and 3D/4D inputs only by CNN-based recognition models (2 studies) and generative synthesis models (2 studies), which indicates that the shift towards Transformer-based, hybrid, and language-grounded architectures observed in Fig.~\ref{fig:bibliometrics}(d) has largely not yet propagated to these modalities. Third, the learning strategies are partly task-dependent, namely fully supervised training remains the most frequent choice (117 studies, 70.1\%), metric/contrastive learning is mostly encountered in MER (4 of 5 studies), where annotated samples are particularly scarce, whereas weakly supervised, few-shot, and zero-shot schemes are spread over five of the seven task categories (11 studies). Fourth, the cross-criteria structure of the corpus also evolves over time; comparing the 74 primary studies published up to 2022 with the 93 studies published in 2023--2026, the share of studies addressing tasks other than categorical recognition increases from 33.8\% (25 studies) to 40.9\% (38 studies), Transformer-based and hybrid designs account for 8 of the 13 recent MER studies (compared with 1 of 9 earlier ones), text-grounded vision--language inputs, not encountered in any study up to 2022, are adopted by 10 recent studies, whereas weakly supervised, few-shot, and zero-shot schemes increase only slightly (from 5 to 6 studies), since most of the recent language-grounded methods are trained with the labels of the target benchmarks. These observations describe the research activity of the analyzed corpus and do not indicate the relative performance of the corresponding approaches, which is discussed in Section~\ref{sec:performance_metrics}; nevertheless, they identify combinations that remain under-explored, such as graph-based and generative modeling beyond categorical recognition and AU detection, and language-grounded modeling of compound expressions and expression intensity, which are revisited in Sections~\ref{sec:challenges} and \ref{sec:future_directions}.

\section{Recognition tasks}
\label{sec:recognition_tasks}

FER methods can be grouped by the expression-level target they predict, which in turn determines the form of the output, the type of supervision, and the temporal scale of operation. In particular, the main FER tasks identified in the literature are: a) Categorical macro-expression recognition, b) Dimensional valence--arousal (VA) estimation, c) Compound expression recognition, d) Micro-expression recognition, e) AU detection, and f) Expression intensity estimation, as discussed in Section~\ref{sec:taxonomy_criteria} and further detailed below. Within each task, methods are further grouped according to their dominant source of inductive bias (representational prior).

\subsection{Categorical macro-expression recognition}
\label{ssec:rec_tasks-categorical}

Categorical macro-expression recognition (also referred to as conventional or prototypical FER) classifies prolonged facial expressions into a small set of discrete expression categories, typically named after the universal basic emotions~\citep{ekman1971constants,li2020deep}, and is conventionally decomposed into SFER on isolated images and DFER on facial video sequences. As the most mature and widely deployed FER setting, it provides the standard front-end for downstream human--computer interaction (HCI), healthcare, education, and social-robotics applications, and is the central reference task against which alternative formulations are typically benchmarked.

The main categories of categorical macro-expression recognition methods are:
\begin{itemize}
    \item \uline{Holistic appearance priors}: Methods extracting expression-discriminative information directly from raw facial pixels and global features. CNN-attention designs combine deep supervision and demographic-aware attention~\citep{fan2020facial}, squeeze-and-excitation/channel attention~\citep{gera2022cern,li2020facial}, scene-context attention~\citep{le2022global}, and multi-head dynamic global-local attention with multi-scale aggregation~\citep{wang2026mhan,xu2024multiscale}. Transformer-based and hybrid CNN--Transformer designs use multi-attention dropping for relation-aware patches~\citep{xue2021transfer}, attentional RGB/LBP fusion~\citep{ma2021facial}, attentive patch/token selection~\citep{xue2022vision}, and quaternion/wavelet representations~\citep{zhou2025delving}. Robust-supervision variants on the same input address label noise, imbalance, and demographic bias through self-attention ranking~\citep{wang2020suppressing}, flip-attention consistency with random erasing~\citep{zhang2022learn}, category-adaptive confidence margins with contrastive learning~\citep{li2022towards,li2025enhanced}, meta-learning circuit-feedback debiasing~\citep{zeng2022face2exp}, and label-distribution, uncertainty-aware, and graph-based noise-robust designs~\citep{le2023uncertainty,ma2023transformer,min2026robust}.

    \item \uline{Geometric and landmark-based priors}: Methods injecting facial geometry through landmarks, landmark graphs, or identity/pose disentanglement. POSTER \citep{zheng2023poster} and POSTER++~\citep{mao2025posterpp} use pyramid landmark-image cross-fusion Transformers; CMCNN~\citep{yu2022co} exploits landmark detection as an auxiliary task (Section~\ref{ssec:arch_hybrid}); IPD-FER~\citep{jiang2022disentangling} removes identity and pose factors from the expression representation (Section~\ref{ssec:arch_generative}); HRL~\citep{han2022devil} encodes landmark geometry as a graph (Section~\ref{ssec:arch_gnn}); Chen et al.~\citep{chen2024dual} target recognition that is invariant to expression intensity through graph-based manifold learning over samples (Section~\ref{ssec:arch_gnn}); and CRN~\citep{ZHU2022116046} casts in-the-wild FER as few-shot relation learning between the deep features of query and support samples.

    \item \uline{Motion and dynamics priors (DFER)}: Video-based methods exploiting the temporal/optical-flow evolution across the neutral--onset--apex--offset cycle. STCAM~\citep{chen2020stcam} attends to the expression-relevant frames and channels of the video (Section~\ref{ssec:mod_dfer}); EST~\citep{liu2023expression} decomposes the video into snippets (Section~\ref{ssec:arch_transformer}); MSSIF~\citep{pan2025learning} performs hierarchical temporal fusion across scales; PSRNet~\citep{wang2020phase} models the temporal dynamics of the expression (Section~\ref{ssec:mod_dfer}); OFR~\citep{poux2021dynamic} addresses occlusion in video-based recognition (Section~\ref{ssec:mod_dfer}); and ST-RDGCN~\citep{huang2025modeling} builds dynamic space-time graphs over local facial regions.

    \item \uline{Language-grounded semantic priors}: Methods aligning visual features with textual emotion or AU descriptions through CLIP-style or prompt-based pipelines. FER-Former~\citep{li2024fer} introduces CLIP-derived textual supervision via heterogeneous domain steering; MPA-FER~\citep{ma2025multimodal} uses large language model (LLM)-generated class descriptions (Section~\ref{ssec:mod_vlm}); MMPL-FER~\citep{pei2025multi} couples LLM expression descriptions with emoji-based visual prompts; and TG-DFER~\citep{jung2025text} extends language grounding to DFER under video-level supervision (Section~\ref{ssec:learn_weak_fewshot}).

    \item \uline{Multi-task and unified facial-representation priors}: Methods based on shared, foundation-style, or multi-modal representations across multiple facial-analysis objectives. S2D~\citep{chen2024static} reuses a model trained for static recognition on video (Section~\ref{ssec:mod_dfer}); MMA-DFER~\citep{chumachenko2024mma} adds the audio stream to in-the-wild DFER (Section~\ref{ssec:mod_audiovisual}); M3DFEL~\citep{wang2023rethinking} learns from video-level labels only (Section~\ref{ssec:learn_weak_fewshot}); HDF~\citep{cui2025learning} improves robustness through distribution-aware optimization; and MLLM-based Emotion-LLaMA~\citep{cheng2024emotionllama} and EMO-LLaMA~\citep{xing2024emollama} unify recognition and reasoning over audio-visual (Emotion-LLaMA) and visual (EMO-LLaMA) inputs via instruction tuning.
\end{itemize}

\subsection{Dimensional valence--arousal (VA) estimation}
\label{ssec:rec_tasks-va}

Dimensional VA estimation abandons the discrete categorical assumption and represents the displayed expression along the continuous, two-dimensional valence--arousal plane of the circumplex model~\citep{li2020deep}, capturing gradual, blended, and ambiguous expressions that the categorical model cannot adequately describe. It has become the de facto target of large-scale in-the-wild affective behavior analysis benchmarks (e.g., AffectNet, Aff-Wild2/ABAW series)~\citep{mollahosseini2017affectnet,kollias2023abaw} and underpins downstream applications requiring fine-grained continuous affect, such as HCI and mental-health monitoring. In line with the definition adopted in this survey (Section~\ref{sec:scope}), dimensional VA estimation is considered here as the estimation of valence and arousal as perceived from the displayed facial behavior, as annotated in AffectNet and Aff-Wild2, rather than of the felt affect of the subject, whose inference falls within the broader scope of FAR.

The main categories of dimensional VA estimation methods are:
\begin{itemize}
    \item \uline{Holistic appearance priors}: Methods estimating continuous VA from holistic facial appearance. CAGE~\citep{wagner2024cage} jointly supervises a MaxViT backbone with discrete and continuous VA targets on AffectNet/EMOTIC, demonstrating that discrete categories partially overlap in VA space.

    \item \uline{Motion and dynamics priors}: Video-based methods exploiting the temporal evolution of continuous affect. Masked autoencoder (MAE)-based pre-training combined with temporal convolutional network (TCN) and Transformer encoders captures Aff-Wild2 temporal dependencies~\citep{zhou2024enhancing}; Mamba-VA~\citep{liang2025mamba} uses state-space (Mamba) blocks for efficient long-sequence modeling; and audio-visual cross-attention fusion~\citep{praveen2022joint} adds prosodic/paralinguistic cues.

    \item \uline{Multi-task and unified facial-representation priors}: Methods built on shared multi-objective facial encoders. EmotiEffNets~\citep{savchenko2023emotieffnets} applies pre-trained facial descriptors to ABAW VA, expression, and AU heads; Savchenko et al.~\citep{savchenko2024leveraging} train lightweight MobileViT/MobileFaceNet/EfficientNet/DDAMFN models and transfer descriptors to VA, compound expression, and intensity tasks without extensive task-specific fine-tuning.

    \item \uline{Language-grounded semantic priors}: Methods grounding continuous affect in language-aligned MLLM-style representations. None of the primary studies of the analyzed corpus reports valence--arousal results with a language-grounded model, where the MLLM-based affect models, such as Emotion-LLaMA~\citep{cheng2024emotionllama}, EMO-LLaMA~\citep{xing2024emollama}, and EmoVerse~\citep{li2025emoverse}, are evaluated on multi-modal categorical emotion recognition, sentiment analysis, and emotion reasoning (Section~\ref{ssec:arch_vlm}), so that language-grounded valence--arousal estimation remains an open direction (Section~\ref{sec:cross_criteria}).
\end{itemize}

\subsection{Compound expression recognition}
\label{ssec:rec_tasks-cfer}

Compound expression recognition (CFER) extends categorical FER from the six/seven basic Ekman emotions to composite, blended facial expressions (e.g., happily surprised, sadly fearful) that combine multiple basic-emotion components~\citep{du2014compound,li2020deep}, naturally calling for a multi-label formulation over overlapping AU activations and ambiguous semantic boundaries. CFER is directly relevant to real-world HCI and clinical scenarios where pure basic emotions are rare, and is evaluated on dedicated benchmarks, such as the RAF-DB compound subset~\citep{li2017reliable} and C-EXPR-DB~\citep{kollias2023multi}.

The main categories of CFER methods are:
\begin{itemize}
    \item \uline{Anatomical/AU-grounded priors}: Methods decomposing compound affect into AU combinations and exploiting AU information as supervision or intermediate representation. MML~\citep{10208819} jointly learns CFER and AU detection through AU-expression alignment and meta-learning; C-EXPR-NET~\citep{kollias2023multi} combines visual and AU semantic features via cross-modal attention and distribution matching on C-EXPR-DB and the RAF-DB compound subset; and distribution-matching cross-task formulations~\citep{kollias2024distribution} enable knowledge transfer across compound, AU, and basic-expression labels.

    \item \uline{Holistic appearance priors with transfer}: Methods adapting strong basic-expression appearance backbones to compound recognition through few-shot, cross-domain, or compound-aware fine-tuning. EGS-Net~\citep{zou2022facial} formulates CFER as a cross-domain few-shot learning problem via an emotion-guided similarity framework.

    \item \uline{Language-grounded semantic priors (MLLM)}: Methods grounding compound recognition in natural-language verbalization of blended affect. Emotion-LLaMA~\citep{cheng2024emotionllama} and EMO-LLaMA~\citep{xing2024emollama} verbalize fine-grained affect through instruction tuning, but are evaluated on categorical recognition and emotion reasoning (Section~\ref{ssec:arch_vlm}); no primary study of the analyzed corpus reports compound expression results with a language-grounded model (Section~\ref{sec:cross_criteria}).
\end{itemize}

\subsection{Micro-expression recognition}
\label{ssec:rec_tasks-mer}

Micro-Expression Recognition (MER) targets spontaneous, low-intensity, and very short facial muscle activations (typically below 0.5\,s) that arise when subjects attempt to conceal their true emotional state~\citep{li2022deep}, and is therefore mostly addressed with motion-aware designs operating on apex frames or onset--apex--offset segments over small specialized corpora. MER provides highly objective, AU-level access to involuntary affect and is central to applications such as deception detection, security screening, and clinical assessment, where deliberately suppressed expressions cannot be reliably analyzed through standard categorical FER.

The main categories of MER methods are:
\begin{itemize}
    \item \uline{Motion and dynamics priors}: Methods explicitly modeling subtle facial motion through optical flow, motion magnification, temporal deformation, or apex/onset/offset evidence. MERASTC~\citep{gupta2021merastc} relies on AU, landmark, and gaze priors (Section~\ref{ssec:mod_dfer}); SE-DenseNet~\citep{cai2022micro} amplifies the low-intensity motion of micro-expressions (Section~\ref{ssec:mod_dfer}); AutoMER~\citep{verma2021automer} searches for efficient 3D CNN architectures for limited data; LTR3O~\citep{zhu2025learning} exploits the temporal ordering of micro-expression phases (Section~\ref{ssec:prep_temporal}); and SODA4MER~\citep{zhang2025dynamic} models the oriented facial deformations of micro-expressions (Section~\ref{ssec:learn_selfsupervised}). Motion-centric contrastive methods further include symmetric contrastive over motion features (SelfME~\citep{fan2023selfme}), supervised contrastive (MER-SupCon~\citep{zhi2022micro}), prototype-based memory contrastive (SRMCL~\citep{bao2024boosting}), and temporal augmentation consistency (TACL~\citep{wang2023temporal}).

    \item \uline{Geometric and region-based priors}: Methods focusing attention on expression-relevant facial regions through region-aware architectures. HTNet~\citep{wang2024htnet} restricts attention to expression-relevant facial regions (Section~\ref{ssec:arch_transformer}); C3DBed~\citep{pan2023c3dbed} and SLSTT~\citep{zhang2022short} combine 3D convolutions with a Transformer encoder (C3DBed) or optical-flow ViT features with LSTM-based short/long-range temporal aggregation (SLSTT); and FRL-DGT~\citep{zhai2023feature} represents micro-expressions as displacements of landmark-anchored facial regions (Section~\ref{ssec:arch_hybrid}).

    \item \uline{Anatomical/AU-grounded priors}: Methods exploiting AU labels or AU-aware supervision as the primary inductive bias. ARL~\citep{shao2019facial} models AU relationships through learnable correlation matrices, providing interpretable priors transferable to MER, while TS-AUCNN~\citep{sun2020dynamic} distills AU knowledge into a lightweight MER model via teacher--student transfer.

    \item \uline{Language-grounded semantic priors}: Methods aligning MER dynamics with textual semantics. MER-CLIP~\citep{liu2025mer} converts AU labels into textual descriptions and aligns them with visual MER dynamics via CLIP-style learning, whereas $\mu$-BERT~\citep{nguyen2023micron} uses BERT-style masked prediction over visual tokens rather than language (Section~\ref{ssec:learn_selfsupervised}).

    \item \uline{Multi-task and unified facial-representation priors}: Methods jointly training MER with related facial-analysis objectives over shared backbones. LightmanNet~\citep{wang2024meta} transfers macro-expression knowledge to MER through a dual-branch meta-auxiliary framework, while MOL~\citep{shao2025mol} jointly learns MER, optical-flow estimation, and landmark detection through a Transformer with graph-style convolutions.
\end{itemize}

\subsection{Action unit detection}
\label{ssec:rec_tasks-au}

Action unit (AU) detection identifies the activation of individual anatomical muscle groups defined by the Facial Action Coding System (FACS)~\citep{friesen1978facial,li2020deep}, typically formulated as multi-label classification (presence/absence) or AU-intensity regression over dedicated benchmarks. As an anatomically grounded, FACS-aligned formulation, AU detection underpins clinical, forensic, and behavioral-science applications, and provides a transferable mid-level representation widely reused as auxiliary supervision across categorical FER, CFER, MER, and expression intensity estimation.

The main categories of AU detection methods are:
\begin{itemize}
    \item \uline{Geometric and region-based priors}: Methods exploiting the spatially localized nature of AUs through landmark-anchored crops, region-specific filters, or region-aware architectures. DRML~\citep{zhao2016deep} learns AU-specific local filters within a unified CNN, while EAC-Net~\citep{li2017eacnet} combines enhancing and cropping branches around facial landmarks, jointly establishing region-anchored geometry as a strong inductive bias on BP4D and DISFA.

    \item \uline{Anatomical and relational AU priors}: Methods modeling the anatomical co-activation/mutual-exclusion structure among AUs through relational or graph-based reasoning. ARL~\citep{shao2019facial} models AU relationships through learnable correlation matrices for interpretable behavior analysis, while ME-GraphAU~\citep{luo2022megraphau} learns an AU relation graph with multi-dimensional edge features to encode structural muscle-group dependencies.

    \item \uline{Multi-task and unified facial-representation priors}: Methods integrating AU detection within shared, multi-objective, or foundation-style backbones. AVT~\citep{jin2022avt} fuses AU features with expression features for categorical recognition (Section~\ref{ssec:arch_transformer}); MAL~\citep{li2021meta} uses meta-auxiliary learning for scarce labels; and LibreFace~\citep{chang2024libreface} provides deployable models for AU detection, AU-intensity estimation, and FER, obtained via knowledge distillation.

    \item \uline{Language-grounded semantic priors}: Methods aligning AU representations with textual descriptions through CLIP-style or LLM-based prompts. No primary study of the analyzed corpus reports AU detection results with a language-grounded model; AU information enters such models only through text, e.g., MER-CLIP~\citep{liu2025mer} converts AU labels into textual descriptions and aligns them with visual micro-expression features, while MMPL-FER~\citep{pei2025multi} employs LLM-generated, AU-based expression descriptions as auxiliary text prompts for expression recognition.
\end{itemize}

\subsection{Expression intensity estimation}
\label{ssec:rec_tasks-intensity}

Expression intensity estimation extends FER from identifying an affective category or AU activation to quantifying how intense an expression or muscle activation is, encompassing Facial Expression Intensity Estimation (FEIE), Emotional Reaction Intensity (ERI) estimation, and AU-intensity regression as an anatomically grounded surrogate~\citep{li2020deep,kollias2023abaw}. It enables fine-grained behavioral analyses beyond presence/absence prediction, and constitutes a key building block for FAR applications, such as clinical pain, depression, and engagement assessment (Section~\ref{sec:app_domains}), as well as for reaction-level affective computing. In line with the definition adopted in this survey (Section~\ref{sec:scope}), intensity is considered here as the strength of the displayed expression or AU activation; ERI, whose labels are self-reported by the recorded subjects rather than assigned by external annotators~\citep{kollias2023abaw}, lies at the boundary between FER and FAR.

The main categories of expression intensity estimation methods are:
\begin{itemize}
    \item \uline{Motion and dynamics priors}: Methods estimating intensity from the explicit temporal evolution of facial behavior. Qian et al.~\citep{qian2023computer} compare CNN--LSTM and CNN--Transformer configurations for ERI, showing that ResNet50-Transformer outperforms LSTM alternatives in their experiments; LDL-EOR~\citep{xu2024facial} combines ordinal regression with label-distribution learning over temporal sequences for noise-robust supervision.

    \item \uline{Multi-task and unified facial-representation priors}: Methods based on shared multi-affect descriptors (expression, AU, VA) from pre-trained encoders. MTL-DAN~\citep{oh2023human} feeds multi-task expression/AU/VA representations into an LSTM regression head, while MMA-MRNNet~\citep{kollias2024mma} uses a Multiple-Models-of-Affect module followed by masked recurrent fusion over variable-length videos.

    \item \uline{Anatomical/AU-grounded priors}: Methods predicting AU intensity as a fine-grained, anatomically grounded surrogate. Relatedly, but without estimating intensity, Chen et al.~\citep{chen2024dual} formulate intensity-invariant FER as graph convolutional network (GCN)-based dual-subspace manifold learning over a graph of samples, while LibreFace~\citep{chang2024libreface} provides AU-intensity regression alongside FER and AU detection within a deployable toolkit.

    \item \uline{Language-grounded semantic priors}: Methods grounding intensity estimation in natural-language verbalization of expression strength. Emotion-LLaMA~\citep{cheng2024emotionllama}, EMO-LLaMA~\citep{xing2024emollama}, and EmoVerse~\citep{li2025emoverse} can describe affect in natural language, but none of them reports intensity results, so that language-grounded intensity estimation remains an open direction (Section~\ref{sec:cross_criteria}).
\end{itemize}

\begin{table*}[!t]
  \caption{Recognition tasks: Comparative analysis and key insights.}
  \label{tab:recognition_tasks_summary}
  \centering
  \scriptsize

  \setlength{\aboverulesep}{0pt}
  \setlength{\belowrulesep}{0pt}
  
  \setlength{\tabcolsep}{4pt}
  
  \renewcommand{\arraystretch}{1.3}

  \rowcolors{2}{gray!20}{gray!2}

  \newlength{\Wrtask}\setlength{\Wrtask}{1.4cm}
  \newlength{\Wrcat}\setlength{\Wrcat}{2.55cm}
  \newlength{\Wrva}\setlength{\Wrva}{2.55cm}
  \newlength{\Wrcfer}\setlength{\Wrcfer}{2.55cm}
  \newlength{\Wrmer}\setlength{\Wrmer}{2.55cm}
  \newlength{\Wrau}\setlength{\Wrau}{2.55cm}
  \newlength{\Wrint}\setlength{\Wrint}{2.55cm}

  \resizebox{\textwidth}{!}{%
  \begin{tabular}{@{}|
    >{\raggedright\arraybackslash}Q{\Wrtask}|
    >{\raggedright\arraybackslash}Q{\Wrcat}|
    >{\raggedright\arraybackslash}Q{\Wrva}|
    >{\raggedright\arraybackslash}Q{\Wrcfer}|
    >{\raggedright\arraybackslash}Q{\Wrmer}|
    >{\raggedright\arraybackslash}Q{\Wrau}|
    >{\raggedright\arraybackslash}Q{\Wrint}|
  @{}}
    \toprule
    \rowcolor{gray!50}
    \headerbreak{Recognition\\task} &
    \headerbreak{Categorical\\macro-expression} &
    \headerbreak{Dimensional\\VA estimation} &
    \headerbreak{Compound\\expression} &
    \headerbreak{Micro-\\expression} &
    \headerbreak{Action unit\\detection} &
    \headerbreak{Expression\\intensity} \\
    \midrule

    Primary function &
    \begin{tabitem}
      \item Discrete expression classification of holistic prototypical expressions
      \item Static\slash dynamic FER (SFER\slash DFER) over coarse expression categories
    \end{tabitem} &
    \begin{tabitem}
      \item Continuous valence--arousal regression over a two-dimensional affect plane
      \item Smooth, ambiguity-tolerant dimensional affect modeling
    \end{tabitem} &
    \begin{tabitem}
      \item Recognition of blended, composite facial expressions
      \item Multi-label modeling of co-activated basic-emotion components
    \end{tabitem} &
    \begin{tabitem}
      \item Recognition of subtle, brief, involuntary expressions
      \item Localized, low-intensity facial-muscle activation analysis
    \end{tabitem} &
    \begin{tabitem}
      \item FACS-grounded detection of individual muscle-group activations
      \item Multi-label, mid-level facial-behavior description
    \end{tabitem} &
    \begin{tabitem}
      \item Estimation of expression\slash AU strength and reaction intensity
      \item ERI and FEIE prediction over reaction-level windows
    \end{tabitem} \\
    \midrule

    Temporal scale &
    \begin{tabitem}
      \item Single frame (SFER)
      \item Short to long video clip (DFER)
    \end{tabitem} &
    \begin{tabitem}
      \item Frame to long sequence
      \item Sliding affective windows
    \end{tabitem} &
    \begin{tabitem}
      \item Single frame or short clip
      \item Onset--apex--offset evidence
    \end{tabitem} &
    \begin{tabitem}
      \item Very short clip ($\leq$0.5\,s)
      \item Onset--apex--offset segments
    \end{tabitem} &
    \begin{tabitem}
      \item Single frame or short clip
      \item Frame-level AU dynamics
    \end{tabitem} &
    \begin{tabitem}
      \item Reaction- or sequence-level windows
      \item Variable-length video
    \end{tabitem} \\
    \midrule

    Dominant representational priors &
    \begin{tabitem}
      \item Holistic appearance
      \item Geometric\slash landmark
      \item Motion\slash dynamics (DFER)
      \item Language semantics
      \item Multi-task\slash unified
    \end{tabitem} &
    \begin{tabitem}
      \item Holistic appearance
      \item Motion\slash dynamics
      \item Multi-task\slash unified
      \item Language semantics
    \end{tabitem} &
    \begin{tabitem}
      \item Anatomical\slash AU grounding
      \item Holistic appearance (with transfer)
      \item Language semantics (MLLM)
    \end{tabitem} &
    \begin{tabitem}
      \item Motion\slash dynamics
      \item Geometric\slash region
      \item Anatomical\slash AU grounding
      \item Language semantics
      \item Multi-task\slash unified
    \end{tabitem} &
    \begin{tabitem}
      \item Geometric\slash region
      \item Anatomical\slash relational AU
      \item Multi-task\slash unified
      \item Language semantics
    \end{tabitem} &
    \begin{tabitem}
      \item Motion\slash dynamics
      \item Multi-task\slash unified
      \item Anatomical\slash AU grounding
      \item Language semantics
    \end{tabitem} \\
    \midrule

    Input &
    \begin{tabitem}
      \item Single facial image (SFER)
      \item Facial video sequence (DFER)
    \end{tabitem} &
    \begin{tabitem}
      \item Facial image, video, or audio-visual stream
    \end{tabitem} &
    \begin{tabitem}
      \item Facial image or video
      \item Optional AU\slash landmark side input
    \end{tabitem} &
    \begin{tabitem}
      \item Short facial clip
      \item Optical-flow or motion-magnified input
    \end{tabitem} &
    \begin{tabitem}
      \item Aligned facial image or short clip
      \item Local AU-region crops
    \end{tabitem} &
    \begin{tabitem}
      \item Facial video or reaction sequence
      \item Optional audio-visual stream
    \end{tabitem} \\
    \midrule

    Output &
    \begin{tabitem}
      \item Single-label discrete category from image/\allowbreak video-level supervision
      \item Optional softmax confidence
    \end{tabitem} &
    \begin{tabitem}
      \item Continuous valence and arousal scores
      \item Per-frame or per-window regression supervision
    \end{tabitem} &
    \begin{tabitem}
      \item Compound-expression labels with multi-label blended affect
      \item Often combined with AU side-supervision
    \end{tabitem} &
    \begin{tabitem}
      \item Micro-expression class with apex\slash temporal-phase annotations
      \item Sparse, short clip-level supervision
    \end{tabitem} &
    \begin{tabitem}
      \item Multi-label, FACS-coded AU activations
      \item Optional AU intensity values
    \end{tabitem} &
    \begin{tabitem}
      \item Continuous intensity score
      \item Ordinal or label-distribution supervision
    \end{tabitem} \\
    \midrule

    Strengths &
    \begin{tabitem}
      \item Mature benchmarks
      \item Strong in-the-wild generalization
      \item Direct downstream integration
    \end{tabitem} &
    \begin{tabitem}
      \item Captures blended/\allowbreak subtle affect
      \item Temporal continuity
      \item Cross-cultural neutrality
    \end{tabitem} &
    \begin{tabitem}
      \item Captures real-world affect complexity
      \item Natural multi-label formulation
      \item Direct HCI relevance
    \end{tabitem} &
    \begin{tabitem}
      \item Objective access to involuntary affect
      \item AU-level interpretability
      \item Motion-aware robustness
    \end{tabitem} &
    \begin{tabitem}
      \item High interpretability
      \item Anatomically grounded
      \item Transferable to other FER tasks
    \end{tabitem} &
    \begin{tabitem}
      \item Fine-grained behavioral characterization
      \item Ordinal\slash label-distribution support
      \item Synergy with VA and AU
    \end{tabitem} \\
    \midrule

    Limitations &
    \begin{tabitem}
      \item Subjective and noisy labels
      \item Class imbalance
      \item Closed taxonomy
      \item Demographic\slash pose sensitivity
    \end{tabitem} &
    \begin{tabitem}
      \item High annotation cost
      \item Noise-sensitive metrics
      \item Weak frame-level evidence
      \item Discrete-continuous mismatch
    \end{tabitem} &
    \begin{tabitem}
      \item Small, imbalanced data
      \item Ambiguous semantic boundaries
      \item Negative transfer from basic FER
    \end{tabitem} &
    \begin{tabitem}
      \item Extreme data scarcity
      \item Apex-frame dependence
      \item Limited cross-corpus generalization
    \end{tabitem} &
    \begin{tabitem}
      \item Expensive expert annotation
      \item Dataset-dependent AU co-occurrence
      \item Long-tailed AU distribution
    \end{tabitem} &
    \begin{tabitem}
      \item Sparse and noisy intensity labels
      \item High annotation subjectivity
      \item Long-tailed intensity distribution
    \end{tabitem} \\
    \midrule

    Indicative models &
    \begin{tabitem}
      \item POSTER++ \citep{mao2025posterpp}, MPA-FER \citep{ma2025multimodal}, EAC \citep{zhang2022learn}, MMA-DFER \citep{chumachenko2024mma}
    \end{tabitem} &
    \begin{tabitem}
      \item CAGE \citep{wagner2024cage}, EmotiEffNets \citep{savchenko2023emotieffnets}, Mamba-VA \citep{liang2025mamba}, MAE+TCN \citep{zhou2024enhancing}
    \end{tabitem} &
    \begin{tabitem}
      \item MML \citep{10208819}, C-EXPR-NET \citep{kollias2023multi}, EGS-Net \citep{zou2022facial}
    \end{tabitem} &
    \begin{tabitem}
      \item HTNet \citep{wang2024htnet}, MOL \citep{shao2025mol}, SelfME \citep{fan2023selfme}, MER-CLIP \citep{liu2025mer}
    \end{tabitem} &
    \begin{tabitem}
      \item DRML \citep{zhao2016deep}, EAC-Net \citep{li2017eacnet}, ARL \citep{shao2019facial}, ME-GraphAU \citep{luo2022megraphau}
    \end{tabitem} &
    \begin{tabitem}
      \item LDL-EOR \citep{xu2024facial}, MTL-DAN \citep{oh2023human}, MMA-MRNNet \citep{kollias2024mma}, Qian et al. \citep{qian2023computer}
    \end{tabitem} \\
    \bottomrule
  \end{tabular}%
  }
\end{table*}

\begin{figure*}[!ht]
    \centering

    \begin{subfigure}[c]{0.30\textwidth}
        \centering
        \includegraphics[width=\textwidth]{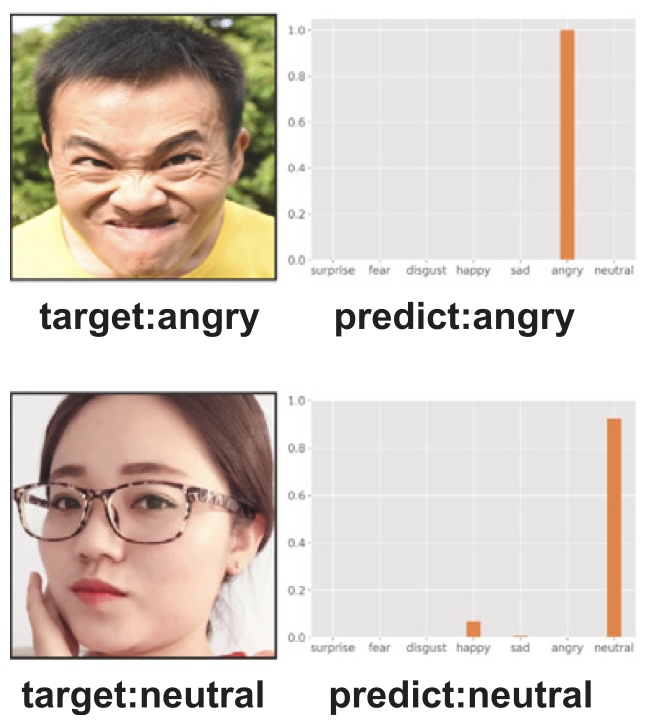}
        \caption{}
    \end{subfigure}
    \hfill
    \begin{subfigure}[c]{0.30\textwidth}
        \centering
        \includegraphics[width=\textwidth]{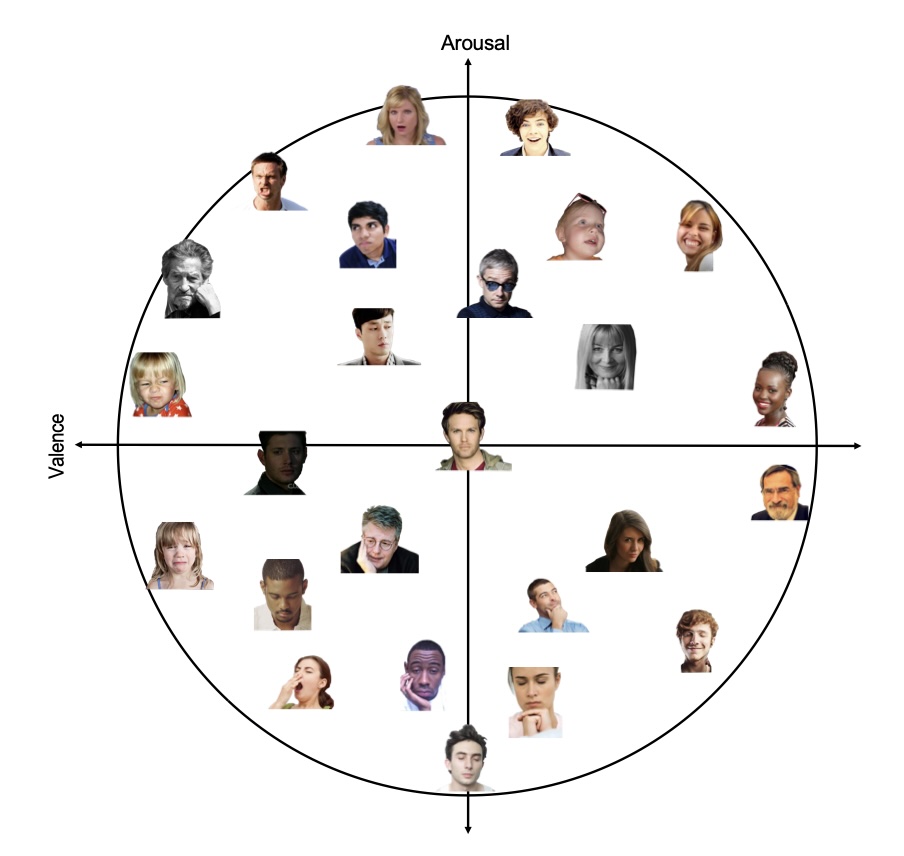}
        \caption{}
    \end{subfigure}
    \hfill
    \begin{subfigure}[c]{0.30\textwidth}
        \centering
        \includegraphics[width=\textwidth]{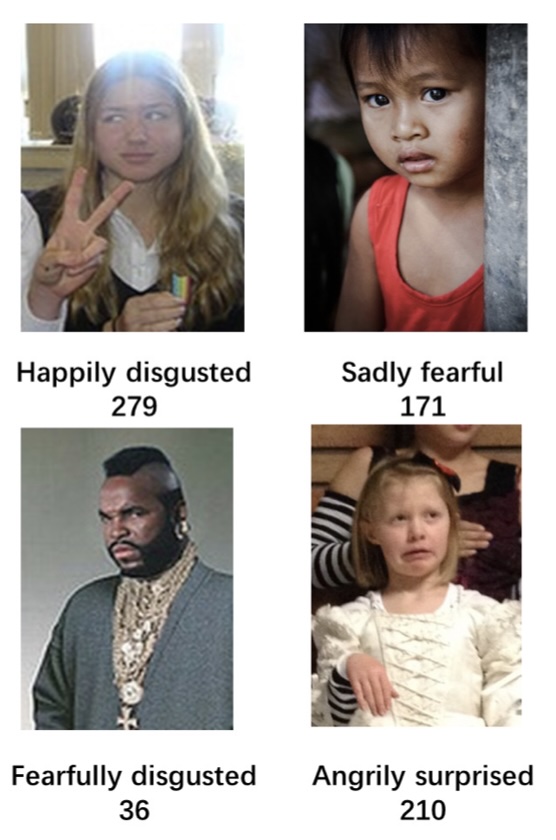}
        \caption{}
    \end{subfigure}

    \vspace{0.25cm}

    \begin{subfigure}[c]{0.30\textwidth}
        \centering
        \includegraphics[width=\textwidth]{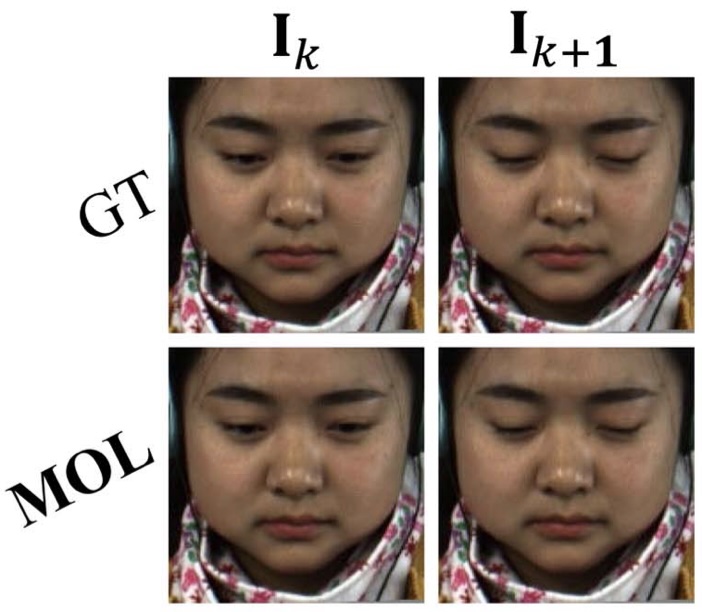}
        \caption{}
    \end{subfigure}
    \hfill
    \begin{subfigure}[c]{0.30\textwidth}
        \centering
        \includegraphics[width=\textwidth]{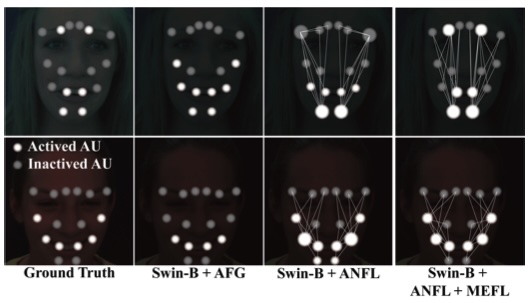}
        \caption{}
    \end{subfigure}
    \hfill
    \begin{subfigure}[c]{0.30\textwidth}
        \centering
        \includegraphics[width=\textwidth]{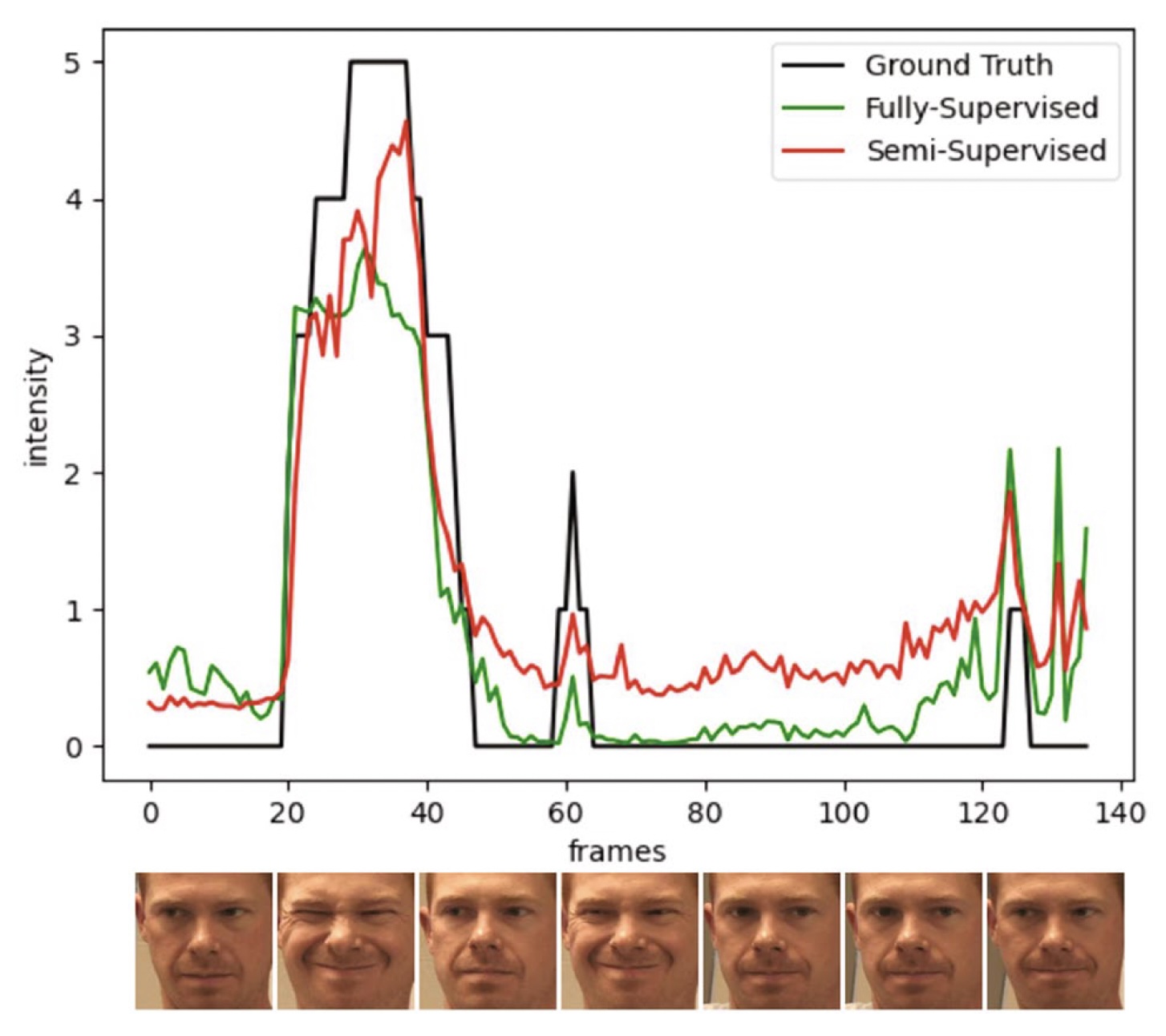}
        \caption{}
    \end{subfigure}

    \caption{Representative literature methods per recognition task: a) Categorical macro-expression recognition (POSTER++~\citep{mao2025posterpp}), b) Dimensional VA estimation (CAGE~\citep{wagner2024cage}), c) Compound expression recognition (MML~\citep{10208819}), d) Micro-expression recognition (MOL~\citep{shao2025mol}), e) Action unit detection (ME-GraphAU~\citep{luo2022megraphau}), and f) Expression intensity estimation (LDL-EOR~\citep{xu2024facial}).}
    \label{fig:recognition_tasks_examples}
\end{figure*}

\subsection{Comparative analysis and key insights}
\label{ssec:rec_tasks_discussion}

Building on the detailed discussion of the recognition tasks (Sections~\ref{ssec:rec_tasks-categorical}--\ref{ssec:rec_tasks-intensity}), this subsection compares them and provides critical insights for each task. Table~\ref{tab:recognition_tasks_summary} summarizes, for each recognition task, its a) Primary function, b) Temporal scale, c) Dominant representational priors, d) Input, e) Output formulation, f) Key strengths, g) Critical limitations, and h) Indicative models. The comparison shows that categorical macro-expression recognition, dimensional VA estimation, and compound expression recognition target holistic-emotion reasoning, whereas MER, AU detection, and expression intensity estimation focus on fine-grained, anatomically grounded analysis. Additionally, an important temporal differentiator across tasks lies in evidence selection and aggregation, ranging from single-frame SFER and frame-level AU detection to extremely short ($\leq$0.5\,s) MER windows and reaction-level intensity sequences. Moreover, AU information is used across all tasks (serving as anatomical supervision, intermediate representation, or text-aligned semantic cue), and landmark or AU priors are over-represented among the highest reported SFER and MER results, although their effect on cross-dataset generalization has not yet been tested (Section~\ref{ssec:perf_discussion}), while language-grounded priors are increasingly adopted, enabling open-vocabulary and interpretable FER and, in a few cases, zero-shot recognition. Furthermore, the main supervision bottlenecks differ in nature rather than in severity, with annotation cost, label noise, demographic shifts, and long-tailed distributions and data scarcity recurrently constraining all tasks. Finally, a trend from isolated, task-specific models towards unified multi-task Transformers and MLLM-based backbones is observed in the reviewed works, although no recognition task is fully solved. Representative literature methods per recognition task are illustrated in Fig.~\ref{fig:recognition_tasks_examples}, while the distribution of the reviewed primary studies over the combinations of recognition task with architecture family and learning strategy is reported in Fig.~\ref{fig:crosscriteria}(a)--(b).

\section{Input modalities}
\label{sec:input_modalities}

FER methods can be grouped by the type and dimensionality of the input signal they process. The input determines which facial cues are available to the model and thus shapes the entire architectural design space. In particular, the main input modalities exploited in modern FER are: a) Static 2D RGB images, b) Dynamic 2D RGB videos, c) 3D/4D geometric and depth data, d) Thermal and near-infrared imagery, e) Audio-visual streams, f) Multi-modal physiological-signal fusion, and g) Text-grounded vision--language inputs, as discussed in Section~\ref{sec:taxonomy_criteria} and further detailed below.

\subsection{Static 2D RGB images}
\label{ssec:mod_sfer}

Static 2D RGB images constitute the most widely adopted input modality for FER, primarily targeting SFER, i.e., the prediction of the expression class from a single facial frame, as well as frame-level AU detection and valence--arousal estimation, through spatial appearance and geometric cues, such as wrinkles, furrows, and landmark configurations~\citep{li2020deep,mollahosseini2017affectnet}. The modality is supported by abundant large-scale in-the-wild benchmarks and is directly compatible with mature pre-trained CNN, Transformer, and hybrid backbones, underpinning the bulk of real-time and on-device FER deployments.

Static 2D RGB-based FER systems can be grouped by the dominant backbone family and auxiliary inductive bias into the following categories:
\begin{itemize}
    \item \uline{Holistic-appearance CNN/Transformer methods}: Methods directly mapping the raw aligned RGB face crop to expression labels through deep CNN, Transformer, or hybrid backbones. CNN-attention designs inject squeeze-and-excitation and multi-scale attention (DSAN~\citep{fan2020facial}, CERN~\citep{gera2022cern}, SPWFA-SE~\citep{li2020facial}, GLAMOR-Net~\citep{le2022global}, MHAN~\citep{wang2026mhan}, MFER~\citep{xu2024multiscale}), while Transformer-based and hybrid CNN--Transformer designs tokenize the aligned face and apply multi-head self-attention (TransFER~\citep{xue2021transfer}, VTFF~\citep{ma2021facial}, APViT~\citep{xue2022vision}, QWTR~\citep{zhou2025delving}). Robust-supervision variants on the same input address label noise, class imbalance, and demographic bias (SCN~\citep{wang2020suppressing}, EAC~\citep{zhang2022learn}, Ada-CM~\citep{li2022towards}, Face2Exp~\citep{zeng2022face2exp}, EACM~\citep{li2025enhanced}).

    \item \uline{Landmark-augmented RGB methods}: Methods complementing the RGB face with auxiliary geometric streams (landmarks, heatmaps, identity/pose codes) through dual-branch or cross-fusion designs. POSTER/POSTER++~\citep{zheng2023poster,mao2025posterpp} use pyramid landmark-image cross-fusion Transformers; CMCNN~\citep{yu2022co} adds a landmark stream to the single RGB image (Section~\ref{ssec:arch_hybrid}); IPD-FER~\citep{jiang2022disentangling} factors out the identity and pose cues of the single image (Section~\ref{ssec:arch_generative}); HRL~\citep{han2022devil} embeds landmark-graph priors, leveraging structural anatomy that the holistic branch tends to underutilize; and Chen et al.~\citep{chen2024dual} and CRN~\citep{ZHU2022116046} instead model relations between samples, through graph-based manifold learning and few-shot relation learning, respectively.

    \item \uline{Region-based AU-oriented methods}: SFER and AU-detection methods operating on local landmark- or AU-anchored crops rather than the global face. DRML~\citep{zhao2016deep} learns AU-specific local filters within a unified CNN, while EAC-Net~\citep{li2017eacnet} combines enhancing and cropping branches around landmarks, encoding the spatially localized nature of facial muscle activations for AU detection and occlusion-robust SFER.

    \item \uline{Generative and synthesized RGB methods}: Methods augmenting, balancing, or de-identifying the static RGB input through GAN- or diffusion-based image synthesis. GAN-based approaches~\citep{jiang2022disentangling,pumarola2018ganimation} synthesize identity-invariant or pose-normalized RGB faces to mitigate dataset bias and long-tailed distributions, while diffusion-based pipelines extend this direction to AU-conditioned image editing.
\end{itemize}

\subsection{Dynamic 2D RGB videos}
\label{ssec:mod_dfer}

Dynamic 2D RGB videos extend the static modality with the temporal dimension, providing access to the entire neutral--onset--apex--offset cycle of an expression through ordered RGB frame sequences that combine spatial appearance with motion dynamics~\citep{li2020deep,jiang2020dfew,wang2022ferv39k}. This modality is the natural substrate for DFER, dimensional VA, and emotional reaction intensity estimation in the wild, and underpins most large-scale video-based affect benchmarks.

With respect to the temporal-modeling architecture, dynamic 2D RGB-based FER systems fall into the following categories:
\begin{itemize}
    \item \uline{3D CNN and spatio-temporal CNN methods}: Methods processing the RGB video as a spatio-temporal volume through 3D convolutions or two-stream CNNs. STCAM~\citep{chen2020stcam} extends 3D CNNs with spatio-temporal/channel attention for DFER; SE-DenseNet~\citep{cai2022micro} couples 3D DenseNet with motion magnification for MER; and multi-stream snippet-based designs (EST~\citep{liu2023expression}, MSSIF~\citep{pan2025learning}) decompose the video into snippets and aggregate hierarchically.

    \item \uline{Video Transformer and adapter-based methods}: Methods tokenizing RGB frames into spatio-temporal patches and applying self-attention or temporal adapters on strong image backbones. S2D~\citep{chen2024static} transfers a landmark-aware SFER model via temporal adapters and emotion-anchor self-distillation; C3DBed~\citep{pan2023c3dbed} and SLSTT~\citep{zhang2022short} combine 3D convolutions with a Transformer encoder (C3DBed) or optical-flow ViT features with LSTM-based short/long-range temporal aggregation (SLSTT); M3DFEL~\citep{wang2023rethinking} aggregates 3D instances generated from the video (Section~\ref{ssec:learn_weak_fewshot}); and HDF~\citep{cui2025learning} introduces heterogeneity-aware distributionally robust optimization.

    \item \uline{Optical-flow and motion-magnified methods}: Methods modeling facial motion through optical flow, motion magnification, or onset--apex--offset evidence. MERASTC~\citep{gupta2021merastc} encodes compact motion-aware AU/landmark/gaze/appearance cues; OFR~\citep{poux2021dynamic} reconstructs occluded motion in the optical-flow domain; LTR3O~\citep{zhu2025learning} uses the ordering of video segments as a learning signal (Section~\ref{ssec:prep_temporal}); SODA4MER~\citep{zhang2025dynamic} extracts deformation cues from short video clips (Section~\ref{ssec:learn_selfsupervised}); and PSRNet~\citep{wang2020phase} reconstructs phase-space trajectories. Motion-centric contrastive methods (SelfME~\citep{fan2023selfme}, MER-SupCon~\citep{zhi2022micro}, SRMCL~\citep{bao2024boosting}, TACL~\citep{wang2023temporal}) regularize motion representations under scarcity.

    \item \uline{State-space and long-sequence methods}: Methods replacing quadratic self-attention with linear-complexity state-space modeling for efficient long-horizon temporal reasoning. Mamba-VA~\citep{liang2025mamba} couples MAE features with TCN and Mamba state-space blocks for continuous VA on Aff-Wild2, while MAE-based pre-training combined with TCN/Transformer encoders captures dependencies across long ABAW clips~\citep{zhou2024enhancing}.
\end{itemize}

\subsection{3D/4D geometric and depth data}
\label{ssec:mod_3d}

3D/4D geometric and depth data extend the 2D RGB modality to the third (geometric) dimension through dense meshes, depth maps, or temporally indexed 4D mesh sequences~\citep{sandbach2012static,li2022cas}. By decoupling facial geometry from photometric variation, this modality provides intrinsic, illumination- and pose-invariant representations and is particularly suited to micro-expression analysis and high-precision biomedical or forensic applications.

Depending on the geometric data representation, 3D/4D-based FER systems can be divided into the following main categories:
\begin{itemize}
    \item \uline{Depth/RGB-D fusion methods}: Methods operating on aligned depth and RGB streams through early-, intermediate-, or late-fusion schemes. CAS(ME)$^3$~\citep{li2022cas} provides RGB-D micro-expression sequences with depth-enhanced apex evidence, while 4DME~\citep{li20224dme} couples DI4D, RGB, depth, and grayscale streams. Such fusion is particularly valuable for MER and AU detection, where subtle deformations are easier to detect in the depth channel.

    \item \uline{Mesh-based and point-cloud methods}: Methods operating directly on the unordered 3D point cloud or surface mesh via permutation-invariant or geometric deep learning operators. Point-cloud consistency learning, proposed for incomplete 3D face (identity) recognition, offers completion strategies transferable to partial 3D FER inputs~\citep{liu2024pointcloud}; deep fusion CNNs combine 2D attribute maps of the 3D geometry and texture for multi-modal 2D+3D FER~\citep{guo20203dfacial}; and full-head 3D morphable models disentangle shape, expression, and texture~\citep{ploumpis2021head}, providing expression parameters that can serve as geometric FER features.

    \item \uline{4D spatio-temporal methods}: Methods exploiting the temporal evolution of 3D meshes for fine-grained micro-motion analysis. Behzad et al.~\citep{behzad2021sparsity} introduce sparsity-aware deep 4D FER, Zhen et al.~\citep{behzad2021magnifying} show with a non-deep Riemannian pipeline that magnifying subtle facial deformations improves 4D FER, while the recent 4D spontaneous MER database~\citep{wang2024spontaneous4D} provides a benchmark for controlled-setting 4D MER evaluation.

    \item \uline{Projection-based 2D-from-3D methods}: Hybrid methods rendering multiple 2D views or depth/normal maps from the 3D mesh and feeding them to standard 2D CNN/Transformer backbones, inheriting 2D-backbone maturity while preserving the geometric robustness of 3D data~\citep{sandbach2012static,yin20063d}.
\end{itemize}

\subsection{Thermal and near-infrared imagery}
\label{ssec:mod_thermal}

Thermal infrared (TIR) and near-infrared (NIR) imagery capture facial heat-emission or NIR-reflectance patterns that are largely invariant to visible-light conditions and that reveal physiological correlates of affect, such as vasoconstriction and perspiration~\citep{bhattacharyya2021deep}. This modality is essential for FER in low-light, nighttime, or privacy-sensitive deployments in security, automotive, healthcare, and smart-classroom settings, and is evaluated on dedicated benchmarks.

Thermal/NIR-based FER systems can be grouped by the degree and form of RGB--infrared (IR) fusion into the following categories:
\begin{itemize}
    \item \uline{Pure-IR deep models}: Methods training CNN, Transformer, or Kolmogorov--Arnold network (KAN)-based backbones directly on thermal/NIR imagery without RGB fusion. CTIFERK~\citep{wang2025ctiferk} combines a MobileViT (CNN--Transformer) backbone with KANs for thermal FER on Tufts, IRIS, and GUET-thermal-face; the NIR Oulu-CASIA benchmark~\citep{zhao2011facial} was introduced with handcrafted LBP-based baselines; and IRFacExNet-style residual networks~\citep{bhattacharyya2021deep} recover expression-discriminative features from low-contrast thermal images.

    \item \uline{RGB-IR early/late-fusion methods}: Methods combining visible and IR streams through early, intermediate, or late fusion in attention-based pipelines. Naseem et al.~\citep{naseem2024msx} systematically compare early and late fusion of ResNet-18 backbones over visible, IR, and multi-spectral dynamic imaging (MSX) streams on VIRI and NVIE, and subsequently extend ResNet-18 with an attention mechanism for the early fusion of visible and IR features on the same benchmarks~\citep{khan2025visir}.

    \item \uline{Multi- and hyper-spectral methods}: Methods exploiting visible/NIR/short-wave infrared facial signatures through specialized deep fusion architectures. HyperFace~\citep{vasquez2024hyperface}, a hyperspectral face (identity) recognition model fusing visible and NIR bands that illustrates spectral-fusion designs transferable to FER, combines pre-fusion, encoder, fusion, and decoder modules with local and global residual learning, exploiting complementary spectral bands over standard RGB, while the spectrum-aware deep metric-learning approach of Wu et al.~\citep{wu2021spectrum} jointly leverages spectrum and class-label information through a spectrum-aware embedding and a multi-spectral discriminant correlation loss.
\end{itemize}

\subsection{Audio-visual streams}
\label{ssec:mod_audiovisual}

Audio-visual streams jointly exploit the facial video and the synchronously recorded speech signal, leveraging the well-known correlation between prosodic/paralinguistic cues and visible facial behavior~\citep{li2020deep,kollias2023abaw}. This modality is the natural substrate for dimensional VA, compound expression, and emotional reaction intensity estimation in large-scale in-the-wild challenges, where audio disambiguates cases in which the face is occluded, in profile, or expressively neutral.

With respect to the cross-modal fusion strategy, audio-visual FER systems fall into the following categories:
\begin{itemize}
    \item \uline{Two-stream CNN/Transformer fusion methods}: Methods processing audio and visual streams through separate backbones and fusing via cross-attention, gating, or concatenation. The Transformer-based multi-modal framework of Zhang et al.~\citep{zhang2022transformer} integrates visual, audio, and textual streams for joint AU/expression on Aff-Wild2; MMA-DFER~\citep{chumachenko2024mma} combines in-the-wild video with its audio track, with 77.92\% versus 67.15\% weighted average recall (WAR) on DFEW (fold~1) for the audio-visual and the visual-only variants (Section~\ref{ssec:arch_transformer}); and the joint cross-attention model of Praveen et al.~\citep{praveen2022joint} refines continuous-affect estimation through audio-visual cross-modal attention.

    \item \uline{Audio-visual MLLM-based methods}: Methods grounding audio-visual signals in multi-modal LLMs for natural-language reasoning over both streams. In Emotion-LLaMA~\citep{cheng2024emotionllama}, the audio stream is encoded by HuBERT and passed to the language model together with the visual features; the corresponding architecture is described in Section~\ref{ssec:arch_vlm}.

    \item \uline{Multi-modal unified-representation methods}: Methods adapting shared visual backbones to multiple FER tasks of the audio-visual ABAW corpora within a single model, in part without exploiting the audio stream. EmotiEffNets~\citep{savchenko2023emotieffnets} and Savchenko et al.~\citep{savchenko2024leveraging} train lightweight pre-trained descriptors in multi-task settings and transfer them to VA, compound, and intensity tasks, while MMA-MRNNet~\citep{kollias2024mma} uses a Multiple-Models-of-Affect module with masked recurrent fusion for ERI.

    \item \uline{Multi-modal emotion databases and benchmarks}: Works constructing and evaluating multi-modal corpora combining facial video with speech, EEG, and other auxiliary signals. The MGEED database~\citep{wang2023mgeed} provides multi-modal genuine-emotion detection in lab conditions, while large-scale audio-visual challenges (Aff-Wild2~\citep{kollias2018aff}, C-EXPR-DB~\citep{kollias2023multi}, ABAW~\citep{kollias2022abaw,kollias2023abaw,kollias20247th}) provide the de facto in-the-wild evaluation setting.
\end{itemize}

\subsection{Multi-modal physiological-signal fusion}
\label{ssec:mod_physio}

Multi-modal physiological-signal fusion extends the facial input with bio-signals (e.g., EEG, ECG, electrodermal activity (EDA)/galvanic skin response (GSR), electrooculography (EOG), photoplethysmography (PPG), respiration), targeting affective phenomena that are not fully observable through visible behavior~\citep{ouzar2022video,yu2025gbvnet}. This modality is central to clinical, automotive, and HCI applications that demand robustness to deliberate expression suppression and access to autonomic correlates of affect (e.g., disambiguating concentration from anxiety), with continuous, sample-level supervision over synchronously recorded facial and bio-signal streams.

Depending on the type of bio-signal coupled with facial features, multi-modal physiological FER systems can be divided into the following main categories:
\begin{itemize}
    \item \uline{Facial-video and remote-photoplethysmography fusion methods}: Methods combining facial appearance with rPPG signals extracted from the same video, supporting fully camera-based multi-modal affect recognition. Ouzar et al.~\citep{ouzar2022video} introduce a video-based spontaneous-emotion framework combining facial expressions with rPPG-derived physiological signals, with consistent gains over single-modality baselines, while Li and Peng~\citep{li2024endtoend} propose an end-to-end network that fuses facial-expression and rPPG features through a Transformer-based cross-modal attention mechanism.

    \item \uline{Facial-EEG fusion methods}: Methods fusing facial video with EEG, exploiting complementarity between cortical and behavioral correlates of affect. The Residual Multimodal Transformer (RMMT)~\citep{jin2024residual} introduces multi-modal attention and residual fusion for continuous emotion recognition from face-EEG streams on the MAHNOB-HCI dataset, outperforming earlier fusion methods, while the MMHA-FNN of Feng et al.~\citep{feng2025transformer} couples EEG and facial features through multi-head self-attention for emotion recognition in hearing-impaired subjects.

    \item \uline{Facial multi-bio-signal hierarchical fusion}: Methods hierarchically fusing facial signals with multiple peripheral bio-signals (EEG, ECG, EOG, GSR) via cross-modal hierarchies. GBV-Net~\citep{yu2025gbvnet} introduces a Gated Biological Visual Network that hierarchically fuses these streams through gated cross-modal attention, addressing the limitations of methods that treat physiological signals in isolation, while Saffaryazdi et al.~\citep{saffaryazdi2022micro} combine facial micro-expressions with EEG, GSR, and PPG signals for valence--arousal recognition on synchronously recorded multi-bio-signal data.

    \item \uline{Application-driven multi-cue methods}: Methods combining facial behavior with further behavioral or physiological cues for application-specific scenarios. EmoTake~\citep{gu2024emotake} derives facial-expression, head-pose, and body-movement cues from a single in-cabin camera to predict takeover readiness, reaction time, and quality in semi-autonomous driving, while FacePsy~\citep{islam2024facepsy}, a non-deep system that classifies CNN-extracted AU and head-gesture features with gradient boosting, shows that longitudinal mental-health monitoring is also feasible from front-camera facial behavior alone, without physiological sensors.
\end{itemize}

\subsection{Text-grounded vision--language inputs}
\label{ssec:mod_vlm}

Text-grounded inputs complement the facial image or video with a textual signal that encodes expression semantics, so that recognition can be formulated as the alignment of visual evidence with language rather than as the prediction of an index over a closed label set~\citep{radford2021clip,zhou2022coop}. This subsection concerns the textual input itself, i.e., its source, granularity, and role; the backbones that process it and the learning schemes that exploit it are discussed in Sections~\ref{ssec:arch_vlm}, \ref{ssec:learn_supervised}, \ref{ssec:learn_weak_fewshot}, and~\ref{ssec:learn_kd_peft}.

Depending on the source of the text, text-grounded FER inputs can be grouped into the following categories:
\begin{itemize}
    \item \uline{Class-level templates}: Fixed prompts built around the expression label, e.g., ``this is a face image of \{expression\}'' in FER-Former~\citep{li2024fer}.

    \item \uline{Class-level descriptions}: LLM-generated descriptions of the facial behavior associated with each class, combined with learnable context tokens in DFER-CLIP~\citep{zhao2023prompting} and with learnable soft prompts in MPA-FER~\citep{ma2025multimodal}.

    \item \uline{AU-level descriptions}: FACS-based descriptions of the AUs associated with each class, as in PE-CLIP~\citep{saadi2025peclip}, or, at the sample level, descriptions generated from the ground-truth AU annotations of each clip, as in MER-CLIP~\citep{liu2025mer}.

    \item \uline{Symbolic prompts}: Non-lexical cues, such as emojis, paired with LLM-generated descriptions in MMPL-FER~\citep{pei2025multi}.

    \item \uline{Instructions and dialogue}: Free-form natural-language questions answered in natural language, as in EMO-LLaMA~\citep{xing2024emollama}, EmoVerse~\citep{li2025emoverse}, and FaceLLM~\citep{shahreza2025facellm}, the latter trained on LLM-generated question--answer pairs; Emotion-LLaMA~\citep{cheng2024emotionllama} additionally receives the audio stream (Section~\ref{ssec:mod_audiovisual}).
\end{itemize}

The role of the text also differs, since it acts as a fixed class-level classifier at inference (FER-Former, DFER-CLIP, PE-CLIP), as a per-sample auxiliary supervision signal used only during training (MER-CLIP), or as a per-query instruction (MLLM-based methods); a text-grounded method therefore does not necessarily receive text at inference time. For video, the text is typically attached at the clip level, so that the relevant frames must be inferred, e.g., through the text-guided temporal modeling of TG-DFER~\citep{jung2025text}. The limitations specific to this modality concern the text itself, where the results depend on the wording and the quality of the prompts~\citep{zhou2022coop}, sample-level AU descriptions require AU annotations that are costly to obtain~\citep{liu2025mer}, and generated descriptions may carry lexical or cultural bias.

\begin{table*}[!t]
  \caption{Input modalities: Comparative analysis and key insights.}
  \label{tab:input_modalities_summary}
  \centering
  \scriptsize

  \setlength{\aboverulesep}{0pt}
  \setlength{\belowrulesep}{0pt}
  
  \setlength{\tabcolsep}{4pt}
  
  \renewcommand{\arraystretch}{1.3}

  \rowcolors{2}{gray!20}{gray!2}

  \newlength{\Wmod}\setlength{\Wmod}{1.4cm}
  \newlength{\Wsta}\setlength{\Wsta}{2.2cm}
  \newlength{\Wdyn}\setlength{\Wdyn}{2.2cm}
  \newlength{\Wgeo}\setlength{\Wgeo}{2.2cm}
  \newlength{\Wthe}\setlength{\Wthe}{2.2cm}
  \newlength{\Wav}\setlength{\Wav}{2.2cm}
  \newlength{\Wphy}\setlength{\Wphy}{2.2cm}
  \newlength{\Wvl}\setlength{\Wvl}{2.2cm}

  \resizebox{\textwidth}{!}{%
  \begin{tabular}{@{}|
    >{\raggedright\arraybackslash}Q{\Wmod}|
    >{\raggedright\arraybackslash}Q{\Wsta}|
    >{\raggedright\arraybackslash}Q{\Wdyn}|
    >{\raggedright\arraybackslash}Q{\Wgeo}|
    >{\raggedright\arraybackslash}Q{\Wthe}|
    >{\raggedright\arraybackslash}Q{\Wav}|
    >{\raggedright\arraybackslash}Q{\Wphy}|
    >{\raggedright\arraybackslash}Q{\Wvl}|
  @{}}
    \toprule
    \rowcolor{gray!50}
    \headerbreak{Input\\modality} &
    \headerbreak{Static 2D\\RGB image} &
    \headerbreak{Dynamic 2D\\RGB video} &
    \headerbreak{3D/4D\\geometric/depth} &
    \headerbreak{Thermal/\\near-infrared} &
    \headerbreak{Audio--\\visual} &
    \headerbreak{Physiological\\fusion} &
    \headerbreak{Vision--\\language} \\
    \midrule

    Signal form &
    \begin{tabitem}
      \item Single aligned RGB image
      \item Optional landmark\slash AU-region side-input
    \end{tabitem} &
    \begin{tabitem}
      \item Sequence of RGB frames
      \item Optional optical-flow or apex frames
    \end{tabitem} &
    \begin{tabitem}
      \item RGB-D frame, depth map, 3D mesh, or 4D mesh sequence
      \item Optional 3D landmarks
    \end{tabitem} &
    \begin{tabitem}
      \item Thermal or NIR image\slash video
      \item Optional aligned RGB stream
    \end{tabitem} &
    \begin{tabitem}
      \item Synchronized RGB video and audio waveform\slash spectrogram
      \item Optional transcript
    \end{tabitem} &
    \begin{tabitem}
      \item Facial video + bio-signals (EEG, ECG, EDA, PPG, EOG)
      \item Optional wearable-sensor data
    \end{tabitem} &
    \begin{tabitem}
      \item Aligned image\slash video and textual prompt
      \item Optional AU verbalizations, emojis, instructions
    \end{tabitem} \\
    \midrule

    Temporal granularity &
    \begin{tabitem}
      \item Single frame (no temporal context)
      \item Peak\slash onset still configurations
    \end{tabitem} &
    \begin{tabitem}
      \item Short to long video clip
      \item Onset--apex--offset cycle modeling
    \end{tabitem} &
    \begin{tabitem}
      \item Single 3D scan or 4D mesh sequence
      \item Subtle micro-motion windows
    \end{tabitem} &
    \begin{tabitem}
      \item Single thermal\slash NIR frame or short clip
      \item Slow physiological-thermal dynamics
    \end{tabitem} &
    \begin{tabitem}
      \item Sequence- or reaction-level window
      \item Continuous synchronized streams
    \end{tabitem} &
    \begin{tabitem}
      \item Continuous synchronized bio-signal traces
      \item Long-horizon, sample-level windows
    \end{tabitem} &
    \begin{tabitem}
      \item Frame or video paired with persistent textual prompt
      \item Flexible context window
    \end{tabitem} \\
    \midrule

    Supported FER tasks &
    \begin{tabitem}
      \item SFER (primary)
      \item AU detection
      \item Intensity from peak frames
    \end{tabitem} &
    \begin{tabitem}
      \item DFER (primary)
      \item Dimensional VA
      \item MER, AU dynamics, intensity
    \end{tabitem} &
    \begin{tabitem}
      \item Categorical 3D/4D FER (primary); 4D MER
      \item AU under pose\slash illumination
    \end{tabitem} &
    \begin{tabitem}
      \item SFER and DFER under low-light\slash night
      \item Privacy-sensitive monitoring
    \end{tabitem} &
    \begin{tabitem}
      \item Categorical DFER (primary)
      \item Dimensional VA
      \item Compound expression
      \item Reaction-intensity estimation
    \end{tabitem} &
    \begin{tabitem}
      \item Categorical affect (primary), also under expression suppression
      \item Continuous VA
    \end{tabitem} &
    \begin{tabitem}
      \item Open-vocabulary SFER\slash DFER (primary)
      \item AU-aware FER and MER
      \item Verbalized affective reasoning
    \end{tabitem} \\
    \midrule

    Strengths &
    \begin{tabitem}
      \item Low cost
      \item Mature benchmarks
      \item Direct compatibility with pre-trained backbones
    \end{tabitem} &
    \begin{tabitem}
      \item Captures kinematic signatures
      \item Robust in the wild
      \item Compatible with self-supervised pre-training
    \end{tabitem} &
    \begin{tabitem}
      \item Illumination\slash pose invariance
      \item Captures subtle geometric cues
      \item Direct biomedical\slash forensic suitability
    \end{tabitem} &
    \begin{tabitem}
      \item Operates under low light\slash night
      \item Privacy-preserving
      \item Physiological correlates
    \end{tabitem} &
    \begin{tabitem}
      \item Complementary prosodic cues
      \item Strong on in-the-wild challenges
      \item Natural fit for VA\slash intensity
    \end{tabitem} &
    \begin{tabitem}
      \item Access to autonomic cues
      \item Robust to expression suppression
      \item Clinically grounded
    \end{tabitem} &
    \begin{tabitem}
      \item Semantic label priors from class text
      \item AU-level descriptive granularity
      \item Supervision from textual descriptions
    \end{tabitem} \\
    \midrule

    Limitations &
    \begin{tabitem}
      \item No temporal evidence
      \item Sensitive to occlusion\slash pose
      \item Subjective labels
    \end{tabitem} &
    \begin{tabitem}
      \item Higher compute cost
      \item Key-frame selection
      \item Weak frame-level evidence
    \end{tabitem} &
    \begin{tabitem}
      \item Specialized hardware
      \item Small benchmarks
      \item Sparse\slash incomplete data
    \end{tabitem} &
    \begin{tabitem}
      \item No color\slash fine texture
      \item Small datasets
      \item Sensor drift
    \end{tabitem} &
    \begin{tabitem}
      \item Requires synchronized audio
      \item Background noise
      \item Speech-content bias
    \end{tabitem} &
    \begin{tabitem}
      \item Multi-sensor acquisition
      \item Inter-subject variability
      \item Privacy concerns
    \end{tabitem} &
    \begin{tabitem}
      \item Dependence on prompt wording and quality
      \item Cost of sample-level AU annotations
      \item Lexical/cultural bias of generated text
    \end{tabitem} \\
    \midrule

    Indicative methods &
    \begin{tabitem}
      \item POSTER++ \citep{mao2025posterpp}, EAC \citep{zhang2022learn}, DRML \citep{zhao2016deep}, MHAN \citep{wang2026mhan}
    \end{tabitem} &
    \begin{tabitem}
      \item S2D \citep{chen2024static}, Mamba-VA \citep{liang2025mamba}, M3DFEL \citep{wang2023rethinking}, EST \citep{liu2023expression}
    \end{tabitem} &
    \begin{tabitem}
      \item CAS(ME)$^3$ \citep{li2022cas}, 4DME \citep{li20224dme}, Behzad et al. \citep{behzad2021sparsity}
    \end{tabitem} &
    \begin{tabitem}
      \item CTIFERK \citep{wang2025ctiferk}, IRFacExNet \citep{bhattacharyya2021deep}, Naseem et al. \citep{naseem2024msx,khan2025visir}
    \end{tabitem} &
    \begin{tabitem}
      \item MMA-DFER \citep{chumachenko2024mma}, Emotion-LLaMA \citep{cheng2024emotionllama}, Praveen et al. \citep{praveen2022joint}, Zhang et al. \citep{zhang2022transformer}
    \end{tabitem} &
    \begin{tabitem}
      \item Ouzar et al. \citep{ouzar2022video}, RMMT \citep{jin2024residual}, GBV-Net \citep{yu2025gbvnet}
    \end{tabitem} &
    \begin{tabitem}
      \item FER-Former \citep{li2024fer}, DFER-CLIP \citep{zhao2023prompting}, PE-CLIP \citep{saadi2025peclip}, MER-CLIP \citep{liu2025mer}
    \end{tabitem} \\
    \bottomrule
  \end{tabular}%
  }
\end{table*}

\subsection{Comparative analysis and key insights}
\label{ssec:input_modalities_discussion}

Building on the detailed discussion of the input modalities (Sections~\ref{ssec:mod_sfer}--\ref{ssec:mod_vlm}), this subsection compares them and provides critical insights for each modality. Table~\ref{tab:input_modalities_summary} summarizes, for each input modality, its a) Signal form, b) Temporal granularity, c) Supported FER tasks, d) Key strengths, e) Critical limitations, and f) Indicative methods. A first observation is that modalities differ along largely distinct information axes (geometric, spectral, acoustic, autonomic, and linguistic) rather than along refinements of a single visual signal, so each modality can contribute affective evidence that is only partly recoverable from the others; the controlled comparisons available in the reviewed works support this mainly for audio-visual input, e.g., 67.15\% versus 77.92\% WAR on DFEW (fold~1) for the visual-only and the audio-visual variants of MMA-DFER~\citep{chumachenko2024mma}. Additionally, temporal granularity is a key axis separating static (RGB, thermal/NIR) from sequence-level modalities (dynamic RGB, audio-visual, physiological), with recent state-space (Mamba) VA and adapter-based DFER approaches indicating a steady move towards efficient, long-horizon temporal reasoning. Moreover, no single modality covers the entire FER task space uniformly, where static RGB inputs are the most frequently used for SFER and AU detection (and thermal/NIR inputs, in the analyzed corpus, exclusively for categorical recognition), dynamic RGB and audio-visual modalities are the natural substrate for DFER and dimensional VA, 3D/4D modalities capture geometric deformations that are relevant to MER and pose/illumination-robust AU detection, although the four 3D/4D studies of the analyzed corpus address categorical expressions (two recognition and two generative synthesis studies) (Section~\ref{sec:cross_criteria}), physiological fusion particularly supports continuous affect under expression suppression, and vision--language inputs natively enable open-vocabulary and verbalized affective outputs. Furthermore, the main bottlenecks differ in nature rather than in severity (missing temporal evidence for static RGB, computational cost for dynamic RGB, hardware and small-benchmark constraints for 3D/4D and thermal/NIR, speech-content bias for audio-visual, multi-sensor complexity for physiological, and prompt sensitivity for vision--language). Finally, several recent works combine the facial signal with a second modality (audio or text), so that the design question shifts from the choice of a single modality to the selection and alignment of complementary signals; the corresponding architectural trend is discussed in Section~\ref{ssec:nn_architectures_discussion}. The distribution of the reviewed primary studies over the combinations of input modality and architecture family is reported in Fig.~\ref{fig:crosscriteria}(c).

\section{Pre-processing}
\label{sec:preprocessing}

FER methods can be grouped by the upstream pre-processing pipeline that converts raw signals into a normalized representation suitable for deep models. This pipeline strongly influences the model's robustness to head pose, illumination, occlusion, acquisition noise, demographic shift, and class imbalance, and often determines the practical reproducibility of the reported results across different benchmarks. In particular, the main pre-processing pipelines used in deep learning-based FER methods are: a) Face detection and localization, b) Facial landmark detection and geometric alignment, c) Illumination and photometric normalization, d) Spatial normalization and data augmentation, e) Temporal sampling, apex selection, and motion magnification, and f) Modality-specific pre-processing, as discussed in Section~\ref{sec:taxonomy_criteria} and further detailed below.

\subsection{Face detection and localization}
\label{ssec:prep_facedet}

Face detection and localization constitute the entry point of virtually every FER pipeline, identifying the spatial extent of faces within a cluttered scene and producing the bounding boxes that subsequent stages crop, align, and resize~\citep{sariyanidi2014automatic,kollias2024abaw6}. As the upstream stage on which the rest of the pipeline depends, it directly determines the quality of in-the-wild, on-device, and challenge-grade FER deployments, and often returns reusable anchor landmarks that downstream alignment and AU-region cropping consume at no extra cost.

With respect to the detector architecture family, face-detection pre-processing pipelines fall into the following categories:
\begin{itemize}
    \item \uline{Handcrafted-feature cascade detectors}: Methods relying on hand-engineered features (Haar, LBP, HOG) with cascaded boosting or sliding-window classifiers. The Viola--Jones detector~\citep{viola2004robust} was the standard real-time face detector of the pre-deep-learning era and was commonly applied to early lab benchmarks in subsequent FER studies (JAFFE~\citep{lyons1998japanese}, CK+~\citep{lucey2010extended}, RaFD~\citep{langner2010presentation}, Oulu-CASIA~\citep{zhao2011facial}), while the dlib HOG detector~\citep{king2009dlib} provides a CPU-friendly desktop baseline; early FAA surveys document their central role in pre-deep-learning FER~\citep{sariyanidi2014automatic}.

    \item \uline{Multi-task CNN cascade detectors}: MTCNN-style multi-task cascaded networks jointly predicting bounding boxes and anchor landmarks via a P-R-O three-stage CNN pipeline. MTCNN~\citep{zhang2016mtcnn} is the default detector in modern in-the-wild SFER/DFER alignment chains of RAF-DB~\citep{li2017reliable}, AffectNet~\citep{mollahosseini2017affectnet}, ExpW, and large-scale benchmarks (DFEW~\citep{jiang2020dfew}, FERV39k~\citep{wang2022ferv39k}, Aff-Wild2~\citep{kollias2018aff,kollias2023abaw}); its five-point landmark output is routinely reused for affine alignment downstream.

    \item \uline{Single-shot dense / anchor-based deep detectors}: Single-shot dense-prediction deep architectures combining bounding-box regression with anchor-landmark or 3D-aware heads on multi-scale feature pyramids. RetinaFace~\citep{deng2020retinaface} introduces single-shot multi-level localization with self-supervised mesh regression, providing high recall on extreme-pose, low-resolution, and occluded data; such detectors are increasingly used in robust in-the-wild FER and AU pipelines for ABAW~\citep{kollias2022abaw,kollias2023abaw,kollias20247th} and unified face-analysis frameworks~\citep{narayan2025facexformer,qin2024faceptor}.

    \item \uline{Compact and anchor-free deep detectors}: Methods targeting small computational budgets or anchor-free/Transformer-style detection. EmotiEffNets~\citep{savchenko2023emotieffnets} and the multi-task models of Savchenko et al.~\citep{savchenko2024leveraging} combine an off-the-shelf face detector (RetinaFace, with MediaPipe as fallback, in~\citep{savchenko2024leveraging}) with compact MobileFaceNet/MobileViT-style backbones; LibreFace~\citep{chang2024libreface} packages face detection and alignment together with its AU/FER models in a deployable toolkit (Section~\ref{ssec:learn_kd_peft}); and anchor-free/detection Transformer (DETR)-style detectors complement this line.
\end{itemize}

\subsection{Facial landmark detection and geometric alignment}
\label{ssec:prep_landmark}

Facial landmark detection and geometric alignment register the detected face to a canonical frame, removing rigid head-pose, translation, and scale variability so that downstream networks can focus on intrinsic muscular activation rather than extrinsic motion~\citep{li2020deep,wu2019facial}. This stage is critical to AU detection, DFER, and MER, where misalignment errors propagate directly to recognition, and its landmark output (sparse 2D, dense 2D heatmap, 3D-aware sparse, or dense 3D/UV-map) is routinely reused as an auxiliary geometric stream in landmark-augmented FER backbones.

Depending on the landmark dimensionality and density, landmark-based FER pre-processing pipelines can be divided into the following main categories:
\begin{itemize}
    \item \uline{Sparse 2D landmark methods}: Methods adopting a sparse set of 5--68 2D anatomical landmarks regressed by fast trees or shallow networks. The dlib ensemble-of-regression-trees 68-point estimator~\citep{kazemi2014one,king2009dlib} and MTCNN's 5-point output~\citep{zhang2016mtcnn} are extensively used in lab benchmarks (CK+~\citep{lucey2010extended}, JAFFE~\citep{lyons1998japanese}, RaFD~\citep{langner2010presentation}, MMI, Oulu-CASIA~\citep{zhao2011facial}) and as a lightweight alignment stage in compressed FER models (EAN~\citep{kong2022real}, LFNSB~\citep{chen2024lightweight}, EmotiEffNets~\citep{savchenko2023emotieffnets}), trading expressiveness under occlusion/pose for CPU-friendly speed.

    \item \uline{Dense 2D heatmap-based landmark methods}: Methods adopting deep CNN/Transformer landmark detectors regressing per-landmark heatmaps over denser 68/98/106-point templates, providing higher accuracy under occlusion and profile views. FAN~\citep{bulat2017far} and PFLD~\citep{guo2019pfld} are the de facto baselines, while landmark features from such detectors (e.g., MobileFaceNet-based) are reused as auxiliary geometric streams or prompts in POSTER~\citep{zheng2023poster}, POSTER++~\citep{mao2025posterpp}, and S2D~\citep{chen2024static}, and landmark detection is learned jointly with expression recognition in CMCNN~\citep{yu2022co} and MOL~\citep{shao2025mol}; landmark-detection surveys~\citep{wu2019facial} review the holistic, constrained local model-based, and regression-based detectors underlying these pipelines.

    \item \uline{3D-aware sparse landmark methods}: Methods lifting the sparse landmark set into a 3D-aware representation for extreme head poses and 3D/4D mesh inputs. 3DDFA~\citep{zhu2016face} introduces CNN-based 3D-aware face alignment across large poses via 3DMM coefficients yielding sparse 3D landmarks, widely used as the alignment backbone of pose-invariant SFER~\citep{jiang2022disentangling}.

    \item \uline{Dense 3D and UV-map alignment methods}: Methods establishing fully dense per-pixel correspondence between the input face and a canonical 3D template, via UV position maps or 3D morphable models. PRNet~\citep{feng2018joint} jointly performs 3D reconstruction and dense alignment via UV position maps; 3DMM fitting~\citep{ploumpis2021head} provides shape/expression disentanglement; registered 3D scans are rendered into 2D geometry and texture maps for multi-modal 2D+3D FER~\citep{guo20203dfacial}, and 2.5D depth-map representations developed for 3D face identification~\citep{li2017deep3d} can be reused for expression analysis; and dense correspondence further underpins 4D FER~\citep{behzad2021sparsity,behzad2021magnifying} and 4D MER~\citep{wang2024spontaneous4D}.
\end{itemize}

\subsection{Illumination and photometric normalization}
\label{ssec:prep_photo}

Illumination and photometric normalization aim to suppress nuisance variation due to lighting, shadows, sensor noise, and color-channel imbalance, which would otherwise be encoded by deep FER backbones as spurious expression cues, especially under in-the-wild and cross-cultural settings~\citep{sariyanidi2014automatic}. This stage is particularly important for AU detection, MER, and cross-corpus FER, and is essential to NIR and thermal pipelines targeting low-light, nighttime, and outdoor deployments, where the photometric signal carries most of the expression-discriminative evidence.

Illumination/photometric pre-processing pipelines can be grouped by the normalization mechanism into the following categories:
\begin{itemize}
    \item \uline{Global histogram-based methods}: Methods applying per-channel histogram equalization, mean-variance normalization, or gamma correction to the entire facial crop. These are the de facto baseline in lab benchmarks~\citep{lucey2010extended,zhao2011facial,langner2010presentation}, remaining widely used as a lightweight normalization stage upstream of CNN/Transformer/hybrid FER backbones.

    \item \uline{Local adaptive methods}: Methods applying locally adaptive contrast enhancement such as contrast-limited adaptive histogram equalization (CLAHE)~\citep{pizer1987adaptive,zuiderveld1994clahe} to recover local texture detail in shadowed or over- or under-exposed face regions. Particularly relevant for AU detection and MER, where local intensity changes carry expression-discriminative information, they are incorporated into NIR and thermal FER pipelines~\citep{zhao2011facial,bhattacharyya2021deep}.

    \item \uline{Retinex and illumination-invariant filtering methods}: Methods decomposing the image into reflectance and illumination components through retinex-style filtering~\citep{land1971lightness}, suppressing the illumination component and retaining the reflectance signal. Such filtering is complementary to active NIR imaging, which reduces the dependence on ambient illumination already at acquisition time~\citep{zhao2011facial}.

    \item \uline{Learned deterministic photometric normalization methods}: Methods replacing fixed photometric operators with deterministic learned mappings applied identically at training and inference time (learned tone mapping, color constancy/white balance, cross-domain photometric translation). In the reviewed works, such operators are rarely isolated or ablated, so their contribution to cross-corpus and cross-spectral FER has not yet been quantified.
\end{itemize}

\subsection{Spatial normalization and data augmentation}
\label{ssec:prep_augment}

Spatial normalization and data augmentation convert the aligned, photometrically normalized face into a fixed-resolution, model-ready tensor, while exposing the network to deterministic spatial standardization (resize, AU-region-anchored cropping) and stochastic geometric, photometric, occlusion-based, sample-mixing, and generative variation~\citep{li2020deep,porcu2020evaluation}. This stage is a widely used regularizer for in-the-wild FER pipelines, mitigating overfitting, class imbalance, occlusion, and demographic bias on small lab and long-tailed in-the-wild benchmarks.

With respect to the spatial-variation handling paradigm, spatial-normalization and augmentation FER pre-processing pipelines fall into the following categories:
\begin{itemize}
    \item \uline{Deterministic spatial normalization methods}: Methods performing a deterministic spatial mapping from the aligned face to the model-ready tensor, either by resizing to a canonical resolution (typically $112\!\times\!112$, $224\!\times\!224$, $256\!\times\!256$) or by cropping AU-anchored local regions. AU-region-anchored cropping is commonly adopted in region-aware FER and AU detection, where DRML~\citep{zhao2016deep} learns AU-specific local filters on landmark-derived regions; EAC-Net~\citep{li2017eacnet} crops AU-centered regions around facial landmarks; and FRL-DGT~\citep{zhai2023feature} operates on the displacements of landmark-anchored facial regions (Section~\ref{ssec:arch_hybrid}).

    \item \uline{Stochastic geometric and photometric augmentation methods}: Methods applying hand-designed or learned stochastic geometric (random crop/flip/rotation/scaling) and photometric (color/brightness/contrast jitter, blur) transformations on aligned crops. AutoAugment~\citep{cubuk2019autoaugment} and RandAugment~\citep{cubuk2020randaugment} provide the de facto policy-search baselines, routinely adopted for in-the-wild SFER (POSTER~\citep{zheng2023poster}, POSTER++~\citep{mao2025posterpp}, MHAN~\citep{wang2026mhan}, MFER~\citep{xu2024multiscale}) and in compressed FER models (EAN~\citep{kong2022real}, EmotiEffNets~\citep{savchenko2023emotieffnets,savchenko2024leveraging}); fairness studies~\citep{xu2020investigating} evaluate standard augmentation together with attribute-aware and disentangled models with respect to demographic bias.

    \item \uline{Occlusion-based regularization methods}: Methods simulating occlusion through random masking, patch erasing, or Cutout-style operators. Random Erasing~\citep{zhong2020random}, Cutout~\citep{devries2017improved}, and EAC~\citep{zhang2022learn} (random erasing with flip-attention consistency) improve robustness to facial occlusion and partial-face scenarios, and are particularly relevant for AU detection on DISFA~\citep{mavadati2013disfa} and BP4D~\citep{zhang2014bp4d}.

    \item \uline{Sample-mixing augmentation methods}: Methods mixing pairs of training samples through linear or patch-level interpolation. MixUp~\citep{zhang2018mixup} linearly interpolates images/labels, while CutMix~\citep{yun2019cutmix} replaces a rectangular patch of one image with another, both used as a regularizer in modern FER pipelines under class imbalance and label noise (SCN~\citep{wang2020suppressing}, EAC~\citep{zhang2022learn}, EACM~\citep{li2025enhanced}, noisy-label FER~\citep{le2023uncertainty,ma2023transformer,min2026robust}).

    \item \uline{Generative augmentation methods}: Methods using GAN- or diffusion-based generative models to synthesize expressive faces, balance long-tailed distributions, or enforce identity invariance. Triple-BigGAN~\citep{gangwar2023triple} synthesizes in-the-wild expressive samples; IPD-FER~\citep{jiang2022disentangling} uses its generative disentanglement branch for identity-preserving augmentation (Section~\ref{ssec:arch_generative}); 4DFM~\citep{zou2024fourdfm} introduces diffusion-based 4D mesh-level expression synthesis; and diffusion-based generative models~\citep{preechakul2022diffusion,zou2024fourdfm} offer identity-preserving, attribute-controllable synthesis that could be used to generate class-balanced samples, although such class balancing has not yet been evaluated in the reviewed works.
\end{itemize}

\subsection{Temporal sampling, apex selection, and motion magnification}
\label{ssec:prep_temporal}

Temporal sampling, apex selection, and motion magnification operate on dynamic FER inputs (RGB videos, 4D meshes, audio-visual streams), isolating the most informative temporal evidence for downstream networks, while keeping memory and computational budgets bounded~\citep{li2022deep,kollias2024abaw6}. This stage is a widely adopted temporal pre-processing step in DFER and MER, directly exposing the kinematic onset--apex--offset signature on which the recognition of subtle, blended, or low-intensity expressions relies, and producing fixed-length sequences, apex/onset--offset clips, or explicit motion descriptors (optical flow, strain, magnified frames) consumed by the recognition backbone.

Depending on the temporal mechanism, temporal-sampling FER pre-processing pipelines can be divided into the following main categories:
\begin{itemize}
    \item \uline{Uniform and random clip-sampling methods}: Methods uniformly or randomly sampling a fixed number of frames from the input clip. Temporal segment network (TSN)-style uniform-segment sampling is the de facto in-the-wild DFER baseline, adopted in DFEW~\citep{jiang2020dfew}, FERV39k~\citep{wang2022ferv39k}, MAFW, M3DFEL~\citep{wang2023rethinking}, EST~\citep{liu2023expression}, S2D~\citep{chen2024static}, and HDF~\citep{cui2025learning}.

    \item \uline{Apex- and onset--offset-selection methods}: Methods estimating the apex frame or onset--apex--offset triplet and feeding only this evidence to the classifier, especially relevant for MER. Liong et al.~\citep{liong2018less} formalize ``less is more'' MER from a single apex frame; LTR3O~\citep{zhu2025learning} learns to rank onset-occurring-offset structures; HTNet~\citep{wang2024htnet} relies on apex-frame evidence (Section~\ref{ssec:arch_transformer}); SODA4MER~\citep{zhang2025dynamic} operates on the whole onset--offset sequence and detects the apex frame itself, without apex annotations (Section~\ref{ssec:learn_selfsupervised}); MERASTC~\citep{gupta2021merastc} encodes compact apex-centered motion-aware cues; and the temporal interpolation model of Pfister et al.~\citep{huang2017beyond} provides a standard interpolation/sampling protocol for short MER clips.

    \item \uline{Optical-flow and strain-based motion-extraction methods}: Methods explicitly extracting motion descriptors (total variation--L1 (TV-L1) optical flow, optical strain, displacement) as a complementary or primary input. TV-L1 optical flow~\citep{zach2007duality} is the canonical estimator adopted by MER pipelines~\citep{gupta2021merastc,verma2021automer,wang2023temporal,zhi2022micro}, whereas other pipelines derive optical strain from differential flow estimates~\citep{liong2014subtle} or rely on Farneb{\"a}ck dense flow~\citep{allaert2019micro}; OFR~\citep{poux2021dynamic} operates on optical-flow input (Section~\ref{ssec:mod_dfer}); and MOL~\citep{shao2025mol} estimates optical flow jointly with the recognition task, instead of pre-computing it (Section~\ref{ssec:arch_hybrid}).

    \item \uline{Eulerian and phase-based motion-magnification methods}: Methods amplifying low-amplitude facial motion through Eulerian video magnification~\citep{wu2012eulerian} or phase-based magnification~\citep{wadhwa2013phase}, particularly relevant for MER, intensity estimation, and 4D MER. Eulerian magnification is incorporated into SE-DenseNet~\citep{cai2022micro}, in the non-deep 4D FER pipeline of Zhen et al.~\citep{behzad2021magnifying}, whereas handcrafted spatio-temporal descriptors such as STCLQP~\citep{huang2018spontaneous} encode subtle motion directly, through quantized local sign, magnitude, and orientation patterns, without magnification.
\end{itemize}

\subsection{Modality-specific pre-processing}
\label{ssec:prep_modality}

Modality-specific pre-processing addresses the distinctive normalization requirements of non-RGB FER input modalities (3D/4D geometric and depth data, thermal/NIR imagery, audio-visual streams, and physiological signals) that cannot be reduced to standard 2D RGB face pipelines~\citep{sariyanidi2014automatic,kollias2024abaw6}. This stage is essential to multi-modal in-the-wild FER, aligning geometric, spectral, acoustic, and autonomic streams with the visible-light face and unlocking robustness to low-light, expression suppression, and clinical/automotive deployment constraints, by producing cleaned meshes, calibrated thermograms, mel-spectrograms or self-supervised audio embeddings, and filtered bio-signal segments consumed by downstream multi-modal fusion.

Modality-specific FER pre-processing pipelines can be grouped by the signal domain being normalized into the following categories:
\begin{itemize}
    \item \uline{Geometric-domain pre-processing methods}: Methods operating in the 3D geometric domain to normalize depth maps, meshes, or point clouds for 3D/4D FER backbones. Standard pipelines combine point-cloud completion and consistency learning (developed for 3D face recognition)~\citep{liu2024pointcloud}, 3DMM fitting and shape/expression disentanglement~\citep{ploumpis2021head}, rendering of 2D geometry and texture maps for multi-modal 2D+3D fusion~\citep{guo20203dfacial}, 2.5D depth-map normalization adopted from 3D face identification~\citep{li2017deep3d}, and dense UV-map alignment with PRNet~\citep{feng2018joint}. Benchmarks such as BU-3DFE~\citep{yin20063d}, BP4D~\citep{zhang2014bp4d}, CAS(ME)$^3$~\citep{li2022cas}, 4DME~\citep{li20224dme}, and the 4D spontaneous MER database~\citep{wang2024spontaneous4D} define canonical protocols (mesh downsampling, depth gap filling, vertex correspondence).

    \item \uline{Spectral-domain pre-processing methods}: Methods operating outside the visible RGB spectrum to calibrate thermal/NIR/multi/hyper-spectral streams (sensor-drift correction, intensity calibration, low-light histogram stretching) and align them with the visible stream. In visible--IR early fusion, the two streams are encoded by attention-augmented ResNet-18 branches and fused at the feature level~\citep{khan2025visir}; early NIR-FER~\citep{zhao2011facial} relies on LBP-style normalization for Oulu-CASIA NIR; and IRFacExNet~\citep{bhattacharyya2021deep}, and CTIFERK~\citep{wang2025ctiferk} integrate spectral-domain normalization with deep CNN and MobileViT/KAN backbones.

    \item \uline{Spectro-temporal acoustic pre-processing methods}: Methods operating in the acoustic spectro-temporal domain, converting audio waveforms into time-frequency representations or self-supervised audio embeddings. Mel-spectrogram/mel-frequency cepstral coefficient (MFCC) representations~\citep{davis1980comparison,hershey2017cnn,mcfee2015librosa} provide standard input to CNN/Transformer audio encoders, while HuBERT/Wav2Vec-based self-supervised speech representations~\citep{cheng2024emotionllama,hsu2021hubert} are frequently adopted in recent in-the-wild audio-visual FER methods (e.g., Emotion-LLaMA~\citep{cheng2024emotionllama}), while MMA-DFER~\citep{chumachenko2024mma} adopts a masked-autoencoder audio encoder (AudioMAE).

    \item \uline{Autonomic-temporal pre-processing methods}: Methods operating on continuous bio-signals (EEG, ECG, EDA/GSR, PPG, EOG, respiration) for multi-modal physiological FER. Standard pipelines (band-pass filtering, artifact removal, sampling alignment, windowing) follow EEG pre-processing surveys~\citep{chaddad2023eeg} and toolboxes such as NeuroKit2~\citep{makowski2021neurokit}, and are adopted in rPPG-fused FER~\citep{ouzar2022video}, EEG-fused continuous-affect FER (RMMT~\citep{jin2024residual}), and multi-bio-signal hierarchical fusion (GBV-Net~\citep{yu2025gbvnet}).
\end{itemize}

\begin{table*}[!t]
  \caption{Pre-processing pipelines: Comparative analysis and key insights.}
  \label{tab:preprocessing_summary}
  \centering
  \scriptsize

  \setlength{\aboverulesep}{0pt}
  \setlength{\belowrulesep}{0pt}
  
  \setlength{\tabcolsep}{4pt}
  
  \renewcommand{\arraystretch}{1.3}

  \rowcolors{2}{gray!20}{gray!2}

  \newlength{\Waspectp}\setlength{\Waspectp}{1.5cm}
  \newlength{\Wfd}\setlength{\Wfd}{2.4cm}
  \newlength{\Wla}\setlength{\Wla}{2.4cm}
  \newlength{\Wph}\setlength{\Wph}{2.4cm}
  \newlength{\Wau}\setlength{\Wau}{2.4cm}
  \newlength{\Wts}\setlength{\Wts}{2.4cm}
  \newlength{\Wms}\setlength{\Wms}{2.4cm}

  \resizebox{\textwidth}{!}{%
  \begin{tabular}{@{}|
    >{\raggedright\arraybackslash}Q{\Waspectp}|
    >{\raggedright\arraybackslash}Q{\Wfd}|
    >{\raggedright\arraybackslash}Q{\Wla}|
    >{\raggedright\arraybackslash}Q{\Wph}|
    >{\raggedright\arraybackslash}Q{\Wau}|
    >{\raggedright\arraybackslash}Q{\Wts}|
    >{\raggedright\arraybackslash}Q{\Wms}|
  @{}}
    \toprule
    \rowcolor{gray!50}
    \headerbreak{Pipeline} &
    \headerbreak{Face detection\\\& localization} &
    \headerbreak{Landmark detection\\\& alignment} &
    \headerbreak{Illumination/\\photometric} &
    \headerbreak{Spatial norm.\\\& augmentation} &
    \headerbreak{Temporal sampling\\\& magnification} &
    \headerbreak{Modality-specific\\pre-processing} \\
    \midrule

    Input-output &
    \begin{tabitem}
      \item In: Raw RGB\slash IR frame or RGB-D map
      \item Out: Face bounding box(es) + optional anchor landmarks\slash confidence scores
    \end{tabitem} &
    \begin{tabitem}
      \item In: Detected face crop
      \item Out: Sparse\slash dense 2D or 3D landmark coordinates + canonical-alignment transform
    \end{tabitem} &
    \begin{tabitem}
      \item In: Aligned face crop (RGB\slash NIR\slash thermal)
      \item Out: Illumination-normalized, photometrically standardized crop
    \end{tabitem} &
    \begin{tabitem}
      \item In: Aligned face crop
      \item Out: Fixed-resolution model-ready tensor (resized\slash cropped, possibly jittered\slash erased\slash mixed/\allowbreak synthesized)
    \end{tabitem} &
    \begin{tabitem}
      \item In: Long video clip (or 4D mesh sequence)
      \item Out: Fixed-length frame sequence, apex\slash onset--offset clip, or motion descriptor (flow\slash strain\slash magni\-fied frames)
    \end{tabitem} &
    \begin{tabitem}
      \item In: Raw non-RGB signal (depth\slash mesh\slash point cloud, thermal\slash NIR, audio waveform, bio-signal trace)
      \item Out: Aligned, model-ready modality-specific representation (cleaned mesh, calibrated thermogram, mel-spectrogram\slash self-supervised learning (SSL) embedding, filtered bio-signal segment)
    \end{tabitem} \\
    \midrule

    Nuisance factor addressed &
    \begin{tabitem}
      \item Spatial localization
      \item Background clutter and multi-face scenes
      \item Scale and in-plane rotation
    \end{tabitem} &
    \begin{tabitem}
      \item Rigid head pose and 3D rotation
      \item Translation and scale residuals
      \item Bounding-box convention drift across detectors
    \end{tabitem} &
    \begin{tabitem}
      \item Global illumination and shadow
      \item Local contrast loss
      \item Color cast and sensor noise
    \end{tabitem} &
    \begin{tabitem}
      \item Overfitting and limited data
      \item Occlusion and partial face views
      \item Class imbalance and demographic bias
    \end{tabitem} &
    \begin{tabitem}
      \item Temporal redundancy
      \item Non-expressive frames
      \item Weak frame-level supervision
      \item Sub-perceptible motion amplitude
    \end{tabitem} &
    \begin{tabitem}
      \item Signal-form heterogeneity (geometric\slash spectral/\allowbreak acoustic\slash autonomic)
      \item Sensor drift and modality-specific noise
      \item Cross-modal asynchrony with RGB
    \end{tabitem} \\
    \midrule

    Main mechanisms &
    \begin{tabitem}
      \item Handcrafted-feature cascades
      \item Multi-task CNN cascades
      \item Single-shot dense\slash anchor-based detectors
      \item Compact\slash anchor-free detectors
    \end{tabitem} &
    \begin{tabitem}
      \item Sparse 2D landmarks
      \item Dense 2D heatmap landmarks
      \item 3D-aware sparse landmarks
      \item Dense 3D\slash UV-map alignment
    \end{tabitem} &
    \begin{tabitem}
      \item Global histogram equalization
      \item CLAHE\slash local adaptive
      \item Retinex\slash illumina\-tion-invariant
      \item Learned photometric normalization
    \end{tabitem} &
    \begin{tabitem}
      \item Deterministic resize\slash AU-anchored crop
      \item Stochastic geometric\slash photometric jitter
      \item Occlusion-based erasing
      \item MixUp\slash CutMix mixing
      \item GAN\slash diffusion synthesis
    \end{tabitem} &
    \begin{tabitem}
      \item Uniform\slash random clip sampling
      \item Apex\slash onset--offset selection
      \item Optical-flow\slash strain extraction
      \item Eulerian\slash phase magnification
    \end{tabitem} &
    \begin{tabitem}
      \item Geometric (3D\slash depth/\allowbreak mesh)
      \item Spectral (thermal\slash NIR)
      \item Spectro-temporal acoustic
      \item Autonomic-temporal bio-signal
    \end{tabitem} \\
    \midrule

    Strengths &
    \begin{tabitem}
      \item Robust to scale\slash rotation\slash occlu\-sion
      \item Reusable anchor landmarks
      \item Mature open-source toolchains
    \end{tabitem} &
    \begin{tabitem}
      \item Removes rigid head pose
      \item Reusable as auxiliary stream
      \item Supports 3D-aware FER
    \end{tabitem} &
    \begin{tabitem}
      \item Reduces lighting\slash shadow bias
      \item Targets cross-corpus transfer
      \item Marginal computational cost
    \end{tabitem} &
    \begin{tabitem}
      \item Strong regularization
      \item Robustness to occlusion\slash pose
      \item Mitigates class imbalance
    \end{tabitem} &
    \begin{tabitem}
      \item Exposes kinematic signature
      \item Reduces input redundancy
      \item Enables MER and VA
    \end{tabitem} &
    \begin{tabitem}
      \item Exploits non-visible cues
      \item Robust under low light\slash suppression
      \item Aligns multi-modal streams
    \end{tabitem} \\
    \midrule

    Limitations &
    \begin{tabitem}
      \item Fails under extreme blur\slash occlusion
      \item Inherits demographic bias
      \item Cost on edge devices
    \end{tabitem} &
    \begin{tabitem}
      \item Misalignment under occlusion
      \item Identity\slash demogra\-phic bias
      \item Cost of dense\slash 3D estimators
    \end{tabitem} &
    \begin{tabitem}
      \item May erase MER-relevant cues
      \item Hand-tuned hyperparameters
      \item Limited under severe color shift
    \end{tabitem} &
    \begin{tabitem}
      \item Distorts AU-specific cues
      \item Hyperparameter sensitivity
      \item Identity leakage in GAN\slash DM
    \end{tabitem} &
    \begin{tabitem}
      \item Fragile apex detection
      \item Discards long-clip evidence
      \item Amplifies head motion\slash noise
    \end{tabitem} &
    \begin{tabitem}
      \item Specialized hardware
      \item Small benchmarks
      \item Subject-specific calibration
    \end{tabitem} \\
    \midrule

    Indicative methods &
    \begin{tabitem}
      \item Viola--Jones \citep{viola2004robust}, dlib HOG \citep{king2009dlib}, MTCNN \citep{zhang2016mtcnn}, RetinaFace \citep{deng2020retinaface}, EmotiEffNets \citep{savchenko2023emotieffnets}
    \end{tabitem} &
    \begin{tabitem}
      \item dlib 68-pt \citep{kazemi2014one}, FAN \citep{bulat2017far}, PFLD \citep{guo2019pfld}, 3DDFA \citep{zhu2016face}, PRNet \citep{feng2018joint}
    \end{tabitem} &
    \begin{tabitem}
      \item Sariyanidi et al. \citep{sariyanidi2014automatic}, CLAHE \citep{zuiderveld1994clahe}, retinex \citep{land1971lightness}
    \end{tabitem} &
    \begin{tabitem}
      \item DRML \citep{zhao2016deep}, EAC-Net \citep{li2017eacnet}, AutoAugment \citep{cubuk2019autoaugment}, CutMix \citep{yun2019cutmix}, Random Erasing \citep{zhong2020random}, EAC \citep{zhang2022learn}, IPD-FER \citep{jiang2022disentangling}, Porcu et al. \citep{porcu2020evaluation}
    \end{tabitem} &
    \begin{tabitem}
      \item Liong et al. \citep{liong2018less}, HTNet \citep{wang2024htnet}, LTR3O \citep{zhu2025learning}, TV-L1 \citep{zach2007duality}, Eulerian magnification \citep{wu2012eulerian}
    \end{tabitem} &
    \begin{tabitem}
      \item PRNet \citep{feng2018joint}, CTIFERK \citep{wang2025ctiferk}, HuBERT-style audio \citep{cheng2024emotionllama}, NeuroKit2 \citep{makowski2021neurokit}
    \end{tabitem} \\
    \bottomrule
  \end{tabular}%
  }
\end{table*}

\subsection{Comparative analysis and key insights}
\label{ssec:preprocessing_discussion}

Building on the detailed discussion of the pre-processing pipelines (Sections~\ref{ssec:prep_facedet}--\ref{ssec:prep_modality}), this subsection compares them and provides critical insights for each pipeline type. Table~\ref{tab:preprocessing_summary} summarizes, for each pre-processing pipeline, its a) Input--output signal transformation, b) Nuisance factor addressed, c) Main mechanisms, d) Key strengths, e) Critical limitations, and f) Indicative methods. The analysis indicates that pre-processing is best understood as a chain of signal transformations in which the outputs of upstream pipelines serve as the inputs of downstream ones, so errors of an upstream stage (e.g., a missed or misaligned face) propagate to all subsequent stages. Additionally, the pipelines tackle largely distinct nuisance factors (spatial localization, rigid head pose, photometric variation, overfitting and class imbalance, temporal redundancy, and signal-form heterogeneity), which is consistent with the fact that in-the-wild FER pipelines typically combine several of these stages; however, since pre-processing is rarely ablated in the reviewed works, the contribution of each individual stage to the reported results is seldom quantified (the input resolution and clip sampling adopted by the methods of Section~\ref{sec:performance_metrics} are reported in Tables~\ref{tab:perf_sfer}--\ref{tab:perf_intensity}). Moreover, a trend towards deep, learned operators is observed in the reviewed works across virtually all pipeline types, with MTCNN/RetinaFace detectors~\citep{zhang2016mtcnn,deng2020retinaface}, FAN/3DDFA/PRNet alignment~\citep{bulat2017far,zhu2016face,feng2018joint}, learned augmentation policies~\citep{cubuk2019autoaugment,cubuk2020randaugment}, learned motion estimation~\citep{shao2025mol}, and self-supervised modality-specific encoders~\citep{cheng2024emotionllama,hsu2021hubert} progressively replacing handcrafted predecessors. Furthermore, pre-processing is itself the locus of several open problems, including demographic and cultural biases entering as early as face detection and landmark estimation, photometric and augmentation operators that may erase the subtle textural cues critical for AU detection and MER, and modality-specific pipelines bottlenecked by hardware, calibration, and small-benchmark constraints. Finally, an increasing entanglement between pre-processing and learning is observed through attentive patch and token pooling that discards uninformative facial regions inside the backbone~\citep{xue2022vision}, identity-preserving generative synthesis~\citep{preechakul2022diffusion,zou2024fourdfm}, and unified face-analysis backbones (e.g., FaceXFormer~\citep{narayan2025facexformer}, Faceptor~\citep{qin2024faceptor}) that are starting to absorb the alignment and landmark estimation steps into the backbone itself.

\section{Neural network architectures}
\label{sec:nn_architectures}

FER methods can be grouped by the neural network architecture they employ, which largely determines their representational capacity, efficiency, and inductive biases. In particular, the main types of neural network architectures used in deep learning-based FER methods are: a) Convolutional neural networks (CNNs), b) Recurrent neural networks (RNNs/LSTMs), c) Transformers, d) Graph neural networks (GNNs), e) Generative and diffusion models, f) Vision--language and multi-modal large language models (VLMs/MLLMs), and g) Hybrid architectures, as discussed in Section~\ref{sec:taxonomy_criteria} and further detailed below.

\subsection{Convolutional Neural Networks (CNNs)}
\label{ssec:arch_cnn}

CNNs constitute the historically dominant architectural family for FER, learning hierarchical spatial features through stacks of learnable convolutional, normalization, and pooling layers applied to aligned facial crops~\citep{lecun2002gradient,li2020deep}. As mature, locality-aware backbones with a rich pre-training ecosystem (ResNet, EfficientNet, MobileNet, DenseNet), CNNs underpin the bulk of real-time and on-device FER deployments and remain the standard SFER backbone against which more recent Transformer-, GNN-, and hybrid-based designs are typically compared.

With respect to how they extend the convolutional baseline, CNN-based FER systems fall into the following categories:
\begin{itemize}
    \item \uline{Plain deep CNN backbone methods}: Methods adopting off-the-shelf deep CNN backbones (ResNet, VGG, DenseNet) with standard supervised cross-entropy. CNN classifiers from the FER-2013 challenge~\citep{goodfellow2013challenges} established the baseline for early in-the-wild SFER, while BReG-NeXt~\citep{hasani2020breg} redesigns the residual unit with bounded-gradient activations for shallower, more stable affective CNNs.
    \item \uline{Multi-scale and region-aware CNN extensions}: Methods aggregating multi-resolution spatial information or operating on local AU-relevant regions. FLEPNet~\citep{karnati2022flepnet} employs parallel multi-scale blocks; MLCNN~\citep{nguyen2019facial} concatenates intermediate feature maps; SPWFA-SE~\citep{li2020facial} and MFER~\citep{xu2024multiscale} build local-global multi-scale aggregators; DRML~\citep{zhao2016deep} learns AU-specific local filters; and EAC-Net~\citep{li2017eacnet} combines enhancing and cropping branches around landmarks.
    \item \uline{Attention-enhanced CNN variants}: Methods injecting channel-, spatial-, or self-attention modules into CNN backbones. DSAN~\citep{fan2020facial} exploits deep supervision and demographic-aware attention; CERN~\citep{gera2022cern} uses squeeze-and-excitation over LightCNN layers; GLAMOR-Net~\citep{le2022global} attends to emotion-informative scene context; MHAN~\citep{wang2026mhan} adds multi-head global-local attention with multi-scale aggregation to a MobileFaceNet-based CNN; EAC~\citep{zhang2022learn} adds an attention-consistency constraint to a standard CNN classifier (Section~\ref{ssec:learn_uncertainty}); and SCN~\citep{wang2020suppressing} adds a self-attention importance-weighting module to a CNN backbone (Section~\ref{ssec:learn_uncertainty}).
    \item \uline{Compression-oriented CNN variants}: Methods targeting mobile/real-time deployment through compression, distillation, or efficient backbone design. EAN~\citep{kong2022real} uses iterative transfer learning and efficient attention; LFNSB~\citep{chen2024lightweight} introduces spatial-bias modules with cosine-harmony loss; LibreFace~\citep{chang2024libreface} uses compact CNN students for deployment (Section~\ref{ssec:learn_kd_peft}); and the EmotiEffNets family~\citep{savchenko2023emotieffnets,savchenko2024leveraging} trains compact MobileViT/MobileFaceNet/EfficientNet backbones in multi-task settings.
\end{itemize}

\subsection{Recurrent Neural Networks (RNNs/LSTMs)}
\label{ssec:arch_rnn}

RNNs and LSTMs explicitly model the temporal evolution of facial behavior by propagating a hidden state across time, capturing inter-frame dependencies in the neutral--onset--apex--offset cycle~\citep{hochreiter1997long,li2020deep}. Most commonly stacked on top of a CNN spatial backbone, they serve as sequence-level classifiers or regressors for DFER, dimensional VA, AU dynamics, and reaction-intensity estimation, natively supporting variable-length video and per-frame outputs with a low memory footprint.

Depending on the temporal-aggregation strategy, RNN/LSTM-based FER systems can be divided into the following main categories:
\begin{itemize}
    \item \uline{CNN--LSTM and CNN-RNN stacks}: Methods pairing a CNN spatial backbone with an LSTM/RNN temporal head for DFER and reaction-intensity estimation. Qian et al.~\citep{qian2023computer} compare CNN--LSTM and CNN--Transformer for ERI; CNN+RNN dynamic FER frameworks~\citep{manalu2024detection} improve robustness on challenging video datasets; and residual CNN--LSTM networks with audio-visual feature fusion~\citep{tzirakis2017end} extend the pattern to multi-modal affect.
    \item \uline{Sequence-level recurrent aggregators}: Methods aggregating clip-level evidence via LSTM-style memory and masked recurrent components. MMA-MRNNet~\citep{kollias2024mma} uses a masked recurrent network on top of Multiple-Models-of-Affect descriptors to handle variable-length videos for ERI prediction, while STC-NLSTM~\citep{yu2018stcnlstm} aggregates multi-level spatio-temporal convolutional features through a nested LSTM that jointly models appearance and temporal dynamics.
    \item \uline{Multi-task and multi-output recurrent heads}: Methods driving LSTM-based heads with multi-task descriptors. MTL-DAN~\citep{oh2023human} extracts expression/AU/VA representations and feeds them to an LSTM regression head for joint intensity estimation, while multi-order networks~\citep{tallec2022multi} offer a non-recurrent alternative for image-based AU detection, learning the order in which AU-specific modules are chained over shared CNN features.
    \item \uline{Recurrent designs for fine-grained MER}: RNN/LSTM-based methods specialized for micro-expression dynamics. Enriched long-term recurrent convolutional networks (ELRCN~\citep{khor2018enriched}) encode per-frame CNN features and model their temporal evolution through an LSTM, providing a learned temporal alternative to non-deep MER pipelines such as kernelized two-groups sparse learning (KTGSL~\citep{wei2022learning}), which selects discriminative features from two groups of handcrafted spatio-temporal descriptors.
\end{itemize}

\subsection{Transformers}
\label{ssec:arch_transformer}

Transformers comprise a rapidly adopted architectural family for in-the-wild FER, used, alone or within hybrid designs, by a growing share of the primary studies published since 2021 (Fig.~\ref{fig:bibliometrics}(d)), replacing locality-bound convolutions with multi-head self-attention over tokenized inputs and supporting long-range, global dependency modeling~\citep{vaswani2017attention,li2020deep}. They are used as ViT-style image encoders for SFER, as video and temporal-adapter encoders for DFER and MER, and as cross-fusion backbones for landmark, AU, and multi-modal inputs.

Transformer-based FER systems can be grouped by the token design and attention scope into the following categories:
\begin{itemize}
    \item \uline{ViT-based image Transformers}: Methods tokenizing aligned facial crops into non-overlapping patches and applying multi-head self-attention. TransFER~\citep{xue2021transfer} uses multi-attention dropping for relation-aware patches; VTFF~\citep{ma2021facial} fuses RGB and LBP through attentional selective fusion; APViT~\citep{xue2022vision} performs attentive patch/token selection; QWTR~\citep{zhou2025delving} incorporates quaternion/wavelet representations (TransFER, VTFF, APViT, and QWTR couple the Transformer with a CNN stem and are therefore counted as hybrids; Section~\ref{ssec:arch_hybrid}); and HLA-ViT~\citep{tian2024facial} complements ViT global attention with local attention (Section~\ref{ssec:arch_hybrid}).
    \item \uline{Cross-fusion and pyramid Transformers}: Methods coupling the RGB-image stream with auxiliary geometric or AU streams through cross-attention. POSTER~\citep{zheng2023poster} and POSTER++~\citep{mao2025posterpp} use pyramid cross-fusion Transformers where the landmark stream guides image tokens; as both build on CNN backbones, they are counted as hybrids (Section~\ref{ssec:arch_hybrid}).
    \item \uline{Video and temporal Transformers}: Methods extending Transformers to tokenized facial video sequences for DFER and MER. EST~\citep{liu2023expression} models intra- and inter-snippet dependencies; SLSTT~\citep{zhang2022short} applies a ViT to optical-flow inputs with an LSTM temporal aggregator, while C3DBed~\citep{pan2023c3dbed} combines 3D convolutions with a Transformer encoder; HTNet~\citep{wang2024htnet} adopts hierarchical local-region self-attention for MER; $\mu$-BERT~\citep{nguyen2023micron} adds micro-attention to a BERT-style Transformer for MER (Section~\ref{ssec:learn_selfsupervised}); Former-DFER~\citep{zhao2021former} couples a convolutional spatial Transformer with a temporal Transformer for in-the-wild DFER; S2D~\citep{chen2024static} couples a ViT image backbone with lightweight temporal adapters; and M3DFEL~\citep{wang2023rethinking} formulates DFER as multiple-instance learning over Transformer-tokenized snippets.
    \item \uline{Multi-modal Transformers}: Methods extending the Transformer backbone to multiple input streams (audio, text, AU, depth) through cross-attention or fusion bottlenecks. Zhang et al.~\citep{zhang2022transformer} integrate visual/audio/textual streams for joint AU and expression analysis; MMA-DFER~\citep{chumachenko2024mma} adapts pre-trained uni-modal audio and visual encoders through progressive prompts and fusion bottlenecks; AVT~\citep{jin2022avt} introduces an AU-assisted visual Transformer, which combines CNN local features, a Transformer encoder, and an AU fusion module; and FaceXFormer~\citep{narayan2025facexformer} and Faceptor~\citep{qin2024faceptor} use task-token Transformers for unified multi-task face analysis.
\end{itemize}

\subsection{Graph Neural Networks (GNNs)}
\label{ssec:arch_gnn}

GNNs explicitly model the face as a sparse topological graph over landmarks, AU-relevant regions, or spatio-temporal nodes, propagating information through learnable message-passing operators~\citep{kipf2016semi,li2020deep}. They are particularly suited to AU detection, MER, and dimensional/intensity estimation, where anatomical co-activation patterns and the spatial localization of expressions are essential, and they provide a naturally interpretable substrate for inter-region and inter-AU reasoning.

With respect to the graph-construction strategy, GNN-based FER systems fall into the following categories:
\begin{itemize}
    \item \uline{Landmark and region-graph methods}: Methods building the graph over facial landmarks or AU-anchored regions and learning geometry-aware features. Geo-GCN~\citep{zhao2021geometry} introduces geometry-aware attentive GCNs over landmark-derived facial graphs; HRL~\citep{han2022devil} performs landmark-guided graph message propagation; Chen et al.~\citep{chen2024dual} build the graph over samples rather than landmarks, casting intensity-invariant FER as GCN-based manifold learning; and GCF~\citep{kassab2024gcf} applies GCN layers to CNN-extracted features for SFER (and is therefore counted as a hybrid; Section~\ref{ssec:arch_hybrid}).
    \item \uline{AU-correlation graph methods}: Methods encoding the FACS co-activation/mutual-exclusion structure among AUs, in some cases through relation modules on CNN backbones rather than an explicit GNN. ARL~\citep{shao2019facial} models AU relationships through learnable correlation matrices; TS-AUCNN~\citep{sun2020dynamic} transfers AU-relation knowledge from a teacher network to an MER student (Section~\ref{ssec:learn_kd_peft}); and C-EXPR-NET~\citep{kollias2023multi} combines visual features with AU semantic information through cross-modality attention and distribution matching.
    \item \uline{Spatio-temporal graph methods}: Methods extending the facial graph along the temporal axis. ST-RDGCN~\citep{huang2025modeling} constructs dynamic space-time graphs over local regions for video-based FER; SpoT-GCN~\citep{deng2024multi} introduces multi-scale spatio-temporal GCNs for macro/micro-expression spotting; and MOL~\citep{shao2025mol} includes a graph-style convolution in a Transformer-based design (Section~\ref{ssec:arch_hybrid}).
    \item \uline{Cross-domain adversarial graph methods}: Methods using graph-based reasoning under domain shift. AGRA~\citep{chen2021cross} introduces adversarial graph learning for cross-domain FER through a unified evaluation benchmark, bridging lab/in-the-wild and cross-cultural gaps via graph-aligned domain adaptation, while GAT-ADA~\citep{ghaedi2025gatada} couples a batch-level graph-attention network with gradient-reversal adversarial alignment and correlation alignment (CORAL)/maximum mean discrepancy (MMD) statistical alignment for unsupervised cross-domain adaptation.
\end{itemize}

\subsection{Generative and diffusion models}
\label{ssec:arch_generative}

Generative architectures (GANs, variational autoencoders) and diffusion models (DMs) extend FER beyond discriminative classification, supporting expression synthesis, identity/pose disentanglement, data augmentation, and AU-conditioned image- and 4D-mesh editing~\citep{li2020deep,zou2024fourdfm}. They are used both as pre-processing/augmentation tools that mitigate class imbalance and long-tailed distributions on AffectNet, RAF-DB, and ExpW, and as conditional generators that explicitly model the underlying expression manifold, with growing relevance to 3D/4D and AU-aware FER.

Depending on the generative paradigm and conditioning signal, generative and diffusion FER systems can be divided into the following main categories:
\begin{itemize}
    \item \uline{Unconditional GAN augmentation methods}: Methods exploiting unconditional or weakly conditioned GANs to mitigate class imbalance and long-tailed FER distributions. Triple-BigGAN~\citep{gangwar2023triple} introduces semi-supervised GANs for joint synthesis and classification.
    \item \uline{Disentanglement-driven GAN methods}: Methods using adversarial or generative objectives to separate expression from identity and pose. IPD-FER~\citep{jiang2022disentangling} explicitly disentangles facial representations into identity/pose/expression components, while DeRL~\citep{yang2018derl} trains a conditional GAN to generate the neutral counterpart of an input face and mines the expressive component from the intermediate de-expression residues.
    \item \uline{Expression/AU-conditioned GAN or DM synthesis methods}: GAN/DM methods synthesizing expressive faces conditioned on a target expression, AU set, or intensity descriptor for cross-corpus augmentation, balancing, and AU-aware diagnostics. Adversarial pipelines for in-the-wild expression synthesis~\citep{gangwar2023triple} and AU-conditioned generators support both pre-training and downstream FER, while the AUD method~\citep{xu2024aud} adapts a latent diffusion model with an AU-aware fine-tuning strategy to synthesize and edit expressions from a single image conditioned on target AUs.
    \item \uline{3D/4D mesh-level diffusion methods}: Recent diffusion-based methods on 3D meshes, 4D sequences, or AU latent codes. The 4D Facial Expression Diffusion Model~\citep{zou2024fourdfm} introduces a denoising diffusion process for 4D mesh sequences conditioned on expression labels, supporting realistic 4D FER synthesis; EmoDiffusion~\citep{zhang2025emodiffusion} performs latent-space diffusion over disentangled upper-face and mouth blendshape regions with an emotion adapter to generate expressive 3D facial-mesh sequences; and related directions extend conditional diffusion to AU-aware image editing and identity-preserving expression transfer.
\end{itemize}

\subsection{Vision--language and multi-modal LLMs (VLMs/MLLMs)}
\label{ssec:arch_vlm}

VLMs and MLLMs introduce language as a first-class architectural component of the FER pipeline, coupling visual encoders with text encoders or large language model backbones and supporting prompt-based and instruction-tuned reasoning~\citep{radford2021clip,cheng2024emotionllama}. They enable open-vocabulary recognition, in a few cases zero-shot FER (Section~\ref{ssec:learn_weak_fewshot}), and verbalized affective reasoning, and they are increasingly adopted for in-the-wild FER, directly exploiting large-scale image--text and instruction-tuning pre-training. Whereas Section~\ref{ssec:mod_vlm} characterizes the textual information that these models receive, this subsection addresses how the visual and the textual streams are composed architecturally.

VLM/MLLM-based FER systems can be grouped by the backbone-composition pattern into the following categories:
\begin{itemize}
    \item \uline{CLIP-based dual-encoder methods}: Methods classifying the visual embedding by its similarity to the text-encoder embeddings of the class prompts in a shared image--text space. DFER-CLIP~\citep{zhao2023prompting} fine-tunes the CLIP image encoder, adds a Transformer temporal model and a learnable class token trained from scratch, and keeps the CLIP text encoder frozen; PE-CLIP~\citep{saadi2025peclip} keeps both CLIP ViT-B/16 encoders frozen and inserts a gated recurrent unit (GRU)-based temporal adapter and shared adapters into them (about 9M trainable parameters); MER-CLIP~\citep{liu2025mer} fine-tunes a UniFormerV2 video encoder initialized from CLIP ViT-B/16, together with a frozen CLIP text encoder and a Transformer emotion head; and FER-Former~\citep{li2024fer} aligns a hybrid IR-50 CNN--Transformer visual branch with a fine-tuned CLIP text encoder (Section~\ref{ssec:arch_hybrid}).
    \item \uline{Prompt-tuning and prompt-learning methods}: Methods introducing learnable soft prompts, LLM-generated hard prompts, or emoji-based visual prompts on frozen or partially tuned VLM backbones, at the input of the text encoder, of the visual encoder, or of both. MPA-FER~\citep{ma2025multimodal} aligns learnable soft prompts with LLM-generated hard prompts and prototype-guided visual alignment; MMPL-FER~\citep{pei2025multi} couples LLM descriptions with emoji-based visual prompts through symmetrical cross-attention; and TG-DFER~\citep{jung2025text} introduces text-guided prompts and multi-grained temporal modeling; PE-CLIP~\citep{saadi2025peclip} additionally learns multi-modal prompts in both CLIP encoders.
    \item \uline{MLLM-based instruction-tuned methods}: Methods building on instruction-tuned MLLMs for unified affective reasoning over RGB images, video, audio, and text. Emotion-LLaMA~\citep{cheng2024emotionllama} couples HuBERT audio with multi-view visual encoders and a LLaMA backbone for multi-modal emotion recognition and reasoning; EMO-LLaMA~\citep{xing2024emollama} and EmoVerse~\citep{li2025emoverse} extend this paradigm with multi-stage instruction tuning, on visual inputs combined with facial priors and on video--text inputs, respectively; and FaceLLM~\citep{shahreza2025facellm} adapts InternVL3, with a frozen InternViT vision encoder and low-rank adaptation (LoRA) on the Qwen2.5 language model, to general face understanding (demographic attributes, head pose, and expression), evaluated on the multiple-choice FaceXBench benchmark with expression recognition on RAF-DB and AffectNet.
\end{itemize}

Unified face-analysis Transformers with task tokens (Faceptor~\citep{qin2024faceptor}, FaceXFormer~\citep{narayan2025facexformer}) share the query-based multi-task design of these models but do not process text; they are therefore discussed in Section~\ref{ssec:arch_transformer}.

\subsection{Hybrid architectures}
\label{ssec:arch_hybrid}

Hybrid architectures combine two or more base NN families (most commonly CNN, Transformer, RNN/LSTM, or GNN) within a single FER pipeline, leveraging their complementary inductive biases, i.e., local pixel-level perception (CNN), global self-attention reasoning (Transformer), temporal recurrence (RNN/LSTM), and structural relational modeling (GNN)~\citep{li2020deep,mao2025posterpp}. They are represented among the highest reported results of several benchmarks (e.g., POSTER~\citep{zheng2023poster} on FER+, QWTR~\citep{zhou2025delving} on AffectNet-7, and SODA4MER~\citep{zhang2025dynamic} and MOL~\citep{shao2025mol} in MER), without being over-represented among them (Section~\ref{ssec:perf_discussion}), and are commonly adopted for multi-task and multi-affect FER.

With respect to the combination pattern of base families, hybrid FER systems fall into the following categories:
\begin{itemize}
    \item \uline{CNN+Transformer hybrids}: Methods coupling a CNN spatial extractor with a Transformer global reasoner. POSTER~\citep{zheng2023poster} and POSTER++~\citep{mao2025posterpp} combine landmark-guided CNN features with cross-fusion Transformer encoders; APViT~\citep{xue2022vision} couples CNN tokenization with attentive Transformer patch selection; HLA-ViT~\citep{tian2024facial} integrates ViT global attention with local CNN attention; and FER-Former~\citep{li2024fer} mixes CNN backbones with a CLIP-style Transformer text head.
    \item \uline{CNN+RNN/LSTM hybrids}: Methods combining a CNN spatial encoder with an LSTM/RNN temporal head for DFER, AU dynamics, or intensity. CNN--LSTM and CNN--Transformer comparison studies~\citep{qian2023computer} provide reference baselines for ERI; MTL-DAN~\citep{oh2023human} couples multi-task CNN descriptors with LSTM regression; and CNN+RNN dynamic FER frameworks~\citep{manalu2024detection} combine the two families for video input (Section~\ref{ssec:arch_rnn}).
    \item \uline{CNN+GNN/landmark hybrids}: Methods combining CNN backbones with explicit landmark/region graphs or AU-relation graphs. CMCNN~\citep{yu2022co} treats landmark detection as a co-attentive auxiliary task; FRL-DGT~\citep{zhai2023feature} fuses displacement-based features of landmark-anchored facial regions through Transformer reasoning, without an explicit graph; Chen et al.~\citep{chen2024dual} couple CNN features with GCN-based manifold learning over a sample graph; SpoT-GCN~\citep{deng2024multi} combines CNN and GCN components (Section~\ref{ssec:arch_gnn}); and GCF~\citep{kassab2024gcf} applies GCN layers to CNN features.
    \item \uline{Multi-branch and state-space hybrids}: Methods combining three or more base families or introducing state-space (Mamba) blocks alongside Transformer/CNN backbones. Mamba-VA~\citep{liang2025mamba} combines an MAE encoder with TCN and Mamba blocks for continuous VA; FER-YOLO-Mamba~\citep{ma2024feryolomamba} couples a YOLO-style CNN detector with selective state-space blocks for joint detection/FER; MOL~\citep{shao2025mol} jointly learns MER, optical flow, and landmark detection through Transformer graph-style convolution; and SODA4MER~\citep{zhang2025dynamic} combines self-supervised oriented deformation learning with a CNN encoder and a Bi-LSTM temporal head.
\end{itemize}

\begin{table*}[!t]
  \caption{Neural network architectures: Comparative analysis and key insights.}
  \label{tab:nn_architectures_summary}
  \centering
  \scriptsize

  \setlength{\aboverulesep}{0pt}
  \setlength{\belowrulesep}{0pt}
  
  \setlength{\tabcolsep}{4pt}
  
  \renewcommand{\arraystretch}{1.3}

  \rowcolors{2}{gray!20}{gray!2}

  \newlength{\WAarch}\setlength{\WAarch}{1.6cm}
  \newlength{\WAcnn}\setlength{\WAcnn}{2.4cm}
  \newlength{\WArnn}\setlength{\WArnn}{2.4cm}
  \newlength{\WAtr}\setlength{\WAtr}{2.4cm}
  \newlength{\WAgnn}\setlength{\WAgnn}{2.4cm}
  \newlength{\WAgen}\setlength{\WAgen}{2.4cm}
  \newlength{\WAvlm}\setlength{\WAvlm}{2.4cm}
  \newlength{\WAhyb}\setlength{\WAhyb}{2.4cm}

  \resizebox{\textwidth}{!}{%
  \begin{tabular}{@{}|
    >{\raggedright\arraybackslash}Q{\WAarch}| %
    >{\raggedright\arraybackslash}Q{\WAcnn}|
    >{\raggedright\arraybackslash}Q{\WArnn}|
    >{\raggedright\arraybackslash}Q{\WAtr}|
    >{\raggedright\arraybackslash}Q{\WAgnn}|
    >{\raggedright\arraybackslash}Q{\WAgen}|
    >{\raggedright\arraybackslash}Q{\WAvlm}|
    >{\raggedright\arraybackslash}Q{\WAhyb}|
  @{}}
    \toprule
    \rowcolor{gray!50}
    \headerbreak{Architecture} &
    \headerbreak{CNNs} &
    \headerbreak{RNNs/LSTMs} &
    \headerbreak{Transformers} &
    \headerbreak{GNNs} &
    \headerbreak{Generative/DMs} &
    \headerbreak{VLMs/MLLMs} &
    \headerbreak{Hybrid} \\
    \midrule

    Primary functions &
    \begin{tabitem}
      \item Hierarchical spatial feature learning
      \item AU\slash region-local pattern detection
      \item Mature SFER backbone
    \end{tabitem} &
    \begin{tabitem}
      \item Temporal evolution modeling
      \item Sequence-level aggregation for DFER\slash VA\slash intensity
      \item AU dynamics tracking
    \end{tabitem} &
    \begin{tabitem}
      \item Global long-range attention modeling
      \item Cross-fusion of multiple streams
      \item Unified multi-task FER backbone
    \end{tabitem} &
    \begin{tabitem}
      \item Inter-region\slash AU-relation reasoning
      \item Permutation-invariant facial-graph modeling
      \item Cross-domain alignment
    \end{tabitem} &
    \begin{tabitem}
      \item Expression synthesis and augmentation
      \item Identity\slash pose disentanglement
      \item AU\slash intensity-conditioned editing
    \end{tabitem} &
    \begin{tabitem}
      \item Open-vocabulary recognition
      \item Verbalized affective reasoning
      \item Multi-task unification
    \end{tabitem} &
    \begin{tabitem}
      \item Joint local-global representation
      \item Multi-task\slash multi-affect modeling
      \item Represented among, but not over-represented in, the highest reported results
    \end{tabitem} \\
    \midrule

    Main mechanisms &
    \begin{tabitem}
      \item Learnable convolutions
      \item Local receptive fields
      \item Parameter sharing
      \item Squeeze-and-excitation\slash attention
    \end{tabitem} &
    \begin{tabitem}
      \item Recurrent hidden state
      \item Gated memory (LSTM\slash GRU)
      \item Masked recurrent aggregation
      \item CNN--LSTM stacking
    \end{tabitem} &
    \begin{tabitem}
      \item Multi-headed self-attention
      \item Patch\slash token embedding
      \item Cross-attention with auxiliary streams
      \item Temporal-adapter modules
    \end{tabitem} &
    \begin{tabitem}
      \item Graph message passing
      \item Learnable AU-correlation matrices
      \item Spatio-temporal graph convolution
      \item Adversarial graph alignment
    \end{tabitem} &
    \begin{tabitem}
      \item Adversarial training
      \item Cycle and reconstruction losses
      \item Score-based denoising diffusion
      \item Latent expression conditioning
    \end{tabitem} &
    \begin{tabitem}
      \item Dual image--text encoders
      \item Prompt learning (soft\slash hard)
      \item Instruction tuning
    \end{tabitem} &
    \begin{tabitem}
      \item CNN+Transfor\-mer\slash RNN\slash GNN coupling
      \item Cross-fusion and adapter modules
      \item State-space (Mamba) blocks
      \item Multi-branch ensembles
    \end{tabitem} \\
    \midrule

    Strengths &
    \begin{tabitem}
      \item Mature pre-training ecosystem
      \item Low compute\slash memory cost
      \item Strong locality and SFER baselines
    \end{tabitem} &
    \begin{tabitem}
      \item Variable-length video support
      \item Medium-range temporal modeling
      \item Low memory footprint
    \end{tabitem} &
    \begin{tabitem}
      \item Long-range global dependencies
      \item Cross-modal alignment
      \item Architectural homogenization
    \end{tabitem} &
    \begin{tabitem}
      \item Interpretable AU\slash region relations
      \item Sample efficiency
      \item Strong inductive bias
    \end{tabitem} &
    \begin{tabitem}
      \item Class-imbalance mitigation
      \item Fine-grained expression control
      \item 3D\slash 4D synthesis capability
    \end{tabitem} &
    \begin{tabitem}
      \item Shared image--text embedding space
      \item Frozen encoders with light adaptation
      \item Natural-language outputs
    \end{tabitem} &
    \begin{tabitem}
      \item Widely adopted in-the-wild backbone
      \item Joint local-global modeling
      \item Multi-task flexibility
    \end{tabitem} \\
    \midrule

    Limitations &
    \begin{tabitem}
      \item Limited long-range context
      \item Sensitive to occlusion\slash pose
      \item Performance saturation
    \end{tabitem} &
    \begin{tabitem}
      \item Long-range gradient decay
      \item Sequential, non-parallel
      \item Weak long-context modeling
    \end{tabitem} &
    \begin{tabitem}
      \item Quadratic compute cost
      \item Data-hungry pre-training
      \item Limited interpretability
    \end{tabitem} &
    \begin{tabitem}
      \item Landmark\slash graph-noise sensitivity
      \item Limited pre-training
      \item Poor pixel-level appearance
    \end{tabitem} &
    \begin{tabitem}
      \item Training instability (GAN) / sampling cost (DM)
      \item Anatomical inconsistency risk
      \item Real\slash synthetic domain gap
    \end{tabitem} &
    \begin{tabitem}
      \item Fine-grained categorical limits
      \item Hallucination in LLM decoders
      \item High compute and memory cost
    \end{tabitem} &
    \begin{tabitem}
      \item Architectural complexity
      \item Higher compute
      \item Branch-conflict risk
    \end{tabitem} \\
    \midrule

    Indicative methods &
    \begin{tabitem}
      \item DSAN \citep{fan2020facial}, FLEPNet \citep{karnati2022flepnet}, BReG-NeXt \citep{hasani2020breg}, EAC \citep{zhang2022learn}
    \end{tabitem} &
    \begin{tabitem}
      \item MMA-MRNNet \citep{kollias2024mma}, MTL-DAN \citep{oh2023human}, Qian et al. \citep{qian2023computer}, Tzirakis et al. \citep{tzirakis2017end}
    \end{tabitem} &
    \begin{tabitem}
      \item MAE-DFER \citep{sun2023maedfer}, S2D \citep{chen2024static}, HTNet \citep{wang2024htnet}, SRMCL \citep{bao2024boosting}
    \end{tabitem} &
    \begin{tabitem}
      \item Geo-GCN \citep{zhao2021geometry}, AGRA \citep{chen2021cross}, ST-RDGCN \citep{huang2025modeling}, SpoT-GCN \citep{deng2024multi}
    \end{tabitem} &
    \begin{tabitem}
      \item IPD-FER \citep{jiang2022disentangling}, Triple-BigGAN \citep{gangwar2023triple}, 4DFM \citep{zou2024fourdfm}
    \end{tabitem} &
    \begin{tabitem}
      \item DFER-CLIP \citep{zhao2023prompting}, MPA-FER \citep{ma2025multimodal}, Emotion-LLaMA \citep{cheng2024emotionllama}, FaceLLM \citep{shahreza2025facellm}
    \end{tabitem} &
    \begin{tabitem}
      \item POSTER++ \citep{mao2025posterpp}, QWTR \citep{zhou2025delving}, Mamba-VA \citep{liang2025mamba}, FER-YOLO-Mamba \citep{ma2024feryolomamba}, MOL \citep{shao2025mol}
    \end{tabitem} \\
    \bottomrule
  \end{tabular}%
  }
\end{table*}

\subsection{Comparative analysis and key insights}
\label{ssec:nn_architectures_discussion}

Building on the detailed discussion of the neural network architectures (Sections~\ref{ssec:arch_cnn}--\ref{ssec:arch_hybrid}), this subsection compares them and provides critical insights for each architectural family. Table~\ref{tab:nn_architectures_summary} summarizes, for each architecture, its a) Primary functions, b) Main mechanisms, c) Key strengths, d) Critical limitations, and e) Indicative methods. The comparison shows that the architectures partition into discriminative-perception families (CNNs, RNNs/LSTMs, Transformers, GNNs), generative-synthesis families (GANs, diffusion models), language-grounded families (VLMs/MLLMs), and complementary-combination families (hybrids), exposing a fundamental trade-off between perceptual depth, generative flexibility, semantic grounding, and architectural complexity. Additionally, the newly introduced mechanisms have progressively shifted from pure locality and recurrence (convolutions, recurrent hidden states) towards global attention and message passing (multi-head self-attention, learnable AU-correlation graphs), and more recently towards score-based denoising, instruction tuning, and state-space modeling, and combinations of these mechanisms are becoming increasingly common (e.g., hybrid designs account for 12 of the 25 primary studies published in 2023; Fig.~\ref{fig:bibliometrics}(d)). Moreover, Transformer-based backbones account for 8 of the 9 top-three in-the-wild DFER entries (e.g., MAE-DFER~\citep{sun2023maedfer}, S2D~\citep{chen2024static}, and MMA-DFER~\citep{chumachenko2024mma}; Section~\ref{ssec:perf_discussion}), although their contribution is confounded with the pre-training source, whereas among the highest SFER results only the Transformer-based S2D stands out, while CNNs remain a widely used, computationally efficient choice for SFER, GNNs are particularly suited to AU-relation modeling, generative models address data scarcity and intensity control, and VLMs/MLLMs natively enable open-vocabulary recognition. Furthermore, the main bottlenecks differ in nature rather than in severity (long-range context for CNNs, gradient decay for RNNs/LSTMs, computational cost and data hunger for Transformers, graph-construction noise for GNNs, training instability and sampling cost for generative models, fine-grained discrimination, computational cost, and hallucination for VLMs/MLLMs, and architectural complexity for hybrids). Finally, a trend towards architectural convergence is observed in the reviewed works across families, with attention-based and Transformer-style modules increasingly absorbed into CNN, GNN, and generative pipelines (e.g., S2D~\citep{chen2024static}, ST-RDGCN~\citep{huang2025modeling}, FaceXFormer~\citep{narayan2025facexformer}, FaceLLM~\citep{shahreza2025facellm}), blurring the boundary between architectural categories, which is consistent with the increasing use of hybrid designs in the reviewed works. The distribution of the reviewed primary studies over the combinations of architecture family with recognition task and input modality is reported in Fig.~\ref{fig:crosscriteria}(a) and (c), respectively.

\section{Learning strategies}
\label{sec:learning_strategies}

FER methods can be grouped by the learning strategy they adopt. This choice determines the type and amount of supervision required, how label noise, scarcity, and distribution shift are handled, and how generic visual or affective priors are reused. In particular, the main learning strategies identified in the FER literature are: a) Supervised learning, b) Semi-supervised learning, c) Self-supervised learning, d) Weakly supervised, few-shot, and zero-shot learning, e) Domain adaptation, f) Uncertainty-aware and noisy-label learning, g) Knowledge distillation and parameter-efficient adaptation, and h) Metric and contrastive learning, as discussed in Section~\ref{sec:taxonomy_criteria} and further detailed below.

\subsection{Supervised learning}
\label{ssec:learn_supervised}

Supervised learning is the most widely adopted training strategy in deep learning-based FER, where models are optimized over facial images, videos, or multi-modal streams paired with annotated affective targets (categorical expressions, AU activations, compound expressions, continuous valence--arousal values, or expression intensities)~\citep{li2020deep,mollahosseini2017affectnet}. As the most commonly adopted objective across SFER, DFER, AU detection, MER, and dimensional VA estimation, it underlies most of the highest results reported on aligned in-the-wild benchmarks and remains the standard reference paradigm against which all other learning strategies are typically compared.

Depending on the target output structure, supervised FER methods can be divided into the following main categories:
\begin{itemize}
    \item \uline{Single-label categorical classification methods}: Methods training the FER model with softmax cross-entropy over a discrete expression label set, optionally with imbalance-aware refinements (focal loss~\citep{lin2017focal}, weighted/class-balanced cross-entropy, label smoothing~\citep{szegedy2016rethinking}). CNN classifiers from the FER-2013 challenge~\citep{goodfellow2013challenges} established the baseline for early in-the-wild SFER, while modern attention-based CNN and hybrid methods (POSTER~\citep{zheng2023poster}, POSTER++~\citep{mao2025posterpp}, MHAN~\citep{wang2026mhan}, MFER~\citep{xu2024multiscale}) and dynamic backbones (M3DFEL~\citep{wang2023rethinking}, EST~\citep{liu2023expression}, S2D~\citep{chen2024static}) extend it with multi-head, pyramid, and snippet-level supervision. BReG-NeXt~\citep{hasani2020breg} uses a weighted loss for under-represented categories; C3DBed~\citep{pan2023c3dbed} adopts focal loss for hard MER samples; MERASTC~\citep{gupta2021merastc} applies class-balanced weighting to sparse-AU imbalance; and spatio-temporal CNN/RNN methods (STCAM~\citep{chen2020stcam}, PSRNet~\citep{wang2020phase}, AutoMER~\citep{verma2021automer}) adopt cross-entropy over per-clip labels.

    \item \uline{Language-supervised classification methods}: Methods computing the classification loss over the similarities between the visual features and the text-encoder embeddings of the class labels, instead of, or in addition to, a linear classifier, while still being trained with the labels of the target benchmark. FER-Former~\citep{li2024fer} combines a text-oriented and an one-hot cross-entropy loss, DFER-CLIP~\citep{zhao2023prompting} uses LLM-generated class descriptions with learnable context tokens, and MER-CLIP~\citep{liu2025mer} additionally aligns each micro-expression clip with a description of its annotated AUs through a contrastive loss (Section~\ref{ssec:arch_vlm}). MMPL-FER~\citep{pei2025multi} is likewise trained on the labels of the target benchmarks, while instruction-tuned MLLMs, such as EMO-LLaMA~\citep{xing2024emollama} and EmoVerse~\citep{li2025emoverse}, are trained on instruction data derived from the ground-truth labels of existing FER and affective datasets.

    \item \uline{Ordinal classification methods}: Methods targeting an ordered set of discrete intensity levels (typically $0$--$5$ for expression/AU intensity, or rank-ordered emotion intensity), enforced through rank-aware label distributions or ordinal-regression heads. Ordinal regression enhanced with label-distribution learning~\citep{xu2024facial} targets expression-intensity estimation, while AU-intensity heads in DISFA~\citep{mavadati2013disfa}- and BP4D~\citep{zhang2014bp4d}-style benchmarks routinely combine ordinal classification with the FACS intensity scale.

    \item \uline{Continuous regression methods}: Methods targeting continuous affective dimensions (valence, arousal, dominance, ERI) with mean squared error (MSE), concordance correlation coefficient (CCC), or sigmoid-based regression heads. Mamba-VA~\citep{liang2025mamba} and the MAE+TCN+Transformer pipeline of Zhou et al.~\citep{zhou2024enhancing} regress continuous VA over Aff-Wild2; MMA-MRNNet~\citep{kollias2024mma} formulates ERI as a sequence regression problem; and continuous-affect ABAW pipelines~\citep{kollias2022abaw,kollias2023abaw,kollias20247th} adopt continuous regression as the main objective.

    \item \uline{Multi-task and unified supervision methods}: Methods jointly supervising multiple FER targets (expression, AU, VA, intensity, landmarks) with shared backbones and task-specific heads. EmotiEffNets~\citep{savchenko2023emotieffnets,savchenko2024leveraging} train EfficientNet/MobileViT/MobileFaceNet backbones with joint expression/AU/VA heads; FaceXFormer~\citep{narayan2025facexformer} and Faceptor~\citep{qin2024faceptor} unify FER/parsing/landmarks/attributes via task-token Transformers; CMCNN~\citep{yu2022co} jointly supervises expression recognition and landmark detection; MTL-DAN~\citep{oh2023human} feeds multi-task expression/AU/VA representations into an LSTM head for reaction-intensity estimation; MOL~\citep{shao2025mol} jointly learns MER, optical flow, and landmarks; and recent in-the-wild multi-task frameworks~\citep{kollias2024distribution} extend the paradigm to several classification tasks (e.g., expressions and AUs) through distribution matching.
\end{itemize}

\subsection{Semi-supervised learning}
\label{ssec:learn_semisupervised}

Semi-supervised learning combines a small labeled FER subset with the much larger pool of unlabeled facial data, exploiting pseudo-labeling, consistency regularization, and graph-propagation mechanisms to compensate for the high cost and subjectivity of expression annotation~\citep{li2020deep,sohn2020fixmatch}. This strategy is essential for scaling FER to web-scale corpora (e.g., AffectNet's unlabeled split, scraped face images) without proportionally scaling annotation effort, and it improves generalization under demographic and acquisition shifts.

Semi-supervised FER methods can be grouped by the form of the signal extracted from unlabeled data into the following categories:
\begin{itemize}
    \item \uline{Explicit confidence-based pseudo-label methods}: Methods explicitly assigning hard pseudo-labels to unlabeled samples whose confidence exceeds a (possibly adaptive) threshold, in the spirit of FixMatch~\citep{sohn2020fixmatch}. Ada-CM~\citep{li2022towards} introduces a category-adaptive confidence margin for per-class pseudo-label selection; EACM~\citep{li2025enhanced} extends this with momentum memory and feature-level contrastive regularization for low-confidence samples; Wang et al.~\citep{wang2026mhan} soften pseudo-label targets through label smoothing within MHAN; and Face2Exp~\citep{zeng2022face2exp} formalizes face-to-expression pseudo-labeling under the Meta-Face2Exp framework with domain-aware bias correction.

    \item \uline{Augmentation-based consistency regularization methods}: Methods penalizing prediction inconsistency between augmented views without explicit pseudo-labels. Mean-Teacher-style exponential moving average (EMA) targets~\citep{tarvainen2017mean} provide the canonical consistency baseline re-instantiated over expression and AU labels; and the flip-attention consistency of EAC~\citep{zhang2022learn} is applied to noisy-label learning (Section~\ref{ssec:learn_uncertainty}).

    \item \uline{Graph/manifold-based label propagation methods}: Methods propagating labels from the labeled to the unlabeled subset via learned feature graphs, similarity manifolds, or AU-correlation graphs. Chen et al.~\citep{chen2024dual} combine locality-preserving projections with a GCN over a sample-similarity graph for similarity-based feature propagation, while landmark/AU-graph constructions~\citep{shao2019facial,zhao2021geometry} provide a substrate for cross-sample propagation in AU detection.
\end{itemize}

\subsection{Self-supervised learning}
\label{ssec:learn_selfsupervised}

Self-supervised learning (SSL) trains FER backbones on pretext tasks that do not require expression labels, leveraging the abundance of unlabeled facial images and videos through reconstruction, motion-estimation, or contrastive objectives that yield expression-discriminative features transferable to downstream SFER, DFER, MER, AU detection, and dimensional VA estimation~\citep{li2020deep,he2022masked}. SSL is a recurring component of recent in-the-wild DFER methods reporting strong results and is particularly important for MER, AU detection, and intensity estimation, where labels are especially scarce.

With respect to the pretext-task family, self-supervised FER methods fall into the following categories:
\begin{itemize}
    \item \uline{Masked reconstruction methods}: Methods adapting masked-autoencoder pre-training~\citep{he2022masked,tong2022videomae} to facial images and videos, reconstructing randomly masked patches or frames. $\mu$-BERT~\citep{nguyen2023micron} applies masked-prediction Transformers with micro-attention for MER; MAE-DFER~\citep{sun2023maedfer} and Zhou et al.~\citep{zhou2024enhancing} use video-MAE pre-training for in-the-wild DFER and continuous VA; Mamba-VA~\citep{liang2025mamba} couples MAE features with TCN/state-space blocks; and MAE-DFER additionally introduces explicit temporal facial motion modeling for dynamic expression pre-training~\citep{sun2023maedfer}.

    \item \uline{Motion- and deformation-aware SSL methods}: Methods exploiting short-clip facial motion as a self-supervised signal, especially for MER and dynamic FER. SelfME~\citep{fan2023selfme} learns inter-frame motion fields as an MER pretext task; FRL-DGT~\citep{zhai2023feature} uses self-supervised displacement generation between onset and apex frames; and SODA4MER~\citep{zhang2025dynamic} introduces self-supervised oriented deformation learning on the whole sequence, without apex annotations.

    \item \uline{Contrastive pre-training methods}: Methods adopting contrastive SSL objectives (SimCLR-style instance discrimination, motion-aware contrastive, temporal contrastive) on unlabeled facial corpora~\citep{chen2020simple}. Gao et al.~\citep{gao2024self} introduce a self-supervised contrastive FER pipeline; SRMCL~\citep{bao2024boosting} couples self-supervised contrastive pre-training with prototype-based memory for MER; and TACL~\citep{wang2023temporal} and MER-SupCon~\citep{zhi2022micro} provide temporal- and supervised-contrastive variants that transfer well to downstream FER fine-tuning.

    \item \uline{Generative self-supervised learning methods}: Methods using generative reconstruction (GAN/diffusion-style) as a pretext task to learn a latent expression manifold. IPD-FER~\citep{jiang2022disentangling} exploits identity/pose/expression disentanglement as a generative pretext; 4DFM~\citep{zou2024fourdfm} provides 4D mesh-level generative pre-training; and diffusion autoencoders~\citep{preechakul2022diffusion} learn semantically meaningful, editable face representations that are candidates for generative pre-training in FER.
\end{itemize}

\subsection{Weakly supervised, few-shot, and zero-shot learning}
\label{ssec:learn_weak_fewshot}

Weakly supervised, few-shot, and zero-shot learning collectively address the scenario where annotated expression labels are coarse, scarce, or entirely unavailable for the target classes, exploiting partial/surrogate supervision, episodic adaptation, or vision--language pre-training to bridge the data gap~\citep{li2020deep,radford2021clip}. These paradigms are particularly important for compound, novel, culture-specific, and clinical-affect categories that fall outside the closed taxonomies of standard benchmarks, and they natively support open-vocabulary, verbalized expression recognition.

Depending on the degree of label scarcity and the supervision modality, weakly supervised, few-shot, and zero-shot FER methods can be divided into the following main categories:
\begin{itemize}
    \item \uline{Weakly supervised methods (coarse-grained labels)}: Methods operating under coarse-grained expression labels, where frame-level supervision is unavailable but clip-, video-, or set-level labels are available. M3DFEL~\citep{wang2023rethinking} formulates DFER as multiple-instance learning under video-level supervision; and TG-DFER~\citep{jung2025text} introduces text-guided prompts and multi-grained temporal modeling for frame-level relevance, exploiting video-level labels to train fine-grained expression heads.

    \item \uline{Few-shot and meta-learning methods (small support set)}: Methods operating under a very small number of expression-labeled samples, typically through episodic support/query frameworks (prototypical networks~\citep{snell2017proto}, MAML~\citep{finn2017maml}). MAL~\citep{li2021meta} applies meta auxiliary learning to AU detection, adaptively weighting auxiliary facial expression samples when AU labels are scarce; LightmanNet~\citep{wang2024meta} adopts a bi-level meta-learning pipeline; ME-PLAN~\citep{zhao2022me} uses prototypical learning for rare MER classes; and CRN~\citep{ZHU2022116046} performs relation-network-style few-shot recognition.

    \item \uline{Weak supervision and zero-shot recognition through language}: Methods in which language replaces or reduces the need for expression labels of the target data. FaceLLM~\citep{shahreza2025facellm} fine-tunes an MLLM on LLM-generated question--answer pairs for face images without expression labels, while Emotion-LLaMA~\citep{cheng2024emotionllama} is pre-trained on automatically generated descriptions of unlabeled video samples and is also applied zero-shot to benchmarks not seen in training (Table~\ref{tab:perf_cross}).
\end{itemize}

By contrast, methods that replace the one-hot targets with textual class embeddings but are trained with the labels of the target benchmark (FER-Former~\citep{li2024fer}, DFER-CLIP~\citep{zhao2023prompting}, MER-CLIP~\citep{liu2025mer}, and MMPL-FER~\citep{pei2025multi}), as well as MLLMs instruction-tuned on label-derived data (EMO-LLaMA~\citep{xing2024emollama}, EmoVerse~\citep{li2025emoverse}), are discussed in Section~\ref{ssec:learn_supervised}, and PE-CLIP~\citep{saadi2025peclip} and MPA-FER~\citep{ma2025multimodal} in Section~\ref{ssec:learn_kd_peft}; for reference, zero-shot CLIP reaches 20.07\% WAR on DFEW, compared with 71.25\% for the supervised DFER-CLIP~\citep{zhao2023prompting}.

\subsection{Domain adaptation}
\label{ssec:learn_da}

Domain adaptation (DA) bridges the distribution gap between source and target FER corpora (e.g., lab vs.\ in-the-wild data, cross-cultural and cross-demographic distributions, cross-acquisition setups) through feature alignment, prototype transfer, source-free adaptation, or distributionally robust optimization~\citep{li2020deep,ganin2015dann}. It directly targets the well-documented generalization gap between heavily curated source benchmarks and the markedly more heterogeneous in-the-wild target distributions, and is critical to real-world FER deployability.

Domain-adaptation FER methods can be grouped by the alignment mechanism into the following categories:
\begin{itemize}
    \item \uline{Adversarial feature-alignment methods}: Methods training a domain discriminator alongside the FER model to align source/target feature distributions, in the spirit of DANN~\citep{ganin2015dann}, whereas moment-matching alternatives, such as DAN~\citep{long2015dan}, align the distributions through multi-kernel maximum mean discrepancy without a discriminator. AGRA~\citep{chen2021cross} introduces adversarial graph learning for cross-domain FER through a unified benchmark, aligning lab and in-the-wild distributions via graph-based reasoning over a domain discriminator.

    \item \uline{Continual domain-adaptation methods}: Methods adapting the FER model to a sequence of domains while retaining the performance on the previously encountered ones. Churamani et al.~\citep{churamani2022domain} treat demographic groups as sequentially encountered domains and apply domain-incremental continual learning (regularization- and replay-based strategies) to mitigate bias in expression and AU recognition.

    \item \uline{Source-free and test-time adaptation methods}: Methods adapting the FER model to a target distribution without source data, often at test time, via entropy minimization, pseudo-label refinement, or batch-norm statistics adaptation. Generic source-free hypothesis-transfer methods, such as SHOT~\citep{liang2020shot}, adapt a model without access to source data, and test-time adaptation methods, such as TENT~\citep{wang2021tent}, adapt it at inference via entropy minimization over its predictions; both were proposed for generic image classification and are applicable to FER under in-the-wild shifts.

    \item \uline{Distributionally robust optimization methods}: Methods optimizing the worst-case loss over a family of perturbed distributions for robustness against unknown in-the-wild shifts. HDF~\citep{cui2025learning} introduces heterogeneity-aware distributionally robust optimization over in-the-wild videos, while demographic-bias metrics~\citep{dominguez2024metrics} and attribute-aware or disentangled representations~\citep{xu2020investigating} support the assessment and mitigation of demographic shift.
\end{itemize}

\subsection{Uncertainty-aware and noisy-label learning}
\label{ssec:learn_uncertainty}

Uncertainty-aware and noisy-label learning explicitly models the inherent subjectivity, ambiguity, and annotator disagreement of FER labels through self-cure mechanisms, label-distribution learning, robust losses, and uncertainty estimation over expression logits~\citep{li2020deep,geng2016label}. It is critical to in-the-wild benchmarks where annotation noise, compound expressions, and ambiguous configurations are pervasive, and natively supports compound and culturally conditioned affective states through soft/distributional labels.

With respect to the intervention point in the training pipeline, uncertainty-aware and noisy-label FER methods fall into the following categories:
\begin{itemize}
    \item \uline{Sample-level intervention methods}: Methods intervening at the sample level by re-weighting, relabeling, or selectively suppressing samples according to their reliability or per-sample uncertainty. SCN~\citep{wang2020suppressing} suppresses annotation noise through self-attention ranking and relabeling; EAC~\citep{zhang2022learn} combines random erasing with flip-attention consistency; relative uncertainty learning (RUL)~\citep{zhang2021rul} weights per-sample contributions by learned relative difficulty through feature mixup; and uncertainty-aware FER frameworks~\citep{le2023uncertainty} weight each given label by its estimated uncertainty, which is derived from neighboring samples in valence--arousal space.

    \item \uline{Label-level intervention methods}: Methods replacing one-hot expression labels with soft, distributional, or otherwise refined targets that encode annotator ambiguity~\citep{geng2016label}. Label-distribution consistency variants~\citep{le2023uncertainty,ma2023transformer} adopt distributional targets capturing compound and culturally conditioned affective states; Transformer-based frameworks~\citep{ma2023transformer} couple label-distribution learning with online label correction; and label smoothing~\citep{szegedy2016rethinking} provides a lightweight label-level intervention widely used for in-the-wild FER pipelines.

    \item \uline{Loss-level intervention methods}: Methods replacing standard cross-entropy with noise-tolerant loss formulations (generalized cross-entropy, symmetric cross-entropy, bounded losses) for robust FER under noisy labels. Min et al.~\citep{min2026robust} pursue noise robustness at the level of the graph structure rather than of the loss, supervising the adjacency matrix of a CNN--GCN model (NHG) to limit the propagation of noisy labels, while BReG-NeXt~\citep{hasani2020breg} complements robust losses by stabilizing gradient flow through bounded residual activations.
\end{itemize}

\subsection{Knowledge distillation and parameter-efficient adaptation}
\label{ssec:learn_kd_peft}

Knowledge distillation (KD) and parameter-efficient fine-tuning (PEFT) jointly target the efficient deployment and adaptation of FER models, either by distilling a heavy teacher (FER-, face-, or vision--language pre-trained) into a compact student, or by adapting a large pre-trained model through a small number of additional parameters (adapters, LoRA, prompts)~\citep{hinton2015distilling,hu2022lora}. KD/PEFT is critical to on-device deployment (ADAS, smart classroom, clinical) and to adapting large MLLM/VLM backbones to fine-grained affective categories without full fine-tuning.

Depending on the injection point of the adaptation signal in the pre-trained model, KD/PEFT FER methods can be divided into the following main categories:
\begin{itemize}
    \item \uline{Output-level injection (soft-label distillation) methods}: Methods injecting the learning signal at the output level, distilling a teacher's soft expression distribution into a student~\citep{hinton2015distilling}. EAN~\citep{kong2022real} uses iterative teacher--student transfer from softened distributions for efficient real-time FER; LibreFace~\citep{chang2024libreface} distills heavy AU/FER teachers into deployable students; and compact single-network pipelines~\citep{savchenko2022classifying} reach on-device efficiency without distillation, by fine-tuning lightweight backbones pre-trained for face identification, for FER and engagement recognition.

    \item \uline{Feature-level injection (intermediate distillation) methods}: Methods injecting the signal at the intermediate-feature level, matching feature maps or pairwise sample relations between teacher and student~\citep{romero2015fitnets}. TS-AUCNN~\citep{sun2020dynamic} distills AU knowledge into a lightweight MER student via feature hints; and EmotiEffNets~\citep{savchenko2023emotieffnets,savchenko2024leveraging} instead reuse the frozen features of lightweight pre-trained backbones (e.g., MobileFaceNet, MobileViT), without distilling a heavy teacher.

    \item \uline{Parameter-level injection (LoRA/adapter PEFT) methods}: Methods injecting the signal at the parameter level via additional learnable parameters (low-rank updates, adapter bottlenecks) on a frozen backbone~\citep{houlsby2019parameter,hu2022lora}. S2D~\citep{chen2024static} couples a frozen ViT image backbone with temporal adapters and emotion-anchor self-distillation; MMA-DFER~\citep{chumachenko2024mma} keeps its pre-trained audio and visual encoders frozen and trains only progressive prompts and fusion bottlenecks, without teacher--student distillation (Section~\ref{ssec:arch_transformer}); PE-CLIP~\citep{saadi2025peclip} inserts a GRU-based temporal adapter, shared adapters, and multi-modal prompts into frozen CLIP encoders; and adapter-based PEFT frameworks~\citep{yuan2024auformer} apply parameter-efficient adapters to AU detection.

    \item \uline{Prompt-level injection (prompt tuning) methods}: Methods injecting the signal at the input-prompt level, learning soft prompts (CoOp~\citep{zhou2022coop}), possibly combined with hand-crafted or LLM-generated textual prompts, on frozen or partially tuned backbones; examples are MPA-FER~\citep{ma2025multimodal}, MMPL-FER~\citep{pei2025multi}, TG-DFER~\citep{jung2025text}, and DFER-CLIP~\citep{zhao2023prompting}, whose architectures are described in Section~\ref{ssec:arch_vlm}.
\end{itemize}

\subsection{Metric and contrastive learning}
\label{ssec:learn_metric}

Metric and contrastive learning supervise the geometry of the learned embedding space rather than (or in addition to) the final classification head, pulling same-expression samples together and pushing different-expression samples apart to address the high intra-class variance and low inter-class margin characteristic of affective data~\citep{khosla2020supervised,wen2016centerloss}. These objectives are widely used in SFER, MER, AU detection, and cross-corpus FER, where the embedding-space structure is at least as important as the per-class output, and they integrate naturally with self-supervised and semi-supervised pre-training.

Metric and contrastive FER methods can be grouped by the embedding-shaping objective into the following categories:
\begin{itemize}
    \item \uline{Triplet and center-loss methods}: Methods adopting the classical triplet~\citep{schroff2015facenet} or center-loss~\citep{wen2016centerloss} formulations to shape the FER embedding space. Early FER pipelines use triplet losses to mitigate class imbalance and intra-class variance; center-loss-based regularization (e.g., DACL~\citep{farzaneh2021facial}) encourages compactness around per-class prototypes, which is particularly useful for MER and compound expression recognition where inter-class margins are narrow.

    \item \uline{Supervised contrastive methods}: Methods adopting the supervised contrastive (SupCon) objective~\citep{khosla2020supervised} over expression-labeled samples. MER-SupCon~\citep{zhi2022micro} extends SupCon to MER features projected through a contrastive head; and EACM~\citep{li2025enhanced} applies an InfoNCE objective to low-confidence samples for feature-level contrastive regularization.

    \item \uline{Prototype- and memory-based contrastive methods}: Methods contrasting samples against stored class prototypes or memory banks rather than in-batch negatives. SRMCL~\citep{bao2024boosting} introduces prototype-based memory contrastive learning for MER; and TACL~\citep{wang2023temporal} provides a temporal-contrastive variant aggregating clip-level prototypes.

    \item \uline{Feature-space regularization methods}: Methods adding embedding-space constraints (stability, compactness, separability) on top of a primary FER objective without a dedicated contrastive head. Ada-CM~\citep{li2022towards} regularizes predictions between weakly and strongly augmented samples via an adaptive confidence margin; EACM~\citep{li2025enhanced} uses a feature-level contrastive term within semi-supervised training (Section~\ref{ssec:learn_semisupervised}); Chen et al.~\citep{chen2024dual} introduce intensity-invariant dual-subspace manifold learning that extends locality-preserving projections with GCN-based fusion; and ME-PLAN~\citep{zhao2022me} uses prototypical learning with local attention to regularize MER embeddings.
\end{itemize}

\begin{table*}[!t]
  \caption{Learning strategies: Comparative analysis and key insights (Part 1 of 2).}
  \label{tab:learning_strategies_summary_part1}
  \centering
  \scriptsize

  \setlength{\aboverulesep}{0pt}
  \setlength{\belowrulesep}{0pt}
  
  \setlength{\tabcolsep}{4pt}
  
  \renewcommand{\arraystretch}{1.3}

  \rowcolors{2}{gray!20}{gray!2}

  \newlength{\Wls}\setlength{\Wls}{1.6cm}
  \newlength{\Wlsf}\setlength{\Wlsf}{2.05cm}
  \newlength{\Wlsm}\setlength{\Wlsm}{2.05cm}
  \newlength{\Wlsd}\setlength{\Wlsd}{2.05cm}
  \newlength{\Wlss}\setlength{\Wlss}{2.05cm}
  \newlength{\Wlsl}\setlength{\Wlsl}{2.05cm}
  \newlength{\Wlsmod}\setlength{\Wlsmod}{2.4cm}

  \resizebox{\textwidth}{!}{%
  \begin{tabular}{@{}|
    >{\raggedright\arraybackslash}Q{\Wls}| %
    >{\raggedright\arraybackslash}Q{\Wlsf}|
    >{\raggedright\arraybackslash}Q{\Wlsm}|
    >{\raggedright\arraybackslash}Q{\Wlsd}|
    >{\raggedright\arraybackslash}Q{\Wlss}|
    >{\raggedright\arraybackslash}Q{\Wlsl}|
    >{\raggedright\arraybackslash}Q{\Wlsmod}|
  @{}}
    \toprule
    \rowcolor{gray!50}
    \headerbreak{Learning\\strategy} &
    \headerbreak{Stage in the\\training life-cycle} &
    \headerbreak{Main\\mechanisms} &
    \headerbreak{Primary\\data source} &
    \headerbreak{Strengths} &
    \headerbreak{Limitations} &
    \headerbreak{Indicative\\methods} \\
    \midrule

    Supervised learning &
    \begin{tabitem}
      \item Train-time fine-tuning (end-to-end on labeled FER data)
      \item Can also be applied from scratch on labeled corpora
    \end{tabitem} &
    \begin{tabitem}
      \item Single-label categorical (cross-entropy, with imbalance-aware\slash label-smoothing refinements)
      \item Ordinal classification
      \item Continuous regression
      \item Multi-task / unified heads
    \end{tabitem} &
    \begin{tabitem}
      \item Labeled FER benchmarks
    \end{tabitem} &
    \begin{tabitem}
      \item Strong, well-established empirical baselines
      \item Stable optimization
      \item Mature and easy to integrate
    \end{tabitem} &
    \begin{tabitem}
      \item Annotation dependence
      \item Class imbalance
      \item No use of unlabeled data
      \item Inherits demographic bias
    \end{tabitem} &
    \begin{tabitem}
      \item POSTER++ \citep{mao2025posterpp}, MHAN \citep{wang2026mhan}, S2D \citep{chen2024static}, Mamba-VA \citep{liang2025mamba}, EmotiEffNets \citep{savchenko2023emotieffnets}
    \end{tabitem} \\
    \midrule

    Semi-supervised learning &
    \begin{tabitem}
      \item Train-time fine-tuning over labeled + unlabeled mini-batches
      \item Often jointly with a self-training or consistency loop
    \end{tabitem} &
    \begin{tabitem}
      \item Explicit confidence-based pseudo-labeling
      \item Augmentation-based consistency regularization
      \item Graph\slash mani\-fold-based label propagation
    \end{tabitem} &
    \begin{tabitem}
      \item Small labeled subset + large unlabeled facial corpus
    \end{tabitem} &
    \begin{tabitem}
      \item Reduced annotation cost
      \item Exploits unlabeled data
      \item Improved generalization
    \end{tabitem} &
    \begin{tabitem}
      \item Pseudo-label errors
      \item Sensitive thresholds
      \item Graph-quality dependence
    \end{tabitem} &
    \begin{tabitem}
      \item Ada-CM \citep{li2022towards}, EACM \citep{li2025enhanced}, Face2Exp \citep{zeng2022face2exp}, Chen et al. \citep{chen2024dual}
    \end{tabitem} \\
    \midrule

    Self-supervised learning &
    \begin{tabitem}
      \item Train-time pre-training stage
      \item Typically followed by supervised fine-tuning on a smaller labeled FER set
    \end{tabitem} &
    \begin{tabitem}
      \item Masked reconstruction
      \item Motion\slash defor\-mation SSL
      \item Contrastive pre-training
      \item Generative SSL
    \end{tabitem} &
    \begin{tabitem}
      \item Unlabeled facial images\slash videos
      \item Web-scale face corpora
    \end{tabitem} &
    \begin{tabitem}
      \item No expression labels needed
      \item Scalable to massive corpora
      \item Strong transfer to MER, AU, intensity
    \end{tabitem} &
    \begin{tabitem}
      \item Pretext-target mismatch
      \item Augmentation sensitivity
      \item Needs supervised fine-tuning
    \end{tabitem} &
    \begin{tabitem}
      \item $\mu$-BERT \citep{nguyen2023micron}, SelfME \citep{fan2023selfme}, FRL-DGT \citep{zhai2023feature}, SRMCL \citep{bao2024boosting}, MAE-DFER \citep{sun2023maedfer}
    \end{tabitem} \\
    \midrule

    Weakly\slash few/\allowbreak zero-shot learning &
    \begin{tabitem}
      \item Weakly supervised: Train-time fine-tuning on coarse labels
      \item Few-shot: Train-time episodic meta-training + test-time adaptation
      \item Language-based: Train-time tuning on generated text or inference-only zero-shot transfer
    \end{tabitem} &
    \begin{tabitem}
      \item Weakly supervised (coarse-grained labels)
      \item Few-shot / meta-learning (small support set)
      \item Weak supervision and zero-shot recognition through language
    \end{tabitem} &
    \begin{tabitem}
      \item Coarse video-level labels
      \item Support\slash query episodes
      \item Image--text pairs (CLIP\slash MLLM corpora)
    \end{tabitem} &
    \begin{tabitem}
      \item Reduced annotation requirements
      \item Open-vocabulary recognition
      \item Interpretable verbalized reasoning
    \end{tabitem} &
    \begin{tabitem}
      \item Limited fine-grained discrimination
      \item Gap to fully supervised training
      \item Cultural\slash demo\-graphic bias
    \end{tabitem} &
    \begin{tabitem}
      \item M3DFEL \citep{wang2023rethinking}, TG-DFER \citep{jung2025text}, LightmanNet \citep{wang2024meta}, FaceLLM \citep{shahreza2025facellm}, Emotion-LLaMA \citep{cheng2024emotionllama}
    \end{tabitem} \\
    \bottomrule
  \end{tabular}%
  }
\end{table*}

\begin{table*}[!t]
  \caption{Learning strategies: Comparative analysis and key insights (Part 2 of 2).}
  \label{tab:learning_strategies_summary_part2}
  \centering
  \scriptsize

  \setlength{\aboverulesep}{0pt}
  \setlength{\belowrulesep}{0pt}

  \setlength{\tabcolsep}{4pt}

  \renewcommand{\arraystretch}{1.3}

  \rowcolors{2}{gray!20}{gray!2}

  \setlength{\Wls}{1.6cm}
  \setlength{\Wlsf}{2.05cm}
  \setlength{\Wlsm}{2.05cm}
  \setlength{\Wlsd}{2.05cm}
  \setlength{\Wlss}{2.05cm}
  \setlength{\Wlsl}{2.05cm}
  \setlength{\Wlsmod}{2.4cm}

  \resizebox{\textwidth}{!}{%
  \begin{tabular}{@{}|
    >{\raggedright\arraybackslash}Q{\Wls}|
    >{\raggedright\arraybackslash}Q{\Wlsf}|
    >{\raggedright\arraybackslash}Q{\Wlsm}|
    >{\raggedright\arraybackslash}Q{\Wlsd}|
    >{\raggedright\arraybackslash}Q{\Wlss}|
    >{\raggedright\arraybackslash}Q{\Wlsl}|
    >{\raggedright\arraybackslash}Q{\Wlsmod}|
  @{}}
    \toprule
    \rowcolor{gray!50}
    \headerbreak{Learning\\strategy} &
    \headerbreak{Stage in the\\training life-cycle} &
    \headerbreak{Main\\mechanisms} &
    \headerbreak{Primary\\data source} &
    \headerbreak{Strengths} &
    \headerbreak{Limitations} &
    \headerbreak{Indicative\\methods} \\
    \midrule

    Domain adaptation &
    \begin{tabitem}
      \item Train-time alignment with source and (un)labeled target data
      \item Source-free / test-time adaptation at inference
    \end{tabitem} &
    \begin{tabitem}
      \item Adversarial feature alignment
      \item Prototype\slash class-level alignment
      \item Source-free/\allowbreak test-time adaptation
      \item Distributionally robust optimization
    \end{tabitem} &
    \begin{tabitem}
      \item Source FER corpus + (un)labeled target corpus
    \end{tabitem} &
    \begin{tabitem}
      \item Reduced target-domain annotation
      \item Robustness to demographic/\allowbreak cultural shift
      \item Continual/\allowbreak source-free deployment
    \end{tabitem} &
    \begin{tabitem}
      \item Negative transfer
      \item Adversarial instability
      \item Multi-shift complexity
    \end{tabitem} &
    \begin{tabitem}
      \item AGRA \citep{chen2021cross}, HDF \citep{cui2025learning}, SHOT \citep{liang2020shot} and TENT \citep{wang2021tent} (generic)
    \end{tabitem} \\
    \midrule

    Uncertainty-aware and noisy-label learning &
    \begin{tabitem}
      \item Train-time fine-tuning with noise-aware intervention
      \item Inference-time uncertainty estimation (Monte-Carlo dropout, ensembles)
    \end{tabitem} &
    \begin{tabitem}
      \item Sample-level intervention (re-weighting\slash self-cure\slash uncertainty-guided)
      \item Label-level intervention (label-distribution learning, soft labels)
      \item Loss-level intervention (robust losses, bounded activations)
    \end{tabitem} &
    \begin{tabitem}
      \item Noisy in-the-wild FER labels
      \item Soft\slash distribu\-tional annotations
    \end{tabitem} &
    \begin{tabitem}
      \item Mitigates annotation noise
      \item Improves calibration
      \item Handles compound\slash ambi\-guous affect
    \end{tabitem} &
    \begin{tabitem}
      \item Risk of overfitting noise
      \item Hyperparameter sensitivity
      \item Aleatoric\slash epi\-stemic entanglement
    \end{tabitem} &
    \begin{tabitem}
      \item SCN \citep{wang2020suppressing}, EAC \citep{zhang2022learn}, LDL-FER \citep{le2023uncertainty}, TAN \citep{ma2023transformer}, NHG \citep{min2026robust}, RUL \citep{zhang2021rul}
    \end{tabitem} \\
    \midrule

    Knowledge distillation and PEFT &
    \begin{tabitem}
      \item KD: Two-stage train-time procedure (teacher--student)
      \item PEFT (LoRA\slash adapter/\allowbreak prompt): Train-time on top of a frozen pre-trained backbone
    \end{tabitem} &
    \begin{tabitem}
      \item Output-level injection (soft-label distillation)
      \item Feature-level injection (intermediate distillation)
      \item Parameter-level injection (LoRA\slash adapter PEFT)
      \item Prompt-level injection (prompt tuning)
    \end{tabitem} &
    \begin{tabitem}
      \item Heavy FER, face, or VLM teacher
      \item Small target FER dataset
    \end{tabitem} &
    \begin{tabitem}
      \item Reduced latency and memory
      \item Reuse of pre-trained capacity
      \item Multi-task FER deployment
    \end{tabitem} &
    \begin{tabitem}
      \item Teacher-quality dependence
      \item Catastrophic forgetting
      \item Two-stage complexity
    \end{tabitem} &
    \begin{tabitem}
      \item LibreFace \citep{chang2024libreface}, TS-AUCNN \citep{sun2020dynamic}, MMA-DFER \citep{chumachenko2024mma}, S2D \citep{chen2024static}, PE-CLIP \citep{saadi2025peclip}
    \end{tabitem} \\
    \midrule

    Metric and contrastive learning &
    \begin{tabitem}
      \item Train-time fine-tuning, typically as an auxiliary regularizer to a primary supervised or self-supervised objective
    \end{tabitem} &
    \begin{tabitem}
      \item Triplet\slash center loss
      \item Supervised contrastive
      \item Prototype\slash mem\-ory contrastive
      \item Feature-space regularization
    \end{tabitem} &
    \begin{tabitem}
      \item Same\slash different-expression sample pairs
      \item Class prototypes or memory banks
    \end{tabitem} &
    \begin{tabitem}
      \item Handles intra-class variance
      \item Suited to low-data\slash MER
      \item Targets cross-corpus transfer
    \end{tabitem} &
    \begin{tabitem}
      \item Sampling\slash min\-ing sensitivity
      \item Training cost
      \item Representation-collapse risk
    \end{tabitem} &
    \begin{tabitem}
      \item MER-SupCon \citep{zhi2022micro}, SRMCL \citep{bao2024boosting}, TACL \citep{wang2023temporal}, DACL \citep{farzaneh2021facial}, EACM \citep{li2025enhanced}
    \end{tabitem} \\
    \bottomrule
  \end{tabular}%
  }
\end{table*}

\subsection{Comparative analysis and key insights}
\label{ssec:learning_strategies_discussion}

Building on the detailed discussion of the learning strategies (Sections~\ref{ssec:learn_supervised}--\ref{ssec:learn_metric}), this subsection compares them and provides critical insights for each strategy. Tables~\ref{tab:learning_strategies_summary_part1} and~\ref{tab:learning_strategies_summary_part2} summarize, for each learning strategy, its a) Stage in the training life-cycle, b) Main mechanisms, c) Primary data source, d) Key strengths, e) Critical limitations, and f) Indicative methods. A first observation is that FER learning strategies operate at fundamentally different points of the training and inference pipeline rather than being interchangeable substitutes (self-supervised pre-training, supervised/semi-supervised/metric/uncertainty-aware fine-tuning, knowledge distillation, parameter-efficient adaptation, train- or test-time domain adaptation, and inference-only zero-shot via image--text alignment), so that modern in-the-wild FER pipelines frequently combine several such stages (e.g., face-specific self-supervised pre-training followed by parameter-efficient or supervised fine-tuning in S2D~\citep{chen2024static} and MMA-DFER~\citep{chumachenko2024mma}, whose ablations are discussed in Section~\ref{ssec:perf_discussion}). Additionally, a trend towards data-efficient and representation-aware mechanisms is observed in the reviewed works across strategies, with imbalance/ordinal-aware and multi-task supervision, confidence-based pseudo-labeling and consistency regularization, masked-reconstruction and contrastive self-supervision, language-based weak supervision and zero-shot recognition, adversarial and source-free domain adaptation, sample/label/loss-level uncertainty interventions, and output/feature/parameter/prompt-level KD/PEFT increasingly complementing the cross-entropy-only paradigm. Moreover, the primary data source spans a continuum from fully labeled FER corpora through partially- and weakly labeled regimes to unlabeled facial data and externally sourced supervision (image--text pairs, teacher-model outputs), making modern FER less and less dependent on per-frame expression annotations and more dependent on large pre-trained backbones, vision--language pairs, and unlabeled facial corpora. Furthermore, the main bottlenecks differ in nature rather than in severity (annotation noise and class imbalance for supervised, pseudo-label error propagation for semi-supervised, pretext-target mismatch for self-supervised, fine-grained discrimination and the gap to fully supervised training for weak/few/zero-shot, negative transfer for domain adaptation, aleatoric/epistemic disentanglement for uncertainty-aware learning, teacher quality and forgetting for KD/PEFT, and sampling and collapse risk for metric/contrastive learning). Finally, several methods reporting strong recent in-the-wild results (S2D~\citep{chen2024static}, MMA-DFER~\citep{chumachenko2024mma}, POSTER++~\citep{mao2025posterpp}, and Mamba-VA~\citep{liang2025mamba}) combine supervised, self/semi-supervised, metric/contrastive, and KD/PEFT components, which is consistent with a shift from purely supervised training over closed taxonomies towards a multi-strategy, foundation model-style learning paradigm. The distribution of the reviewed primary studies over the combinations of recognition task and learning strategy is reported in Fig.~\ref{fig:crosscriteria}(b).

\section{Acquisition settings}
\label{sec:acquisition_settings}

FER methods can be grouped by the conditions under which their training and evaluation data are captured, which determine the difficulty of the recognition task and the conclusions that the reported performance supports. In particular, the main acquisition settings encountered in the FER literature are: a) Controlled (laboratory) acquisition and b) In-the-wild acquisition, as discussed in Section~\ref{sec:taxonomy_criteria} and further detailed below. Since the acquisition setting is a property of the data rather than of the method, the methods of each setting are grouped below by the benchmarks on which they are evaluated, namely by sensing modality for controlled acquisition and by recognition task for in-the-wild acquisition, whereas the two settings themselves are compared in Section~\ref{ssec:acq_discussion}.

\subsection{Controlled (laboratory) acquisition}
\label{ssec:acq_controlled}

Controlled (laboratory) acquisition is the historical default protocol of FER research, in which expressions are elicited or posed under uniform lighting, frontal head pose, and a constrained set of subjects, and recognition performance is reported on these tightly curated corpora. As the canonical ablation and isolation environment for new architectural, learning, and pre-processing contributions, it covers specialized acquisition modalities (3D/4D, thermal, NIR), FACS-coded MER, and AU-intensity ground truth, which are largely absent from in-the-wild corpora.

With respect to the sensing modality of the employed benchmarks, controlled-acquisition FER methods fall into the following categories:
\begin{itemize}
    \item \uline{2D RGB lab benchmarks}: Methods evaluated on 2D RGB lab corpora under tight acquisition control. The family splits along the elicitation-type axis into a posed-expression subset (CK+~\citep{lucey2010extended}, JAFFE~\citep{lyons1998japanese}, RaFD~\citep{langner2010presentation}, MMI-style for SFER/AU) and a spontaneous-expression subset (DISFA~\citep{mavadati2013disfa} and BP4D~\citep{zhang2014bp4d} for spontaneous-AU, and SMIC~\citep{li2013spontaneous}, CASME~\citep{yan2013casme}, CASME~II~\citep{yan2014casme}, CAS(ME)$^2$~\citep{qu2016cas}, and SAMM~\citep{davison2016samm} for spontaneous-MER). Early FAA surveys~\citep{sariyanidi2014automatic} document their central role in pre-deep-learning FER, and spontaneous-MER methods (HTNet~\citep{wang2024htnet}, $\mu$-BERT~\citep{nguyen2023micron}, MOL~\citep{shao2025mol}, LTR3O~\citep{zhu2025learning}, MER-CLIP~\citep{liu2025mer}, MER-SupCon~\citep{zhi2022micro}, SelfME~\citep{fan2023selfme}, SODA4MER~\citep{zhang2025dynamic}) report on two or more of CASME~II, SAMM, and SMIC-HS (Table~\ref{tab:perf_mer}).

    \item \uline{3D/4D geometric lab benchmarks}: Methods evaluated on lab corpora of dense 3D meshes and 4D mesh sequences, both posed and spontaneous. BU-3DFE~\citep{yin20063d} and BU-4DFE (posed) and BP4D (spontaneous) 3D/4D corpora support 3D FER and 4D MER; CAS(ME)$^3$~\citep{li2022cas} adds RGB-D micro-expression sequences; 4DME~\citep{li20224dme} combines DI4D, RGB, depth, and grayscale streams for spontaneous 4D MER; and the recent 4D spontaneous MER database~\citep{wang2024spontaneous4D} extends 4D evaluation to spontaneous micro-expressions, with these benchmarks being the default for sparsity-aware 4D FER~\citep{behzad2021sparsity}, magnification-based 4D FER~\citep{behzad2021magnifying}, and 3D mesh-level diffusion synthesis~\citep{zou2024fourdfm}.

    \item \uline{Thermal/NIR and multi-spectral lab benchmarks}: Methods evaluated on lab corpora of thermal-IR, NIR, or multi-spectral acquisitions, both posed and spontaneous. The Oulu-CASIA NIR sequences~\citep{zhao2011facial} are the canonical NIR-FER lab benchmark, while IRFacExNet~\citep{bhattacharyya2021deep}, CTIFERK~\citep{wang2025ctiferk}, and Naseem et al.~\citep{naseem2024msx,khan2025visir} report on thermal/NIR or multi-spectral lab corpora (Tufts Face, IRIS, GUET-thermal-face, VIRI, NVIE), where visible--IR fusion improves over single-modality models.
\end{itemize}

\subsection{In-the-wild acquisition}
\label{ssec:acq_wild}

In-the-wild acquisition is the dominant protocol in modern FER, in which models are benchmarked on unconstrained imagery and video scraped from the internet, recorded in unconstrained environments, or collected through reaction-eliciting stimuli, exposing the model to real-world variability in pose, occlusion, illumination, identity, culture, and annotation~\citep{li2020deep,mollahosseini2017affectnet}. As the closest available approximation to deployment conditions, it is the de facto benchmark protocol for new in-the-wild FER methods and natively supports categorical, dimensional, compound, and AU FER tasks within a single corpus family.

Depending on the target task of the employed benchmarks, in-the-wild FER methods can be divided into the following main categories:
\begin{itemize}
    \item \uline{Categorical SFER benchmarks}: Methods evaluated on unconstrained static 2D RGB benchmarks for categorical expression recognition. RAF-DB~\citep{li2017reliable}, AffectNet~\citep{mollahosseini2017affectnet}, FER+~\citep{barsoum2016training}, ExpW~\citep{zhang2018facial}, and SFEW~\citep{dhall2011static} are the canonical corpora; modern attention-based CNN and hybrid methods (POSTER~\citep{zheng2023poster}, POSTER++~\citep{mao2025posterpp}, MHAN~\citep{wang2026mhan}, MFER~\citep{xu2024multiscale}, EAC~\citep{zhang2022learn}, SCN~\citep{wang2020suppressing}, EACM~\citep{li2025enhanced}, and the deep/handcrafted local-learning approach of Georgescu et al.~\citep{georgescu2019local}) are systematically evaluated on them, while CLIP-based methods (FER-Former~\citep{li2024fer}, MPA-FER~\citep{ma2025multimodal}, MMPL-FER~\citep{pei2025multi}) are trained on them with language-grounded supervision (Section~\ref{ssec:learn_supervised}).

    \item \uline{Categorical DFER benchmarks}: Methods evaluated on unconstrained dynamic 2D RGB video benchmarks for categorical expression recognition. DFEW~\citep{jiang2020dfew}, FERV39k~\citep{wang2022ferv39k}, and MAFW~\citep{liu2022mafw} are the canonical corpora, evaluated by S2D~\citep{chen2024static}, M3DFEL~\citep{wang2023rethinking}, EST~\citep{liu2023expression}, and HDF~\citep{cui2025learning}; DFER-CLIP~\citep{zhao2023prompting}, PE-CLIP~\citep{saadi2025peclip}, and TG-DFER~\citep{jung2025text} use them for language-guided and prompt-driven DFER, with MAFW additionally supporting audio-visual evaluation.

    \item \uline{AU detection and compound-expression benchmarks}: Methods evaluated on unconstrained corpora targeting AU detection or compound expression recognition. EmotioNet~\citep{fabian2016emotionet,benitez2017emotionet} is the canonical in-the-wild AU benchmark; the RAF-DB compound split~\citep{li2017reliable} and C-EXPR-DB~\citep{kollias2023multi} support compound evaluation; and multi-task in-the-wild AU/expression pipelines~\citep{kollias2024distribution} jointly report on AU and expression targets.

    \item \uline{Continuous valence--arousal benchmarks}: Methods evaluated on unconstrained corpora targeting continuous VA estimation. Aff-Wild2~\citep{kollias2018aff,kollias2022abaw,kollias2023abaw,kollias20247th,kollias2025abaw8} and AFEW-VA~\citep{kossaifi2017afew} are the dominant corpora, evaluated by Mamba-VA~\citep{liang2025mamba}, MAE+TCN~\citep{zhou2024enhancing}, and EmotiEffNets~\citep{savchenko2023emotieffnets,savchenko2024leveraging}; many of these corpora additionally provide synchronized audio, making audio-visual evaluation a cross-cutting property.
\end{itemize}

\subsection{Comparative analysis and key insights}
\label{ssec:acq_discussion}

The two acquisition settings differ mainly in the nature, rather than in the amount, of the variability that they contain. Controlled corpora (Table~\ref{tab:datasets_fer_lab}) fix the illumination, the head pose, and often the elicitation protocol, so that they isolate the expression signal; they therefore remain the reference setting for MER, AU detection, and 3D/4D and thermal/NIR inputs, for which in-the-wild data are scarce. In-the-wild corpora (Table~\ref{tab:datasets_fer_itw}) introduce pose, occlusion, resolution, and demographic variability, but also label noise; for instance, the annotators of AffectNet agree on only 60.7\% of the doubly annotated images (Section~\ref{ssec:perf_discussion}). The acquisition setting of the training data also bounds the conclusions that the reported results support, since the cross-dataset results collected in Table~\ref{tab:perf_cross} are in most cases lower than the corresponding within-dataset values (e.g., 76.16\% instead of 92.21\% on RAF-DB for S2D~\citep{chen2024static}, when trained on AffectNet). As the acquisition setting is a property of the employed datasets rather than of the method, it is not included in the cross-criteria analysis of Section~\ref{sec:cross_criteria}.

\section{Application domains}
\label{sec:app_domains}

FER methods can be grouped by the application domain in which they are deployed; each domain imposes its own constraints on latency, robustness, fairness, modality availability, and the required granularity of the affective signal. In particular, the main FER application domains identified in the literature are: a) Healthcare and clinical assessment, b) Education and learning environments, c) Driver monitoring and intelligent transportation, d) Security, surveillance, and content moderation, e) Human--robot and human--computer interaction, f) Entertainment, gaming, and immersive media, and g) Marketing and consumer experience analytics, as discussed in Section~\ref{sec:taxonomy_criteria} and further detailed below.

\subsection{Healthcare and clinical assessment}
\label{ssec:app_healthcare}

Healthcare is a prominent FER application domain, since facial behavior offers a non-invasive cue for pain, mood, cognitive, and neurological states when verbal self-report is unreliable~\citep{kumar2023artificial,poria2017review,mentalhealth2025bmc}. Pain estimation uses AU-coded muscle dynamics through multiple-instance and spatio-temporal pipelines~\citep{chen2019learning,demelo2022spatiotemporal}, with neonatal extensions using ViTs on hospital neonatal image sets~\citep{neonatal2024scireports} and CNN--LSTM fusion over neonatal intensive care unit (NICU) recordings~\citep{wuetal2024neonatalpain}. Depression and mood assessment exploit deep spatio-temporal~\citep{de2020deep}, multi-scale temporal-difference attention~\citep{zhang2023mtdan}, smartphone-camera longitudinal (non-deep, gradient boosting on AU features)~\citep{islam2024facepsy}, and multi-modal video/audio/text~\citep{videodepression2025frontiers} pipelines (reviewed in~\citep{depression2025neurocomputing}), while complementary work targets mild cognitive impairment~\citep{fei2022novel} and facial-nerve-paralysis stress~\citep{xu2022stress}. For autism spectrum disorder (ASD), attention-augmented YOLOv8 detectors~\citep{hosney2024autyolo}, genetic-algorithm-optimized CNNs with fog/Internet of Things (IoT) notification of parents~\citep{talaat2023autismiot}, few-shot trait models~\citep{zhang2022autism}, and Transformer atypical-expression backbones~\citep{autistic2025appsci} support objective screening and assistive teaching, with cross-corpus generalization, demographic fairness, and personalization as critical deployment requirements.

\subsection{Education and learning environments}
\label{ssec:app_education}

In educational settings, FER infers learner engagement, attention, confusion, frustration, and satisfaction, feeding intelligent tutoring platforms and classroom analytics~\citep{savchenko2022classifying,guo2022smartedu}. Under unconstrained classroom and remote-learning conditions, recent works frequently adopt efficient deep models such as lightweight EmotiEffNet backbones~\citep{savchenko2022classifying} and multi-region divided-attention networks for smart-education clouds~\citep{guo2022smartedu}, with online-learning extensions targeting composite engagement emotions (confusion, satisfaction, disappointment, frustration)~\citep{onlineedu2024complex} and well-being monitoring in higher-education cohorts~\citep{collegemental2025natsci}. Related threads target group-level affect through non-volume-preserving fusion over crowd videos~\citep{quach2022grouplevel}, while domain shift across ages, classrooms, and cultures motivates cross-domain FER pipelines that transfer expression knowledge from in-the-wild corpora~\citep{chen2023crosscorpusfer,zhang2023jointlocalglobal,xie2022crossdomain}.

\subsection{Driver monitoring and intelligent transportation}
\label{ssec:app_driver}

Within intelligent transportation, FER is central to advanced driver-assistance systems (ADAS), estimating affective state, attention, fatigue, distraction, takeover readiness, and road-rage onset jointly with gaze, head pose, posture, and vehicle-state signals~\citep{duongthang2025driver,saadi2024driver,gu2024emotake}. Part of this literature abbreviates driver facial expression recognition as DFER~\citep{saadi2024driver}; in this survey, DFER denotes exclusively dynamic FER, and driver facial expression recognition is written in full. Driver-oriented pipelines must meet real-time embedded constraints and cope with severe illumination variability and partial occlusion; EmoTake derives facial-expression, head-pose, and body-movement cues from an in-cabin camera to predict takeover readiness, reaction time, and quality~\citep{gu2024emotake}, while an emotion-aware ADAS survey frames the link between affective state and driving safety~\citep{saadi2024driver}. To meet on-vehicle latency budgets, recent works exploit dual-attention lightweight CNNs with SqueezeNext-style backbones~\citep{lightdriver2025designs} and edge-AI Inception/single-shot detector (SSD) combinations for production driver-state inference~\citep{driveredge2025sensors}, with recurring trends being multi-modality, lightweight parameter-efficient adaptation, and closed-loop integration into driver-state controllers.

\subsection{Security, surveillance, and content moderation}
\label{ssec:app_security}

FER has been investigated in security-oriented scenarios including smart-home surveillance, crowd monitoring, mask-aware recognition, content moderation, and deepfake forensics. iSecureHome~\citep{kaushik2022isecurehome} combines real-time FER with multi-modal surveillance evidence for detecting abnormal, aggressive, or distressed behavior, while public-space group-level FER aggregates facial and spatio-temporal information across individuals for crowd-affect estimation and event monitoring~\citep{quach2022grouplevel}. Mask-aware FER tackles lower-face occlusion through face-parsing-aware ViT pipelines that route attention to non-occluded upper-face regions~\citep{yang2022maskfer}, and content-moderation pipelines exploit FER-derived affective cues to identify inappropriate or harmful video content~\citep{gangwar2023triple}. A growing security-adjacent application is deepfake forensics, where AU-guided spatio-temporal representations and mask-completion/AU-detection cross-attention target localized expression edits~\citep{deepfakeau2025}, turning AU dynamics into a signature for facial-behavior authenticity, liveness detection, and anti-spoofing.

\subsection{Human--robot and human--computer interaction}
\label{ssec:app_hri}

Affect-aware human--robot interaction (HRI) and human--computer interaction (HCI) are among the original and still most active FER application domains~\citep{poria2017review}, providing a real-time channel through which social robots, conversational agents, and intelligent tutoring systems perceive the user's emotional state and adapt their behavior. Representative pipelines couple a face-detection front-end with a deep FER backbone and a behavior-generation module, ranging from compact two-stage CNN expression classifiers suitable for such front-ends~\citep{mehendale2020facialhri} to MSAFNet-style multi-scale attention and convolution--Transformer fusion for embodied AI agents~\citep{embodied2025msafnet}, with assistive HCI extensions for children with autism, such as an augmented-reality game that uses FER in imitation exercises~\citep{lyu2024emooly} and home-based interventions on situation-based emotional understanding~\citep{lyu2024dailyconnect}. Recent MLLM-based affect systems (Emotion-LLaMA~\citep{cheng2024emotionllama}, EMO-LLaMA~\citep{xing2024emollama}, EmoVerse~\citep{li2025emoverse}) further extend HCI to natural-language reasoning over multi-modal affect, with personalized, low-latency, and culture-aware models needed to avoid systematic interaction failures.

\subsection{Entertainment, gaming, and immersive media}
\label{ssec:app_entertainment}

FER is increasingly embedded in entertainment, gaming, and immersive-media pipelines, driving expressive avatars, emotion-responsive game logic, and real-time facial animation for AR/VR and metaverse applications. Bellenger et al.~\citep{bellenger2024avatar} target avatar control through a real-time facial-feature extractor and reduced AU-aligned feature set mapped onto standard rigging conventions, while the integration of FER with ViT- and MLLM-based facial-analysis backbones (FaceXFormer~\citep{narayan2025facexformer}, FaceLLM~\citep{shahreza2025facellm}) supports unified perception for animation and content-creation tools that jointly estimate expression, head pose, facial attributes, and, in the case of FaceXFormer, landmarks. Ambient and interactive entertainment further exploits MLLM-grounded affect systems for context-aware audio-visual responses~\citep{cheng2024emotionllama}, with the common trend being a shift from purpose-built FER networks towards unified, foundation model-style facial encoders serving recognition, animation, and synthesis simultaneously.

\subsection{Marketing and consumer experience analytics}
\label{ssec:app_marketing}

An additional FER application area is marketing and consumer experience analytics, where FER infers engagement, sentiment, and reaction strength from facial behavior towards advertising stimuli, retail interactions, and consumer products~\citep{poria2017review}. Marketing deployments operate in semi-controlled environments (focus groups, in-store cameras, smart mirrors) and prioritize aggregate, population-level trends over per-subject diagnostic accuracy, aggregating FER outputs (categorical expressions, continuous VA, attention-related AUs) over time and across audiences to obtain reaction curves and engagement indices complementing survey-based metrics. Methodologically, the same in-the-wild SFER, DFER, and continuous-VA backbones (POSTER++~\citep{mao2025posterpp}, MHAN~\citep{wang2026mhan}, S2D~\citep{chen2024static}, Mamba-VA~\citep{liang2025mamba}, MMA-DFER~\citep{chumachenko2024mma}) are reused with adaptations for on-device inference, demographic fairness, and temporal smoothing, while recent reviews highlight physiological and audio-visual multi-modality as a key enabler~\citep{mentalhealth2025bmc}.

\subsection{Comparative analysis and key insights}
\label{ssec:app_discussion}

The application domains differ less in the recognition methods that they employ than in the constraints that they impose. Healthcare prioritizes clinically meaningful targets and interpretability; education and consumer analytics aggregate affect over groups of subjects and over time, rather than per frame; driver monitoring and human--robot interaction require real-time operation under head-pose and illumination variation; security and surveillance raise the strongest privacy and fairness concerns; and entertainment emphasizes latency and the joint estimation of expression and facial geometry. In the reviewed works, the domain-specific contribution typically lies in the recognition target, the sensing setup, or the deployment pipeline, rather than in the recognition architecture itself (e.g., the camera-based combination of facial expression, head pose, and body movement for takeover prediction in EmoTake~\citep{gu2024emotake}, or the fusion of facial and cry-sound cues for neonatal pain assessment~\citep{wuetal2024neonatalpain}), and many of these works address the ``other affective states'' of Section~\ref{sec:scope} (e.g., pain, engagement, or driver state), rather than the six expression-level tasks. Since such requirements are only partly reported in the primary studies, the application domain is not included in the cross-criteria analysis of Section~\ref{sec:cross_criteria}; the corresponding open issues are discussed in Sections~\ref{sec:challenges} and~\ref{sec:future_directions}.

\newlength{\Dds}\setlength{\Dds}{1.5cm}
\newlength{\Dyr}\setlength{\Dyr}{0.6cm}
\newlength{\Dscale}\setlength{\Dscale}{2.4cm}
\newlength{\Dclass}\setlength{\Dclass}{1.7cm}
\newlength{\Dmoda}\setlength{\Dmoda}{1.8cm}
\newlength{\Dann}\setlength{\Dann}{2.4cm}
\newlength{\Dtask}\setlength{\Dtask}{1.2cm}
\newlength{\Dtemp}\setlength{\Dtemp}{0.6cm}
\newlength{\Dsubj}\setlength{\Dsubj}{1.6cm}
\newlength{\Ddesc}\setlength{\Ddesc}{3.0cm}

\begin{table*}[!t]
  \caption{Public FER datasets acquired under laboratory (controlled) settings. Abbreviations used: Temporality (Static images (ST), Video (V), 3D/4D (3D)), and Recognition task (static FER (SFER), dynamic FER (DFER), action unit detection (AUD), micro-expression recognition (MER), compound expression recognition (CFER), dimensional valence--arousal estimation (VA)).}
  \label{tab:datasets_fer_lab}
  \centering
  \scriptsize

  \setlength{\aboverulesep}{0pt}
  \setlength{\belowrulesep}{0pt}
  
  \setlength{\tabcolsep}{4pt}
  
  \renewcommand{\arraystretch}{1.3}
  
  \setlength{\tabcellpad}{0.5pt}
  \rowcolors{2}{gray!20}{gray!2}

  \resizebox{\textwidth}{!}{%
  \begin{tabular}{@{}|
    >{\raggedright\arraybackslash}Q{\Dds}| %
    >{\centering\arraybackslash}Q{\Dyr}|
    >{\raggedright\arraybackslash}Q{\Dscale}|
    >{\raggedright\arraybackslash}Q{\Dclass}|
    >{\raggedright\arraybackslash}Q{\Dmoda}|
    >{\raggedright\arraybackslash}Q{\Dann}|
    >{\centering\arraybackslash}Q{\Dtask}|
    >{\centering\arraybackslash}Q{\Dtemp}|
    >{\raggedright\arraybackslash}Q{\Dsubj}|
    Q{\Ddesc}| %
  @{}}
    \toprule
    \rowcolor{gray!50}
    \headerbreak{Dataset} &
    \headerbreak{Year} &
    \headerbreak{Scale} &
    \headerbreak{Classes} &
    \headerbreak{Modalities} &
    \headerbreak{Annotation\\type} &
    \headerbreak{Task} &
    \headerbreak{Temp.} &
    \headerbreak{Subjects} &
    \headerbreak{Description} \\
    \midrule

    \textbf{JAFFE} \citep{lyons1998japanese,lyons1997jaffe} &
    1997 &
    \begin{tabitem}
      \item 213 images
    \end{tabitem} &
    \begin{tabitem}
      \item 7 requested poses (6 basic + neutral)
    \end{tabitem} &
    \begin{tabitem}
      \item Grayscale images
    \end{tabitem} &
    \begin{tabitem}
      \item Requested-pose labels
      \item Semantic ratings (6 emotion adjectives, 60 raters)
    \end{tabitem} &
    SFER &
    ST &
    \begin{tabitem}
      \item 10 (all F)
    \end{tabitem} &
    Posed dataset of Japanese female subjects created for expression-perception research, later widely used as an SFER benchmark \\
    \midrule

    \textbf{MMI} \citep{pantic2005web} &
    2005 &
    \begin{tabitem}
      \item 1{,}500 images and videos
    \end{tabitem} &
    \begin{tabitem}
      \item 7 basic classes
      \item AU set
    \end{tabitem} &
    \begin{tabitem}
      \item RGB images and videos
    \end{tabitem} &
    \begin{tabitem}
      \item Categorical labels
      \item AU annotations
    \end{tabitem} &
    SFER, AUD &
    ST,V &
    \begin{tabitem}
      \item 19
    \end{tabitem} &
    Early web-based lab benchmark supporting both expression and AU evaluation \\
    \midrule

    \textbf{BU-3DFE} \citep{yin20063d} &
    2006 &
    \begin{tabitem}
      \item 2{,}500 3D scans
    \end{tabitem} &
    \begin{tabitem}
      \item 7 classes
      \item Intensity levels
    \end{tabitem} &
    \begin{tabitem}
      \item 3D face scans
    \end{tabitem} &
    \begin{tabitem}
      \item Categorical labels
      \item Intensity levels
      \item Facial landmarks
    \end{tabitem} &
    SFER &
    3D &
    \begin{tabitem}
      \item 100 (60F, 40M)
    \end{tabitem} &
    Canonical 3D FER benchmark with multi-intensity prototypical expressions \\
    \midrule

    \textbf{RaFD} \citep{langner2010presentation} &
    2010 &
    \begin{tabitem}
      \item Multi-view image collection
    \end{tabitem} &
    \begin{tabitem}
      \item 8 classes
    \end{tabitem} &
    \begin{tabitem}
      \item RGB images
    \end{tabitem} &
    \begin{tabitem}
      \item Categorical labels
      \item Gaze direction
      \item Camera-angle metadata
    \end{tabitem} &
    SFER &
    ST &
    \begin{tabitem}
      \item 67
    \end{tabitem} &
    Posed multi-view, multi-gaze face stimulus set for controlled FER \\
    \midrule

    \textbf{CK+} \citep{lucey2010extended} &
    2010 &
    \begin{tabitem}
      \item 593 sequences
    \end{tabitem} &
    \begin{tabitem}
      \item 7 classes
      \item AU set
    \end{tabitem} &
    \begin{tabitem}
      \item Grayscale videos and images
    \end{tabitem} &
    \begin{tabitem}
      \item Categorical labels
      \item AU annotations
      \item Facial landmarks
    \end{tabitem} &
    SFER, AUD &
    V &
    \begin{tabitem}
      \item 123 (from a pool of 210 adults, 69\% female)
    \end{tabitem} &
    Extended Cohn-Kanade benchmark with both expression and AU ground truth \\
    \midrule

    \textbf{Oulu-CASIA} \citep{zhao2011facial} &
    2011 &
    \begin{tabitem}
      \item 2{,}400 sequences\slash images
    \end{tabitem} &
    \begin{tabitem}
      \item 6 classes
    \end{tabitem} &
    \begin{tabitem}
      \item VIS and NIR videos\slash images
    \end{tabitem} &
    \begin{tabitem}
      \item Categorical labels
      \item Illumination conditions
    \end{tabitem} &
    SFER, DFER &
    V &
    \begin{tabitem}
      \item 80
    \end{tabitem} &
    VIS\slash NIR multi-spectral FER under three controlled illumination settings \\
    \midrule

    \textbf{DISFA} \citep{mavadati2013disfa} &
    2013 &
    \begin{tabitem}
      \item Spontaneous AU videos
    \end{tabitem} &
    \begin{tabitem}
      \item 12 AUs
      \item AU intensity
    \end{tabitem} &
    \begin{tabitem}
      \item RGB video
    \end{tabitem} &
    \begin{tabitem}
      \item AU intensities
      \item Facial landmarks
    \end{tabitem} &
    AUD &
    V &
    \begin{tabitem}
      \item 27 (12F, 15M)
    \end{tabitem} &
    Spontaneous AU benchmark with frame-level intensity annotations \\
    \midrule

    \textbf{CASME} \citep{yan2013casme} &
    2013 &
    \begin{tabitem}
      \item 195 clips
    \end{tabitem} &
    \begin{tabitem}
      \item 8 classes
      \item AU set
    \end{tabitem} &
    \begin{tabitem}
      \item RGB video
    \end{tabitem} &
    \begin{tabitem}
      \item Categorical labels
      \item AU annotations
      \item Onset/apex/offset
    \end{tabitem} &
    MER &
    V &
    \begin{tabitem}
      \item 35 (13F, 22M)
    \end{tabitem} &
    Spontaneous micro-expression corpus with apex annotations \\
    \midrule

    \textbf{SMIC} \citep{li2013spontaneous} &
    2013 &
    \begin{tabitem}
      \item 164 clips
    \end{tabitem} &
    \begin{tabitem}
      \item 3 classes (Pos/\allowbreak Neg/\allowbreak Surp)
    \end{tabitem} &
    \begin{tabitem}
      \item RGB and infrared images
    \end{tabitem} &
    \begin{tabitem}
      \item Categorical labels
    \end{tabitem} &
    MER &
    V &
    \begin{tabitem}
      \item 16 (3F, 13M)
    \end{tabitem} &
    Spontaneous MER benchmark with RGB and infrared streams \\
    \midrule

    \textbf{BP4D} \citep{zhang2014bp4d} &
    2014 &
    \begin{tabitem}
      \item 328 clips
    \end{tabitem} &
    \begin{tabitem}
      \item 27 AUs
      \item AU intensity
    \end{tabitem} &
    \begin{tabitem}
      \item RGB-D
      \item 2D texture
      \item Audio
    \end{tabitem} &
    \begin{tabitem}
      \item AU intensities
      \item Onset\slash offset
      \item 3D landmarks
    \end{tabitem} &
    AUD &
    V,3D &
    \begin{tabitem}
      \item 41 (23F, 18M)
    \end{tabitem} &
    Spontaneous 3D and RGB-D AU benchmark with 3D landmark annotations \\
    \midrule

    \textbf{CASME~II} \citep{yan2014casme} &
    2014 &
    \begin{tabitem}
      \item 247 clips
    \end{tabitem} &
    \begin{tabitem}
      \item 5 classes
      \item AU set
    \end{tabitem} &
    \begin{tabitem}
      \item RGB video
    \end{tabitem} &
    \begin{tabitem}
      \item Categorical labels
      \item AU annotations
      \item Onset/apex/offset
    \end{tabitem} &
    MER &
    V &
    \begin{tabitem}
      \item 26
    \end{tabitem} &
    Canonical high-frame-rate spontaneous MER benchmark \\
    \midrule

    \textbf{CAS(ME)$^2$} \citep{qu2016cas} &
    2016 &
    \begin{tabitem}
      \item 250 macro
      \item 53 micro clips
    \end{tabitem} &
    \begin{tabitem}
      \item 3 classes
      \item AU set
    \end{tabitem} &
    \begin{tabitem}
      \item RGB video
    \end{tabitem} &
    \begin{tabitem}
      \item Categorical labels
      \item AU annotations
    \end{tabitem} &
    MER &
    V &
    \begin{tabitem}
      \item 22 (16F, 6M)
    \end{tabitem} &
    Macro- and micro-expression spotting and recognition benchmark \\
    \midrule

    \textbf{SAMM} \citep{davison2016samm} &
    2016 &
    \begin{tabitem}
      \item 159 clips
      \item 338 macro\slash micro movements
    \end{tabitem} &
    \begin{tabitem}
      \item AU set
    \end{tabitem} &
    \begin{tabitem}
      \item RGB video
    \end{tabitem} &
    \begin{tabitem}
      \item AU annotations
      \item Onset\slash apex\slash offset
    \end{tabitem} &
    MER &
    V &
    \begin{tabitem}
      \item 32 (16F, 16M)
    \end{tabitem} &
    Spontaneous micro-movement benchmark covering diverse ethnicities \\
    \midrule

    \textbf{4DME} \citep{li20224dme} &
    2022 &
    \begin{tabitem}
      \item $\sim$278 videos
      \item 1{,}068 micro
      \item 492 macro
    \end{tabitem} &
    \begin{tabitem}
      \item 5 classes
      \item 22 AUs
    \end{tabitem} &
    \begin{tabitem}
      \item DI4D mesh
      \item RGB video
      \item Depth
      \item Grayscale
    \end{tabitem} &
    \begin{tabitem}
      \item AU annotations
      \item Onset\slash apex\slash offset
      \item Categorical labels
    \end{tabitem} &
    MER &
    3D &
    \begin{tabitem}
      \item 65 (27F, 38M)
    \end{tabitem} &
    4D spontaneous MER benchmark combining DI4D, RGB, depth, and grayscale streams \\
    \midrule

    \textbf{CAS(ME)$^3$} \citep{li2022cas} &
    2022 &
    \begin{tabitem}
      \item 3{,}490 macro
      \item 1{,}109 micro
      \item 1{,}508 unlabeled clips
    \end{tabitem} &
    \begin{tabitem}
      \item 7 classes
      \item AU set
    \end{tabitem} &
    \begin{tabitem}
      \item RGB-D videos
    \end{tabitem} &
    \begin{tabitem}
      \item Categorical labels
      \item AU annotations
      \item Onset\slash apex\slash offset
    \end{tabitem} &
    MER &
    V,3D &
    \begin{tabitem}
      \item 247 (135F, 112M)
    \end{tabitem} &
    Largest RGB-D macro- and micro-expression benchmark with depth-aware annotations at the time of its release \\
    \bottomrule
  \end{tabular}%
  }
\end{table*}

\begin{table*}[!t]
  \caption{Public FER datasets acquired under in-the-wild (unconstrained) settings. Abbreviations used: Temporality (Static images (ST), Video (V), 3D/4D (3D)), and Recognition task (static FER (SFER), dynamic FER (DFER), action unit detection (AUD), micro-expression recognition (MER), compound expression recognition (CFER), dimensional valence--arousal estimation (VA)).}
  \label{tab:datasets_fer_itw}
  \centering
  \scriptsize

  \setlength{\aboverulesep}{0pt}
  \setlength{\belowrulesep}{0pt}
  
  \setlength{\tabcolsep}{4pt}
  
  \renewcommand{\arraystretch}{1.3}
  
  \setlength{\tabcellpad}{0.5pt}
  \rowcolors{2}{gray!20}{gray!2}

  \resizebox{\textwidth}{!}{%
  \begin{tabular}{@{}|
    >{\raggedright\arraybackslash}Q{\Dds}| %
    >{\centering\arraybackslash}Q{\Dyr}|
    >{\raggedright\arraybackslash}Q{\Dscale}|
    >{\raggedright\arraybackslash}Q{\Dclass}|
    >{\raggedright\arraybackslash}Q{\Dmoda}|
    >{\raggedright\arraybackslash}Q{\Dann}|
    >{\centering\arraybackslash}Q{\Dtask}|
    >{\centering\arraybackslash}Q{\Dtemp}|
    >{\raggedright\arraybackslash}Q{\Dsubj}|
    Q{\Ddesc}| %
  @{}}
    \toprule
    \rowcolor{gray!50}
    \headerbreak{Dataset} &
    \headerbreak{Year} &
    \headerbreak{Scale} &
    \headerbreak{Classes} &
    \headerbreak{Modalities} &
    \headerbreak{Annotation\\type} &
    \headerbreak{Task} &
    \headerbreak{Temp.} &
    \headerbreak{Subjects} &
    \headerbreak{Description} \\
    \midrule

    \textbf{AFEW} \citep{dhall2011acted} &
    2011 &
    \begin{tabitem}
      \item 957 clips
    \end{tabitem} &
    \begin{tabitem}
      \item 7 classes
    \end{tabitem} &
    \begin{tabitem}
      \item RGB video
    \end{tabitem} &
    \begin{tabitem}
      \item Categorical labels
      \item Movie metadata
    \end{tabitem} &
    DFER &
    V &
    \begin{tabitem}
      \item 220 actors
    \end{tabitem} &
    Early in-the-wild DFER benchmark mined from movie clips \\
    \midrule

    \textbf{SFEW 2.0} \citep{dhall2011static} &
    2011 &
    \begin{tabitem}
      \item 1{,}766 images (v2.0; about 700 in the original 2011 release)
    \end{tabitem} &
    \begin{tabitem}
      \item 7 classes
    \end{tabitem} &
    \begin{tabitem}
      \item RGB images
    \end{tabitem} &
    \begin{tabitem}
      \item Categorical labels
    \end{tabitem} &
    SFER &
    ST &
    \begin{tabitem}
      \item N\slash A
    \end{tabitem} &
    Static frames from AFEW for in-the-wild SFER evaluation; version 2.0 was released for the EmotiW 2015 challenge \\
    \midrule

    \textbf{FER-2013} \citep{goodfellow2013challenges} &
    2013 &
    \begin{tabitem}
      \item 35{,}887 images
    \end{tabitem} &
    \begin{tabitem}
      \item 7 classes
    \end{tabitem} &
    \begin{tabitem}
      \item Grayscale images
    \end{tabitem} &
    \begin{tabitem}
      \item Categorical labels
    \end{tabitem} &
    SFER &
    ST &
    \begin{tabitem}
      \item N\slash A
    \end{tabitem} &
    Challenge corpus; one of the first large-scale in-the-wild SFER corpora \\
    \midrule

    \textbf{FER+} \citep{barsoum2016training} &
    2016 &
    \begin{tabitem}
      \item 35{,}887 images
    \end{tabitem} &
    \begin{tabitem}
      \item 8 classes
    \end{tabitem} &
    \begin{tabitem}
      \item Grayscale images
    \end{tabitem} &
    \begin{tabitem}
      \item Crowdsourced label distributions
    \end{tabitem} &
    SFER &
    ST &
    \begin{tabitem}
      \item N\slash A
    \end{tabitem} &
    Crowdsourced relabeling of FER-2013 supporting label-distribution learning \\
    \midrule

    \textbf{EmotioNet} \citep{fabian2016emotionet} &
    2016 &
    \begin{tabitem}
      \item $\sim$1{,}000{,}000 images
    \end{tabitem} &
    \begin{tabitem}
      \item 11 AUs
      \item 16 basic\slash compound classes
    \end{tabitem} &
    \begin{tabitem}
      \item RGB images
    \end{tabitem} &
    \begin{tabitem}
      \item AU annotations
      \item Categorical and compound labels
    \end{tabitem} &
    SFER, AUD, CFER &
    ST &
    \begin{tabitem}
      \item N\slash A
    \end{tabitem} &
    Million-image AU and compound-expression benchmark for in-the-wild FER, with automatically annotated AUs and a smaller manually annotated subset \\
    \midrule

    \textbf{RAF-DB} \citep{li2017reliable} &
    2017 &
    \begin{tabitem}
      \item 30{,}000 images
    \end{tabitem} &
    \begin{tabitem}
      \item 7 basic classes
      \item Compound classes
    \end{tabitem} &
    \begin{tabitem}
      \item RGB images
    \end{tabitem} &
    \begin{tabitem}
      \item Categorical labels
      \item Compound labels
      \item Facial landmarks
    \end{tabitem} &
    SFER, CFER &
    ST &
    \begin{tabitem}
      \item N\slash A
    \end{tabitem} &
    Real-world Affective Faces Database with reliable crowdsourced multi-label annotations \\
    \midrule

    \textbf{AffectNet} \citep{mollahosseini2017affectnet} &
    2017 &
    \begin{tabitem}
      \item $\sim$1{,}000{,}000 images
      \item 404K labeled
    \end{tabitem} &
    \begin{tabitem}
      \item 8 classes
      \item Continuous VA
    \end{tabitem} &
    \begin{tabitem}
      \item RGB images
    \end{tabitem} &
    \begin{tabitem}
      \item Categorical labels
      \item Continuous VA values
    \end{tabitem} &
    SFER, VA &
    ST &
    \begin{tabitem}
      \item N\slash A
    \end{tabitem} &
    Largest manually annotated in-the-wild SFER and continuous VA benchmark at the time of its release \\
    \midrule

    \textbf{ExpW} \citep{zhang2018facial} &
    2018 &
    \begin{tabitem}
      \item 91{,}793 images
    \end{tabitem} &
    \begin{tabitem}
      \item 7 classes
    \end{tabitem} &
    \begin{tabitem}
      \item RGB images
    \end{tabitem} &
    \begin{tabitem}
      \item Categorical labels
    \end{tabitem} &
    SFER &
    ST &
    \begin{tabitem}
      \item N\slash A
    \end{tabitem} &
    Web-image SFER corpus with strong demographic and pose variation \\
    \midrule

    \textbf{Aff-Wild2} \citep{kollias2018aff} &
    2018 &
    \begin{tabitem}
      \item 564 videos
    \end{tabitem} &
    \begin{tabitem}
      \item 8 classes
      \item Continuous VA
      \item AU set
    \end{tabitem} &
    \begin{tabitem}
      \item RGB video
      \item Audio
    \end{tabitem} &
    \begin{tabitem}
      \item Categorical labels
      \item Continuous VA values
      \item AU annotations
    \end{tabitem} &
    SFER, DFER, AUD, VA &
    V &
    \begin{tabitem}
      \item 455 (277M, 178F)
    \end{tabitem} &
    Unified multi-task in-the-wild affect benchmark underpinning the ABAW challenges; scale of the extended version used in ABAW 2022 \citep{kollias2022abaw} \\
    \midrule

    \textbf{RAF-ML} \citep{li2019blended} &
    2019 &
    \begin{tabitem}
      \item 4{,}908 images
    \end{tabitem} &
    \begin{tabitem}
      \item Multi-label blended emotions
    \end{tabitem} &
    \begin{tabitem}
      \item RGB images
    \end{tabitem} &
    \begin{tabitem}
      \item Multi-label distributions
      \item Facial landmarks
    \end{tabitem} &
    CFER &
    ST &
    \begin{tabitem}
      \item N\slash A
    \end{tabitem} &
    Multi-label blended-emotion extension of RAF-DB \\
    \midrule

    \textbf{DFEW} \citep{jiang2020dfew} &
    2020 &
    \begin{tabitem}
      \item 16{,}372 clips
    \end{tabitem} &
    \begin{tabitem}
      \item 7 classes
      \item Distribution vectors
    \end{tabitem} &
    \begin{tabitem}
      \item RGB video
    \end{tabitem} &
    \begin{tabitem}
      \item Categorical labels
      \item Soft label distributions
    \end{tabitem} &
    DFER &
    V &
    \begin{tabitem}
      \item N\slash A
    \end{tabitem} &
    Large-scale in-the-wild DFER benchmark with multi-annotator label distributions \\
    \midrule

    \textbf{FERV39k} \citep{wang2022ferv39k} &
    2022 &
    \begin{tabitem}
      \item 39{,}000 videos
    \end{tabitem} &
    \begin{tabitem}
      \item 7 classes
    \end{tabitem} &
    \begin{tabitem}
      \item RGB video
    \end{tabitem} &
    \begin{tabitem}
      \item Categorical labels
    \end{tabitem} &
    DFER &
    V &
    \begin{tabitem}
      \item N\slash A
    \end{tabitem} &
    Largest in-the-wild DFER benchmark at the time of its release, covering diverse scenes and demographics \\
    \midrule

    \textbf{MAFW} \citep{liu2022mafw} &
    2022 &
    \begin{tabitem}
      \item 10{,}045 clips
    \end{tabitem} &
    \begin{tabitem}
      \item 11 single-expression classes
      \item Compound emotions
    \end{tabitem} &
    \begin{tabitem}
      \item RGB video
      \item Audio
      \item Captions
    \end{tabitem} &
    \begin{tabitem}
      \item Categorical and compound labels
      \item Captions
    \end{tabitem} &
    DFER, CFER &
    V &
    \begin{tabitem}
      \item N\slash A
    \end{tabitem} &
    Multi-modal compound-emotion DFER benchmark with textual descriptions \\
    \midrule

    \textbf{RAF-CE} \citep{10208819} &
    2023 &
    \begin{tabitem}
      \item 4{,}549 images
    \end{tabitem} &
    \begin{tabitem}
      \item 14 compound emotions
      \item 32 AUs
    \end{tabitem} &
    \begin{tabitem}
      \item RGB images
    \end{tabitem} &
    \begin{tabitem}
      \item Compound labels
      \item AU annotations
    \end{tabitem} &
    CFER, AUD &
    ST &
    \begin{tabitem}
      \item N\slash A
    \end{tabitem} &
    In-the-wild compound-expression benchmark with extensive AU coverage \\
    \midrule

    \textbf{C-EXPR-DB} \citep{kollias2023multi} &
    2023 &
    \begin{tabitem}
      \item 400 videos
      \item 200K annotated frames
    \end{tabitem} &
    \begin{tabitem}
      \item 12 compound classes
      \item Continuous VA
      \item AU set
    \end{tabitem} &
    \begin{tabitem}
      \item Audio-visual (video + speech)
    \end{tabitem} &
    \begin{tabitem}
      \item Compound labels
      \item Continuous VA values
      \item AU annotations
    \end{tabitem} &
    CFER, AUD, VA &
    V &
    \begin{tabitem}
      \item N\slash A
    \end{tabitem} &
    Multi-task compound-affect audio-visual benchmark for ABAW-style evaluation \\

    \bottomrule
  \end{tabular}%
  }
\end{table*}

\section{Public datasets}
\label{sec:Datasets}

Having systematically investigated the FER literature using the criteria of Sections~\ref{sec:recognition_tasks}--\ref{sec:app_domains}, this section outlines the main public datasets that have been introduced so far for developing and evaluating deep learning-based FER methods. Tables~\ref{tab:datasets_fer_lab}--\ref{tab:datasets_fer_itw} group the benchmarks by acquisition setting (laboratory vs.\ in-the-wild). For each entry, the tables report: a) Year (publication year of the cited reference), b) Scale, c) Semantic classes/intensity targets, d) Modalities involved, e) Annotation type, f) Main recognition task (categorical SFER/DFER, dimensional VA estimation, compound expression recognition, MER, and AU detection), g) Temporality (static (ST), video (V), 3D/4D (3D)), h) Subjects, and i) Short description. Apart from per-task remarks already discussed in Sections~\ref{sec:recognition_tasks}--\ref{sec:input_modalities}, the following global observations can be drawn:
\begin{itemize}
    \item \uline{Shift from controlled lab corpora to large-scale in-the-wild benchmarks}: Early lab corpora provided posed material from a few dozen subjects, whereas modern web-scraped or reaction-elicited benchmarks deliver up to about one million images and tens of thousands of video clips under realistic pose, occlusion, illumination, and demographic variability.
    \item \uline{Move from a single recognition task to unified, multi-task affect benchmarks}: Recent benchmarks deliver multiple co-annotated targets (categorical, continuous VA, compound, and AU) within a single corpus, directly supporting unified multi-objective backbones.
    \item \uline{Expansion from static images to dynamic, audio-visual, and multi-modal recordings}: Dataset creation has progressed from grayscale/RGB images to large-scale video corpora and, more recently, to audio-visual and multi-modal (e.g., depth and thermal) benchmarks, supporting temporally aware Transformer, state-space, and MLLM-based FER methods.
    \item \uline{Emergence of 3D, 4D, RGB-D, and thermal/NIR FER datasets}: Several lab benchmarks introduce richer geometric and spectral modalities (3D scans, visible (VIS) and NIR sequences, RGB-D micro-expression recordings, 4D mesh streams), although their scale remains substantially smaller than that of 2D in-the-wild corpora.
    \item \uline{Coverage of compound, micro-, and fine-grained affect categories}: Recent corpora extend FER beyond Ekman's basic emotions with blended-emotion labels, onset/apex/offset MER annotations, and continuous valence--arousal values.
    \item \uline{Persistent demographic and cultural under-representation}: Coverage of demographic axes (ethnicity, age, gender, skin tone) is limited and uneven, since lab corpora draw on narrow participant pools, whereas web-scraped corpora inherit the demographic skew of their sources, motivating cross-cultural evaluation and fairness-aware learning.
    \item \uline{Lack of physiologically and contextually annotated FER data}: The FER landscape still lacks corpora pairing facial expressions with synchronized physiological signals (e.g., EEG, EDA, heart-rate variability) or rich contextual metadata, limiting physiologically grounded and context-aware pipelines.
\end{itemize}

\section{Performance metrics and quantitative analysis}
\label{sec:performance_metrics}

A principled comparison of deep learning-based FER methods requires both a clear definition of the performance metrics adopted in the literature and a systematic quantitative analysis of representative methods on widely used benchmarks. This section first describes the main performance metrics reported in the FER literature, then provides a quantitative comparison of representative recent methods grouped by recognition task (Section~\ref{sec:recognition_tasks}), and subsequently summarizes the main observations and key insights that emerge from the reported results.

\subsection{Performance metrics}
\label{ssec:perf_metrics}

The FER literature relies on a compact but well-established set of metrics, whose exact choice depends on the recognition task (categorical macro-expression recognition, dimensional VA estimation, compound expression recognition, MER, AU detection, or expression intensity estimation), the degree of class imbalance, and the adopted acquisition setting. The most commonly reported performance metrics across deep learning-based FER methods are summarized below:
\begin{itemize}
    \item \uline{Accuracy}: Measures the overall proportion of correctly classified samples among all predictions~\citep{sokolova2009systematic}.
    \item \uline{Precision}: Measures the fraction of samples predicted as positive that are actually positive, capturing the model's reliability in its positive predictions~\citep{sokolova2009systematic}.
    \item \uline{Recall (Sensitivity)}: Measures the fraction of actual positive samples that are correctly identified, capturing the model's coverage of the target class~\citep{sokolova2009systematic}.
    \item \uline{Specificity}: Measures the fraction of actual negative samples that are correctly identified, complementing recall in binary and one-vs-rest classification settings~\citep{sokolova2009systematic}.
    \item \uline{F1-score}: Combines precision and recall through their harmonic mean, providing a single balanced metric whose macro variant (unweighted class-wise average) is preferred over micro-F1 under class imbalance~\citep{sokolova2009systematic}.
    \item \uline{Unweighted F1 (UF1)}: Class-balanced F1 variant computed as the unweighted mean of per-class F1 scores, suitable for highly imbalanced multi-class settings~\citep{see2019megc}.
    \item \uline{Unweighted Average Recall (UAR) / Mean Accuracy}: Class-balanced average of per-class recall, giving equal weight to each class regardless of its prevalence~\citep{schuller2009interspeech}.
    \item \uline{Weighted Average Recall (WAR)}: Class-prevalence-weighted recall, emphasizing the recall of more frequent classes and complementing UAR~\citep{schuller2009interspeech}.
    \item \uline{Standard deviation of accuracy}: Quantifies the variation of accuracy values across repeated runs or cross-validation folds, characterizing the stability of the reported performance~\citep{sokolova2009systematic}.
    \item \uline{Confusion matrix and per-class accuracy}: Tabular and per-class summaries of correct and incorrect predictions, providing fine-grained diagnostic information beyond a single scalar score~\citep{stehman1997selecting}.
    \item \uline{ROC-AUC}: Threshold-independent ranking metric that measures the area under the Receiver Operating Characteristic curve, quantifying the trade-off between true-positive and false-positive rates~\citep{fawcett2006introduction}.
    \item \uline{PR-AUC}: Area under the Precision-Recall curve, more informative than ROC-AUC under strong class imbalance~\citep{davis2006relationship}.
    \item \uline{Cohen's kappa}: Chance-corrected agreement metric between two label sets (e.g., predictions vs.\ ground truth or two annotators), accounting for the agreement expected by random labeling~\citep{cohen1960coefficient}.
    \item \uline{Concordance Correlation Coefficient (CCC)}: Jointly measures the linear correlation and the deviation from the diagonal between predicted and ground-truth continuous values~\citep{lin1989ccc}.
    \item \uline{Pearson Correlation Coefficient (PCC)}: Measures the strength of the linear relationship between predicted and ground-truth continuous values, capturing precision without penalizing systematic bias~\citep{benesty2009pearson}.
    \item \uline{Sign Agreement Metric (SAGR)}: Fraction of predictions whose sign matches that of the ground truth, capturing the directional/quadrant-level correctness of continuous-valued predictions~\citep{nicolaou2011continuous}.
    \item \uline{Root Mean Square Error (RMSE) and mean absolute error}: Quantify the average error magnitude of continuous-valued predictions, with RMSE penalizing larger errors more strongly than the mean absolute error~\citep{willmott2005advantages}.
    \item \uline{Intra-Class Correlation Coefficient (ICC)}: Measures the agreement between predicted and ground-truth values on an ordinal or continuous scale, accounting for both correlation and rater bias~\citep{shrout1979intraclass}.
    \item \uline{Efficiency metrics (Params, FLOPs, FPS, latency)}: Quantify the computational footprint of a model in terms of trainable parameters, floating-point operations, frames per second (FPS), and per-frame latency, complementing predictive-accuracy metrics with deployment-oriented information~\citep{canziani2016analysis}.
\end{itemize}

\newlength{\Pyr}\setlength{\Pyr}{0.55cm}
\newlength{\Pmth}\setlength{\Pmth}{2.6cm}
\newlength{\Pds}\setlength{\Pds}{1.4cm}
\newlength{\Pdsw}\setlength{\Pdsw}{1.7cm}
\newlength{\Pdsnar}\setlength{\Pdsnar}{1.25cm}
\newlength{\Pdstwo}\setlength{\Pdstwo}{6.0cm}
\newlength{\Pdsone}\setlength{\Pdsone}{12.0cm}

\begin{table*}[!t]
\newcommand{\hdr}[1]{\begin{minipage}[c][1.15cm][c]{\linewidth}\centering\textbf{#1}\end{minipage}}
\caption{Quantitative comparison of representative SFER methods on widely adopted in-the-wild benchmarks. For each method, top-1 classification accuracy (\%) is reported as in the original publications. Since the values rest on heterogeneous experimental protocols, bold denotes the highest reported value per column and does not constitute a ranking. AffectNet results are reported separately for the 7- and the 8-class protocol (validation set), while RAF-DB, FER+, and FER-2013 results refer to the official test sets; bold is computed only over fully supervised entries with a known class set, and `Setting' reports the training regime or metric whenever it deviates from full supervision and overall accuracy. *: Class set (7 or 8) not stated in the accessible sources. Backbone (pre-training): Network and source of its pre-trained weights (MS1M: MS-Celeb-1M; IN: ImageNet; K400: Kinetics-400; VC2: VoxCeleb2; WIT: CLIP image--text pre-training data; MAE: Masked autoencoding); Mod.: Input modality (V: Visual; A: Audio; L: Language); Prior: Landmark (Lmk) or action unit (AU) prior used as input, supervision, or prompt; Eval.: W, Within-dataset (training and test on the same corpus), C,Cross-dataset; n.r.: not reported; --: Not applicable or not evaluated.}
\label{tab:perf_sfer}
\centering
\scriptsize
\setlength{\aboverulesep}{0pt}
\setlength{\belowrulesep}{0pt}
\setlength{\tabcolsep}{2pt}
\renewcommand{\arraystretch}{1.3}
\rowcolors{2}{gray!20}{gray!2}
\begin{adjustbox}{max width=\textwidth}
\begin{tabular}{@{}|>{\raggedright\arraybackslash}m{2.05cm}|>{\centering\arraybackslash}m{0.60cm}|>{\raggedright\arraybackslash}m{2.90cm}|>{\centering\arraybackslash}m{1.20cm}|>{\centering\arraybackslash}m{0.60cm}|>{\centering\arraybackslash}m{0.65cm}|>{\raggedright\arraybackslash}m{2.00cm}|>{\centering\arraybackslash}m{1.15cm}|>{\centering\arraybackslash}m{1.15cm}|>{\centering\arraybackslash}m{1.15cm}|>{\centering\arraybackslash}m{1.15cm}|>{\centering\arraybackslash}m{1.15cm}|>{\centering\arraybackslash}m{0.60cm}|@{}}
\toprule
\rowcolor{gray!50}
\hdr{Method} & \hdr{Year} & \hdr{Backbone (pre-training)} & \hdr{Input} & \hdr{Mod.} & \hdr{Prior} & \hdr{Setting} & \hdr{AffectNet~\citep{mollahosseini2017affectnet}\\(7 cl.)} & \hdr{AffectNet~\citep{mollahosseini2017affectnet}\\(8 cl.)} & \hdr{RAF-DB\\\citep{li2017reliable}} & \hdr{FER+\\\citep{barsoum2016training}} & \hdr{FER-2013\\\citep{goodfellow2013challenges}} & \hdr{Eval.} \\
\midrule
MLCNN~\citep{nguyen2019facial} & 2019 & Multi-level CNN ensemble & n.r. & V & -- & Fully sup. & -- & -- & -- & -- & \textbf{74.09} & W \\
\midrule
BReG-NeXt~\citep{hasani2020breg} & 2020 & BReG-NeXt-50 (scratch) & 64$\times$64 & V & -- & Fully sup. & -- & \textbf{68.50} & -- & -- & 71.53 & W \\
\midrule
DSAN~\citep{fan2020facial} & 2020 & VGG-16 (pre-trained) + race labels & 224$\times$224 & V & -- & Fully sup. & -- & -- & 85.37 & -- & -- & W \\
\midrule
SPWFA-SE~\citep{li2020facial} & 2020 & ResNet-50 (IN) + SE & n.r. & V & -- & Fully sup. & 59.23 & -- & 86.31 & -- & -- & W \\
\midrule
Deep SNN~\citep{hayale2021deep} & 2021 & Siamese DenseNet-121 (n.r.) & 128$\times$128 & V & -- & Fully sup. & 64.40 & -- & -- & -- & 73.00 & W \\
\midrule
VTFF~\citep{ma2021facial} & 2021 & 2$\times$ResNet-18 (MS1M) + Transf. & 224$\times$224 & V & -- & Fully sup.; AffectNet mean class acc. & -- & 61.85 & 88.14 & 88.81 & -- & W \\
\midrule
Ada-CM~\citep{li2022towards} & 2022 & ResNet-18 (MS1M) & 224$\times$224 & V & -- & Semi-sup.\ (4k/10k labels) & -- & 57.42 & 84.42 & -- & -- & W \\
\midrule
AVT~\citep{jin2022avt} & 2022 & ResNet-26 + ViT (pre-trained) & 224$\times$224 & V & AU & Fully sup. & -- & -- & 89.83 & 90.45 & -- & W \\
\midrule
CERN~\citep{gera2022cern} & 2022 & LightCNN-29v2 (face rec.) & n.r. & V & -- & Fully sup. & 62.06$^{*}$ & -- & 87.09 & 88.17 & -- & W \\
\midrule
CMCNN~\citep{yu2022co} & 2022 & Multi-task CNN (scratch) & 100$\times$100 & V & Lmk & Fully sup. & -- & -- & 85.22 & -- & -- & W \\
\midrule
CRN~\citep{ZHU2022116046} & 2022 & Relation CNN (n.r.) & n.r. & V & -- & Few-shot; mean over unseen classes & -- & -- & 56.25 & -- & 67.32 & W \\
\midrule
EAC~\citep{zhang2022learn} & 2022 & ResNet-18 (MS1M) & 224$\times$224 & V & -- & Fully sup. & 65.32 & -- & 89.99 & 89.64 & -- & W \\
\midrule
Face2Exp~\citep{zeng2022face2exp} & 2022 & ResNet-50 (n.r.) & 224$\times$224 & V & -- & 10\% AffectNet labels; mean class acc. & 64.23 & -- & 88.54 & -- & -- & W \\
\midrule
FLEPNet~\citep{karnati2022flepnet} & 2022 & Multi-scale CNN ensemble & n.r. & V & -- & 10-fold CV (not official split) & -- & -- & 87.56 & -- & 80.72 & W \\
\midrule
IPD-FER~\citep{jiang2022disentangling} & 2022 & ResNet-18 (CASIA-WebFace) & 112$\times$112 & V & -- & Fully sup. & 62.23 & -- & 88.89 & 88.42 & -- & W \\
\midrule
APViT~\citep{xue2022vision} & 2022 & IR-50 (MS1M) + ViT-S (IN) & 112$\times$112 & V & -- & Fully sup. & 66.91 & -- & 91.98 & 90.86 & -- & W \\
\midrule
POSTER~\citep{zheng2023poster} & 2023 & IR-50 (MS1M) + MobileFaceNet & n.r. & V & Lmk & Fully sup. & 67.31 & 63.34 & 92.05 & \textbf{91.62} & -- & W \\
\midrule
TAN~\citep{ma2023transformer} & 2023 & ResNet-18 (MS1M) + Transformers & 224$\times$224 & V & -- & Fully sup.; AffectNet mean class acc. & 65.17 & -- & 89.12 & 90.67 & -- & W \\
\midrule
MFER~\citep{xu2024multiscale} & 2024 & n.r. & n.r. & V & -- & Fully sup. & 67.06 & 63.15 & 92.08 & 91.09 & -- & W \\
\midrule
S2D~\citep{chen2024static} & 2024 & ViT-B/16 (MAE, AffectNet) & 224$\times$224 & V & Lmk & Fully sup. & 67.62 & 63.06 & \textbf{92.57} & 91.17 & -- & W \\
\midrule
FER-Former~\citep{li2024fer} & 2024 & IR-50 (MS1M) + CLIP text enc. & 112$\times$112 & V+L & -- & Fully sup. & -- & -- & 91.30 & 90.96 & -- & W \\
\midrule
POSTER++~\citep{mao2025posterpp} & 2025 & IR-50 (MS1M) + MobileFaceNet & n.r. & V & Lmk & Fully sup. & 67.49 & 63.77 & 92.21 & -- & -- & W \\
\midrule
QWTR~\citep{zhou2025delving} & 2025 & QResNet-18 (MS1M) + quaternion ViT & 100$\times$100 & V & -- & Fully sup. & \textbf{68.37} & -- & 91.31 & -- & -- & W \\
\midrule
EACM~\citep{li2025enhanced} & 2025 & ResNet-18; WRN-28-2 for FER+ & 224$\times$224 & V & -- & Semi-sup.\ (4k/10k/4k labels) & -- & 57.80 & 85.03 & 80.28 & -- & W \\
\midrule
NHG~\citep{min2026robust} & 2026 & ResNet-18 (MS1M) + GCN & n.r. & V & -- & Fully sup. & 65.14 & -- & 90.09 & 88.94 & -- & W \\
\midrule
MHAN~\citep{wang2026mhan} & 2026 & MobileFaceNet-based (MS1M) & n.r. & V & -- & Fully sup. & 67.74 & 65.08 & \textbf{92.57} & 90.90 & -- & W \\
\bottomrule
\end{tabular}
\end{adjustbox}
\end{table*}

\begin{table*}[!t]
\newcommand{\hdr}[1]{\begin{minipage}[c][1.15cm][c]{\linewidth}\centering\textbf{#1}\end{minipage}}
\caption{Quantitative comparison of representative DFER methods on widely adopted lab-controlled (MMI, Oulu-CASIA; top-1 accuracy) and in-the-wild (AFEW; top-1 accuracy; DFEW, FERV39k, MAFW; Weighted Average Recall, WAR) dynamic FER benchmarks. Performance values (\%) are reported as in the original publications. Since the values rest on heterogeneous experimental protocols, bold denotes the highest reported value per column and does not constitute a ranking. Protocols: MMI and Oulu-CASIA, subject-independent 10-fold cross-validation (6 classes); AFEW, official validation set (7 classes); DFEW, 5-fold cross-validation (7 classes); FERV39k, official split (7 classes); MAFW, 5-fold cross-validation (11 classes). Input: Number of frames (fr) per clip and spatial resolution. $^{\ddag}$Two clips sampled per test video and scores averaged; $^{\S}$value reported by the dataset papers or by later works (re-implementation), not by the original publication. Abbreviations as in Table~\ref{tab:perf_sfer}.}
\label{tab:perf_dfer}
\centering
\scriptsize
\setlength{\aboverulesep}{0pt}
\setlength{\belowrulesep}{0pt}
\setlength{\tabcolsep}{2pt}
\renewcommand{\arraystretch}{1.3}
\rowcolors{2}{gray!20}{gray!2}
\begin{adjustbox}{max width=\textwidth}
\begin{tabular}{@{}|>{\raggedright\arraybackslash}m{2.05cm}|>{\centering\arraybackslash}m{0.60cm}|>{\raggedright\arraybackslash}m{2.80cm}|>{\centering\arraybackslash}m{1.60cm}|>{\centering\arraybackslash}m{0.60cm}|>{\centering\arraybackslash}m{0.60cm}|>{\centering\arraybackslash}m{1.45cm}|>{\centering\arraybackslash}m{1.45cm}|>{\centering\arraybackslash}m{1.45cm}|>{\centering\arraybackslash}m{1.45cm}|>{\centering\arraybackslash}m{1.45cm}|>{\centering\arraybackslash}m{1.45cm}|>{\centering\arraybackslash}m{0.60cm}|@{}}
\toprule
\rowcolor{gray!50}
\hdr{Method} & \hdr{Year} & \hdr{Backbone (pre-training)} & \hdr{Input} & \hdr{Mod.} & \hdr{Prior} & \hdr{MMI~\citep{pantic2005web}\\(acc.)} & \hdr{Oulu-CASIA~\citep{zhao2011facial}\\(acc.)} & \hdr{AFEW~\citep{dhall2011acted}\\(acc.)} & \hdr{DFEW~\citep{jiang2020dfew}\\(WAR)} & \hdr{FERV39k~\citep{wang2022ferv39k}\\(WAR)} & \hdr{MAFW~\citep{liu2022mafw}\\(WAR)} & \hdr{Eval.} \\
\midrule
C3D~\citep{tran2015c3d} & 2015 & C3D (n.r.) & 16 fr, 112$^2$ & V & -- & -- & -- & -- & 53.54$^{\S}$ & 31.69$^{\S}$ & 42.25$^{\S}$ & W \\
\midrule
PSRNet~\citep{wang2020phase} & 2020 & Spatio-temporal CNN & 30/18 fr (MMI/Oulu) & V & -- & 85.23 & \textbf{92.50} & -- & -- & -- & -- & W \\
\midrule
STCAM~\citep{chen2020stcam} & 2020 & 3D-Incep.-ResNet (K400) & 7 fr, 224$^2$ & V & -- & 82.21 & 91.25 & -- & -- & -- & -- & W \\
\midrule
Former-DFER~\citep{zhao2021former} & 2021 & Conv.\ + Transformer (n.r.) & 16 fr, 112$^2$ & V & -- & -- & -- & -- & 65.70 & 46.85$^{\S}$ & 43.27$^{\S}$ & W \\
\midrule
NR-DFERNet~\citep{li2022nrdfernet} & 2022 & Conv.\ + Transformer (n.r.) & 16 fr, 112$^2$ & V & -- & -- & -- & -- & 68.19 & -- & -- & W \\
\midrule
EST~\citep{liu2023expression} & 2023 & ResNet-18 + Transformer & 7$\times$5 fr, 224$^2$ & V & -- & \textbf{92.50} & -- & \textbf{54.26} & 65.85 & -- & -- & W \\
\midrule
IAL~\citep{li2023ial} & 2023 & ResNet-18 + Transf.\ (scratch) & 16 fr, 112$^2$ & V & -- & -- & -- & -- & 69.24 & 48.54 & -- & W \\
\midrule
LOGO-Former~\citep{ma2023logoformer} & 2023 & ResNet-18 (MS1M) + Transf. & n.r. & V & -- & -- & -- & -- & 66.98 & 48.13 & -- & W \\
\midrule
M3DFEL~\citep{wang2023rethinking} & 2023 & R3D-18 (K400) + BiLSTM & 16 fr & V & -- & -- & -- & -- & 69.25 & 47.67 & -- & W \\
\midrule
DFER-CLIP~\citep{zhao2023prompting} & 2023 & CLIP ViT-B/32 (WIT) & 16 fr, 224$^2$ & V+L & -- & -- & -- & -- & 71.25 & 51.65 & 52.55 & W \\
\midrule
MAE-DFER~\citep{sun2023maedfer} & 2023 & ViT (MAE, VC2 video) & 16 fr, 160$^2$ & V & -- & -- & -- & -- & 74.43 & 52.07 & 54.31 & W \\
\midrule
S2D~\citep{chen2024static} & 2024 & ViT-B/16 (MAE, AffectNet) & 16 fr, 224$^2$ & V & Lmk & -- & -- & -- & 76.03$^{\ddag}$ & \textbf{52.56$^{\ddag}$} & 57.37$^{\ddag}$ & W \\
\midrule
MMA-DFER~\citep{chumachenko2024mma} & 2024 & ViT-B (MAE-Face) + AudioMAE & 16 fr, 224$^2$ + audio & A+V & -- & -- & -- & -- & \textbf{77.51} & -- & \textbf{58.45$^{\ddag}$} & W \\
\midrule
TG-DFER~\citep{jung2025text} & 2025 & CLIP ViT-B/32 (WIT) & 16 fr, 224$^2$ & V+L & -- & -- & -- & -- & 71.62 & 51.67 & -- & W \\
\midrule
PE-CLIP~\citep{saadi2025peclip} & 2026 & CLIP ViT-B/16 (WIT) & 16 fr, 224$^2$ & V+L & AU & -- & -- & -- & 74.04 & 51.26 & -- & W \\
\midrule
HDF~\citep{cui2025learning} & 2025 & X3D (K400) & 16 fr, 160$^2$ & V & -- & -- & -- & -- & 71.60 & 50.30 & -- & W \\
\bottomrule
\end{tabular}
\end{adjustbox}
\end{table*}

\begin{table*}[!t]
\newcommand{\hdr}[1]{\begin{minipage}[c][1.15cm][c]{\linewidth}\centering\textbf{#1}\end{minipage}}
\caption{Quantitative comparison of representative dimensional VA estimation methods on Aff-Wild2 (ABAW competition validation split) and AffectNet (continuous VA). For each method, the Concordance Correlation Coefficient (CCC) of valence (V) and arousal (A) is reported as in the original publications. Since the values rest on heterogeneous experimental protocols, bold denotes the highest reported value per column and does not constitute a ranking. Protocol: ABAW edition and split, which differ in the number and partitioning of the Aff-Wild2 videos; all values are obtained by single models. Abbreviations as in Table~\ref{tab:perf_sfer}.}
\label{tab:perf_va}
\centering
\scriptsize
\setlength{\aboverulesep}{0pt}
\setlength{\belowrulesep}{0pt}
\setlength{\tabcolsep}{2pt}
\renewcommand{\arraystretch}{1.3}
\rowcolors{2}{gray!20}{gray!2}
\begin{adjustbox}{max width=\textwidth}
\begin{tabular}{@{}|>{\raggedright\arraybackslash}m{2.20cm}|>{\centering\arraybackslash}m{0.60cm}|>{\raggedright\arraybackslash}m{3.10cm}|>{\centering\arraybackslash}m{1.50cm}|>{\centering\arraybackslash}m{0.55cm}|>{\raggedright\arraybackslash}m{2.30cm}|>{\centering\arraybackslash}m{0.85cm}|>{\centering\arraybackslash}m{0.85cm}|>{\centering\arraybackslash}m{0.85cm}|>{\centering\arraybackslash}m{0.85cm}|>{\centering\arraybackslash}m{0.55cm}|@{}}
\toprule
\rowcolor{gray!50}
\multicolumn{1}{|c|}{\multirow{2}{*}{\textbf{Method}}} & \multirow{2}{*}{\textbf{Year}} & \multirow{2}{*}{\textbf{Backbone (pre-training)}} & \multirow{2}{*}{\textbf{Input}} & \multirow{2}{*}{\textbf{Mod.}} & \multirow{2}{*}{\textbf{Protocol}} & \multicolumn{2}{c|}{\textbf{Aff-Wild2~\citep{kollias2018aff}}} & \multicolumn{2}{c|}{\textbf{AffectNet~\citep{mollahosseini2017affectnet}}} & \multirow{2}{*}{\textbf{Eval.}} \\[2pt]
\rowcolor{gray!50}
 & & & & & & \textbf{V} & \textbf{A} & \textbf{V} & \textbf{A} & \\
\midrule
EmotiEffNets~\citep{savchenko2023emotieffnets} & 2023 & EfficientNet-B0 (VGGFace2, AffectNet) & Frame crops & V & ABAW5 val.; single model & 0.490 & 0.604 & -- & -- & W \\
\midrule
Savchenko et al.~\citep{savchenko2024leveraging} & 2024 & MT-DDAMFN (VGGFace2, AffectNet) & Frame crops & V & ABAW6 val.; AffectNet val. & 0.519 & 0.616 & \textbf{0.729} & \textbf{0.643} & W \\
\midrule
MAE+TCN~\citep{zhou2024enhancing} & 2024 & ViT-B (MAE, face images) + TCN & 300-fr segments & V & ABAW6 val.\ (fold 0) & 0.539 & \textbf{0.622} & -- & -- & W \\
\midrule
CAGE~\citep{wagner2024cage} & 2024 & MaxViT-T (IN) & n.r. & V & AffectNet-8 val. & -- & -- & 0.716 & 0.642 & W \\
\midrule
Mamba-VA~\citep{liang2025mamba} & 2025 & ViT-L (MAE) + TCN + Mamba & Frame segments & V & ABAW8 val.\ (fold 0) & \textbf{0.545} & 0.385 & -- & -- & W \\
\bottomrule
\end{tabular}
\end{adjustbox}
\end{table*}

\begin{table*}[!t]
\newcommand{\hdr}[1]{\begin{minipage}[c][1.15cm][c]{\linewidth}\centering\textbf{#1}\end{minipage}}
\caption{Quantitative comparison of representative compound expression recognition methods. Since the reviewed methods are evaluated on different compound-expression benchmarks, class sets, and learning settings, each row reports the benchmark and protocol together with the corresponding metric (\%), as in the original publications; hence, the values are not comparable across rows and no entry is highlighted. CFEE: Compound Facial Expressions of Emotion database~\citep{du2014compound}; RAF-CE: Compound set built by the authors of MML from RAF-ML and RAF-AU. Abbreviations as in Table~\ref{tab:perf_sfer}.}
\label{tab:perf_cfer}
\centering
\scriptsize
\setlength{\aboverulesep}{0pt}
\setlength{\belowrulesep}{0pt}
\setlength{\tabcolsep}{2pt}
\renewcommand{\arraystretch}{1.3}
\rowcolors{2}{gray!20}{gray!2}
\begin{adjustbox}{max width=\textwidth}
\begin{tabular}{@{}|>{\raggedright\arraybackslash}m{1.80cm}|>{\centering\arraybackslash}m{0.55cm}|>{\raggedright\arraybackslash}m{3.00cm}|>{\centering\arraybackslash}m{1.30cm}|>{\centering\arraybackslash}m{0.55cm}|>{\centering\arraybackslash}m{0.60cm}|>{\raggedright\arraybackslash}m{4.40cm}|>{\centering\arraybackslash}m{0.80cm}|>{\centering\arraybackslash}m{0.90cm}|>{\centering\arraybackslash}m{0.55cm}|@{}}
\toprule
\rowcolor{gray!50}
\hdr{Method} & \hdr{Year} & \hdr{Backbone (pre-training)} & \hdr{Input} & \hdr{Mod.} & \hdr{Prior} & \hdr{Benchmark and protocol} & \hdr{Metric} & \hdr{Value} & \hdr{Eval.} \\
\midrule
\rowcolor{gray!20}
EGS-Net~\citep{zou2022facial} & 2022 & ResNet-18 (n.r.) & 224$\times$224 & V & -- & Basic-expression corpora $\rightarrow$ CFEE compound, 5-shot & Acc. & 73.79 & C \\
\midrule
\rowcolor{gray!2}
 &  &  &  &  &  & RAF-DB compound, 11 classes & UAR & 55.3 & W \\
\cmidrule[\lightrulewidth]{7-10}
\rowcolor{gray!2}
\multirow{-2}{=}{C-EXPR-NET~\citep{kollias2023multi}} & \multirow{-2}{*}{2023} & \multirow{-2}{=}{ResNet-50 + AU text enc.} & \multirow{-2}{*}{n.r.} & \multirow{-2}{*}{V} & \multirow{-2}{*}{AU} & C-EXPR-DB, multi-label & F1 & 60.0 & W \\
\midrule
\rowcolor{gray!20}
MML~\citep{10208819} & 2023 & ResNet-34 (CASIA-WebFace) & 112$\times$112 & V & AU & RAF-CE (RAF-ML $\cap$ RAF-AU), 14 classes & Acc. & 55.23 & W \\
\bottomrule
\end{tabular}
\end{adjustbox}
\end{table*}

\begin{table*}[!t]
\newcommand{\hdr}[1]{\begin{minipage}[c][1.15cm][c]{\linewidth}\centering\textbf{#1}\end{minipage}}
\caption{Quantitative comparison of representative MER methods on canonical micro-expression benchmarks, as reported in the original publications (\%). Since the values rest on heterogeneous experimental protocols, bold denotes the highest reported value per column and does not constitute a ranking. Protocol: Number of classes for CASME II/SAMM/SMIC-HS and evaluation split (LOSO: Leave-one-subject-out); Metric: Top-1 accuracy (Acc.) or unweighted F1-score (UF1). Values obtained with different class sets, splits, or metrics are not comparable; bold is therefore computed only over accuracies obtained under LOSO with the standard class sets (5 classes for CASME II and SAMM, 3 classes for SMIC-HS) and verified against the original publications. Input: Onset--apex optical flow (flow), frames (fr), or image patches, with the spatial resolution when reported. $^{\dagger}$Handcrafted-feature (non-deep) method, included only as a reference point. Abbreviations as in Table~\ref{tab:perf_sfer}.}
\label{tab:perf_mer}
\centering
\scriptsize
\setlength{\aboverulesep}{0pt}
\setlength{\belowrulesep}{0pt}
\setlength{\tabcolsep}{2pt}
\renewcommand{\arraystretch}{1.3}
\rowcolors{2}{gray!20}{gray!2}
\begin{adjustbox}{max width=\textwidth}
\begin{tabular}{@{}|>{\raggedright\arraybackslash}m{2.15cm}|>{\centering\arraybackslash}m{0.60cm}|>{\raggedright\arraybackslash}m{3.00cm}|>{\centering\arraybackslash}m{1.80cm}|>{\centering\arraybackslash}m{0.95cm}|>{\centering\arraybackslash}m{1.70cm}|>{\centering\arraybackslash}m{0.85cm}|>{\centering\arraybackslash}m{1.35cm}|>{\centering\arraybackslash}m{1.35cm}|>{\centering\arraybackslash}m{1.35cm}|>{\centering\arraybackslash}m{0.60cm}|@{}}
\toprule
\rowcolor{gray!50}
\hdr{Method} & \hdr{Year} & \hdr{Backbone (pre-training)} & \hdr{Input} & \hdr{Prior} & \hdr{Protocol} & \hdr{Metric} & \hdr{CASME II~\citep{yan2014casme}} & \hdr{SAMM~\citep{davison2016samm}} & \hdr{SMIC-HS~\citep{li2013spontaneous}} & \hdr{Eval.} \\
\midrule
TS-AUCNN~\citep{sun2020dynamic} & 2020 & Teacher--student CNN (AU-pre-trained) & GT apex & AU & 5/4 cl.\ LOSO & Acc. & 72.61 & 86.74 & -- & W \\
\midrule
MERASTC~\citep{gupta2021merastc} & 2021 & 2D CNN (scratch) & AU/landmark/ gaze encodings & Lmk+AU & 5/5/3 cl.\ LOSO & Acc. & \textbf{85.40} & \textbf{83.80} & 79.30 & W \\
\midrule
AutoMER~\citep{verma2021automer} & 2021 & Architecture-searched 3D CNN & n.r. & -- & 4/4 cl.\ LOSO & Acc. & 74.15 & 72.45 & -- & W \\
\midrule
FeatRef~\citep{zhou2022feature} & 2022 & 2-stream Inception (scratch) & Flow, 28$^2$ & -- & 5/5/3 cl.\ LOSO & Acc. & 62.85 & 60.13 & 57.90 & W \\
\midrule
KTGSL$^{\dagger}$~\citep{wei2022learning} & 2022 & Handcrafted features & n.r. & -- & 5/5/3 cl.\ LOSO & Acc. & 72.58 & 56.11 & 75.64 & W \\
\midrule
ME-PLAN~\citep{zhao2022me} & 2022 & 3D residual proto.\ net (macro-expr.) & 10 fr, 112$^2$ & -- & 3/3 cl.\ LOSO (composite) & UF1 & 89.41 & 73.58 & -- & W \\
\midrule
MER-SupCon~\citep{zhi2022micro} & 2022 & CNN (sup.\ contrastive) & Dual flow maps & -- & 5/5 cl.\ LOSO & Acc. & 73.58 & 67.65 & -- & W \\
\midrule
SE-DenseNet~\citep{cai2022micro} & 2022 & 3D DenseNet-SE (n.r.) & Apex-augm.\ seq. & -- & Hold-out & Acc. & 82.74 & -- & -- & W \\
\midrule
SLSTT~\citep{zhang2022short} & 2022 & ViT-B/16 (IN) + LSTM & Flow, 384$^2$ & -- & 5/5/3 cl.\ LOSO & Acc. & 75.81 & 72.39 & 73.17 & W \\
\midrule
C3DBed~\citep{pan2023c3dbed} & 2023 & C3D + Transformer (n.r.) & Apex patches & -- & 5/5 cl.\ LOSO & Acc. & 77.64 & 75.73 & -- & W \\
\midrule
FRL-DGT~\citep{zhai2023feature} & 2023 & Conv.\ + Transf.\ (self-sup.) & Onset--apex pair & AU & 5 cl.\ LOSO & Acc. & 75.7 & -- & -- & W \\
\midrule
$\mu$-BERT~\citep{nguyen2023micron} & 2023 & ViT enc.--dec.\ (CASME3 self-sup.) & Frame pairs, 224$^2$ & -- & 5/5 cl.\ LOSO & UF1 & 85.53 & 83.86 & -- & W \\
\midrule
TACL~\citep{wang2023temporal} & 2023 & 3D CNN (MoCo self-sup.) & Flow, 56$^2$ & -- & 7/8/3 cl.\ LOSO & Acc. & 76.30 & 68.38 & 75.61 & W \\
\midrule
SelfME~\citep{fan2023selfme} & 2023 & ViT-S/16 (IN) + motion net & Onset + apex & -- & 3/--/3 cl.\ LOSO & UF1 & 90.78 & -- & 69.72 & W \\
\midrule
HTNet~\citep{wang2024htnet} & 2024 & Hier.\ Transformer (scratch) & Flow, 28$^2$ & Lmk & 3/3/3 cl.\ LOSO & UF1 & 95.32 & 81.31 & 80.49 & W \\
\midrule
SRMCL~\citep{bao2024boosting} & 2024 & ViT (CelebA) + masked recon. & Flow, 224$^2$ & -- & 5/5/3 cl.\ LOSO & Acc. & 83.20 & 74.63 & 78.98 & W \\
\midrule
LTR3O~\citep{zhu2025learning} & 2025 & ResNet-18 & Flow, 112$^2$ & -- & 5/5/3 cl.\ LOSO & Acc. & 81.78 & 80.15 & \textbf{81.10} & W \\
\midrule
MER-CLIP~\citep{liu2025mer} & 2025 & UniFormerV2 (CLIP) + CLIP text & 16 fr, 224$^2$ & Lmk+AU & 5/5 cl.\ LOSO & Acc. & 82.33 & 77.21 & -- & W \\
\midrule
MOL~\citep{shao2025mol} & 2025 & Conv.\ F5C (Aff-Wild2) & 8 fr, 128$^2$ & Lmk & 5/5/3 cl.\ LOSO & Acc. & 79.23 & 76.68 & 80.71 & W \\
\midrule
SODA4MER~\citep{zhang2025dynamic} & 2025 & Deform.\ CNN (self-sup.) + Bi-LSTM & Full sequence & AU & 5/5 cl.\ LOSO & Acc. & 84.18 & 80.30 & -- & W \\
\midrule
FMEDC-MMEL~\citep{zhang2025towards} & 2025 & DenseNet + RNN/ELM ens. & Images & -- & 5 cl.\ (split n.r.) & Acc. & 60.76 & -- & -- & W \\
\midrule
LightmanNet~\citep{wang2024meta} & 2026 & Conv4 + 3D CNN & Consec.\ frames & -- & Few-shot episodes & Acc. & 93.48 & 81.83 & -- & W \\
\bottomrule
\end{tabular}
\end{adjustbox}
\end{table*}

\begin{table*}[!t]
\newcommand{\hdr}[1]{\begin{minipage}[c][1.15cm][c]{\linewidth}\centering\textbf{#1}\end{minipage}}
\caption{Quantitative comparison of representative AU detection methods on the BP4D and DISFA benchmarks. For each method, the average F1-score (\%) across the AU set used in the original publication is reported. Since the values rest on heterogeneous experimental protocols, bold denotes the highest reported value per column and does not constitute a ranking. Protocol: 12 AUs on BP4D and 8 AUs on DISFA, subject-exclusive 3-fold cross-validation. $^{\S}$DISFA model initialized from, or built on, the corresponding BP4D-trained model. Input: Spatial resolution of the aligned face. Abbreviations as in Table~\ref{tab:perf_sfer}.}
\label{tab:perf_au}
\centering
\scriptsize
\setlength{\aboverulesep}{0pt}
\setlength{\belowrulesep}{0pt}
\setlength{\tabcolsep}{2pt}
\renewcommand{\arraystretch}{1.3}
\rowcolors{2}{gray!20}{gray!2}
\begin{adjustbox}{max width=\textwidth}
\begin{tabular}{@{}|>{\raggedright\arraybackslash}m{2.20cm}|>{\centering\arraybackslash}m{0.60cm}|>{\raggedright\arraybackslash}m{3.40cm}|>{\centering\arraybackslash}m{1.40cm}|>{\centering\arraybackslash}m{0.60cm}|>{\centering\arraybackslash}m{1.60cm}|>{\centering\arraybackslash}m{1.60cm}|>{\centering\arraybackslash}m{0.60cm}|@{}}
\toprule
\rowcolor{gray!50}
\hdr{Method} & \hdr{Year} & \hdr{Backbone (pre-training)} & \hdr{Input} & \hdr{Prior} & \hdr{BP4D~\citep{zhang2014bp4d}} & \hdr{DISFA~\citep{mavadati2013disfa}} & \hdr{Eval.} \\
\midrule
DRML~\citep{zhao2016deep} & 2016 & Region CNN (scratch) & 170$\times$170 & -- & 48.30 & 26.70$^{\S}$ & W \\
\midrule
EAC-Net~\citep{li2017eacnet} & 2018 & VGG-19 (IN) & 224$\times$224 & Lmk & 55.90 & 48.50$^{\S}$ & W \\
\midrule
JAA-Net~\citep{shao2018jaa} & 2018 & Multi-scale region CNN (scratch) & 176$\times$176 & Lmk & 60.0 & 56.0$^{\S}$ & W \\
\midrule
ARL~\citep{shao2019facial} & 2019 & VGG-style CNN (scratch) & 176$\times$176 & Lmk & 61.1 & 58.70 & W \\
\midrule
MAL~\citep{li2021meta} & 2021 & ResNet-34 (IN) & 224$\times$224 & -- & 62.2 & 58.0$^{\S}$ & W \\
\midrule
ME-GraphAU~\citep{luo2022megraphau} & 2022 & ResNet-50 (IN) & 224$\times$224 & AU & \textbf{64.7} & \textbf{63.10} & W \\
\midrule
LibreFace~\citep{chang2024libreface} & 2024 & ResNet-18 (MAE-distilled) & 224$\times$224 & -- & 62.0 & -- & W \\
\bottomrule
\end{tabular}
\end{adjustbox}
\end{table*}

\begin{table*}[!t]
\newcommand{\hdr}[1]{\begin{minipage}[c][1.15cm][c]{\linewidth}\centering\textbf{#1}\end{minipage}}
\caption{Quantitative comparison of representative expression intensity estimation methods on the Hume-Reaction benchmark (Emotional Reaction Intensity, ABAW ERI sub-challenge). For each method, the average Pearson Correlation Coefficient (PCC) across the seven reaction categories is reported as in the original publications. Since the values rest on heterogeneous experimental protocols, bold denotes the highest reported value and does not constitute a ranking; since some works report results only on the validation set and others on the test set, bold is computed over test-set values. Abbreviations as in Table~\ref{tab:perf_sfer}. Since the labels of Hume-Reaction are self-reported by the recorded subjects, this benchmark lies at the boundary between FER and FAR (Section~\ref{ssec:rec_tasks-intensity}).}
\label{tab:perf_intensity}
\centering
\scriptsize
\setlength{\aboverulesep}{0pt}
\setlength{\belowrulesep}{0pt}
\setlength{\tabcolsep}{2pt}
\renewcommand{\arraystretch}{1.3}
\rowcolors{2}{gray!20}{gray!2}
\begin{adjustbox}{max width=\textwidth}
\begin{tabular}{@{}|>{\raggedright\arraybackslash}m{2.20cm}|>{\centering\arraybackslash}m{0.60cm}|>{\raggedright\arraybackslash}m{3.20cm}|>{\centering\arraybackslash}m{1.70cm}|>{\centering\arraybackslash}m{0.60cm}|>{\centering\arraybackslash}m{0.60cm}|>{\centering\arraybackslash}m{0.90cm}|>{\centering\arraybackslash}m{1.90cm}|>{\centering\arraybackslash}m{0.60cm}|@{}}
\toprule
\rowcolor{gray!50}
\hdr{Method} & \hdr{Year} & \hdr{Backbone (pre-training)} & \hdr{Input} & \hdr{Mod.} & \hdr{Prior} & \hdr{Split} & \hdr{Hume-Reaction~\citep{kollias2023abaw}} & \hdr{Eval.} \\
\midrule
Qian et al.~\citep{qian2023computer} & 2023 & ResNet-50 (n.r.) + Transf. & 32 fr, 112$^2$ & V & -- & Test & 0.3120 & W \\
\midrule
MTL-DAN~\citep{oh2023human} & 2023 & ResNet-18 DAN (Aff-Wild2) + LSTM & $\approx$12 fr/video & V & -- & Val. & 0.3254 & W \\
\midrule
MMA-MRNNet~\citep{kollias2024mma} & 2025 & Multi-task CNN + GRU & all fr, 112$^2$ & V & AU & Test & \textbf{0.5310} & W \\
\bottomrule
\end{tabular}
\end{adjustbox}
\end{table*}

\begin{table*}[!t]
\newcommand{\hdr}[1]{\begin{minipage}[c][1.15cm][c]{\linewidth}\centering\textbf{#1}\end{minipage}}
\caption{Cross-dataset results reported by the reviewed studies (training on the source corpus and testing on a different target corpus, without target labels). Setting: UDA, unsupervised domain adaptation with unlabeled target data; direct transfer, model trained on the source only; zero-shot, no training on the expression labels of the target task. Within: Result of the same model when trained and tested on the target corpus, as reported in the same publication, where available (for S2D, the SFER model without AffectNet-8 pre-training, whose within-dataset values differ from those of Table~\ref{tab:perf_sfer}). Values (\%) are reported as in the original publications. $^{a}$ResNet-50 baseline trained on the source only, as reported by AGRA~\citep{chen2021cross}. Abbreviations as in Table~\ref{tab:perf_sfer}.}
\label{tab:perf_cross}
\centering
\scriptsize
\setlength{\aboverulesep}{0pt}
\setlength{\belowrulesep}{0pt}
\setlength{\tabcolsep}{2pt}
\renewcommand{\arraystretch}{1.3}
\rowcolors{2}{gray!20}{gray!2}
\begin{adjustbox}{max width=\textwidth}
\begin{tabular}{@{}|>{\raggedright\arraybackslash}m{1.90cm}|>{\raggedright\arraybackslash}m{2.80cm}|>{\centering\arraybackslash}m{1.50cm}|>{\raggedright\arraybackslash}m{4.90cm}|>{\centering\arraybackslash}m{0.80cm}|>{\centering\arraybackslash}m{3.20cm}|>{\centering\arraybackslash}m{1.50cm}|@{}}
\toprule
\rowcolor{gray!50}
\hdr{Method} & \hdr{Backbone (pre-training)} & \hdr{Setting} & \hdr{Source $\rightarrow$ target} & \hdr{Metric} & \hdr{Cross-dataset} & \hdr{Within} \\
\midrule
\rowcolor{gray!20}
Source only$^{a}$~\citep{chen2021cross} & ResNet-50 (MS1M) & Direct transfer & RAF-DB $\rightarrow$ CK+ / JAFFE / SFEW~2.0 / FER-2013 / ExpW & Acc. & 71.32 / 50.23 / 50.46 / 54.49 / 67.45 & -- \\
\midrule
\rowcolor{gray!2}
AGRA~\citep{chen2021cross} & ResNet-50 (MS1M) & UDA & RAF-DB $\rightarrow$ CK+ / JAFFE / SFEW~2.0 / FER-2013 / ExpW & Acc. & 85.27 / 61.50 / 56.43 / 58.95 / 68.50 & -- \\
\midrule
\rowcolor{gray!20}
CGLRL~\citep{xie2022crossdomain} & ResNet-50 & UDA & RAF-DB $\rightarrow$ SFEW~2.0 / FER-2013 / ExpW & Acc. & 56.88 / 59.30 / 70.02 & -- \\
\midrule
\rowcolor{gray!2}
VTFF~\citep{ma2021facial} & 2$\times$ResNet-18 (MS1M) & Direct transfer & RAF-DB / FER+ / AffectNet-8 $\rightarrow$ CK+ & Acc. & 81.88 / 83.79 / 86.24 & -- \\
\midrule
\rowcolor{gray!20}
TAN~\citep{ma2023transformer} & ResNet-18 (MS1M) + Transf. & Direct transfer & RAF-DB / FER+ / AffectNet-7 $\rightarrow$ CK+ & Acc. & 82.64 / 85.13 / 87.98 & -- \\
\midrule
\rowcolor{gray!2}
 &  &  & RAF-DB $\rightarrow$ CK+ / MMI / FER-2013 & Acc. & 81.72 / 65.63 / 48.68 & -- \\
\cmidrule[\lightrulewidth]{4-7}
\rowcolor{gray!2}
\multirow{-2}{=}{SPWFA-SE~\citep{li2020facial}} & \multirow{-2}{=}{ResNet-50 (IN) + SE} & \multirow{-2}{=}{\centering Direct transfer} & AffectNet-7 $\rightarrow$ CK+ / MMI / FER-2013 & Acc. & 85.44 / 70.43 / 50.29 & -- \\
\midrule
\rowcolor{gray!20}
Ada-CM~\citep{li2022towards} & ResNet-18 (MS1M) & Direct transfer & RAF-DB (4k labels) $\rightarrow$ CK+ & Acc. & 85.32 & -- \\
\midrule
\rowcolor{gray!2}
 &  &  & AffectNet-7 $\rightarrow$ RAF-DB / FER+ & Acc. & 76.16 / 64.32 & 92.21 / 91.01 \\
\cmidrule[\lightrulewidth]{4-7}
\rowcolor{gray!2}
\multirow{-2}{=}{S2D~\citep{chen2024static}} & \multirow{-2}{=}{ViT-B/16 (MAE, AffectNet)} & \multirow{-2}{=}{\centering Direct transfer} & RAF-DB $\rightarrow$ AffectNet-7 & Acc. & 47.57 & 66.42 \\
\midrule
\rowcolor{gray!20}
FeatRef~\citep{zhou2022feature} & 2-stream Inception & Direct transfer & CASME II $\rightarrow$ SMIC-HS / SMIC-HS $\rightarrow$ CASME II & Acc. & 71.49 / 59.68 & -- \\
\midrule
\rowcolor{gray!2}
MOL~\citep{shao2025mol} & Conv.\ F5C (Aff-Wild2) & Direct transfer & CASME II / SAMM $\rightarrow$ SMIC-HS & Acc. & 47.13 / 44.58 & 80.71 \\
\midrule
\rowcolor{gray!20}
Emotion-LLaMA~\citep{cheng2024emotionllama} & LLaMA2-7B + visual/audio enc. & Zero-shot & MERR $\rightarrow$ DFEW & WAR & 59.37 & -- \\
\bottomrule
\end{tabular}
\end{adjustbox}
\end{table*}

\subsection{Quantitative comparison per recognition task}
\label{ssec:perf_quantitative}

Tables~\ref{tab:perf_sfer}--\ref{tab:perf_intensity} consolidate the quantitative performance of representative deep learning-based FER methods, organized in one table per recognition task. Note that the reported values are taken from the original publications and should not be regarded as directly comparable across methods, since they rest on heterogeneous experimental conditions, including different train/validation/test splits, varying pre-training corpora and backbones, task-specific protocols, and partly diverging metric variants. They should therefore be read as indicative of methodological trends and order-of-magnitude differences rather than as a strict ranking. Accordingly, bold entries denote the highest value reported for each benchmark column and do not constitute a ranking under a common protocol, while the observations of Section~\ref{ssec:perf_discussion} describe recurring patterns among the reported results rather than causal effects of specific design choices. In addition, Tables~\ref{tab:perf_sfer}--\ref{tab:perf_intensity} report, for every method, the backbone and the source of its pre-trained weights, the input resolution or clip sampling, the input modality (where several modalities occur), the use of landmark or AU priors (where applicable), the evaluation protocol whenever it deviates from the standard protocol of the benchmark, and whether the result is obtained within or across datasets, so that the comparability of any two entries can be assessed directly. For this purpose, the reported values and protocols were checked against the original publications; values obtained under a different class set, split, or metric are labeled explicitly (e.g., 7- versus 8-class AffectNet, 5- versus 3-class and accuracy versus UF1 for MER), and bold is computed only among the entries that share the standard protocol of each benchmark. Notably, 184 of the 185 values of Tables~\ref{tab:perf_sfer}--\ref{tab:perf_intensity} are obtained within-dataset, the only exception being a cross-domain few-shot compound expression recognition result (Table~\ref{tab:perf_cfer}); the cross-dataset results reported by the reviewed studies are therefore collected separately in Table~\ref{tab:perf_cross}. The year column of each table gives the publication year of the cited reference, as in Table~\ref{tab:FERmilestones}. The pre-training of the backbones ranges from training from scratch and ImageNet initialization to face-recognition corpora (e.g., MS-Celeb-1M), masked-autoencoder pre-training on face images or videos, and image--text (CLIP) pre-training, which constitutes a further source of heterogeneity across the entries of each table.

\subsection{Comparative analysis and key insights}
\label{ssec:perf_discussion}

Building on the most widely adopted performance metrics (Section~\ref{ssec:perf_metrics}) and the per-task quantitative comparison of recent deep learning-based FER methods (Section~\ref{ssec:perf_quantitative}), this subsection synthesizes the main cross-task observations that emerge from Tables~\ref{tab:perf_sfer}--\ref{tab:perf_cross}. Each observation is stated together with the evidence that supports it and with the counter-evidence or limitations that constrain it. Four types of evidence are used: a) Counts of entries in Tables~\ref{tab:perf_sfer}--\ref{tab:perf_cross}, where a pattern is retained only if it is more frequent among the highest reported values (i.e., the three highest entries of each benchmark column, among the entries that share its standard protocol) than among all entries of the same table; b) Counts of primary studies from the cross-criteria analysis of Section~\ref{sec:cross_criteria}; c) Controlled ablations reported in the original publications, which isolate a single design choice under a fixed protocol; and d) The cross-dataset results of Table~\ref{tab:perf_cross}. Since no re-evaluation under a common protocol was performed, the observations describe associations in the reported results and should not be read as causal conclusions.
\begin{itemize}
    \item \uline{No single architecture family is consistently associated with the highest reported results}: Among the 15 top-three entries of the five SFER columns of Table~\ref{tab:perf_sfer}, CNN-based, hybrid CNN--Transformer, and Transformer-based models account for 8, 4, and 3 entries, respectively, compared with 22, 20, and 4 of the 52 fully supervised SFER entries with a known class set and the standard metric (53\%, 27\%, and 20\% versus 42\%, 38\%, and 8\%); hence, CNN-based models are moderately over-represented, hybrid models are not, and the three Transformer-based top-three entries all stem from a single model (S2D), and the highest values are distributed across a quaternion CNN--Transformer (QWTR~\citep{zhou2025delving} on AffectNet-7), lightweight CNNs (MHAN~\citep{wang2026mhan} on RAF-DB, tied with the Transformer-based S2D~\citep{chen2024static}; BReG-NeXt~\citep{hasani2020breg} on AffectNet-8; MLCNN~\citep{nguyen2019facial} on FER-2013), and a hybrid model (POSTER~\citep{zheng2023poster} on FER+). For in-the-wild DFER, Transformer-based backbones account for 8 of the 9 top-three entries on DFEW, FERV39k, and MAFW (MMA-DFER~\citep{chumachenko2024mma}, S2D~\citep{chen2024static}, MAE-DFER~\citep{sun2023maedfer}) and a CLIP-based model for the remaining one (TG-DFER~\citep{jung2025text}). For MER, the highest LOSO accuracies under the standard class sets are reported by CNN-based models (MERASTC~\citep{gupta2021merastc} on CASME II and SAMM, LTR3O~\citep{zhu2025learning} on SMIC-HS), followed by hybrid (SODA4MER~\citep{zhang2025dynamic}, MOL~\citep{shao2025mol}) and ViT-based (SRMCL~\citep{bao2024boosting}) models, whereas the highest AU detection F1-scores on both BP4D and DISFA are reported by a graph-based model (ME-GraphAU~\citep{luo2022megraphau}). \textit{Counter-evidence and limitations:} The more recent hybrid designs are not over-represented among the highest SFER values, and Transformer-based ones only through a single model (S2D), whereas Transformer-based backbones are over-represented among the highest in-the-wild DFER values; most benchmark columns contain fewer than 15 entries; and the architecture family is confounded with the pre-training source and the input resolution (Tables~\ref{tab:perf_sfer}--\ref{tab:perf_dfer}). Hence, the reported results do not single out an architecture family, and the task-dependent choices are consistent with, rather than evidence for, the inductive biases discussed in Section~\ref{ssec:nn_architectures_discussion}.
    \item \uline{Masked-autoencoder pre-training on face data as a recurring component of in-the-wild DFER methods}: Masked-autoencoder (MAE) self-supervised pre-training on face images or videos is used by the three methods that account for 8 of the 9 top-three in-the-wild DFER entries (MMA-DFER, S2D, MAE-DFER; Table~\ref{tab:perf_dfer}), as well as by the two methods reporting the highest Aff-Wild2 CCC values (Mamba-VA~\citep{liang2025mamba}, MAE+TCN~\citep{zhou2024enhancing}; Table~\ref{tab:perf_va}). Controlled comparisons within the original publications point in the same direction, namely for the VideoMAE baseline reported with MAE-DFER, pre-training the same backbone on VoxCeleb2 instead of Kinetics-400 increases WAR from 70.61\% to 74.60\% on DFEW, from 49.61\% to 52.39\% on FERV39k, and from 51.74\% to 53.51\% on MAFW~\citep{sun2023maedfer}, while initializing S2D from face-specific (AffectNet) instead of ImageNet MAE weights increases WAR from 70.91\% to 76.16\% on DFEW (fold~1) and from 47.36\% to 52.56\% on FERV39k~\citep{chen2024static}. \textit{Counter-evidence and limitations:} Video-scale pre-training is used only by MAE-DFER, whereas S2D and MMA-DFER rely on image-level pre-training; the highest DFEW and MAFW values are obtained by an audio-visual model, whose visual encoder alone (frozen, with a linear classifier) reports 67.15\% WAR on DFEW fold~1, compared with 77.92\% for the full audio-visual model~\citep{chumachenko2024mma}, which indicates that the input modality and the adaptation scheme contribute substantially as well; and since these factors have not been compared under a common protocol, the size of the contribution of pre-training cannot be isolated from the reported values.
    \item \uline{Persistent lab/in-the-wild gap and saturation on categorical SFER}: The highest values reported under lab-controlled DFER conditions (92.50\% top-1 accuracy on MMI and Oulu-CASIA) remain well above those reported on in-the-wild benchmarks (52.56\% and 58.45\% WAR on FERV39k and MAFW, respectively), although the two groups also differ in metric and in the number of reported methods. At the same time, the five highest RAF-DB accuracies lie within 0.52 percentage points (92.05--92.57\%) and the three highest FER+ accuracies within 0.53 percentage points (91.09--91.62\%). This is consistent with an annotation-induced ceiling rather than with a limitation of the models alone, where on AffectNet, two independent annotators agreed on the categorical label of only 60.7\% of the doubly annotated images~\citep{mollahosseini2017affectnet}, which is of the same order as the highest AffectNet accuracies of Table~\ref{tab:perf_sfer} (68.37\% and 68.50\% for the 7- and 8-class protocols). The cross-dataset results of Table~\ref{tab:perf_cross} further show that within-dataset values overestimate the performance on unseen corpora, where the accuracy of the SFER model of S2D drops from 92.21\% (within RAF-DB) to 76.16\% when trained on AffectNet and tested on RAF-DB, and from 66.42\% (within AffectNet-7) to 47.57\% in the opposite direction, while a ResNet-50 trained on RAF-DB reaches an average accuracy of 58.79\% on five target corpora, which unsupervised domain adaptation raises to 66.13\% (AGRA~\citep{chen2021cross}). \textit{Counter-evidence and limitations:} The cross-dataset results are available for few methods and label sets, so the size of the gap cannot be generalized across architectures; nevertheless, cross-corpus and robustness-oriented evaluation appears more informative than further gains on saturated within-dataset benchmarks.
    \item \uline{Landmark and AU priors are over-represented among the highest MER and, to a lesser extent, SFER results}: Landmark or AU priors are used by 6 of the 9 top-three MER entries, compared with 12 of the 29 MER entries under the standard protocols (67\% versus 41\%), mainly through MERASTC, SODA4MER, and MOL, and by 6 of the 15 top-three SFER entries, compared with 14 of the 52 SFER entries (40\% versus 27\%), through S2D, POSTER, and POSTER++. Within-paper ablations report consistent gains, where removing the landmark stream of POSTER reduces the accuracy from 92.05\% to 90.51\% on RAF-DB and from 67.31\% to 64.95\% on AffectNet-7~\citep{zheng2023poster}; removing the landmark prompts of S2D reduces it from 92.21\% to 91.07\% on RAF-DB (SFER model without AffectNet-8 pre-training) and the WAR from 76.16\% to 75.48\% on DFEW (fold~1)~\citep{chen2024static}; removing the landmark-estimation task of MOL reduces the CASME II accuracy from 79.23\% to 76.60\%~\citep{shao2025mol}; and the full AU-relation modeling pipeline of ME-GraphAU (including its losses) raises the BP4D F1-score of its ResNet-50 backbone from 59.1\% to 64.7\%~\citep{luo2022megraphau}. \textit{Counter-evidence and limitations:} In AU detection, such priors are less frequent among the top-three entries than among all entries (3 of 6 versus 8 of 13), and in DFER they are rare (3 of 14 versus 5 of 36); several of the highest values are obtained without any prior (e.g., QWTR, MHAN, BReG-NeXt, and MLCNN in SFER, and LTR3O in MER); the ablation gains in SFER, DFER, and MER (0.7--2.6 percentage points) are all measured within-dataset; and none of the cross-dataset results of Table~\ref{tab:perf_cross} isolates the effect of the prior. Whether anatomical priors improve generalization across corpora, as is often assumed, therefore remains a hypothesis to be tested under controlled cross-dataset protocols.
    \item \uline{Language-grounded models are competitive, rarely the highest, and the only ones reporting zero-shot results}: CLIP- or LLM-based models account for 11 of the 138 entries of Tables~\ref{tab:perf_sfer}--\ref{tab:perf_va}, \ref{tab:perf_mer}, and~\ref{tab:perf_au}, and for 1 of their 54 top-three entries (TG-DFER, third on FERV39k; MER-CLIP ranks fourth on CASME II); on DFEW, CLIP-based models (71.25--74.04\% WAR) report lower values than MAE-pre-trained ones (74.43--77.51\%). Within-paper comparisons indicate that the textual stream mainly adds small within-dataset gains, e.g., from 90.45\% to 91.30\% on RAF-DB when adding text supervision to FER-Former~\citep{li2024fer} and from 70.29\% to 71.25\% WAR on DFEW when replacing class-name prompts by LLM-generated descriptions in DFER-CLIP~\citep{zhao2023prompting}, whereas its distinctive contribution is the ability to recognize expressions without task-specific training, where an instruction-tuned MLLM reaches 59.37\% WAR on DFEW in a zero-shot setting (Emotion-LLaMA~\citep{cheng2024emotionllama}; Table~\ref{tab:perf_cross}), compared with 20.07\% for zero-shot CLIP~\citep{zhao2023prompting}. In the reported results, these models therefore appear complementary to, rather than a replacement for, task-specific models, while additionally enabling open-vocabulary and low-resource deployment. \textit{Counter-evidence and limitations:} No language-grounded result on compound expression recognition, dimensional VA estimation, or expression intensity estimation is available in the reviewed tables, and the few CFER results are obtained on different benchmarks, class sets, and metrics (Table~\ref{tab:perf_cfer}). Moreover, the metric--task coupling remains asymmetric, in the sense that while categorical SFER/DFER/MER/AU are consistently evaluated with accuracy/UAR/WAR/F1, VA estimation, CFER, and intensity estimation would benefit from ABAW-style standardization (CCC for VA, F1 for AU/compound, PCC for ERI) for reproducible quantitative analyses.
\end{itemize}

\section{Current challenges}
\label{sec:challenges}

Despite the large body of recent work in the field of FER and the substantial advances achieved across categorical, dimensional, compound, micro-expression, AU detection, and intensity-estimation tasks (Sections~\ref{sec:recognition_tasks}--\ref{sec:performance_metrics}), significant challenges and open research problems remain; if addressed, they would further increase the generalization ability, reliability, fairness, interpretability, and real-world adoption of FER systems. Unlike conventional visual recognition tasks, FER is intrinsically subjective, context-dependent, anatomically grounded, temporally variable, and demographically sensitive, while the underlying labels span multiple incompatible taxonomies (categorical, dimensional, FACS-based AUs, ordinal intensities). In the remainder of this section, the main challenges identified in the literature are systematically examined and outlined along six complementary axes, namely data and annotation, generalization and robustness, representation and temporal modeling, computation and deployment, evaluation, and trustworthiness/fairness/ethical aspects.

\subsection{Data and annotation aspects}
\label{ssec:challenges_data}

In contrast to generic visual recognition data, FER data are subjective, context-dependent, and unevenly annotated across recognition settings. In particular, SFER typically relies on image-level expression labels, DFER on video-level labels, MER on onset--apex--offset annotations, AU detection on FACS-based AU labels, dimensional VA estimation on continuous valence--arousal values, and intensity estimation on ordinal labels. The main data and annotation challenges are:

\begin{itemize}
    \item \uline{Limited and imbalanced annotated data}: Although SFER benefits from relatively large in-the-wild datasets, data availability remains uneven across FER tasks. DFER requires costly video annotation, MER and AU detection rely on finer temporal or AU-level labels, and dimensional VA requires per-frame continuous annotations~\citep{jiang2020dfew,wang2022ferv39k,yan2014casme,davison2016samm,mavadati2013disfa,zhang2014bp4d}. As a result, rare expressions (e.g., fear, disgust, contempt), low-intensity affective states, compound expressions, and under-represented AU combinations remain difficult to model robustly, and the resulting long-tailed label distributions further degrade in-the-wild performance~\citep{zhang2021rul,kollias2024distribution}.
    \item \uline{Annotation ambiguity and noisy supervision}: Facial expression labels vary depending on the annotator, cultural context, available scene information, and adopted emotion taxonomy. This becomes particularly challenging in the wild, where samples may contain mixed, subtle, low-confidence, or partially visible expressions. The transition from FER-2013 to FER+ illustrates this issue, since FER+ was introduced to refine the original single-label annotations through label distributions~\citep{goodfellow2013challenges,barsoum2016training}. The high inter-class similarity between confusable categories (e.g., fear--surprise, anger--disgust) and the inherent subjectivity of crowdsourced labels further amplify label noise~\citep{min2026robust,zhang2021rul}.
    \item \uline{Weak, biased, and incomplete supervision}: In video-based FER, a single label may be assigned to an entire sequence, although only a few frames contain discriminative affective evidence~\citep{wang2023rethinking}. Similarly, MER depends on accurate onset, apex, and offset localization, which is difficult to annotate consistently and often requires expert-level inspection~\citep{yan2014casme,davison2016samm}. In parallel, demographic, cultural, and acquisition biases related to age, gender, ethnicity, pose, illumination, and recording conditions may lead models to learn dataset-specific or identity-related correlations instead of robust affective representations~\citep{xu2020investigating,churamani2022domain}.
    \item \uline{Cross-taxonomy label heterogeneity}: Different FER tasks are annotated under largely incompatible label spaces (basic-6/7/8 categorical, compound, FACS AUs, valence--arousal, ordinal intensities, language-based descriptions). The absence of a unified annotation protocol limits direct dataset fusion, complicates multi-task supervision, and forces methods to rely on lossy label mappings or task-specific heads~\citep{ekman1997what,kollias2024distribution,zhang2022transformer}.
    \item \uline{Scarcity of multi-modal, in-context affective data}: Affect-relevant cues frequently extend beyond the face, including speech, body motion, scene context, and physiological signals. Nevertheless, large-scale datasets that jointly provide synchronized, high-quality annotations across these modalities, in the wild, and at scale remain scarce~\citep{cheng2024emotionllama,liu2025mer}.
\end{itemize}

\subsection{Generalization and robustness aspects}
\label{ssec:challenges_generalization}

FER methods are commonly developed and evaluated on specific benchmarks; however, real-world deployment requires robust operation across diverse subjects, acquisition protocols, environmental conditions, and even adversarial perturbations. The following generalization and robustness challenges can be identified:

\begin{itemize}
    \item \uline{Cross-dataset and cross-domain generalization}: FER models often exhibit reduced performance when transferred across datasets, mainly due to differences in annotation protocols, subject populations, camera settings, expression intensity, and recording conditions~\citep{li2020deep,xie2022crossdomain,churamani2022domain}. This issue becomes more evident in DFER, MER, and AU detection, where the available datasets are smaller and more sensitive to protocol-specific biases~\citep{jiang2020dfew,yan2014casme,mavadati2013disfa}, while source-free and test-time adaptation under continually shifting target distributions remains largely unsolved~\citep{cui2025learning}.
    \item \uline{In-the-wild visual variability}: Real-world FER scenarios involve pose changes, illumination variation, partial occlusion, motion blur, low resolution, head movement, and background clutter. These factors may hide expression-relevant facial regions or introduce nuisance visual cues, making the extraction of robust affective representations particularly challenging~\citep{gera2022cern,zhang2022learn,wang2020region}.
    \item \uline{Subject and identity dependence}: Facial morphology, expressiveness, habitual motion patterns, and identity-related appearance cues vary significantly across individuals. Consequently, FER models may rely on subject-specific patterns rather than expression-related variations, reducing their ability to generalize to unseen subjects and diverse populations~\citep{zeng2022face2exp,jiang2022disentangling}. This issue is particularly critical in MER and intensity estimation, where person-specific baseline expressiveness strongly modulates the observed signal.
    \item \uline{Cross-cultural and cross-demographic transfer}: Cultural display rules, social norms, and age- and ethnicity-related differences modulate both the production and perception of facial expressions, while most large-scale FER corpora over-represent specific demographic groups~\citep{churamani2022domain,xu2020investigating}. As a consequence, models trained predominantly on Western, adult, lab-collected data may underperform or exhibit systematic biases on under-represented populations~\citep{li2020deep,dominguez2024metrics}.
    \item \uline{Adversarial and corruption robustness}: Even minor, imperceptible perturbations of the input, common image corruptions (e.g., compression, blur, noise), and deepfake/synthetic manipulations can lead to significant drops in FER accuracy or to systematic misclassifications, raising concerns for security-critical deployment scenarios~\citep{kim2024adversarial,deepfakeau2025,min2026robust,wang2020region}.
\end{itemize}

\subsection{Representation and temporal modeling aspects}
\label{ssec:challenges_representation}

A central challenge in FER is the extraction of reliable affective representations from facial signals that are subtle, localized, temporally variable, and often semantically ambiguous. This issue appears differently across the considered tasks, where SFER mainly depends on static appearance cues, DFER and MER require the modeling of facial motion over time, AU detection focuses on anatomically grounded muscle activations, while dimensional and intensity estimation tasks require continuous, ordinal-aware regression. Regarding representation and temporal modeling, the main challenges are:

\begin{itemize}
    \item \uline{Subtle and localized affective cues}: Facial expressions are often conveyed through small deformations in specific regions, such as the eyes, eyebrows, nose, and mouth. Global representations may fail to capture these weak local variations, whereas overly local models may ignore the overall facial configuration. This limitation is particularly important in MER and AU detection, where discriminative evidence is frequently restricted to low-intensity muscle activations~\citep{ekman1997what,li2020deep,luo2022megraphau}.
    \item \uline{Temporal dynamics and motion representation}: Video-based FER requires the analysis of expression evolution, including onset, apex, offset, and frame-to-frame facial motion. However, useful temporal cues can be short, weak, noisy, or entangled with irrelevant head movement, making simple frame aggregation insufficient. This is especially critical in DFER and MER, where models need to separate expression-related dynamics from neutral, transitional, or non-discriminative temporal segments~\citep{jiang2020dfew,yan2014casme,davison2016samm,sun2023maedfer,wang2023rethinking}.
    \item \uline{Relationships between expressions, AUs, and semantic affect}: Categorical expressions, AUs, valence--arousal values, and language-based affect descriptions represent the displayed facial behavior from different but complementary perspectives. However, these targets are still commonly modeled in isolation, limiting the exploitation of anatomical, dimensional, and semantic relations. Bridging these representations remains an important challenge for improving interpretability, cross-task transfer, and robust affect reasoning~\citep{ekman1997what,kollias2024distribution,narayan2025facexformer,shahreza2025facellm}.
    \item \uline{Context and multi-modal affect ambiguity}: Facial appearance alone may not provide sufficient evidence for reliable affect interpretation, since similar facial displays can have different meanings depending on speech, body motion, scene context, social interaction, and cultural background. Nevertheless, many FER methods still rely mainly on facial appearance or facial motion, limiting their ability to resolve ambiguous, compound, or context-dependent affect. This challenge becomes increasingly important with the emergence of multi-modal and vision--language FER models, where visual, textual, audio, and AU-level information~\citep{zhang2022transformer,li2024fer,liu2025mer,cheng2024emotionllama}, or facial and scene-context cues~\citep{lee2019context}, need to be aligned without introducing noisy or misleading semantic cues.
    \item \uline{Hallucinations and semantic-anatomical mismatches in MLLM-based FER}: Recent MLLM- and CLIP-based FER methods can produce linguistically plausible but anatomically unsupported affect descriptions, mainly because they rely on general-purpose vision--language priors that are not grounded in FACS or facial dynamics~\citep{cheng2024emotionllama,shahreza2025facellm,ma2025multimodal,pei2025multi}. Aligning open-vocabulary affect reasoning with anatomically faithful facial behavior remains an open challenge.
\end{itemize}

\subsection{Computational and deployment aspects}
\label{ssec:challenges_computation}

In real-world FER applications, inference speed and resource requirements are critical, particularly for on-device, mobile, and embedded deployment scenarios (e.g., driver monitoring, robotics, healthcare wearables). The main computational and deployment challenges are:

\begin{itemize}
    \item \uline{Scale of recent foundation-style backbones}: Recent FER models often employ large visual backbones, temporal aggregation modules, multi-modal encoders, MLLM/CLIP integrations, or ensemble mechanisms. Although these designs improve recognition performance, they impose substantial memory and compute requirements that may be difficult to satisfy in resource-constrained settings, especially for video-based and frame-level tasks~\citep{li2020deep,chen2024static,chumachenko2024mma,liang2025mamba}.
    \item \uline{Real-time and low-latency operation}: Many FER deployment scenarios (e.g., human--robot interaction, driver monitoring, live affective computing) require frame- or sub-frame-level inference latency. Heavy attention-based, MAE-pre-trained, and MLLM-based pipelines typically exceed these latency budgets without careful optimization~\citep{kong2022real,chang2024libreface,driveredge2025sensors,duongthang2025driver}.
    \item \uline{On-device and edge constraints}: Edge devices exhibit limitations in available memory, processing time, energy budget, and thermal envelope, which preclude the direct deployment of full-scale FER backbones. Lightweight architectures, knowledge distillation, parameter-efficient adaptation, and quantization-friendly designs are therefore needed, but typically incur measurable accuracy drops with respect to their full-scale counterparts~\citep{chen2024lightweight,savchenko2022classifying,hinton2015distilling,lightdriver2025designs}.
    \item \uline{Continual and on-the-fly personalization}: Once deployed, FER systems frequently encounter new subjects, new environments, and gradual user-specific drifts (e.g., habitual expressions, aging, accessories). Most current methods are not designed for safe, low-cost, on-device continual or personalized adaptation, which raises stability--plasticity and catastrophic-forgetting concerns under realistic operating conditions~\citep{wang2021tent,liang2020shot,yuan2024auformer}.
\end{itemize}

\subsection{Evaluation aspects}
\label{ssec:challenges_evaluation}

The evaluation of FER methods remains strongly influenced by the selected datasets, protocols, and metrics. Although existing benchmarks have enabled measurable progress, they do not always capture cross-domain robustness, temporal localization, fairness, calibration, or real-world reliability. The following evaluation challenges can be identified:

\begin{itemize}
    \item \uline{Protocol and metric inconsistency}: FER studies often adopt different pre-processing pipelines, data splits, label mappings, frame sampling strategies, and validation protocols, making direct comparison between methods difficult~\citep{li2020deep,khan2025visir,kollias2024abaw6}. Moreover, overall accuracy can be misleading under class imbalance, sparse AU activations, or rare micro-expression categories, while metrics such as macro-F1, unweighted/weighted average recall, balanced accuracy, CCC/PCC, and average precision are not used consistently across tasks~\citep{jiang2020dfew,yan2014casme,kollias2024distribution}.
    \item \uline{Lack of a unified evaluation framework}: No holistic evaluation protocol currently exists for diagnosing the multiple factors behind FER failures. The available metrics are often coarse, failing to clearly highlight the underlying factors leading to low performance (e.g., subtle AU misdetections, micro-expression apex mislocalization, intensity underestimation in low-arousal regions), while task- and corpus-specific metrics make cross-architecture and cross-modality comparisons inherently asymmetric~\citep{dominguez2024metrics,chen2021cross}.
    \item \uline{Imprecise assessment of generalization ability}: Despite the increasing emphasis on cross-corpus, cross-cultural, and zero-shot evaluation, the comprehensive, accurate, and robust assessment of FER generalization remains difficult. Even small distribution shifts in capture conditions, prompts, or demographic composition can strongly affect model behavior, while leakage between large-scale pre-training corpora and downstream benchmarks further obscures the actual transfer ability of recent foundation-style FER backbones~\citep{xie2022crossdomain,churamani2022domain,chen2021cross}.
    \item \uline{Limited assessment of reliability, fairness, and calibration}: Most benchmarks mainly measure recognition performance, while fewer evaluations consider uncertainty calibration, demographic robustness, failure cases, or sensitivity to distribution shift~\citep{dominguez2024metrics,xu2020investigating,churamani2022domain}. As a consequence, current evaluation practices provide only a partial view of whether a model is sufficiently reliable and fair for practical deployment, and explicit fairness-aware, calibration-aware, and worst-group metrics are needed alongside conventional accuracy/F1/CCC scores.
\end{itemize}

\subsection{Trustworthiness, fairness, and ethical aspects}
\label{ssec:challenges_trust}

When FER methods are transferred from benchmark evaluation to real-world use, additional constraints arise regarding privacy, transparency, demographic equity, and responsible decision-making, particularly given the increasing regulatory scrutiny of emotion-recognition technologies. Regarding trustworthiness, fairness, and ethics, the main challenges are:

\begin{itemize}
    \item \uline{Demographic bias and fairness}: FER datasets and models exhibit systematic biases across age, gender, ethnicity, and skin-tone groups, which translate into uneven recognition performance and may amplify socio-cultural inequities~\citep{xu2020investigating,churamani2022domain,dominguez2024metrics}. The extent of such biases varies across architectures and training regimes, with high-accuracy models not necessarily being the fairest, and explicit fairness-aware training, evaluation, and reporting remain underdeveloped in the FER literature.
    \item \uline{Privacy and biometric sensitivity}: Facial expression data inherently contain identity-related information. Therefore, even models trained for expression, AU, or VA estimation may preserve sensitive biometric cues, raising concerns in continuous monitoring, cloud processing, multi-modal logging, and long-term storage scenarios~\citep{zeng2022face2exp,jiang2022disentangling}. Privacy-preserving, federated, and on-device FER paradigms are still in their early stages, and standard FER pipelines do not natively enforce identity decoupling.
    \item \uline{Interpretability and explainability}: FER outputs correspond to probabilistic predictions of visible facial behavior, rather than direct measurements of internal emotional states. In contrast to modular FACS-based pipelines, modern end-to-end and MLLM-based FER models implement implicit, entangled representations that are difficult to attribute to specific facial regions, AUs, or temporal segments. This complicates the diagnosis of failure modes and limits user/operator trust in sensitive domains~\citep{li2020deep,dominguez2024metrics,narayan2025facexformer}.
    \item \uline{Regulatory and responsible-use constraints}: Emotion-recognition technologies (as termed in the regulatory texts) are increasingly subject to regulatory restrictions (e.g., workplace, education, public-space monitoring), reflecting concerns about scientific validity, consent, and potential misuse. FER systems therefore need to clearly communicate their probabilistic nature, scope of validity, and known limitations, and to support application-aware, transparent, and responsibly deployed designs~\citep{li2020deep,dominguez2024metrics}. The recent emergence of dedicated initiatives on trustworthy facial affect analysis further highlights this need, particularly along the axes of fairness, explainability, and safety.
    \item \uline{Synthetic-media and impersonation risks}: The wide availability of generative models enables the production of high-quality synthetic faces and expression manipulations, which can degrade FER accuracy on real samples and be weaponized to spoof affective monitoring systems~\citep{deepfakeau2025,kim2024adversarial}. Robust FER under deepfake-style manipulation, together with the joint detection of synthetic content and expression labels, remains an open research direction.
\end{itemize}

\section{Future research directions}
\label{sec:future_directions}

The introduction of deep learning-based, multi-modal, and foundation-style methods in the field of FER has led to substantial advances across all core FER tasks, as discussed in Sections~\ref{sec:Evolution}--\ref{sec:performance_metrics}. Despite these developments, several open challenges remain, which restrict the wider deployment of FER solutions in real-world scenarios, as outlined in Section~\ref{sec:challenges}. This section discusses the most promising future research directions, largely mirroring these challenges, towards data-efficient, robust, semantically grounded, fair, interpretable, and deployment-aware FER systems.

\subsection{Data-centric and annotation-efficient learning}
\label{ssec:research_data}

A first important direction concerns the improvement of the data and supervision mechanisms used for training FER models. Although large-scale datasets have contributed significantly to recent progress, FER remains strongly affected by scarce labels, class imbalance, subjective annotations, and task-dependent supervision requirements. The following research directions emerge:

\begin{itemize}
    \item \uline{Uncertainty-aware annotation protocols}: Future FER datasets should move beyond single-label annotations, especially in the wild, where facial displays may be ambiguous, mixed, subtle, or context-dependent. Label distributions, annotator-disagreement modeling, confidence-aware supervision, and uncertainty-aware training objectives can provide a more realistic representation of affective ambiguity than hard categorical labels~\citep{goodfellow2013challenges,barsoum2016training,le2023uncertainty,kollias2024distribution,geng2016label}.
    \item \uline{Learning with limited and imperfect supervision}: Self-supervised, semi-supervised, contrastive, and few/zero-shot learning paradigms are expected to become increasingly important for DFER, MER, AU detection, intensity estimation, and other tasks where dense annotations are costly~\citep{li2022towards,fan2023selfme,zhi2022micro,li2025enhanced,bao2024boosting,sun2023maedfer}. Future methods should better exploit unlabeled facial images and videos, and prevent noisy pseudo-labels or unreliable temporal cues from being reinforced during training.
    \item \uline{Bias-aware data construction}: Future datasets should better represent demographic, cultural, and acquisition diversity, while also reducing hidden correlations between affective labels and identity-related factors. This is critical because FER models may otherwise learn shortcuts related to age, gender, ethnicity, pose, illumination, or recording conditions instead of expression-related information~\citep{zeng2022face2exp,jiang2022disentangling,dominguez2024metrics,xu2020investigating,churamani2022domain}.
    \item \uline{Generative and synthetic data augmentation}: High-fidelity generative models (GANs, diffusion models, 3D/4D morphable face models) increasingly enable the synthesis of identity-controlled, AU-conditioned, and intensity-controlled training samples that can mitigate class imbalance and demographic gaps~\citep{preechakul2022diffusion,zou2024fourdfm}. Future research should focus on faithful, AU-grounded synthesis with reliable automatic label calibration, and on the joint use of synthetic and real data for in-the-wild FER, ensuring that synthesis does not amplify pre-existing biases.
\end{itemize}

\subsection{Robustness and generalization}
\label{ssec:research_generalization}

A second direction concerns the development of FER models that remain reliable beyond the corpora on which they were trained. Current methods often achieve strong performance under standard benchmark protocols, but their reliability decreases under cross-dataset evaluation, unseen subjects, different acquisition conditions, demographic shifts, and adversarial perturbations. Promising research directions include:

\begin{itemize}
    \item \uline{Cross-dataset and domain-generalizable learning}: Future FER research should place stronger emphasis on domain adaptation, domain generalization, and cross-dataset testing. Instead of optimizing only for within-dataset performance, models should be trained to learn affective representations that remain stable across different annotation protocols, subject populations, camera settings, and recording environments, exploiting modern source-free, test-time, and continually adaptive paradigms~\citep{li2020deep,xie2022crossdomain,churamani2022domain,cui2025learning,wang2021tent,liang2020shot}.
    \item \uline{Disentanglement of affective and nuisance factors}: A promising direction is to explicitly separate expression-related information from nuisance factors, including identity, pose, illumination, head motion, background context, and dataset-specific artifacts. Disentangled, adversarial, causal, and identity-invariant representation learning can reduce the reliance on spurious correlations and improve generalization to unseen subjects and demographic groups~\citep{zeng2022face2exp,jiang2022disentangling,wang2020region}.
    \item \uline{Robustness under unconstrained visual conditions}: Future methods should be optimized and evaluated under realistic perturbations, such as occlusion, low resolution, motion blur, illumination variation, pose changes, and background clutter, which is particularly important for video-based FER, where expression evidence may be weak, temporally irregular, or mixed with irrelevant head and body motion~\citep{gera2022cern,zhang2022learn,chen2024static}.
    \item \uline{Adversarial robustness and synthetic-media resilience}: Given the proliferation of imperceptible perturbations, common corruptions, and deepfake-style facial manipulations, future FER systems should incorporate adversarial training, certified-robust losses, manipulation-aware features, and joint detection of synthetic content together with expression labels~\citep{kim2024adversarial,deepfakeau2025,min2026robust}. This is essential for safety-critical and security-sensitive deployment scenarios.
\end{itemize}

\subsection{Fine-grained and temporal affect representation}
\label{ssec:research_representation}

Future FER systems need to better capture the fine-grained facial evidence that differentiates affective states. This requirement is particularly important for MER, AU detection, intensity estimation, compound expressions, and DFER, where discriminative cues may be subtle, localized, short-lived, or physiologically structured. Addressing these challenges suggests the following directions:

\begin{itemize}
    \item \uline{Region-focused and AU-guided modeling}: Future models should incorporate informative facial regions and AU priors more explicitly. AU-guided attention, landmark-aware representations, graph-based facial structures, and muscle-region relation modeling can improve fine-grained discrimination and interpretability, especially when affective evidence is localized or low in intensity~\citep{ekman1997what,shao2019facial,luo2022megraphau,wang2024htnet,mao2025posterpp}.
    \item \uline{Stronger temporal modeling for dynamic FER}: DFER and MER require models that can capture short-term facial movements, long-range temporal dependencies, onset--apex--offset evolution, and irregular expression transitions. Future approaches should combine local motion extraction with temporal attention, snippet-level reasoning, graph-based dynamics, state-space backbones, and sequence modeling, while reducing dependence on manually selected apex frames or fixed temporal assumptions~\citep{wang2023rethinking,zhang2025dynamic,zhu2025learning,sun2023maedfer,liang2025mamba}.
    \item \uline{Unified expression, AU, dimensional, and intensity representations}: Categorical expressions, AUs, valence--arousal values, ordinal intensities, and language-based affect descriptions provide different but complementary views of facial behavior. Future FER methods should exploit these connections through shared latent spaces, multi-task learning, label-distribution learning, and cross-representation alignment, rather than treating each prediction target as an isolated task~\citep{ekman1997what,kollias2024distribution,narayan2025facexformer,kollias2023multi}.
    \item \uline{Long-horizon and continual affect modeling}: Realistic FER deployment requires the modeling of affective dynamics over minutes to hours (e.g., engagement, fatigue, pain, depression), as well as gradual adaptation to new subjects, environments, and emerging expression categories without catastrophic forgetting~\citep{wang2023rethinking}. Continual, replay-based, and parameter-efficient adaptation strategies are therefore expected to become first-class FER research components.
\end{itemize}

\subsection{Multi-modal and foundation-model-based affect reasoning}
\label{ssec:research_multimodal}

Another important direction concerns the integration of facial analysis with broader multi-modal and semantic reasoning capabilities. Facial appearance alone may not be sufficient for reliable affect interpretation, since the same facial behavior can have different meanings depending on speech, body motion, scene context, social interaction, and cultural background. Several research directions appear promising:

\begin{itemize}
    \item \uline{Context-aware multi-modal FER}: Future systems should integrate complementary cues, including audio, gaze, head motion, body posture, scene context, text, physiological signals, and interaction history~\citep{le2022global,chumachenko2024mma,zhang2022transformer,gu2024emotake}. Such information can help disambiguate weak facial evidence, improve robustness in unconstrained settings, and support application-specific tasks, such as depression detection, autism monitoring, pain estimation, and engagement analysis.
    \item \uline{Vision--language and prompt-based affect modeling}: VLMs/MLLMs provide a promising route to connect visual facial patterns with semantic descriptions of expressions, AUs, and affective states. Future work should further investigate prompt learning, AU-grounded textual descriptions, LLM-generated affective explanations, and CLIP-style alignment, while ensuring that textual priors do not introduce misleading stereotypes or oversimplified emotion assumptions~\citep{li2024fer,ma2025multimodal,pei2025multi,liu2025mer,zhao2023prompting,jung2025text}.
    \item \uline{FER-specific foundation models}: A longer-term direction is the development of foundation models specialized for facial expression analysis. Such models should be trained on heterogeneous facial data, including static images, videos, AU annotations, expression labels, valence--arousal annotations, intensity labels, and multi-modal interaction signals~\citep{narayan2025facexformer,cheng2024emotionllama,xing2024emollama}. If properly designed, they could provide reusable representations across SFER, DFER, MER, AU detection, compound expression recognition, expression intensity estimation, and application-domain affective tasks, reducing the need for isolated task-specific pipelines.
    \item \uline{Grounded and faithful affect reasoning}: As MLLMs enter FER, future work should ensure that generated affect explanations are grounded in visible facial evidence and AU-level cues, since general-purpose MLLMs may produce plausible but anatomically incorrect explanations~\citep{cheng2024emotionllama,xing2024emollama,shahreza2025facellm}. Verifiable-reward fine-tuning, FACS-grounded chain-of-thought supervision, and reasoning-trace auditing emerge as promising routes towards unified, interpretable FER foundation backbones.
\end{itemize}

\subsection{Reliable and standardized evaluation}
\label{ssec:research_evaluation}

Progress in FER also depends on stronger evaluation practices. Existing benchmarks have enabled measurable improvements, but they do not always provide sufficient evidence regarding robustness, fairness, temporal localization, uncertainty, calibration, or real-world reliability. The following directions can be identified:

\begin{itemize}
    \item \uline{More realistic evaluation protocols}: Future studies should report cross-dataset, cross-domain, and subject-independent evaluation more consistently. Protocols should also account for temporal localization in video-based tasks, demographic diversity, low-quality visual conditions, and distribution shifts, rather than relying only on standard within-dataset splits~\citep{li2020deep,jiang2020dfew,yan2014casme,khan2025visir,kollias2024abaw6,chen2021cross}.
    \item \uline{Task-appropriate and imbalance-aware metrics}: Evaluation should move beyond overall accuracy when datasets are imbalanced or labels are multi-label, continuous, ordinal, or temporally sparse. Metrics such as macro-F1, unweighted/weighted average recall, balanced accuracy, average precision, CCC/PCC, and calibration error should be selected according to the target FER task and reported consistently across studies~\citep{kollias2024distribution,dominguez2024metrics}.
    \item \uline{Fairness, uncertainty, and failure analysis}: Future benchmarks should incorporate fairness-oriented and uncertainty-aware evaluation. This requires reporting performance across demographic groups, expression categories, subject conditions, and acquisition settings, as well as analyzing failure cases and confidence calibration. Such evaluation is essential for determining whether FER systems are reliable enough for practical deployment~\citep{dominguez2024metrics,xu2020investigating,churamani2022domain}.
    \item \uline{Standardized challenge-driven and reproducibility-aware benchmarking}: ABAW-style standardization (CCC for VA, F1 for AU/compound, PCC for ERI) is a critical step towards comparable, reproducible quantitative analyses~\citep{kollias2022abaw,kollias2023abaw,kollias20247th,kollias2025abaw8}. Future work should further consolidate task-specific protocols, release reproducibility kits (data splits, pre-processing pipelines, evaluation code), and establish shared leakage-aware test partitions for foundation-style backbones.
\end{itemize}

\subsection{Efficient, private, and responsible deployment}
\label{ssec:research_deployment}

Future FER research should also consider deployment requirements from the early stages of model design. Real-world systems need to operate under computational constraints, preserve privacy, provide interpretable outputs, and comply with ethical and regulatory expectations. Relevant research directions include:

\begin{itemize}
    \item \uline{Lightweight and adaptive inference}: Future FER methods should further explore lightweight backbones, efficient attention mechanisms, knowledge distillation, pruning, quantization, early-exit strategies, and adaptive frame selection. These techniques are necessary for real-time, mobile, embedded, automotive, assistive, and human--computer interaction applications, especially for video-based FER where temporal processing increases computational cost~\citep{kong2022real,chen2024static,chen2024lightweight,savchenko2022classifying,hinton2015distilling,chang2024libreface,driveredge2025sensors,lightdriver2025designs}.
    \item \uline{Parameter-efficient and on-device adaptation}: Future FER systems should leverage parameter-efficient fine-tuning (LoRA, adapters, prompt tuning) and on-device personalization to enable safe, low-cost adaptation to new subjects and environments, while mitigating catastrophic forgetting and identity leakage~\citep{saadi2025peclip,chen2024static,yuan2024auformer,wang2021tent}. Joint compression--adaptation strategies are particularly relevant for resource-constrained, long-term deployments.
    \item \uline{Privacy-preserving FER}: Since facial expression data contain sensitive biometric and identity-related information, future systems should explore privacy-preserving training and inference mechanisms~\citep{zeng2022face2exp,jiang2022disentangling}. Federated learning, anonymized representations, identity-disentangled embeddings, differential privacy, secure on-device processing, and minimal-retention pipelines are expected to become standard components of FER systems deployed in continuous monitoring or cloud-based scenarios.
    \item \uline{Transparent and responsible use}: FER predictions should be communicated as uncertain estimates of visible facial behavior, rather than direct measurements of internal emotional states. Future systems should therefore include interpretable explanations, AU-level attributions, confidence estimates, human-in-the-loop safeguards, and application-specific usage constraints, especially in sensitive domains such as healthcare, education, recruitment, surveillance, and affect-aware human--computer interaction~\citep{li2020deep,dominguez2024metrics,narayan2025facexformer}.
    \item \uline{Application-aware co-design}: Future research should co-design FER pipelines with their target application domains, so that the choice of recognition task, modality, model size, latency budget, and trustworthiness mechanisms is driven by the actual deployment context (e.g., driver monitoring, clinical assessment, education, immersive media). This would shift the field from generic-benchmark optimization towards application-grounded, responsible FER systems~\citep{driveredge2025sensors,duongthang2025driver,depression2025neurocomputing}.
\end{itemize}

\section{Conclusion}
\label{sec:conclusion}

This survey provided a systematic, cross-task, and cross-criteria review of recent deep learning-based Facial Expression Recognition, with explicit links to the wider Facial Affect Recognition landscape. Following a structured literature review methodology, the field was examined along several complementary axes. Starting from its historical evolution, the research progress of the field was structured into five successive phases that reposition FER from anatomical coding and handcrafted descriptors towards large-scale, pre-trained, multi-modal, and language-grounded systems. Building on this perspective, a seven-criterion taxonomy of complementary criteria, each addressing a distinct question about a FER method, was introduced, organizing the literature along the targeted recognition task, input modality, face-analysis pre-processing pipeline, neural network architecture, learning strategy, acquisition setting, and application domain, and a per-criterion comparative analysis was performed to clarify the relative strengths and limitations of the corresponding method families. The discussion was complemented by a task-organized review of the main public FER datasets, a systematic compilation of the performance metrics adopted in the FER literature, and a per-task quantitative comparison of representative recent methods on the most widely adopted benchmarks. Beyond the separate analysis of each criterion, their joint instantiation in the reviewed literature was examined (Section~\ref{sec:cross_criteria}), revealing a strongly concentrated corpus, in which categorical recognition combined with CNN-, Transformer-, or hybrid backbones accounts for 81 of the 167 primary studies, while 22 of the 49 task--architecture combinations are not represented; notably, no language-grounded models were encountered for valence--arousal estimation, compound expression recognition, AU detection, or expression intensity estimation, whereas thermal/NIR inputs are still addressed mostly by CNN-based models (3 of 4 studies, the fourth being a CNN--Transformer hybrid) and 3D/4D inputs only by CNN-based recognition and generative synthesis models.

Several broad trends emerge from this review. First, a growing share of recent works adopts Transformer-based and hybrid backbones (Fig.~\ref{fig:bibliometrics}(d)), with attention, cross-fusion, and state-space components increasingly absorbed into CNN, GNN, and generative pipelines, blurring the boundary between architectural families. Second, AU- and landmark-grounded supervision is over-represented among the methods reporting the highest MER and, to a lesser extent, SFER results (e.g., 6 of the 9 top-three MER entries, compared with 12 of the 29 MER entries under the standard protocols), and within-paper ablations report gains of 0.7--2.6 percentage points in SFER, DFER, and MER, whereas its effect on cross-dataset generalization remains untested. Third, MAE-style self-supervised pre-training, parameter-efficient adaptation, and CLIP/MLLM-based language grounding are increasingly complementing purely supervised training over closed taxonomies, repositioning FER towards open-vocabulary, instruction-following, and explanation-aware systems. Fourth, no single input modality covers the entire FER task space uniformly, with static, dynamic, geometric, spectral, audio-visual, physiological, and language-grounded inputs contributing partly complementary affective evidence. Moreover, despite the substantial progress reported across all FER tasks, no recognition setting is fully solved, and the main bottlenecks differ in nature rather than in severity across tasks, modalities, architectures, and learning strategies. These trends are derived from results reported under heterogeneous experimental protocols (Section~\ref{ssec:perf_quantitative}) and from the composition of the analyzed corpus (Section~\ref{sec:data_extraction}); they therefore describe recurring patterns rather than a ranking of method families, and their confirmation requires common-protocol, cross-dataset benchmarking (Section~\ref{ssec:research_evaluation}); in this respect, the few cross-dataset results reported by the reviewed studies (Table~\ref{tab:perf_cross}), e.g., accuracy drops of 16--19 percentage points when the same model is transferred between AffectNet and RAF-DB, indicate that within-dataset values overestimate the performance on unseen corpora.

The present review also has limitations, which are detailed in Section~\ref{sec:review_limitations} and should be considered when using its findings. First, its quantitative synthesis relies on the results and ablations reported in the original publications, rather than on a controlled re-evaluation of the methods under a common protocol. Second, the heterogeneity of the evaluation protocols (class sets, data splits, metrics, and pre-training data) prevents a strict quantitative comparison or a meta-analysis of the reported values; the performance tables should therefore be read as a structured record of reported results, rather than as a leaderboard. Third, the seven criteria of the taxonomy are complementary but not orthogonal, and the assignment of each study to a single primary category per criterion simplifies methods that combine several design choices. Finally, the analyzed corpus is criteria-driven and prioritized, rather than exhaustive, and its coverage of the 2026 literature is partial.

Building on these observations, the survey systematically discussed the current challenges of the field along six complementary axes (data and annotation, generalization and robustness, representation and temporal modeling, computation and deployment, evaluation, and trustworthiness/fairness/ethical aspects), and outlined the most promising corresponding research directions. These directions span data-centric and annotation-efficient learning, robust and domain-generalizable models, fine-grained and temporal affect representation, multi-modal and foundation-model-based affect reasoning, reliable and standardized evaluation, and efficient, private, and responsible deployment. Overall, the field is moving from task-specific, closed-taxonomy pipelines towards data-efficient, semantically grounded, fair, interpretable, and deployment-aware FER systems, and the present survey aims to serve as a structured reference and roadmap for researchers and practitioners working towards this goal.

\printcredits

\bibliographystyle{cas-model2-names}

\bibliography{main}

\balance

\end{document}